\documentclass[lettersize,journal]{IEEEtran}
\usepackage{lineno}
\modulolinenumbers[5]
\usepackage{url}
\usepackage{xpatch}
\usepackage{tabularx}
\usepackage{threeparttable}
\usepackage{bm}
\usepackage{array}
\usepackage{graphicx}
\usepackage{amsmath,amssymb,amsthm}
\usepackage{enumerate}
\usepackage{graphicx}
\usepackage{caption}
\usepackage{multirow}
\usepackage{float}
\usepackage[table]{xcolor}
\usepackage{enumitem}
\usepackage{makecell}
\usepackage{subcaption}
\usepackage{algorithm}
\usepackage{algorithmic}
\usepackage{booktabs}
\usepackage{xcolor}
\usepackage{colortbl}
\usepackage{relsize}
\def\BibTeX{{\rm B\kern-.05em{\sc i\kern-.025em b}\kern-.08em
    T\kern-.1667em\lower.7ex\hbox{E}\kern-.125emX}}
\usepackage{balance}

\begin{document}

\captionsetup{font={footnotesize}}
\captionsetup[table]{labelformat=simple, labelsep=newline, textfont=sc, justification=centering}

\setlength{\heavyrulewidth}{1.5pt}
\setlength{\lightrulewidth}{0.7pt}

\title{Beyond Speedups: Hardware-Aware Evaluation of Evolutionary Algorithms on GPUs}

\author{
        Xinmeng Yu, Tao Jiang, Ran Cheng, Yaochu Jin, and Kay Chen Tan

\thanks{
        Xinmeng Yu and Tao Jiang are with the Department of Computer Science and Engineering, Southern University of Science and Technology, Shenzhen 518055, China.
        They are also with the Pengcheng Laboratory, Shenzhen 518000, China. (E-mails: yxm981203@163.com, jiangt1997@gmail.com.)

        Ran Cheng is with the Department of Data Science and Artificial Intelligence, The Hong Kong Polytechnic University, Hong Kong SAR, China. He is also with The Hong Kong Polytechnic University Shenzhen Research Institute, Shenzhen, Guangdong Province, China; The Hong Kong Polytechnic University-Daya Bay Technology and Innovation Research Institute, Huizhou,Guangdong Province, China. (E-mail: ranchengcn@gmail.com)  (\emph{Corresponding author: Ran Cheng}).

}
}

\markboth{Beyond Speedups}{Yu \MakeLowercase{\textit{et al.}}: Hardware-Aware Evaluation of Evolutionary Algorithms on GPUs}

\maketitle

\begin{abstract}
Evolutionary algorithms (EAs) are increasingly executed on graphics processing units (GPUs) to exploit population-level parallelism. This shift changes the resource model under which EAs are designed and evaluated. However, many existing GPU-based EA studies still focus mainly on implementation-level speedup after porting CPU-oriented algorithms to GPUs. As a result, they provide limited insight into how algorithmic mechanisms, function-evaluation (FE) budgets, population scales, and hardware utilization jointly affect optimization behavior. In response, this paper goes beyond speedup measurement and studies the scaling behavior of EAs on GPUs from a hardware-aware evaluation perspective. We evaluate 16 representative EAs on 30 benchmark problems across CPU and GPU platforms, covering single-objective optimization, multi-objective optimization, numerical benchmarks, and neuroevolution tasks. The study leads to four findings. First, GPU acceleration is highly heterogeneous across algorithms because different evolutionary mechanisms expose different degrees of batched computation, memory regularity, and synchronization. Second, FE-budgeted evaluation remains useful for measuring sample efficiency, but it can provide only a limited observation window under GPU execution; time-budgeted evaluation is therefore necessary for assessing practical time-to-solution and long-horizon search behavior. Third, GPU effectiveness depends on scaling regimes induced by problem dimension and population size, where parallelism may be underutilized, effective, or saturated. Fourth, GPU execution makes very large populations practically affordable, and several evolutionary mechanisms can convert this increased population scale into improved optimization performance. These results indicate that GPU parallelism should not be treated only as a post hoc acceleration tool. It should be considered as part of the evaluation and design assumptions of scalable EAs.
\end{abstract}


\IEEEpeerreviewmaketitle
\section{Introduction}
\IEEEPARstart{E}{volutionary} algorithms (EAs) are population-based, derivative-free optimization methods that are widely used when gradient information is unavailable, unreliable, or difficult to exploit~\cite{Baeck1993a, VanVeldhuizen1998}. Representative paradigms, including genetic algorithms (GAs)~\cite{Holland1975a}, particle swarm optimization (PSO)~\cite{Kennedy1995}, differential evolution (DE)~\cite{Storn1997}, and covariance matrix adaptation evolution strategies (CMA-ES)~\cite{Hansen2003}, have been applied to domains such as industrial process optimization and neural architecture search~\cite{Liang2024, Neri2007, Grantham2022, Zhao2024a, Wang2021, Xue2021}. However, the practical use of EAs is often constrained by computational cost. Conventional CPU-based studies typically adopt moderate population sizes and fixed function-evaluation (FE) budgets, which limit the observable search horizon and restrict the scale of problems that can be explored. Consequently, the empirical behavior of many EAs has been studied under resource assumptions that are shaped by CPU execution.

Recent advances in parallel computing hardware, especially graphics processing units (GPUs), have created new opportunities for scaling population-based search. GPUs provide high-throughput data-level parallelism and are well aligned with batched operations over candidate solutions. Early studies showed that computationally expensive components, especially fitness evaluation, can be offloaded to GPUs to obtain substantial runtime reduction~\cite{Wong2006,Souza2011, Qin2012}. More recent frameworks, such as EvoJAX~\cite{Tang2022}, evosax~\cite{Lange2023}, and EvoX~\cite{Huang2024}, further support vectorized and batched implementations of EAs, which enables more complete end-to-end GPU execution. These developments make larger populations, longer evaluation horizons, and more demanding benchmark settings practically feasible.

However, treating GPUs only as acceleration devices gives an incomplete view of their role in evolutionary computation. GPU execution changes the relationship among FE budgets, wall-clock time, population size, and algorithmic structure. An FE budget that is expensive on CPUs may be consumed quickly on GPUs, and a population size that is prohibitive on CPUs may become practical when evaluated in parallel. Therefore, the central question is not only whether GPUs accelerate EAs, but also how GPU execution changes the way EAs should be evaluated, compared, and designed. The conventional FE-centered evaluation protocol remains useful for measuring sample efficiency, but it is no longer sufficient as the only basis for comparing GPU-enabled EAs. A hardware-aware evaluation perspective is needed to understand not only whether an EA runs faster on GPUs, but also how it uses parallel hardware to produce search progress.

Several issues remain insufficiently understood under this perspective. First, GPU acceleration is not uniform across algorithms. Different EAs rely on different mechanisms for candidate generation, state update, population interaction, and selection. These mechanisms expose different levels of batched computation and synchronization to the hardware. Therefore, a direct CPU-to-GPU port does not necessarily imply comparable acceleration across algorithms.

Second, GPU execution changes the empirical meaning of an FE budget. Conventional EA studies often use a fixed number of FEs as the main evaluation budget. Under GPU execution, the same FE budget may cover only a short wall-clock interval and may terminate the run before long-horizon adaptation becomes visible. As a result, FE-budgeted evaluation should be complemented by time-budgeted evaluation, which measures the solution quality achieved under the same practical runtime constraint.

Third, the benefit of GPU execution depends on scale. GPU parallelism becomes useful only when the problem dimension, population size, and algorithmic workload provide enough parallel computation to amortize kernel, memory, and synchronization overheads. When the workload is too small, the GPU may be underutilized. When the workload becomes too large, memory pressure and coordination costs may lead to saturation. Identifying these regimes is necessary for interpreting GPU-based EA results.

Finally, GPUs make very large populations practical, and this may change the search dynamics of EAs. Larger populations can improve search-space coverage and diversity, but their final benefit depends on whether the algorithm can convert additional samples into useful search information. Thus, population size should not be viewed only as a hardware-utilization parameter. It is also an algorithmic parameter that affects convergence, diversity maintenance, and the exploration-exploitation balance.

In response, this paper goes beyond speedup reporting and studies the scaling behavior of EAs on GPUs from a hardware-aware evaluation perspective. The study covers 16 representative EAs, including both single-objective and multi-objective algorithms, and evaluates them on 30 benchmark problems, including numerical optimization problems and neuroevolution tasks. The experiments are conducted on CPU and GPU platforms under different FE budgets, time budgets, problem dimensions, and population sizes. Rather than treating speedup as the final conclusion, we examine how GPU execution changes the observable behavior and practical evaluation of EAs.

The main contributions of this paper are summarized as follows:
\begin{itemize}
    \item We characterize why GPU acceleration is mechanism-dependent rather than uniform across EAs. The results show that GPU compatibility is determined by the degree of batched computation, memory regularity, and synchronization.
    \item We compare FE-budgeted and time-budgeted evaluation protocols in GPU-enabled settings. The results show that FE-only evaluation may truncate long-horizon behavior, while time-budgeted evaluation better reflects practical time-to-solution.
    \item We analyze scaling behavior across problem dimensions, population sizes, and hardware platforms. The results identify regimes in which GPU execution is underutilized, effective, or saturated.
    \item We examine large-population dynamics enabled by GPU execution. The results show that large populations can improve coverage, diversity, and final performance only when the underlying evolutionary mechanism can use the additional samples effectively.
\end{itemize}

The remainder of this paper is organized as follows. Section~II reviews the basic workflow of EAs and the characteristics of GPU execution. Section~III presents the experimental study and analyzes the main findings. Section~IV discusses implications for hardware-aware EA design, and Section~V concludes the paper.

\section{Background}
\subsection{Basis of Evolutionary Algorithms }
\begin{figure}[htbp]
    \centering
    \includegraphics[width=0.49\textwidth]{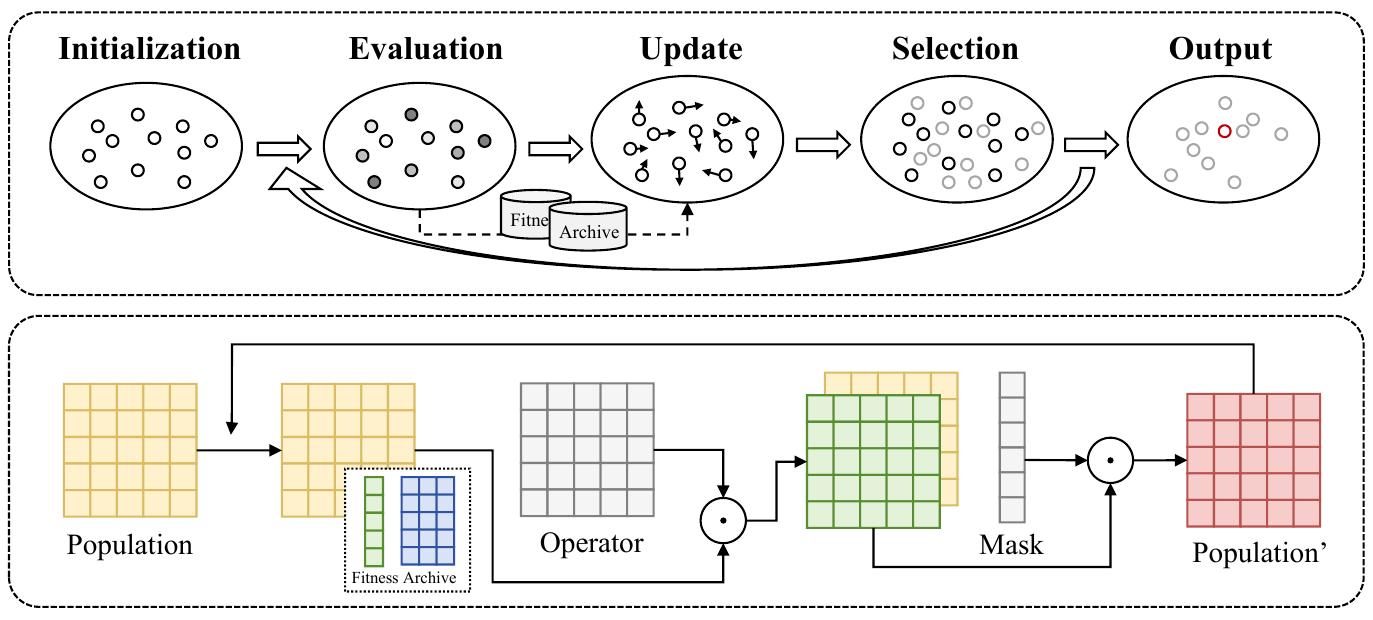}
\caption{
General workflow of an evolutionary algorithm and its batched execution structure.
The upper panel illustrates the algorithmic process, where a population is initialized, evaluated, updated by evolutionary operators, selected according to fitness or archive information, and finally returned as the output population or solution set.
The feedback arrows indicate that fitness values, archive states, and selected individuals jointly affect subsequent generations.
The lower panel shows the corresponding tensorized execution view used in GPU-oriented implementations.
The population is represented as a batched array, while operators, masks, fitness values, and archive information are organized as structured data objects that can be processed in parallel.
This representation highlights the population-level parallelism of EAs, but also shows that different algorithmic components expose different degrees of regularity and synchronization to the underlying hardware.
}
    \label{fig:process}
\end{figure}

EAs are a class of population-based optimization methods that iteratively improve a set of candidate solutions through mechanisms inspired by natural evolution. Unlike gradient-based methods, EAs do not require derivative information and are therefore well suited for non-convex, discontinuous, or black-box optimization problems. Although different EA variants employ different search mechanisms, most of them follow a common iterative workflow consisting of population initialization, fitness evaluation, variation, and selection or population update, as illustrated in Fig.~\ref{fig:process}.

In a typical evolutionary cycle, an initial population is first generated to provide diverse starting points for search. Each individual is then evaluated according to the objective function, which often constitutes the dominant computational cost, especially when the objective involves expensive simulations or high-dimensional decision variables. Variation operators, such as mutation, crossover, or recombination, generate new candidate solutions, while selection or update mechanisms determine which individuals or algorithmic states are retained for the next generation. Through repeated iterations, these components jointly shape the convergence behavior of algorithm.

In practical applications, the computational demand of EAs increases rapidly with problem scale and complexity. Maintaining large populations and performing many function evaluations can become prohibitively expensive, especially under conventional CPU-based execution. As a result, CPU-oriented settings often restrict the feasible population size and evaluation budget, limiting both the observable exploration capacity of EAs and their applicability to large-scale or time-sensitive optimization problems.
At the same time, the population-based structure of EAs makes them naturally amenable to parallelization. Fitness evaluations are usually independent across individuals, and many variation or sampling operations can be expressed in a batched form. This creates a strong connection between EAs and parallel computing platforms. However, the extent to which an EA can benefit from parallel hardware depends not only on the amount of computation, but also on whether its computational structure matches the execution model of the hardware. This motivates the need to examine EAs from the perspective of GPU parallelism.

\subsection{GPU Architecture and Parallelization Characteristics for EAs}
GPUs are designed for high-throughput parallel computation. Unlike CPUs, which are optimized for low-latency sequential execution and complex control flow, GPUs devote a much larger fraction of hardware resources to lightweight arithmetic units and rely on large-scale data-level parallelism~\cite{Owens2008, Anderson2008}. This architectural feature is well aligned with population-based optimization, where similar operations are repeatedly applied to a large number of candidate solutions.

From the perspective of EA execution, GPU parallelism offers two main advantages. First, it enables a large number of individuals to be evaluated simultaneously, thereby increasing the number of function evaluations that can be completed within a given wall-clock time. Second, it makes larger population sizes practically feasible, allowing EAs to explore the search space with denser sampling than is usually affordable in CPU-based studies. These advantages make GPUs particularly attractive when fitness evaluation, variation, and population update can be expressed in a batched and vectorized form.

The use of GPUs for evolutionary computation has been explored for about two decades. Early work by Ling~\cite{Ling2005} analyzed the architectural advantages of GPUs over CPUs and implemented GPU-accelerated variants of GAs, providing one of the earliest demonstrations of GPU-based EAs. Subsequent studies extended GPU acceleration to different EA paradigms. For example, Zhou~\textit{et al.} developed a GPU-based PSO with an $11\times$ speedup over its CPU counterpart~\cite{Zhou2009}, and later extended this idea to multi-objective optimization through a GPU-accelerated MOPSO~\cite{Zhou2011}. Veronese~\textit{et al.}~\cite{P.Veronese2010} implemented DE using CUDA and achieved up to $35\times$ acceleration, while Zhu~\textit{et al.}~\cite{Zhu2011} proposed a GPU-accelerated MOEA on the Fermi architecture with speedups of up to $25\times$. GPU acceleration has also been investigated for genetic programming (GP), where Robilliard~\textit{et al.}~\cite{Robilliard2009} presented a high-performance GPU-based GP framework that achieved a $3\times$ speedup and processed more than four billion GP operations per second.

More recent studies have benefited from advances in GPU hardware and parallel programming frameworks, which have increased speedups and broadened application settings. For example, GPU-accelerated GAs have been applied to vehicle routing, where speedups of up to $1700\times$ have been reported~\cite{Abdelatti2021}. In genetic programming for image classification, recent GPU implementations have reduced training time from weeks to hours~\cite{Zeng2022}, with reported speedups reaching $1600\times$ over CPU baselines~\cite{Zhang2024a}. In parallel, dedicated libraries such as EvoJAX~\cite{Tang2022}, evosax~\cite{Lange2023}, and EvoX~\cite{Huang2024} have simplified the construction of GPU-based EA pipelines by providing modular support for batched population evaluation, variation, and fitness tracking. These developments have lowered the implementation barrier and expanded the practical use of GPU-based EAs across increasingly complex optimization scenarios.



\begin{figure}[htbp]
    \centering
    \includegraphics[width=0.46\textwidth]{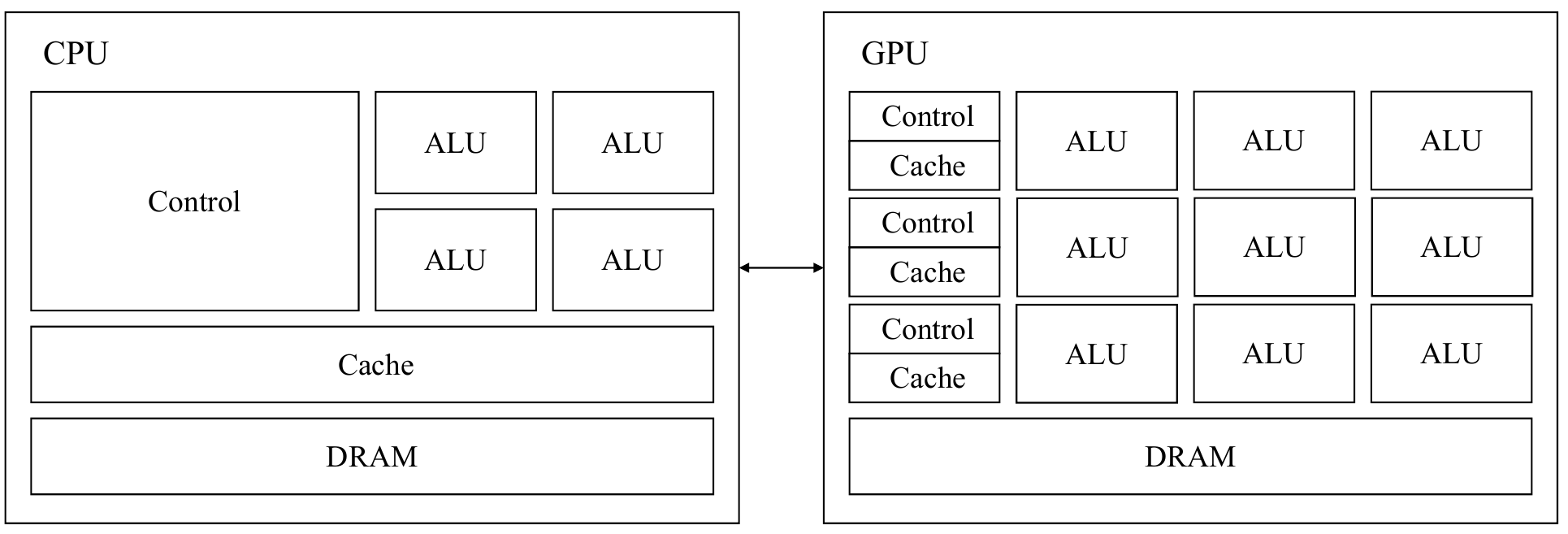}
\caption{
Architectural comparison between CPUs and GPUs from the perspective of evolutionary computation.
CPUs devote a larger fraction of hardware resources to control logic and cache hierarchy, which supports low-latency sequential execution and irregular control flow.
In contrast, GPUs contain many lightweight arithmetic units and are designed for high-throughput data-parallel computation over large batches.
This architectural difference explains why population-level operations in EAs, such as batched fitness evaluation, sampling, mutation, and vectorized state updates, can benefit from GPU execution.
However, EA components involving sorting, archive maintenance, neighborhood interaction, or data-dependent selection may introduce irregular memory access and synchronization, which can limit the achievable GPU speedup.
}
    \label{fig:gpu_cpu}
\end{figure}

\section{Hardware-Aware Evaluation of GPU-Enabled EAs}
EAs have traditionally been evaluated under CPU-oriented resource assumptions, where evaluations are expensive, population sizes are moderate, and runtime is shaped by limited parallelism. GPU execution changes this setting by enabling population-level computation at a much larger scale. This change affects not only runtime, but also the empirical conditions under which EAs are compared. In particular, it changes how FE budgets, time budgets, problem dimension, and population size should be interpreted.

This section evaluates GPU-enabled EAs beyond speedup measurement from four linked perspectives. First, we examine heterogeneous acceleration across algorithms to identify which evolutionary mechanisms map naturally to GPU execution. Second, we compare FE-budgeted and time-budgeted evaluation to separate sample efficiency from practical time-to-solution. Third, we analyze scaling behavior with respect to problem dimension and population size to identify underutilized, effective, and saturated regimes. Fourth, we study large-population dynamics to determine when increased population scale improves search behavior rather than only increasing throughput.

Together, these analyses operationalize the hardware-aware evaluation perspective used in this paper. The goal is not only to measure whether GPUs accelerate existing EAs, but also to clarify how GPU execution changes the resource model, observable search horizon, and design space of evolutionary computation.

\subsection{Experimental Setting}
We evaluate 16 representative EAs, 30 benchmark problems, and three hardware platforms. This setting is designed to cover different algorithmic mechanisms, optimization tasks, and workload scales so that GPU effects can be analyzed at the level of acceleration, evaluation protocol, scaling behavior, and population-level search dynamics.

\subsubsection{Selected Algorithms}

We select 16 representative EAs, including eight single-objective EAs (SOEAs) and eight multi-objective EAs (MOEAs). The SOEAs cover four major families. The swarm-based algorithms include PSO~\cite{Kennedy1995} and CSO~\cite{Cheng2015}. The differential evolution algorithms include DE (rand/1/bin)~\cite{Storn1997} and SaDE~\cite{Qin2009}. The GA variants include GA-SBX/PM~\cite{Agrawal1994}, which employs simulated binary crossover (SBX) with polynomial mutation (PM), and GA-UR/GM, which uses uniform random crossover (UR)~\cite{Syswerda1989} with Gaussian mutation (GM)~\cite{Goldberg1987}. The evolution strategies include CMA-ES~\cite{Hansen2003} and IPOP-CMA-ES~\cite{Auger2005a}. For MOEAs, we consider Pareto-based algorithms, including NSGA-II~\cite{Deb2002}, NSGA-III~\cite{Deb2014}, and SPEA2~\cite{Kim2004}; indicator-based algorithms, including HypE~\cite{Bader2011} and IBEA~\cite{Zitzler2004}; reference-guided algorithms, including MOEA/D~\cite{Zhang2007} and RVEA~\cite{Liang2024a}; and a swarm-based MOEA, LMOCSO~\cite{Tian2020}. These algorithms provide diverse computational patterns for analyzing heterogeneous GPU acceleration across different evolutionary mechanisms.

\subsubsection{Benchmark Problems}
The benchmark problems are divided into two categories: numerical optimization problems ($f_{a}$) and neuroevolution tasks ($f_{b}$). The numerical optimization category contains 20 benchmark problems.
For single-objective optimization, we use five 20-dimensional CEC2022 functions (F1--F5), denoted as $f_{a_1}$--$f_{a_5}$~\cite{Kumar2021}, and five classical test functions with 50 decision variables, including Ackley, Griewank, Rosenbrock, Schwefel, and Sphere, denoted as $f_{a_6}$--$f_{a_{10}}$. For multi-objective problems, we selected seven 50-dimensional DTLZ problems (DTLZ1--DTLZ7) with three objectives, denoted as $f_{a_{11}}$--$f_{a_{17}}$~\cite{Deb2002a}, and three 50-dimensional ZDT problems (ZDT1--ZDT3) with two objectives, denoted as $f_{a_{18}}$-$f_{a_{20}}$~\cite{Zitzler2000a}.
The neuroevolution category contains 10 tasks based on the Brax physics simulation engine~\cite{Freeman2021}. The five single-objective tasks are HalfCheetah, Hopper, Reacher, Swimmer, and Walker2d, denoted as $f_{b_1}$--$f_{b_5}$. To support multi-objective neuroevolution, we extended the Brax framework following the methodology in~\cite{Liang2024a} and construct five multi-objective neuroevolution tasks: MoHopper-m2, MoHopper-m3, MoReacher, MoSwimmer, and MoWalker2d, denoted as $f_{b_6}$--$f_{b_{10}}$.  A consistent policy architecture is used across all neuroevolution tasks, consisting of a multilayer perceptron (MLP) with three fully connected layers. Detailed descriptions of these benchmark problems and task settings are provided in Supplementary Documents~\ref{app:neuroevolution} and~\ref{app:experiments}.

\subsubsection{Computing Platforms}
The experiments are conducted on both GPU and CPU platforms. The key specifications are summarized below:
\begin{itemize}
\item \textbf{GPU Platforms}:
\begin{itemize}
\item NVIDIA GeForce RTX 3090: 24GB GDDR6X memory, 10496 CUDA cores
\item NVIDIA GeForce RTX 2080 Ti: 11GB GDDR6 memory, 4352 CUDA cores
\end{itemize}
\item \textbf{CPU Platform}:
\begin{itemize}
\item Intel Xeon Gold 6226R: 16-core processor, 2.90GHz base frequency, 64GB DDR4 RAM
\end{itemize}
\end{itemize}

All experiments are implemented using the \textit{EvoX} framework~\cite{Huang2024}, which provides unified execution pipelines and consistent metric logging across devices. This setup allows the same algorithmic configurations to be evaluated on different hardware platforms, which reduces implementation-dependent variation in the CPU--GPU comparison. The reported runtimes should therefore be interpreted as measurements under this unified implementation rather than as universal hardware constants. Accordingly, the analysis focuses on relative scaling trends and mechanism-level differences rather than on absolute speedup values alone.

\subsection{Heterogeneous Acceleration Across Algorithms}
GPU acceleration depends on how an EA decomposes its computation across individuals, operators, and population-level interactions. Fitness evaluation is often parallelizable, but the total runtime of an EA also includes algorithm-specific operations such as sorting, neighborhood interaction, archive maintenance, and state adaptation. Therefore, the achieved GPU benefit depends on whether the dominant operations can be expressed as regular, batched, and weakly synchronized computation.
To examine this heterogeneity, we evaluate 16 EAs under a fixed population size of 128 and an evaluation budget of $1,000,000$ FEs. This setting uses a common population scale and provides a sufficiently long budget for stable runtime measurement. For each algorithm and benchmark problem, we conduct 15 independent runs and report the averaged wall-clock runtime on CPU and on an NVIDIA GeForce RTX 2080 Ti GPU, together with the corresponding GPU speedup.

Tables~\ref{tab:speedup-so} and~\ref{tab:speedup-mo} show that GPU execution accelerates most EAs, but the magnitude of acceleration varies substantially across algorithms. Because all EAs are tested under the same FE budget, these differences mainly reflect how their algorithmic mechanisms map to GPU execution. Table~\ref{tab:acceleration_patterns} summarizes the results into four representative acceleration patterns.
Algorithms with dense and batched dominant operations achieve the largest speedups. For SOEAs, CMA-ES and IPOP-CMA-ES obtain average speedups of $31.45\times$ and $28.44\times$, respectively. Their covariance-based sampling and adaptation introduce high CPU cost, but these matrix- and vector-based operations can be executed efficiently on GPUs, especially on high-dimensional problems. For example, on $f_{a_6}$, CMA-ES reduces runtime from 741.90 seconds on CPU to 13.30 seconds on GPU, with a speedup of $55.77\times$. A similar pattern is observed for HypE among MOEAs. Its indicator-estimation procedure involves many repeated and batchable computations, which leads to an average speedup of $97.01\times$ and a maximum speedup of $159.71\times$ on $f_{a_{20}}$.

\begin{table}[htbp]
\centering
\caption{Averaged runtime and speedups between GPU and CPU tested under 1,000,000 FEs across eight SOEAs.}
\label{tab:speedup-so}
\small
\resizebox{0.5\textwidth}{!}{
\begin{tabular}{llcccccccc}
\toprule
Func. & Metric & \makecell{PSO} & \makecell{CSO} & \makecell{DE} & \makecell{SaDE} &
\makecell{CMA-\\ES} & \makecell{IPOP-\\CMA-ES} & \makecell{GA-\\SBX/PM} & \makecell{GA-\\UR/GM} \\
\midrule
\multirow{3}{*}{$f_{a_1}$}
  & $T_{\text{CPU}}$ & 14.34 & 14.51 & 55.68 & 83.92 & 153.49 & 128.99 & 23.83 & 19.86 \\
  & $T_{\text{GPU}}$ &  8.36 &  8.20 &  9.21 & 24.73 &  30.76 &  15.35 &  9.71 &  9.06 \\
  & Speedup         &  1.71 &  1.77 &  6.04 &  3.39 &  4.99 & \colorbox{gray!30}{8.41} &  2.45 &  2.19 \\[2pt]

\multirow{3}{*}{$f_{a_2}$}
  & $T_{\text{CPU}}$ & 14.85 & 14.82 & 55.74 & 84.26 & 155.66 & 131.09 & 23.97 & 20.82 \\
  & $T_{\text{GPU}}$ &  8.55 &  8.59 &  9.12 & 24.80 &  15.73 &  15.07 & 10.48 &  8.95 \\
  & Speedup         &  1.74 &  1.73 &  6.11 &  3.40 & \colorbox{gray!30}{9.90} &  8.70 &  2.29 &  2.33 \\[2pt]

\multirow{3}{*}{$f_{a_3}$}
  & $T_{\text{CPU}}$ & 15.62 & 15.43 & 56.56 & 84.57 & 167.40 & 174.10 & 25.21 & 22.67 \\
  & $T_{\text{GPU}}$ & 10.32 & 10.08 & 10.98 & 26.43 &  36.91 &  14.73 & 11.58 & 10.72 \\
  & Speedup         &  1.51 &  1.53 &  5.15 &  3.20 &  4.54 & \colorbox{gray!30}{11.82} & 2.18 & 2.12 \\[2pt]

\multirow{3}{*}{$f_{a_4}$}
  & $T_{\text{CPU}}$ & 14.92 & 15.01 & 56.45 & 84.20 & 123.40 & 129.61 & 23.87 & 21.03 \\
  & $T_{\text{GPU}}$ &  8.70 &  8.27 &  9.11 & 24.22 &  18.51 &  14.85 &  9.40 &  9.01 \\
  & Speedup         &  1.71 &  1.82 &  6.20 &  3.48 &  6.67 & \colorbox{gray!30}{8.73} & 2.54 & 2.33 \\[2pt]

\multirow{3}{*}{$f_{a_5}$}
  & $T_{\text{CPU}}$ & 16.63 & 14.88 & 57.07 & 83.48 & 162.05 & 133.42 & 25.47 & 21.45 \\
  & $T_{\text{GPU}}$ &  9.31 &  8.64 &  9.18 & 25.11 &  13.97 &  14.58 & 10.42 &  8.91 \\
  & Speedup         &  1.79 &  1.72 &  6.22 &  3.32 & \colorbox{gray!30}{11.60} & 9.15 & 2.44 & 2.41 \\[2pt]

\multirow{3}{*}{$f_{a_6}$}
  & $T_{\text{CPU}}$ & 36.11 & 15.27 & 59.27 & 90.71 & 741.90 & 750.62 & 55.18 & 41.12 \\
  & $T_{\text{GPU}}$ &  8.07 &  8.32 &  9.09 & 24.02 &  13.30 &  16.56 &  9.40 &  8.66 \\
  & Speedup         &  4.47 &  1.84 &  6.52 &  3.78 & \colorbox{gray!30}{55.77} & 45.33 & 5.87 & 4.75 \\[2pt]

\multirow{3}{*}{$f_{a_7}$}
  & $T_{\text{CPU}}$ & 36.74 & 15.51 & 66.01 & 98.44 & 712.59 & 717.00 & 55.43 & 40.46 \\
  & $T_{\text{GPU}}$ &  8.17 &  8.61 &  9.12 & 24.65 &  13.13 &  16.45 &  8.99 &  8.91 \\
  & Speedup         &  4.49 &  1.80 &  7.46 &  3.99 & \colorbox{gray!30}{54.27} & 43.59 & 6.16 & 4.54 \\[2pt]

\multirow{3}{*}{$f_{a_8}$}
  & $T_{\text{CPU}}$ & 30.47 & 16.49 & 63.39 & 95.25 & 759.80 & 750.42 & 52.60 & 35.06 \\
  & $T_{\text{GPU}}$ &  8.51 &  9.98 &  8.85 & 24.79 &  14.73 &  17.79 & 11.25 &  8.97 \\
  & Speedup         &  3.58 &  1.65 &  6.95 &  3.84 & \colorbox{gray!30}{51.58} & 42.18 & 4.68 & 3.91 \\[2pt]

\multirow{3}{*}{$f_{a_9}$}
  & $T_{\text{CPU}}$ & 33.58 & 14.83 & 64.01 & 95.49 & 738.20 & 770.70 & 55.91 & 40.67 \\
  & $T_{\text{GPU}}$ &  8.50 &  8.46 &  8.93 & 23.88 &   12.16 &   12.33 &  2.54 &  2.30 \\
  & Speedup         &  3.95 &  1.75 &  7.17 &  4.00 & \colorbox{gray!30}{60.70} & 62.51 & 22.03 & 17.65 \\[2pt]

\multirow{3}{*}{$f_{a_{10}}$}
  & $T_{\text{CPU}}$ & 34.66 & 14.43 & 64.99 & 96.65 & 726.80 & 735.40 & 54.91 & 38.70 \\
  & $T_{\text{GPU}}$ &  7.95 &  8.42 &  8.98 & 24.97 &  13.34 &  16.72 &  9.86 &  8.60 \\
  & Speedup         &  4.36 &  1.71 &  7.23 &  3.87 & \colorbox{gray!30}{54.47} & 43.98 &  5.57 &  4.50 \\
\bottomrule
\end{tabular}}
\end{table}

\begin{table}[htbp]
\centering
\caption{Averaged runtime and speedups between GPU and CPU tested under 1,000,000 FEs across eight MOEAs.}
\label{tab:speedup-mo}
\small
\resizebox{0.5\textwidth}{!}{
\begin{tabular}{llcccccccc}
\toprule
Func. & Metric & NSGA-II & NSGA-III & SPEA2 & IBEA & HypE & MOEA/D & RVEA & LMOCSO \\
\midrule

\multirow{3}{*}{$f_{a_{11}}$}
& $T_{\text{CPU}}$ & 32.49 & 41.56  & 223.63 & 46.47 & 575.32          & 26.88 & 40.88 & 50.73 \\
& $T_{\text{GPU}}$ & 7.10  & 56.95  & 12.48  & 10.96 & 7.91            & 13.02 & 5.34  & 7.13  \\
& Speedup         & 4.58  & 0.73   & 17.91  & 4.24  & \colorbox{gray!30}{72.76}  & 2.07  & 7.65  & 7.11  \\ [2pt]
\multirow{3}{*}{$f_{a_{12}}$}
& $T_{\text{CPU}}$ & 71.24 & 79.23  & 324.48 & 85.57 & 608.32          & 70.08 & 86.76 & 88.57 \\
& $T_{\text{GPU}}$ & 6.05  & 56.96  & 18.05  & 10.74 & 7.14            & 13.31 & 6.15  & 6.97  \\
& Speedup         & 11.77 & 1.39   & 17.98  & 7.97  & \colorbox{gray!30}{85.16}  & 5.26  & 14.11 & 12.71 \\ [2pt]
\multirow{3}{*}{$f_{a_{13}}$}
& $T_{\text{CPU}}$ & 72.12 & 80.43  & 263.79 & 85.71 & 673.91          & 66.40 & 44.55 & 91.37 \\
& $T_{\text{GPU}}$ & 7.21  & 55.73  & 12.72  & 10.59 & 8.43            & 12.74 & 5.25  & 7.06  \\
& Speedup         & 10.00 & 1.44   & 20.73  & 8.09  & \colorbox{gray!30}{79.95}  & 5.21  & 8.48  & 12.94 \\ [2pt]
\multirow{3}{*}{$f_{a_{14}}$}
& $T_{\text{CPU}}$ & 72.65 & 81.38  & 327.35 & 86.22 & 675.47          & 67.44 & 88.33 & 89.55 \\
& $T_{\text{GPU}}$ & 6.19  & 57.88  & 17.42  & 10.71 & 7.00            & 12.73 & 6.03  & 7.03  \\
& Speedup         & 11.74 & 1.41   & 18.79  & 8.05  & \colorbox{gray!30}{96.54}  & 5.30  & 14.64 & 12.75 \\ [2pt]
\multirow{3}{*}{$f_{a_{15}}$}
& $T_{\text{CPU}}$ & 71.78 & 103.10 & 286.72 & 86.28 & 617.48          & 66.86 & 39.27 & 40.56 \\
& $T_{\text{GPU}}$ & 6.59  & 73.59  & 14.30  & 10.70 & 7.81            & 12.90 & 5.19  & 6.10  \\
& Speedup         & 10.89 & 1.40   & 20.06  & 8.06  & \colorbox{gray!30}{79.04}  & 5.18  & 7.56  & 6.64  \\ [2pt]
\multirow{3}{*}{$f_{a_{16}}$}
& $T_{\text{CPU}}$ & 72.91 & 102.76 & 303.79 & 85.80 & 560.12          & 20.35 & 41.01 & 90.63 \\
& $T_{\text{GPU}}$ & 6.91  & 71.79  & 15.48  & 10.82 & 6.24            & 12.36 & 5.22  & 6.82  \\
& Speedup         & 10.56 & 1.43   & 19.62  & 7.93  & \colorbox{gray!30}{89.81}  & 1.65  & 7.86  & 13.29 \\ [2pt]
\multirow{3}{*}{$f_{a_{17}}$}
& $T_{\text{CPU}}$ & 71.74 & 96.16  & 347.76 & 85.35 & 610.66          & 65.98 & 37.82 & 38.66 \\
& $T_{\text{GPU}}$ & 6.21  & 69.09  & 19.51  & 10.95 & 7.05            & 13.08 & 5.51  & 5.74  \\
& Speedup         & 11.54 & 1.39   & 17.83  & 7.80  & \colorbox{gray!30}{86.64}  & 5.04  & 6.87  & 6.73  \\ [2pt]
\multirow{3}{*}{$f_{a_{18}}$}
& $T_{\text{CPU}}$ & 26.62 & 33.55  & 245.91 & 38.37 & 816.41          & 23.69 & 45.12 & 46.04 \\
& $T_{\text{GPU}}$ & 5.80  & 55.59  & 14.14  & 10.62 & 6.98            & 12.73 & 6.18  & 6.59  \\
& Speedup         & 4.59  & 0.60   & 17.39  & 3.61  & \colorbox{gray!30}{117.01} & 1.86  & 7.30  & 6.98  \\ [2pt]
\multirow{3}{*}{$f_{a_{19}}$}
& $T_{\text{CPU}}$ & 26.66 & 33.94  & 242.65 & 38.29 & 755.25          & 23.48 & 44.32 & 45.52 \\
& $T_{\text{GPU}}$ & 6.19  & 55.53  & 13.73  & 10.78 & 7.30            & 12.65 & 5.83  & 6.60  \\
& Speedup         & 4.30  & 0.61   & 17.68  & 3.55  & \colorbox{gray!30}{103.50} & 1.86  & 7.60  & 6.89  \\ [2pt]
\multirow{3}{*}{$f_{a_{20}}$}
& $T_{\text{CPU}}$ & 26.72 & 40.04  & 233.17 & 38.57 & 756.19          & 23.47 & 44.59 & 45.84 \\
& $T_{\text{GPU}}$ & 3.02  & 53.91  & 10.28  & 8.19  & 4.73            & 9.95  & 3.37  & 3.62  \\
& Speedup         & 8.84  & 0.74   & 22.68  & 4.71  & \colorbox{gray!30}{159.71} & 2.36  & 13.25 & 12.65  \\
\bottomrule
\end{tabular}
}
\end{table}

\begin{table}[htbp]
\centering
\caption{Representative GPU acceleration patterns observed.}
\label{tab:acceleration_patterns}
\resizebox{0.48\textwidth}{!}{
\setlength{\tabcolsep}{4pt}
\begin{tabular}{>{\centering\arraybackslash}p{1.5cm}>{\centering\arraybackslash}p{2.2cm}>{\centering\arraybackslash}p{2.4cm}>{\centering\arraybackslash}p{4.6cm}}
\toprule
\textbf{\larger[1.05]Pattern} & \textbf{\larger[1.05]Example} & \textbf{\larger[1.05]Mean Speedup} & \textbf{\larger[1.05]Dominant Characteristics} \\
\midrule
\multirow{2}{*}{\larger[1.05] High} & HypE, CMA-ES, IPOP-CMA-ES & $97.01\times$, $31.45\times$, $28.44\times$ & \larger[1.05] Dense and batched numerical operations. \\

\midrule
\multirow{2}{*}{\larger[1.05] Moderate} & SPEA2, RVEA, DE, IBEA & $19.07\times$, $9.53\times$, $6.51\times$, $6.40\times$ & \larger[1.05] Parallelizable operators with partial coordination costs. \\

\midrule
\multirow{2}{*}{\larger[1.05] Limited} & PSO, CSO, MOEA/D & $2.93\times$, $1.73\times$, $3.58\times$ & \larger[1.05] Low arithmetic intensity or neighborhood communication. \\
\midrule
\multirow{2}{*}{\larger[1.05] Marginal} & \multirow{2}{*}{NSGA-III} & \multirow{2}{*}{$1.11\times$} & \larger[1.05] Sorting, niching, and irregular synchronization. \\

\bottomrule
\end{tabular}}
\end{table}

By contrast, algorithms whose dominant costs come from data-dependent selection or structured interaction benefit less from direct GPU execution. NSGA-III is a representative case, achieving only a marginal average speedup of $1.11\times$. Its environmental selection involves non-dominated sorting, reference-point association, and niching. These steps are difficult to batch efficiently because the number and size of fronts vary across generations, the critical front is determined dynamically, and niche counts are updated during population filling. As a result, the computation contains irregular branching and synchronization, which weakens the advantage of GPU parallelism. MOEA/D exhibits another source of limitation. Although its decomposition-based search is population-based, each solution mainly interacts with its neighborhood through local replacement. Such neighborhood-level dependencies and non-contiguous memory access patterns restrict the achievable speedup compared with algorithms dominated by dense batched operations.

PSO and CSO show a different form of limited acceleration. Their particle-wise updates are naturally parallel, but their average speedups are only $2.93\times$ and $1.73\times$, respectively. This does not indicate inefficient GPU execution, since their GPU runtimes are among the lowest in Table~\ref{tab:speedup-so}. Instead, the limited speedup mainly comes from their low CPU baselines and lightweight update rules. PSO mainly performs velocity and position updates with global-best information sharing, while CSO relies on simple winner--loser competitions. These operations are easy to parallelize, but each individual contributes only a small amount of arithmetic work. Therefore, GPU execution reduces the absolute runtime, but the relative speedup is less pronounced than in algorithms with heavier batched numerical workloads.

Taken together, these results show that GPU-friendliness is largely determined by the computational form of EA mechanisms. Algorithms benefit most when their dominant operations are regular, batched, and weakly synchronized. In contrast, mechanisms involving swarm coordination, neighborhood interaction, sorting, or archive maintenance may require more than a direct GPU implementation. They should be redesigned toward batched state updates, fixed-size neighborhood computations and parallelized selection procedures to better exploit parallel hardware.
The substantial speedups observed in GPU-compatible EAs further suggest that parallel hardware can expand the feasible evaluation budget and search horizon. Therefore, GPU implementation should not be regarded as a post hoc engineering step. GPU compatibility should instead be treated as a design consideration for evolutionary computation on modern parallel hardware.

\subsection{FE-Budgeted and Time-Budgeted Evaluation}
FE-budgeted evaluation has long been the standard protocol for benchmarking EAs because it assigns the same number of function evaluations to all algorithms. This protocol remains useful for comparing sample efficiency. However, GPU execution weakens the conventional equivalence between FEs and computational cost. With population-level parallelism, the same FE budget can be completed much faster, and the runtime required to consume a fixed number of FEs can vary substantially across algorithms. Therefore, FE-budgeted evaluation provides only one view of GPU-enabled performance. It should be complemented by time-budgeted evaluation, which measures how much solution quality an EA can obtain under the same wall-clock constraint.


\subsubsection{Fixed-FE Truncation under GPU Execution}
\begin{figure}[htbp]
    \centering

    \begin{minipage}[b]{0.24\textwidth}
        \centering
        \includegraphics[width=\textwidth]{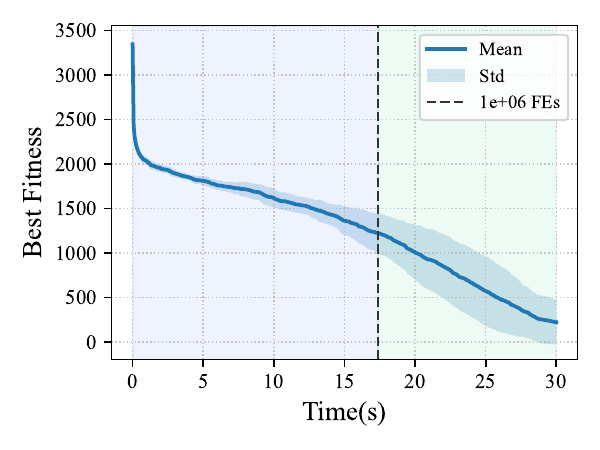}
        \centering
        \vspace{-0.6cm}
        \subcaption{CSO on 200D Rastrigin}
    \end{minipage}
   \hfill
    \begin{minipage}[b]{0.24\textwidth}
        \centering
        \includegraphics[width=\textwidth]{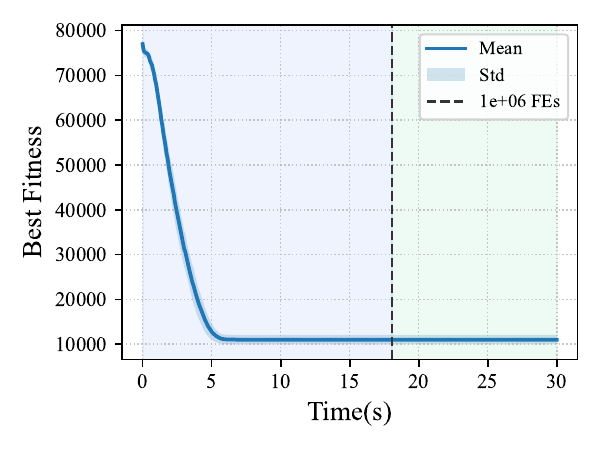}
        \centering
        \vspace{-0.6cm}
        \subcaption{CSO on 200D Schwefel}
    \end{minipage}

    \vspace{0.2cm}

    \begin{minipage}[b]{0.24\textwidth}
        \centering
        \includegraphics[width=\textwidth]{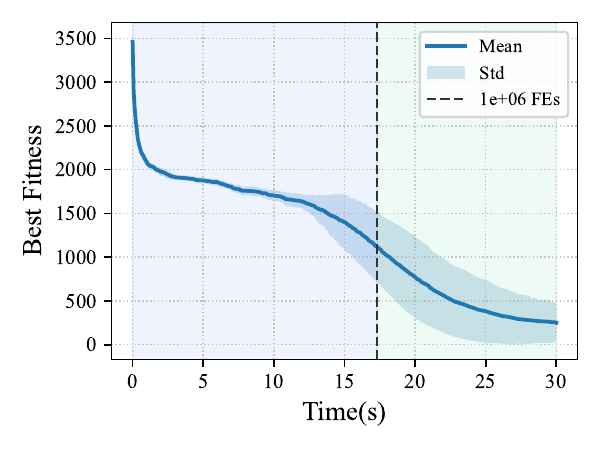}
        \centering
        \vspace{-0.6cm}
        \subcaption{DE on 200D Rastrigin}
    \end{minipage}
   \hfill
    \begin{minipage}[b]{0.24\textwidth}
        \centering
        \includegraphics[width=\textwidth]{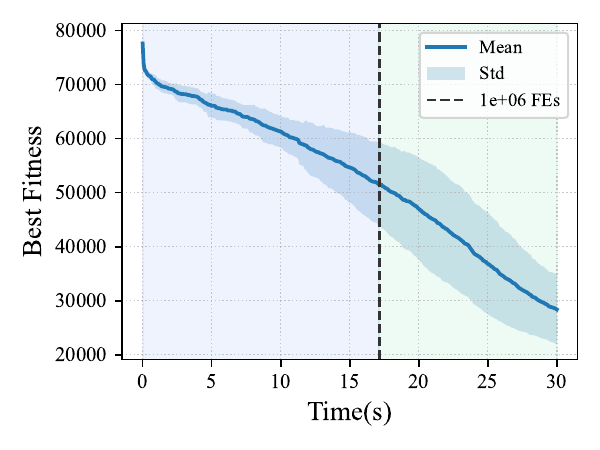}
        \centering
        \vspace{-0.6cm}
        \subcaption{DE on 200D Schwefel}
    \end{minipage}
    \caption{Truncation effect of fixed-FE evaluation under GPU execution. Each curve shows the mean best fitness over 15 independent runs, with shaded regions indicating the standard deviation. The vertical dashed line marks the wall-clock time at which $10^6$ FEs are completed.
    }
 \label{fig:truncation_effect}
\end{figure}

We first examine whether a conventional FE budget may prematurely truncate the optimization process under GPU execution. Fig.~\ref{fig:truncation_effect} presents representative convergence curves on 200-dimensional Rastrigin and Schwefel functions. The vertical dashed line marks the wall-clock time at which $10^6$ FEs are completed on an NVIDIA GeForce RTX 3090 GPU. The region before this line corresponds to the conventional fixed-FE observation window, while the region after it shows the additional search trajectory made observable by GPU acceleration.

The results reveal that the truncation effect is both algorithm- and problem-dependent. On Rastrigin, both CSO and DE continue to improve substantially after the $10^6$-FE cutoff. In particular, a large portion of their fitness reduction occurs in the GPU-extended observation window. This indicates that the conventional fixed-FE budget would hide meaningful late-stage progress and may underestimate algorithms that require a longer search horizon. In these cases, GPU execution does not merely complete the same benchmark budget faster; it makes later-stage search behavior observable within a practical runtime.
By contrast, CSO on Schwefel quickly reaches a plateau, and little improvement is observed after the cutoff. This suggests that the algorithm is not evaluation-limited on this problem, but has already stagnated due to its search dynamics.

The enlarged shaded regions in the later stage further expose the difficulty of long-horizon evolution. Once the population has partially converged, further progress depends on whether the algorithm can preserve diversity while continuing effective exploitation.
For CSO, winner--loser competition can drive fast convergence but may also weaken diversity, making later progress sensitive to whether particles can still escape local regions. For DE, differential mutation remains effective only when the population retains sufficient spread; otherwise, mutation steps become less informative and the search may stagnate. Thus, the increased late-stage variability reflects the challenge of maintaining a stable exploration--exploitation balance over an extended search horizon.

These observations highlight the limitation of relying only on conventional FE budgets in GPU-based EA studies. Although fixed-FE evaluation remains fair for comparing sample efficiency, it may capture only the early stage of the evolutionary process on modern parallel hardware. Using CPU-oriented FE scales on GPUs can underutilize the available computational capacity and obscure important late-stage behaviors, thereby underestimating algorithms with delayed adaptation, long-term exploration, or late-stage refinement.

\subsubsection{Time-budgeted Performance}
\begin{figure}[htbp]
    \centering
    \includegraphics[width=0.48\textwidth]{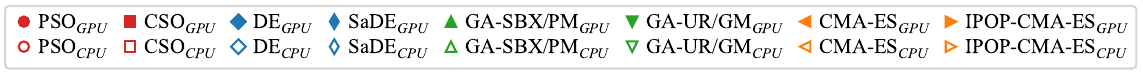}

    \vspace{0.2cm}

    \begin{minipage}[b]{0.235\textwidth}
        \centering
        \includegraphics[width=\textwidth]{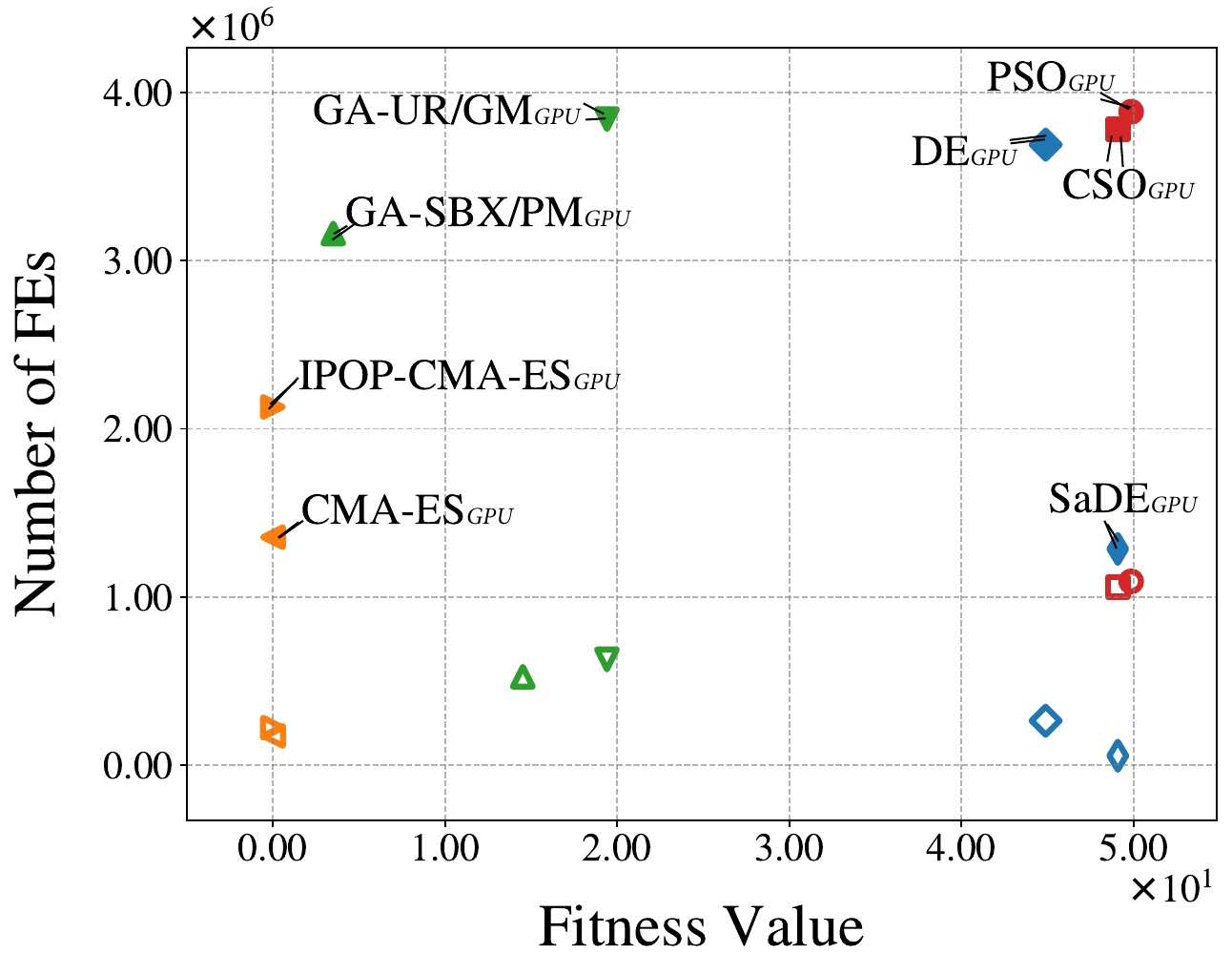}
        \centering
        \vspace{-0.5cm}
        \subcaption{ $f_{a_{2}}$: 20D CEC2022-F2}
    \end{minipage}
   \hfill
    \begin{minipage}[b]{0.235\textwidth}
        \centering
        \includegraphics[width=\textwidth]{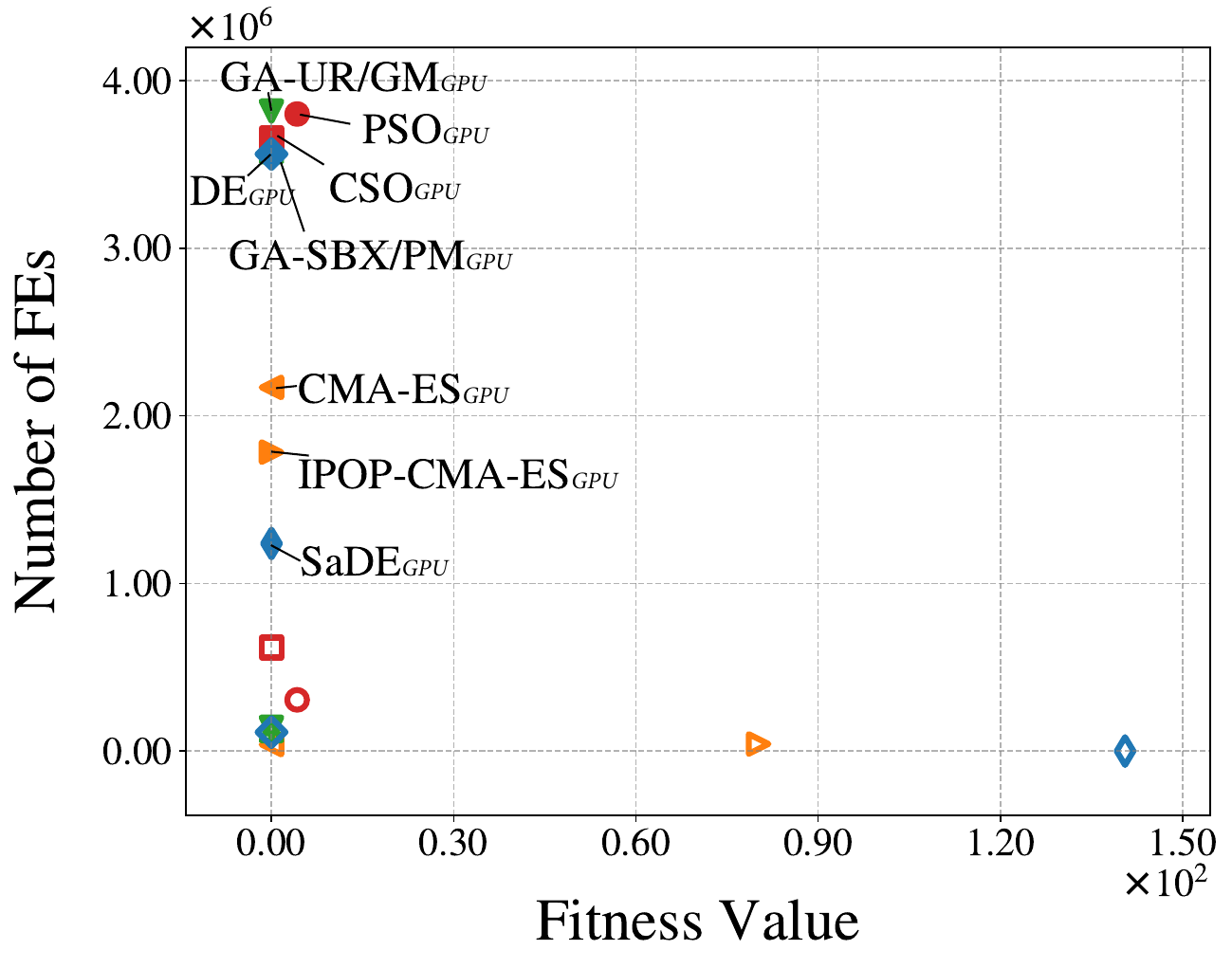}
        \centering
        \vspace{-0.5cm}
        \subcaption{ $f_{a_{10}}$: 50D Sphere}
    \end{minipage}

    \vspace{0.2cm}

    \includegraphics[width=0.48\textwidth]{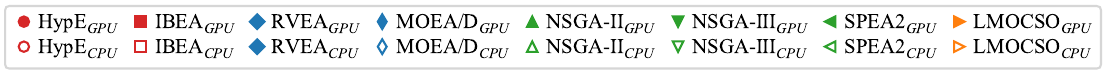}

    \vspace{0.2cm}

    \begin{minipage}[b]{0.235\textwidth}
        \centering
        \includegraphics[width=\textwidth]{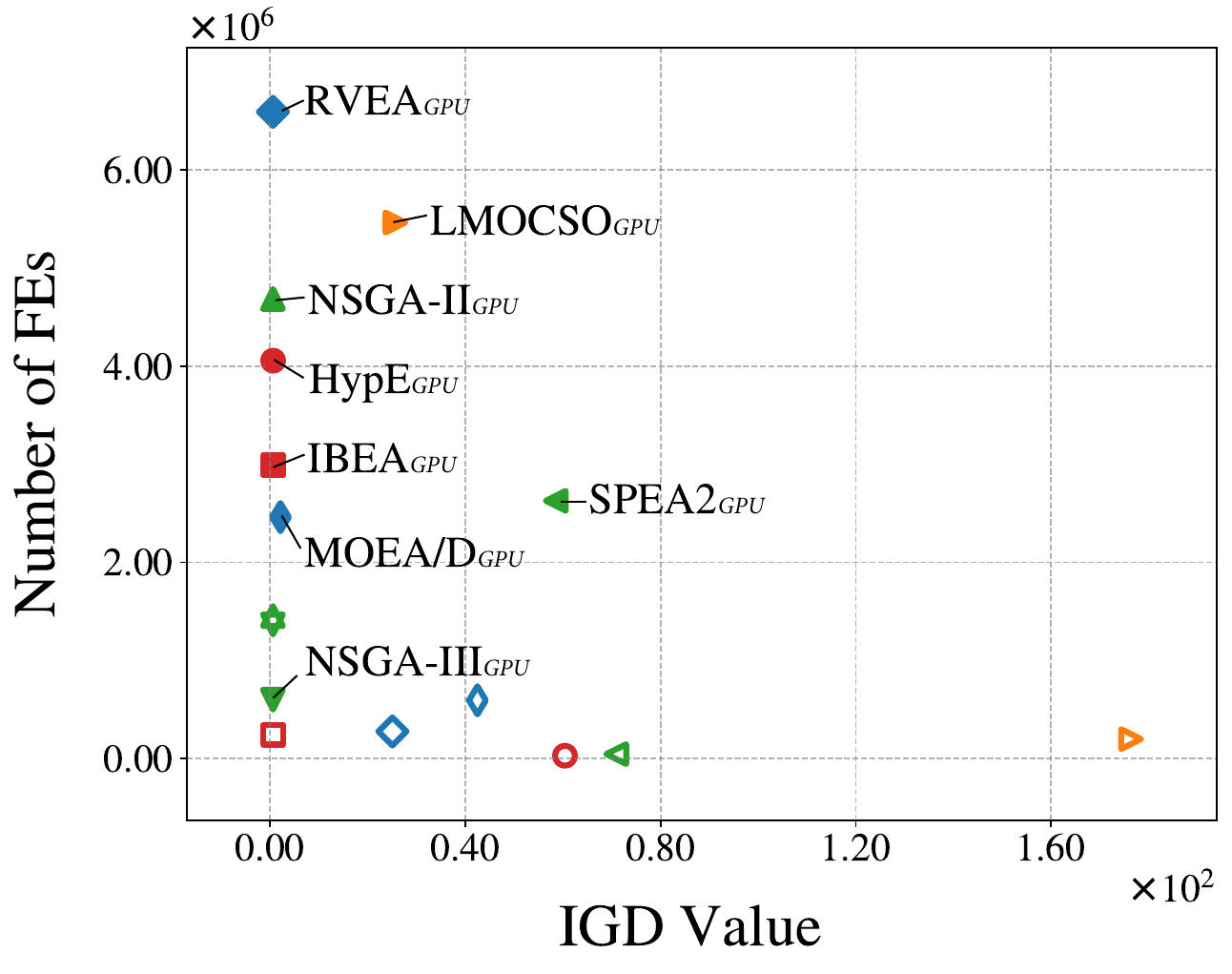}
        \centering
        \vspace{-0.5cm}
        \subcaption{$f_{a_{11}}$: 50D DTLZ1}
    \end{minipage}
   \hfill
    \begin{minipage}[b]{0.235\textwidth}
        \centering
        \includegraphics[width=\textwidth]{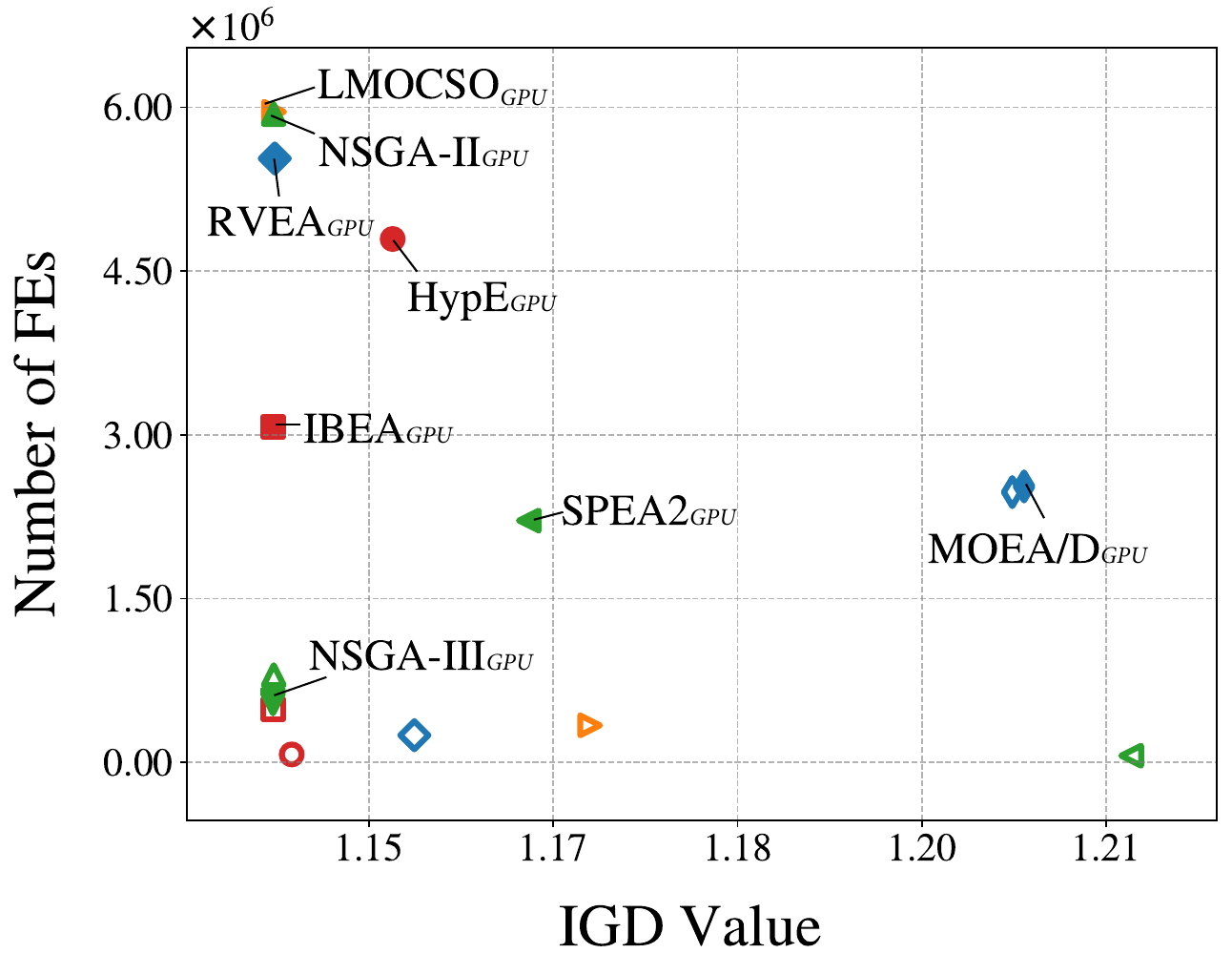}
        \centering
        \vspace{-0.5cm}
        \subcaption{$f_{a_{18}}$: 50D ZDT1}

    \end{minipage}
    \caption{Performance of EAs tested on CPU and GPU, evaluated in terms of solution quality and number of FEs completed within 30 seconds. Lower fitness/IGD values denote better performance. Results represent averaged performance values across 15 independent runs. Solid markers denote GPU-based implementations; hollow markers indicate CPU-based counterparts.
    }
    \label{fig:d-fig1}
\end{figure}

We then evaluate all algorithms under a 30-second time budget with a fixed population size of 128. For each run, we record the number of completed FEs and the final solution quality on both CPU and an NVIDIA GeForce RTX 3090 GPU. Each experiment is repeated 15 times, and the averaged results are reported.
Fig.~\ref{fig:d-fig1} presents the quality--throughput trade-off under the fixed-time setting. The horizontal axis denotes solution quality, and the vertical axis reports the number of FEs completed within 30 seconds.
Solid markers represent GPU implementations, while hollow markers denote their CPU counterparts. Since lower fitness or IGD values indicate better performance, points closer to the upper-left region are preferred.

The results show that GPU execution substantially increases the number of FEs completed within the same wall-clock budget. In most cases, solid markers are located above their CPU counterparts, indicating that GPU parallelism extends the effective search horizon without increasing practical runtime cost. However, higher throughput does not automatically lead to better solutions. In the single-objective cases, several algorithms show clear quality improvement under the fixed-time setting. For example, GA-SBX/PM on $f_{a_2}$ and IPOP-CMA-ES and SaDE on $f_{a_{10}}$ move leftward on GPU, indicating better final fitness within the same 30-second budget. These results suggest that their variation or adaptation mechanisms can benefit from a longer iterative process. In contrast, some swarm-based algorithms complete many FEs but show limited improvement in final quality, indicating that their performance is more constrained by search dynamics than by the available number of evaluations.
The multi-objective cases exhibit a similar pattern. On $f_{a_{11}}$, GPU-based RVEA and HypE are located in favorable upper-left regions, achieving both high FE throughput and competitive IGD values.
On $f_{a_{18}}$, however, the IGD values of several algorithms remain close, even though their completed FE counts differ substantially. In this case, GPU execution mainly changes the computational throughput, while the final solution quality is more strongly shaped by the algorithmic selection and diversity-maintenance mechanisms.

Overall, these results show that time-budgeted evaluation offers a complementary and more practical perspective for GPU-based EA benchmarking.
Fixed-FE evaluation remains necessary for comparing sample efficiency under the same number of evaluations. However, under GPU execution, the same FE budget may correspond to a much shorter observation window and may not reflect how well an algorithm uses the available computational resources. Time-budgeted evaluation addresses this limitation by measuring the solution quality achieved under the same wall-clock budget, which better reflects practical time-to-solution. This distinction becomes increasingly important on modern parallel hardware, where algorithmic mechanisms and hardware execution jointly affect evaluation throughput, runtime efficiency, and final solution quality. Therefore, GPU-based EA benchmarking should not simply inherit CPU-oriented evaluation assumptions; instead, it should assess how well an algorithm exploits parallel hardware in terms of practical resource utilization, and long-horizon search capability.

\subsection{Scaling Regimes with Problem Dimension and Population Size}
GPU benefit is determined by both algorithmic mechanism and workload scale. Even when an EA contains parallelizable operators, the optimization setting must provide enough parallel work to utilize the device effectively. In EAs, this workload is mainly controlled by problem dimension and population size. Problem dimension increases the computation associated with each individual, whereas population size increases population-level parallelism. Because GPUs favor throughput-oriented batched computation rather than small or sequential workloads, their advantage is scale-dependent. When the workload is too small, GPU resources may be underutilized. When the workload becomes large, memory pressure, synchronization, and hardware capacity may lead to saturation.

To characterize this behavior, we vary either the problem dimensionality or the population size from 16 to 8192 on a geometric scale while keeping the other factor fixed. Experiments are conducted on two GPU platforms, NVIDIA GeForce RTX 3090 and RTX 2080 Ti, and one CPU platform, Intel Xeon Gold 6226R. For each configuration, we record the wall-clock time required to complete 100 generations and the number of FEs completed within a fixed 30-second time budget. The first metric measures the cost of completing a prescribed evolutionary trajectory, while the second measures the effective search budget achievable under a practical runtime constraint. All results are averaged over 15 independent runs. Since the tested platforms and implementations are necessarily limited, the observed transition points should not be interpreted as universal thresholds. Instead, we focus on relative scaling trends and their mechanism-level causes.

\subsubsection{Scaling with Problem Dimension}
\begin{figure}[htbp]
    \centering
    \begin{minipage}[b]{0.48\textwidth}
        \includegraphics[width=\textwidth]{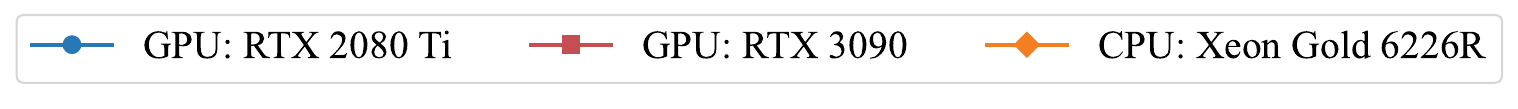}
    \end{minipage}

 \vspace{0.1cm}

 \begin{minipage}[b]{0.48\textwidth}
        \centering
        \includegraphics[width=0.49\textwidth]{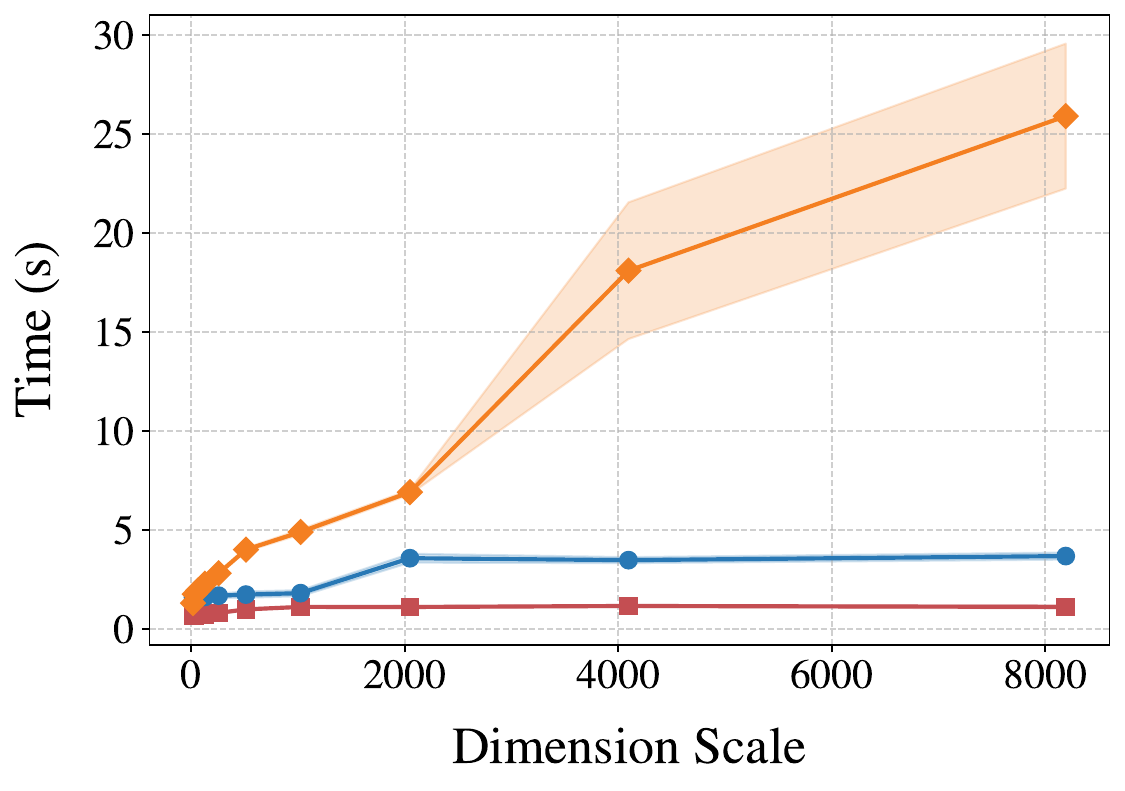}
 \hspace{-0.1cm}
        \centering
        \includegraphics[width=0.48\textwidth]{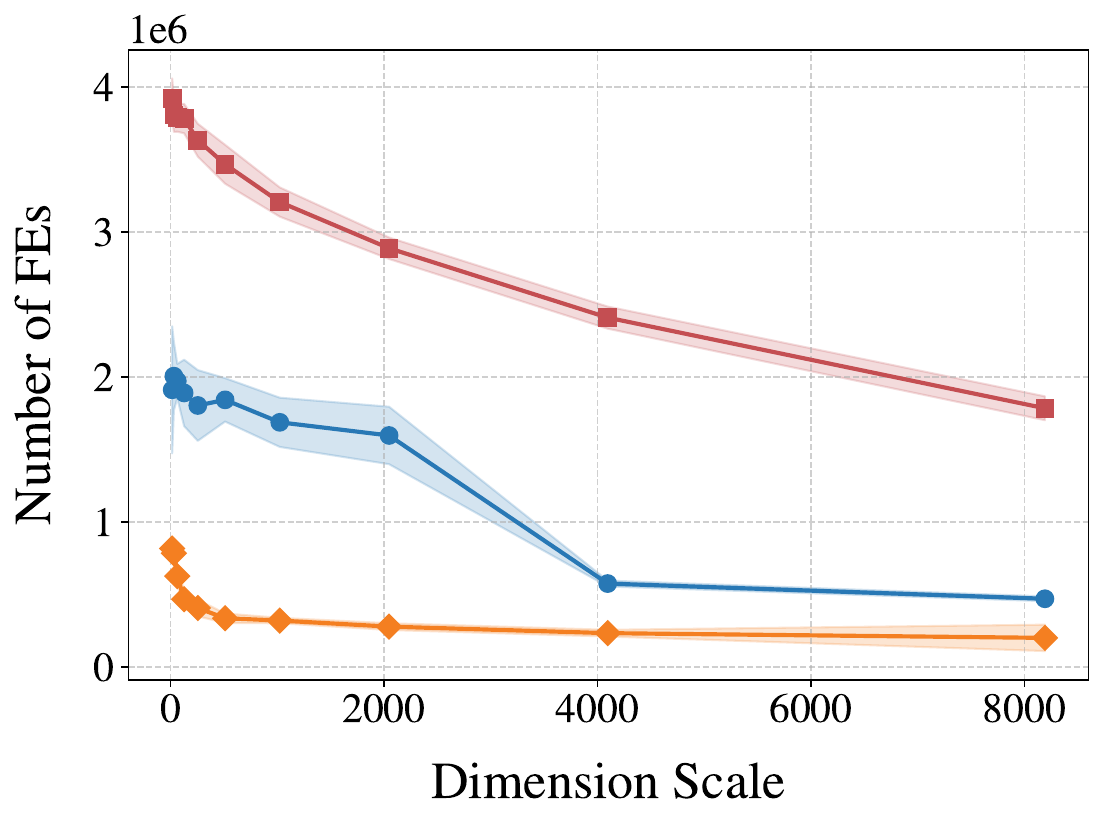}

        \centering
        \subcaption{PSO ($N$ = 128) on Ackley }
   \end{minipage}

    \vspace{0.1cm}

  \begin{minipage}[b]{0.49\textwidth}
        \centering
        \includegraphics[width=0.48\textwidth]{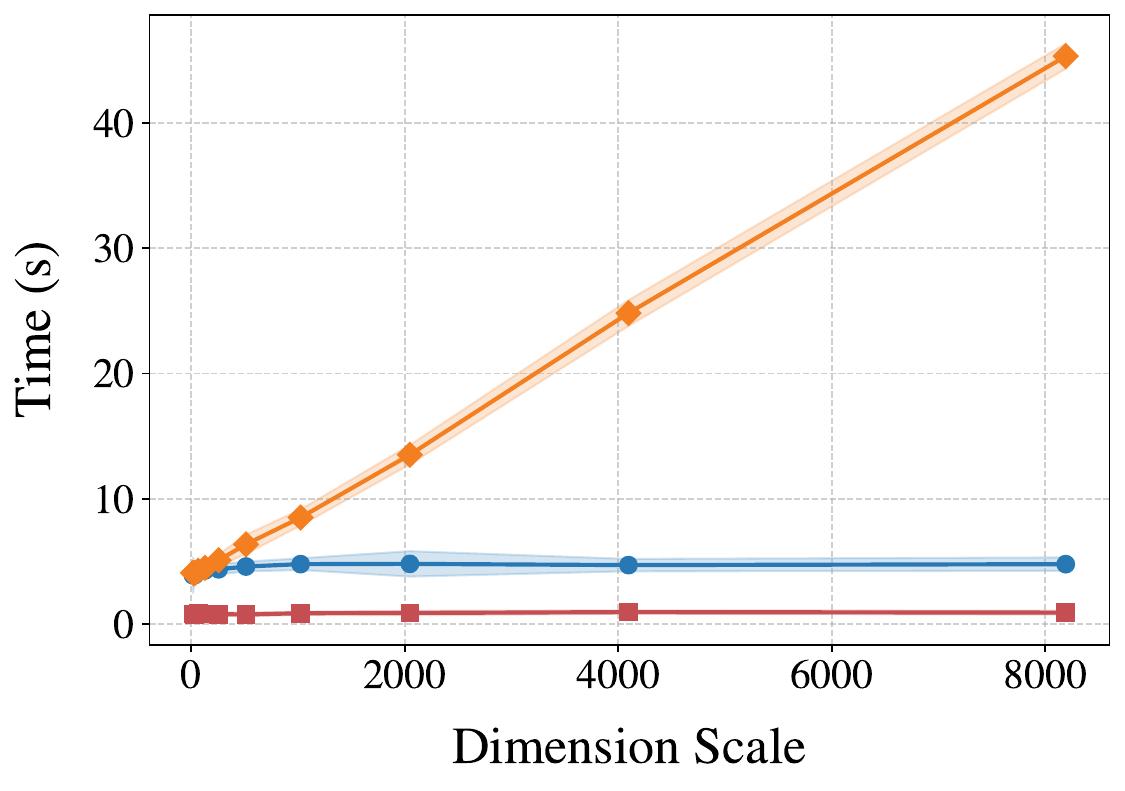}
         \hspace{-0.1cm}
        \centering
     \includegraphics[width=0.48\textwidth]{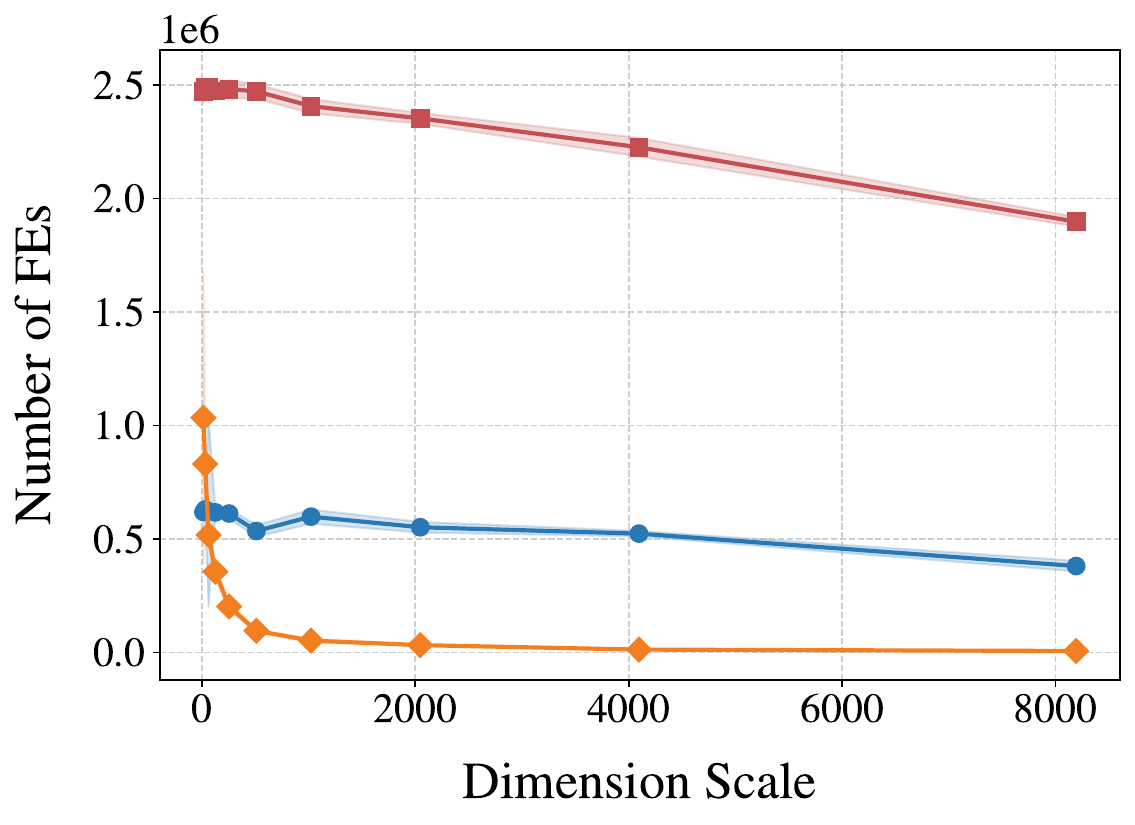}
        \centering
        \subcaption{MOEA/D ($N$ = 128) on DTLZ1}
    \end{minipage}

    \caption{Computational performance tested across varying problem dimensions on three hardware platforms. \textbf{Left}: Total runtime over 100 generations. \textbf{Right}: Number of FEs completed within a 30-second time budget. Results are averaged over 15 independent runs; solid lines indicate mean values, and shaded regions represent standard deviations.}

    \label{fig:varying dimension}
\end{figure}

Fig.~\ref{fig:varying dimension} examines how problem dimensionality affects computational efficiency when the population size is fixed at $N=128$. The two representative cases show that dimensional scaling influences GPU efficiency in different ways, depending on whether the additional workload can be mapped to regular parallel computation. More importantly, the figure also illustrates the absolute computational advantage of GPU execution in high-dimensional settings. As the dimension increases, CPU execution becomes increasingly expensive, whereas GPUs are able to keep the runtime within a much more practical range and maintain substantially higher FE throughput under the same 30-second time budget. This capability is important for EAs because it extends their applicability to high-dimensional and computationally demanding optimization problems that would otherwise be difficult to handle within acceptable wall-clock time.

For PSO on Ackley, increasing dimensionality mainly enlarges the vector operations associated with particle positions, velocities, and fitness evaluations. These operations are regular and naturally compatible with batched GPU execution. Consequently, the CPU runtime increases noticeably with dimensionality, while both GPUs exhibit much slower runtime growth. Under the fixed-time setting, this advantage becomes more pronounced: the GPUs complete far more FEs than the CPU, indicating that GPU execution not only reduces runtime but also preserves a much larger effective search budget as the representation dimension grows. In this sense, GPU acceleration changes the practical boundary of EA applications by making large-dimensional search more computationally feasible.
The behavior of MOEA/D on DTLZ1 is more nuanced. Compared with PSO, MOEA/D involves neighborhood-based cooperation and subproblem interactions, which introduce coordination and memory-access overheads. As a result, GPU acceleration is less direct, especially in small-scale settings where the available workload is insufficient to amortize synchronization and kernel overheads, as shown in the right panel of Fig.~\ref{fig:varying dimension}(b). Nevertheless, as dimensionality increases, the GPUs still provide a clear advantage in terms of feasible runtime and fixed-time FE throughput. This suggests that even for algorithms with stronger interaction structures, higher-dimensional workloads can increase GPU utilization.

\subsubsection{Scaling with Population Size}

\begin{figure}[htbp]
    \centering
    \begin{minipage}[b]{0.48\textwidth}
        \includegraphics[width=\textwidth]{Figures/parameter/legend.pdf}
    \end{minipage}

 \vspace{0.1cm}

 \begin{minipage}[b]{0.48\textwidth}
        \centering
    \includegraphics[width=0.49\textwidth]{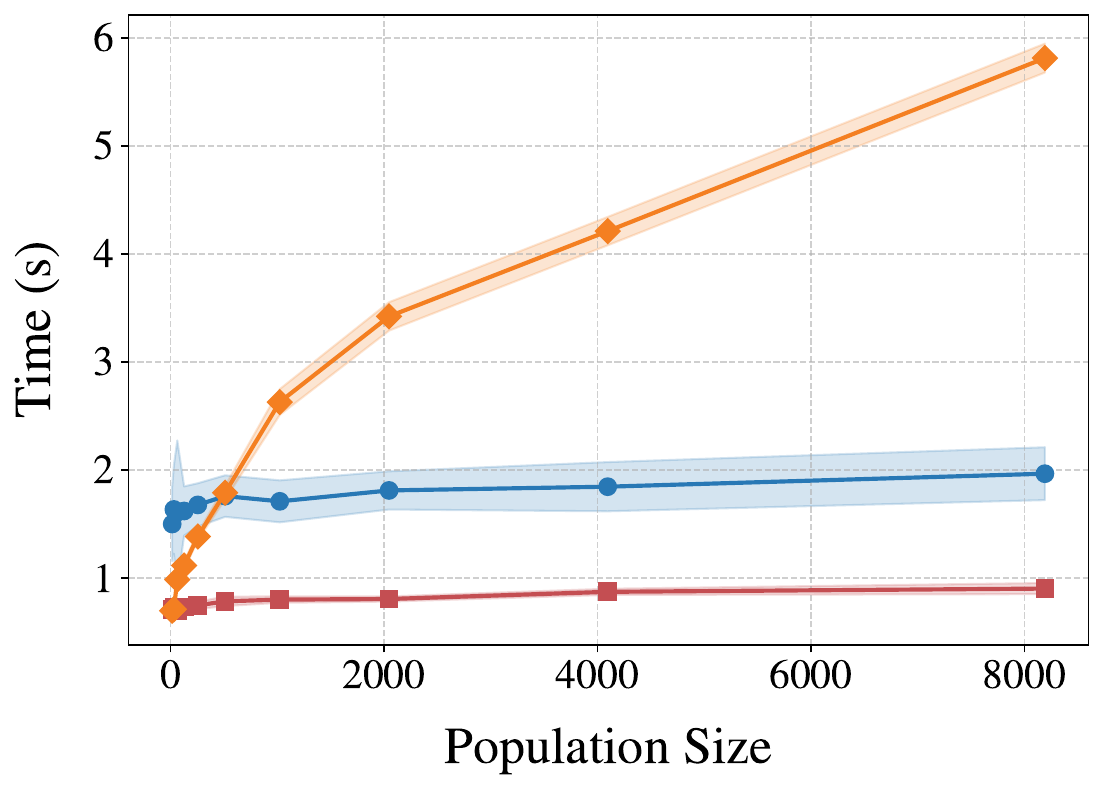}
 \hspace{-0.1cm}
        \centering
    \includegraphics[width=0.49\textwidth]{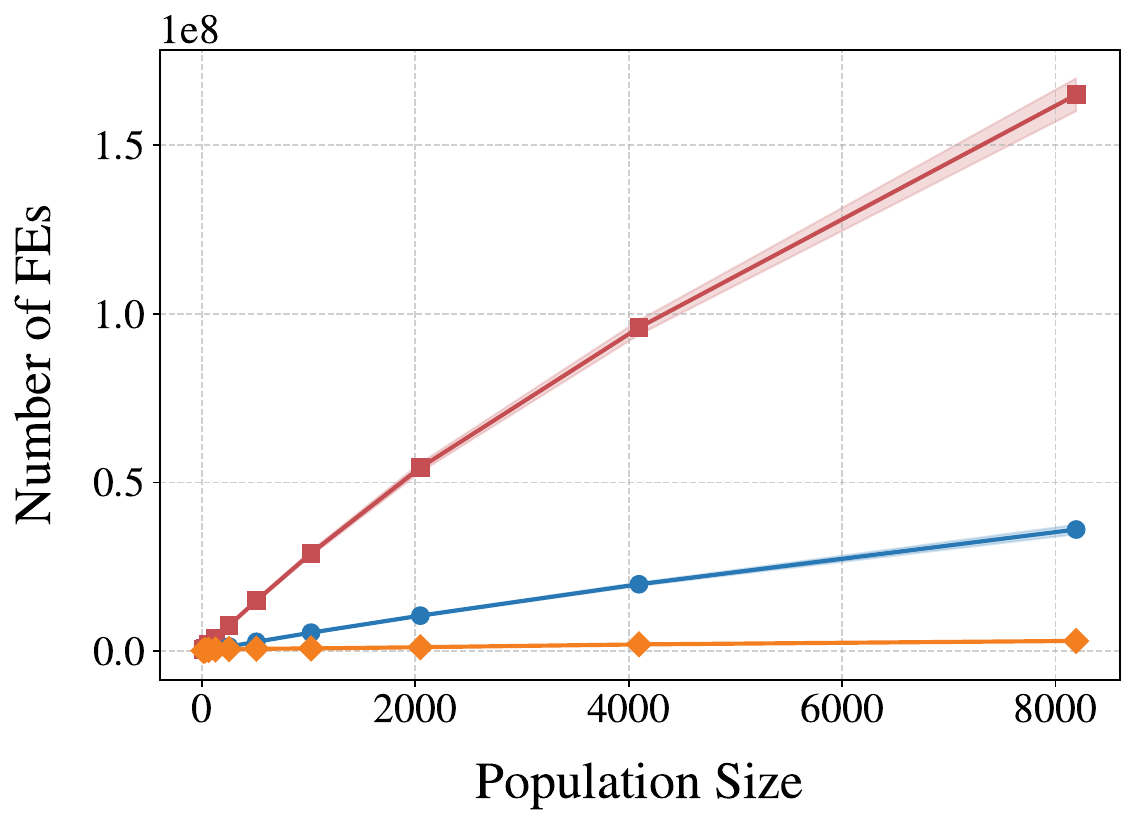}
        \centering
        \subcaption{PSO on Ackley (D = 50) }
   \end{minipage}

    \vspace{0.1cm}

  \begin{minipage}[b]{0.49\textwidth}
        \centering
    \includegraphics[width=0.48\textwidth]{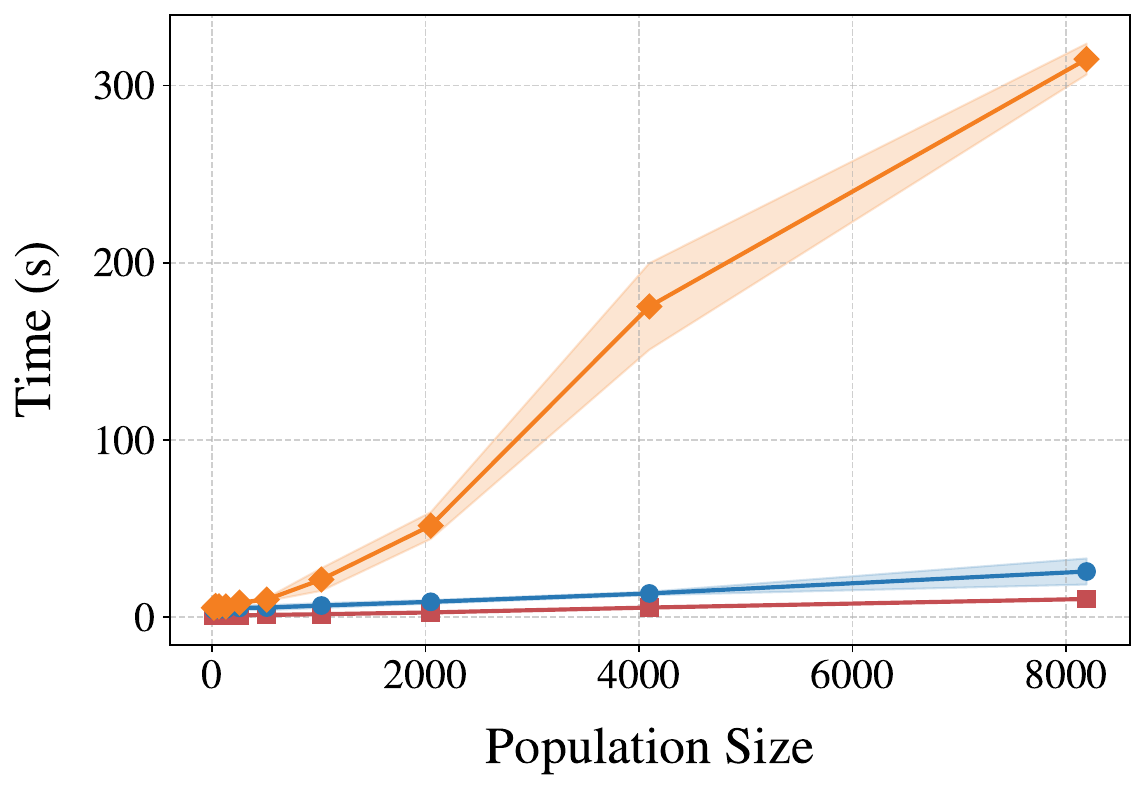}
         \hspace{-0.1cm}
        \centering
     \includegraphics[width=0.48\textwidth]{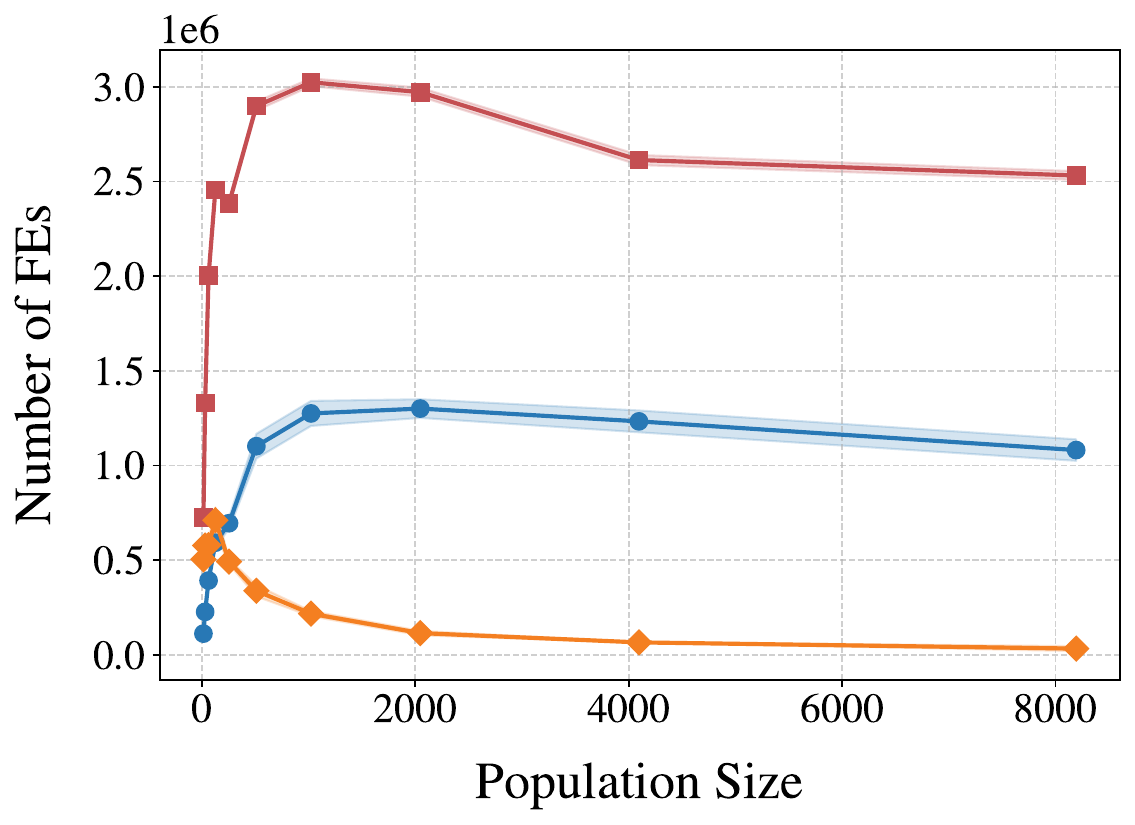}
        \centering
        \subcaption{MOEA/D on DTLZ1 (D = 50)}
    \end{minipage}

    \caption{Computational performance of two EAs configured with varying population sizes across three hardware platforms. \textbf{Left}: Total runtime over 100 generations. \textbf{Right}: Number of FEs completed within a 30-second time budget. Each curve represents the average of 15 independent runs; solid lines denote mean values, and shaded areas indicate standard deviations.}

    \label{fig:varying popsize}
\end{figure}

Fig.~\ref{fig:varying popsize} further investigates population-size scaling with fixed dimensionality at $D=50$. Compared with problem dimensionality, population size more directly controls the number of candidate solutions processed in each generation and therefore represents the most natural source of data parallelism in EAs. The results show that GPU execution provides not only better scaling trends but also a substantial absolute advantage: populations that would be costly or impractical on CPUs can be processed within a manageable runtime range on GPUs. This relaxes the traditional constraint on population-size settings and makes it possible to explore large-population configurations that were previously rarely considered in CPU-based studies.

Specifically, for PSO on Ackley, increasing the population size provides more concurrent particles for GPU execution. Since the update rule is regular and weakly synchronized, the additional individuals are effectively converted into exploitable data parallelism. Consequently, the GPU runtime over 100 generations grows much more slowly than the CPU runtime, while the number of FEs completed within 30 seconds increases substantially. This result indicates that larger populations can better fill GPU resources and significantly improve effective FE throughput. From an algorithmic perspective, GPU execution therefore opens a wider setting space for algorithms like PSO, allowing larger populations to be used not merely for hardware utilization but also for broader exploration of the search space.
In contrast, MOEA/D exhibits a different population-scaling pattern. Although increasing the population size initially provides more parallel workload, the FE throughput does not continue to grow proportionally and may plateau at larger scales. This indicates a transition from insufficient workload to bottleneck-dominated execution. As the population grows, neighborhood updates, subproblem coordination, and memory movement become increasingly significant, limiting the benefit of additional individuals. Even so, the GPU maintains a clear practical advantage over the CPU in handling large populations within a feasible time budget.

\begin{table}[t]
\centering
\caption{Interpretable scaling regimes observed in GPU-based EAs.}
\label{tab:scaling_regimes}
\footnotesize
\resizebox{\columnwidth}{!}{
\begin{tabular}{>{\centering\arraybackslash}p{1.5cm}>{\centering\arraybackslash}p{3.6cm}>{\centering\arraybackslash}p{4.2cm}>{\centering\arraybackslash}p{3.6cm}}
\toprule
\textbf{\larger[1.05]Regime} & \textbf{\larger[1.05]Empirical symptom} & \textbf{\larger[1.05]Main cause} & \textbf{\larger[1.05]Implication} \\

\midrule
Under-utilization
& CPU remains competitive
& Insufficient parallel workload; GPU overhead is not fully amortized
& CPU execution or task batching may be sufficient \\

Effective parallelism
& GPU runtime grows slowly; fixed-time FE throughput remains high
& Sufficient regular and batched workload exposes data parallelism
& GPU execution substantially improves time-to-solution \\

Saturation / bottleneck
& Throughput plateaus or declines
& Memory bandwidth, synchronization, communication, or irregular access becomes dominant
& Mechanism-level restructuring may be required \\
\bottomrule
\end{tabular}}
\end{table}

In summary, Table~\ref{tab:scaling_regimes} reports three representative scaling regimes. At small dimensions or population sizes, the workload is often insufficient to amortize GPU overhead, allowing CPU execution to remain competitive. As the scale increases, more regular population-level computation becomes available, enabling GPUs to achieve lower runtime and substantially higher FE throughput. When the scale becomes very large, however, the gain may saturate as memory bandwidth, synchronization, communication, or algorithm-specific dependencies become the dominant bottlenecks.

These regimes show that the effectiveness of GPU execution is jointly determined by problem scale, population scale, algorithmic structure, and hardware capacity. For EA applications, this implies that GPUs can extend the practical scope of evolutionary optimization by making high-dimensional problems and large-population settings feasible within acceptable wall-clock time. For EA design, the implication is more specific: scalability on GPUs depends on whether the dominant evolutionary operations can be formulated as regular, batched, and weakly synchronized computations. Therefore, GPU-aware EA design should expose population-level parallelism, reduce communication, regularize memory access, and minimize sequential dependencies in critical update steps.

\subsection{Large-Population Dynamics Beyond Throughput}
The preceding analysis shows that increasing population size can enhance GPU efficiency by exposing additional population-level parallelism. However, population size is not merely a hardware-utilization parameter; it directly affects search-space coverage, selection pressure, diversity preservation, and the balance between exploration and exploitation of EAs.
Traditional EA studies have typically employed small to moderate populations, as larger populations increase computational cost and, under a fixed-FE budget, reduce the number of executable generations.
The high-throughput capabilities of modern GPUs relax these constraints and enable us to revisit a key question: when large populations become computationally feasible, do they merely increase throughput, or do they change the search dynamics of EAs?

We analyze this question from three perspectives. First, we examine the overall impact of population scaling across sixteen EAs on standard numerical benchmarks. Second, we employ  mechanism-oriented case studies to explain how large populations affect coverage, contraction, diversity retention, and optimization improvement. Third, we evaluate neuroevolution tasks to examine whether the observed effects remain relevant in expensive, high-dimensional policy-search settings.

\subsubsection{Effects on Numerical Benchmarks}
We first benchmark sixteen EAs across ten population sizes, exponentially increasing from 16 to 8192. Each configuration is evaluated on six numerical functions with 15 independent runs on an NVIDIA GeForce RTX 3090 GPU. To isolate the effect of population size, all algorithms are run for a fixed budget of 100 iterations. Thus, different population settings experience the same number of evolutionary update cycles, and the observed differences mainly reflect how each algorithm utilizes additional individuals. The complete results are reported in Supplementary Material Section VI-C.

\begin{figure}[htbp]
    \centering
    \begin{minipage}[b]{0.24\textwidth}
        \centering
        \includegraphics[width=\textwidth]{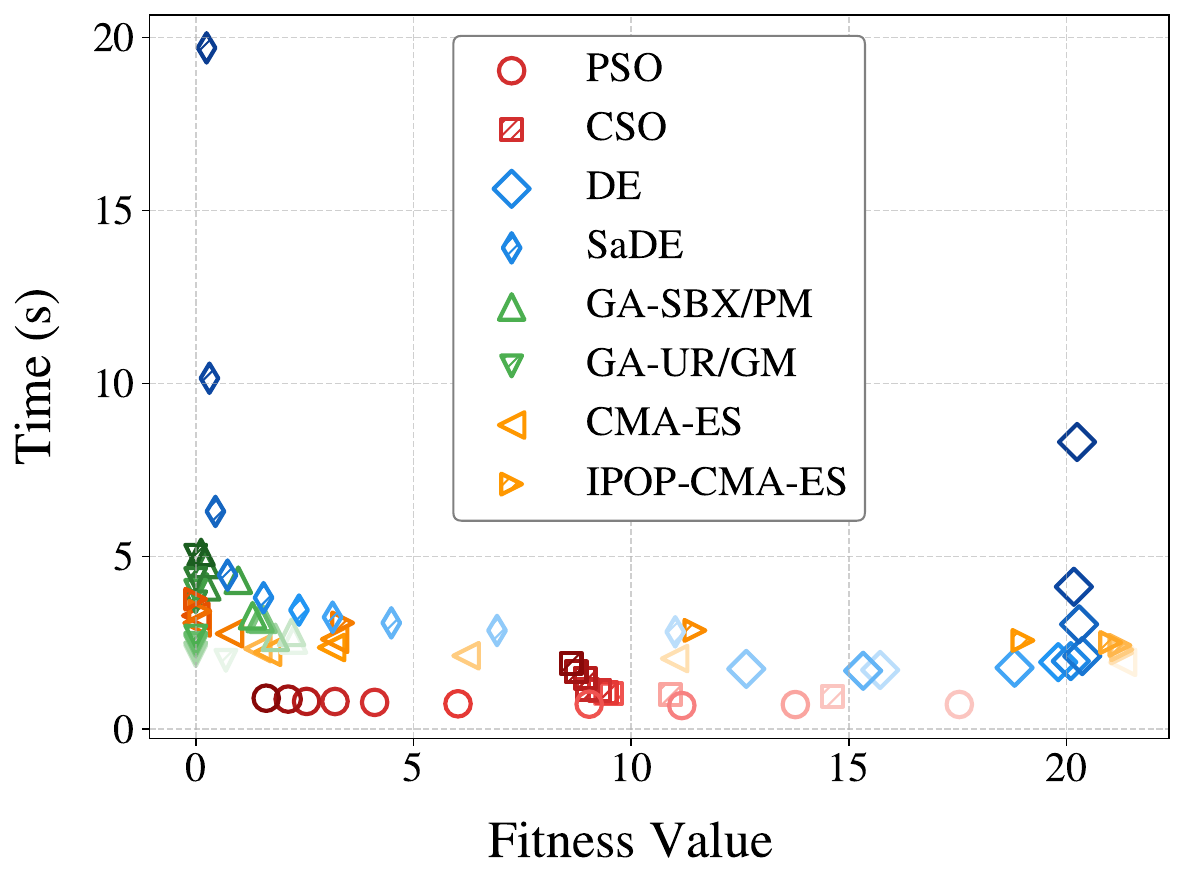}
        \centering
        \vspace{-0.4cm}
        \subcaption{$f_{a_{6}}$: 50D Ackley}
    \end{minipage}
  \hspace{-0.1cm}
    \begin{minipage}[b]{0.24\textwidth}
        \centering
        \includegraphics[width=\textwidth]{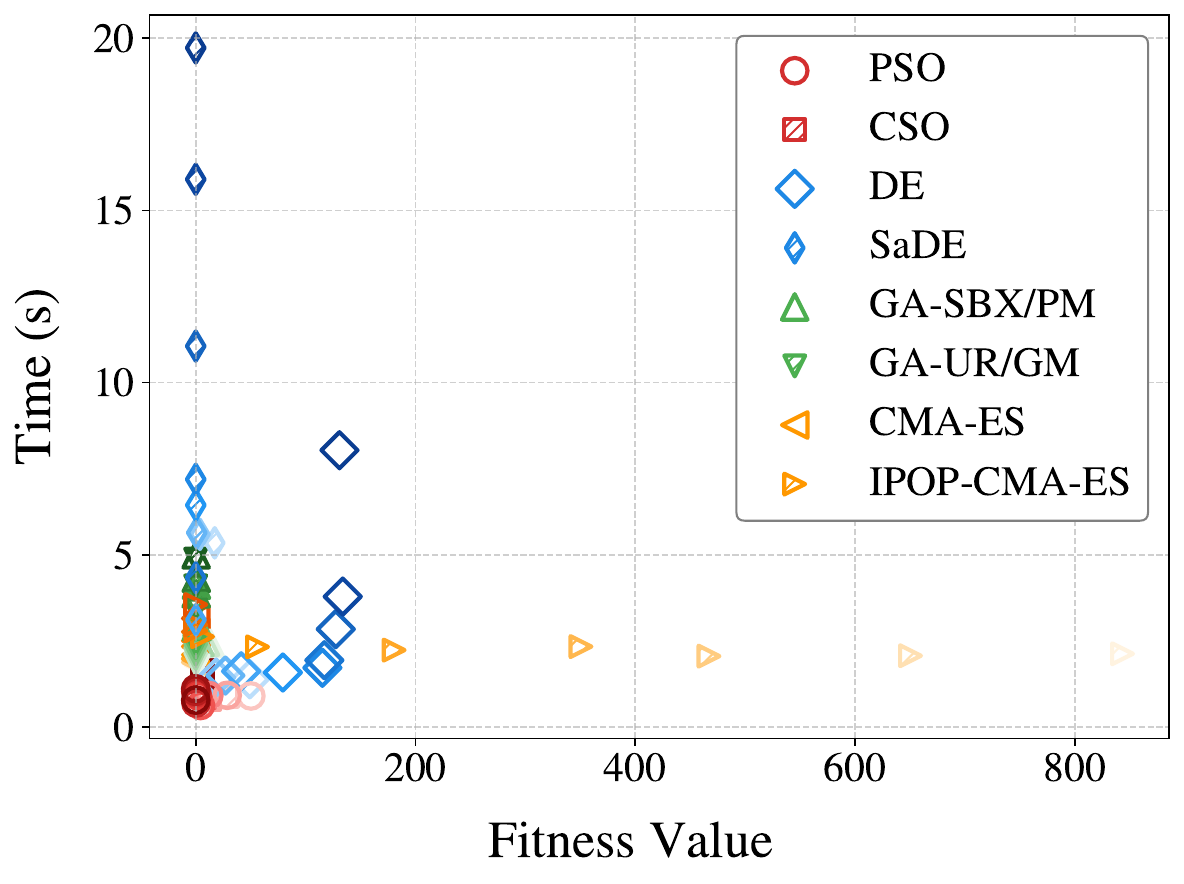}
        \centering
        \vspace{-0.4cm}
        \subcaption{$f_{a_{10}}$: 50D Sphere}
    \end{minipage}

   \vspace{0.1cm}

  \begin{minipage}[b]{0.24\textwidth}
        \centering
        \includegraphics[width=\textwidth]{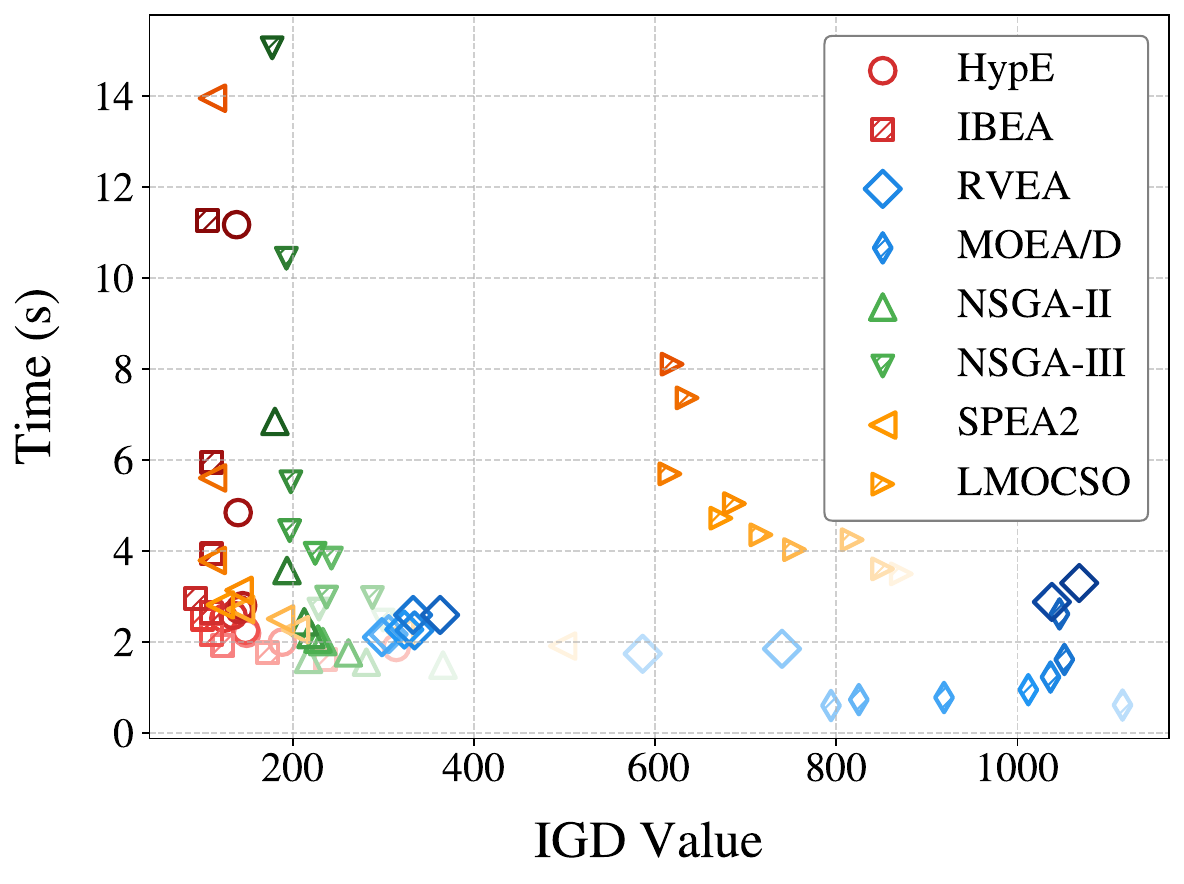}
        \centering
        \vspace{-0.4cm}
        \subcaption{$f_{a_{11}}$: 50D DTLZ1}
     \end{minipage}
  \hspace{-0.1cm}
    \begin{minipage}[b]{0.24\textwidth}
        \centering
        \includegraphics[width=\textwidth]{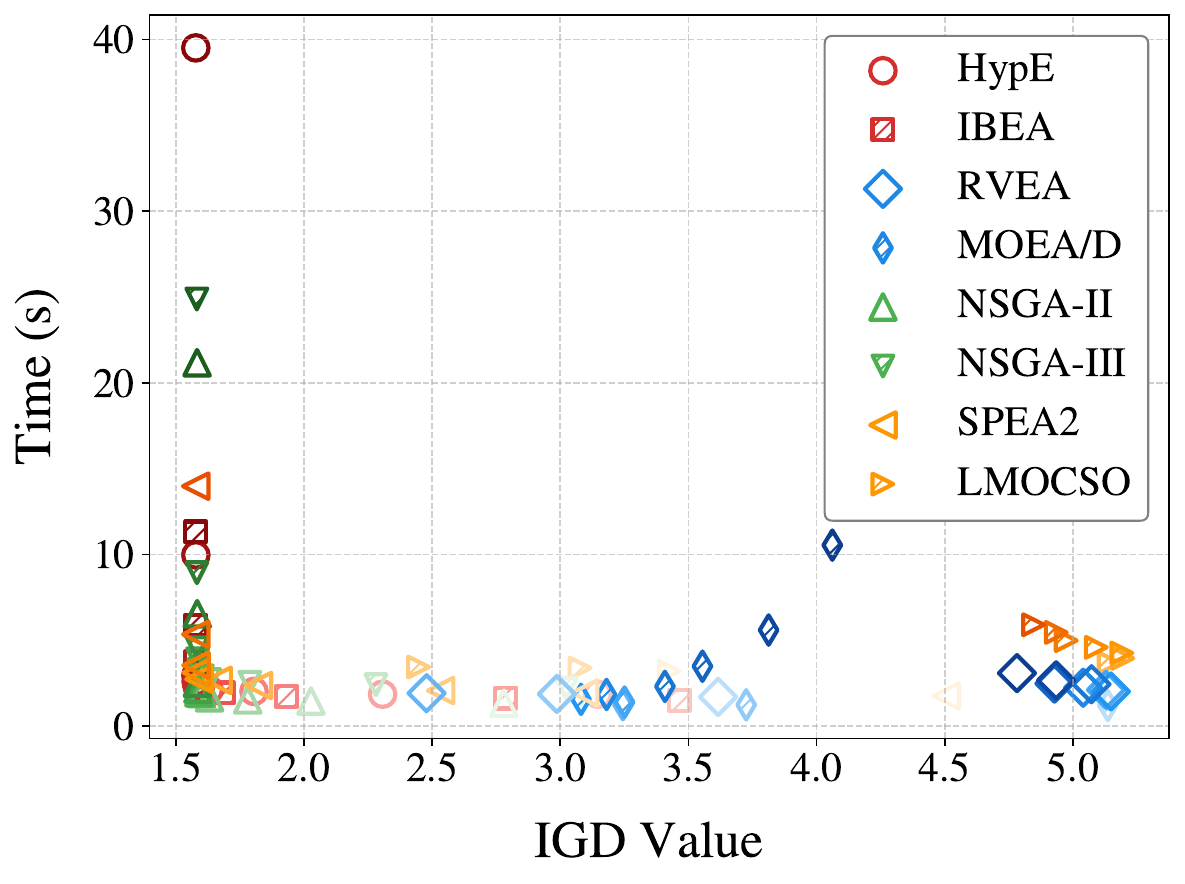}
        \centering
        \vspace{-0.4cm}
        \subcaption{$f_{a_{19}}$: 50D ZDT2}
    \end{minipage}
    \centering

    \caption{Performance comparison of EAs under varying population sizes, evaluated in terms of solution quality and time consumed over 100 iterations. Lower fitness/IGD values denote better performance. Results represent averaged performance values across 15 independent runs. Markers represent individual algorithms, color-coded from light to dark to indicate increasing population sizes (from 16 to 8192).}

    \label{fig:pop-curve}
\end{figure}

Fig.~\ref{fig:pop-curve} summarizes the trade-off between solution quality and runtime under different population sizes. Overall, the GPU implementation supports large populations with a moderate runtime increase, even when the population size grows by more than 500-fold. However, the resulting performance gain is not uniform across algorithms.
Specifically, PSO benefits markedly from larger populations, achieving lower fitness values with minimal additional time cost. In contrast, the two GA variants exhibit only modest improvement with large-population configurations.
On $f_{a_{10}}$, IPOP-CMA-ES demonstrates a pronounced advantage, rapidly converging to near-optimal solutions as population size grows, while DE tends to show stagnation or slight regression with additional individuals.
Similar heterogeneity is observed on multi-objective benchmarks.
On $f_{a_{11}}$, LMOCSO and HypE achieve clear IGD reductions, indicating improved Pareto-front approximation and coverage with larger populations. By contrast, NSGA-II and SPEA2 are relatively insensitive to further population enlargement, and MOEA/D even exhibits degraded performance under larger configurations. These results indicate that a large population is not intrinsically beneficial. Although more individuals improve the chance of covering promising regions, the final effect depends on whether the algorithm can organize, select, and exploit the additional search information effectively.

\begin{figure*}[htbp]
    \centering
    \begin{minipage}[b]{0.48\textwidth}
        \centering
        \includegraphics[width=\textwidth]{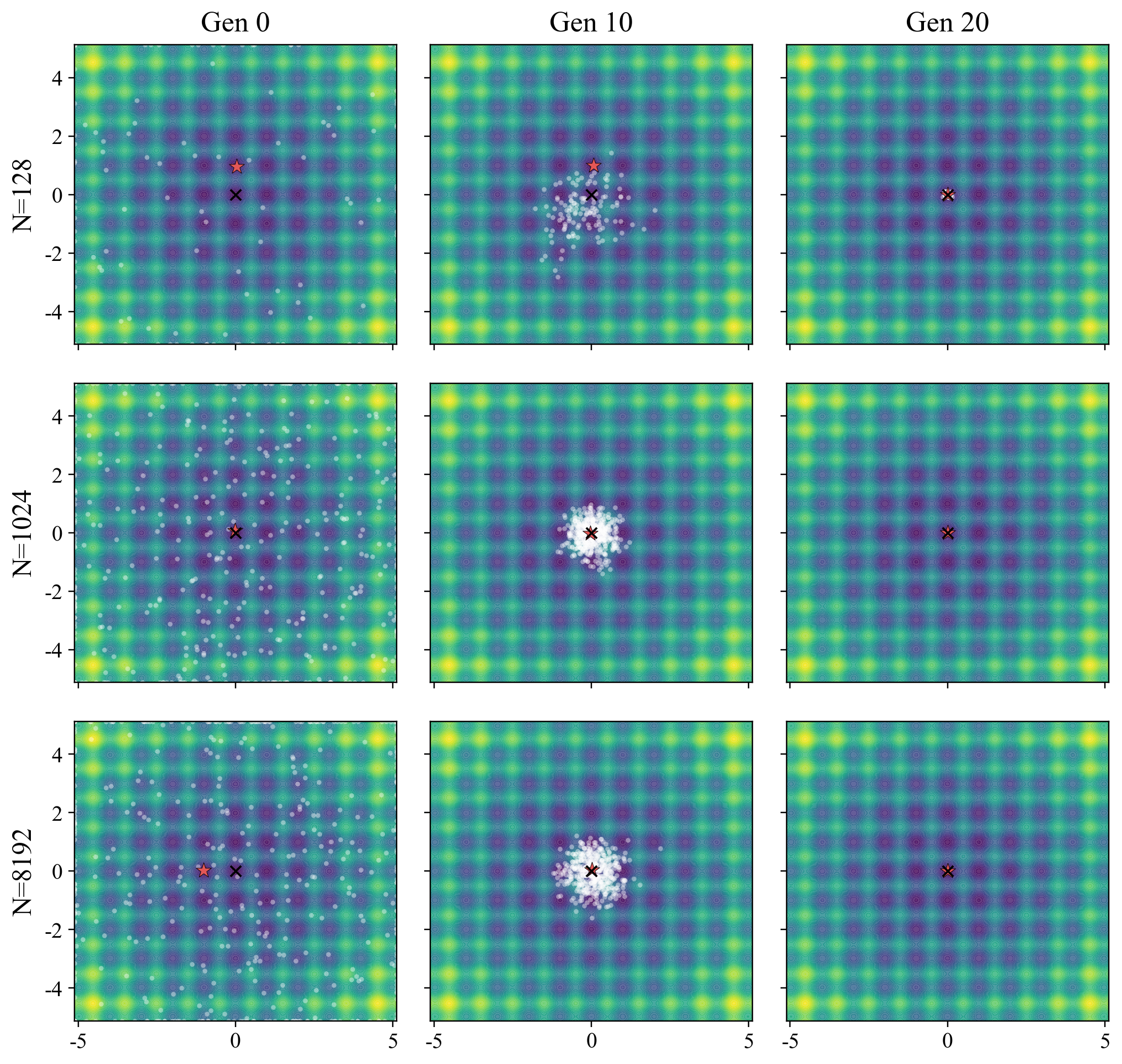}

        \centering
        \vspace{-0.1cm}
        \subcaption{CMA-ES}
    \end{minipage}
    \vspace{0.3cm}
    \centering
    \begin{minipage}[b]{0.48\textwidth}
        \centering
        \includegraphics[width=\textwidth]{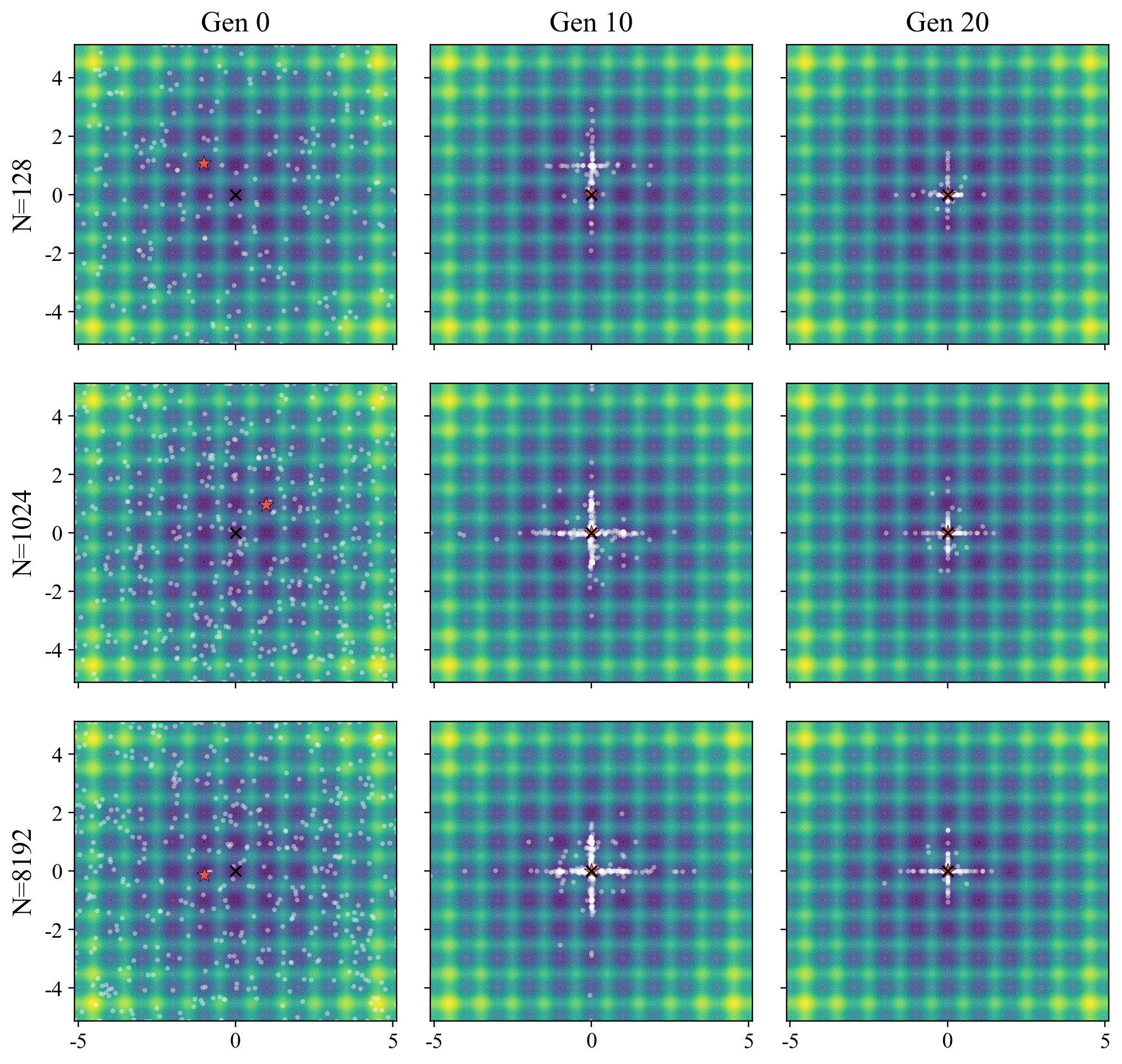}

        \centering
        \vspace{-0.1cm}
        \subcaption{GA-SBX/PM}
    \end{minipage}
     \vspace{0.3cm}
    \centering
    \begin{minipage}[b]{0.48\textwidth}
        \centering
        \includegraphics[width=\textwidth]{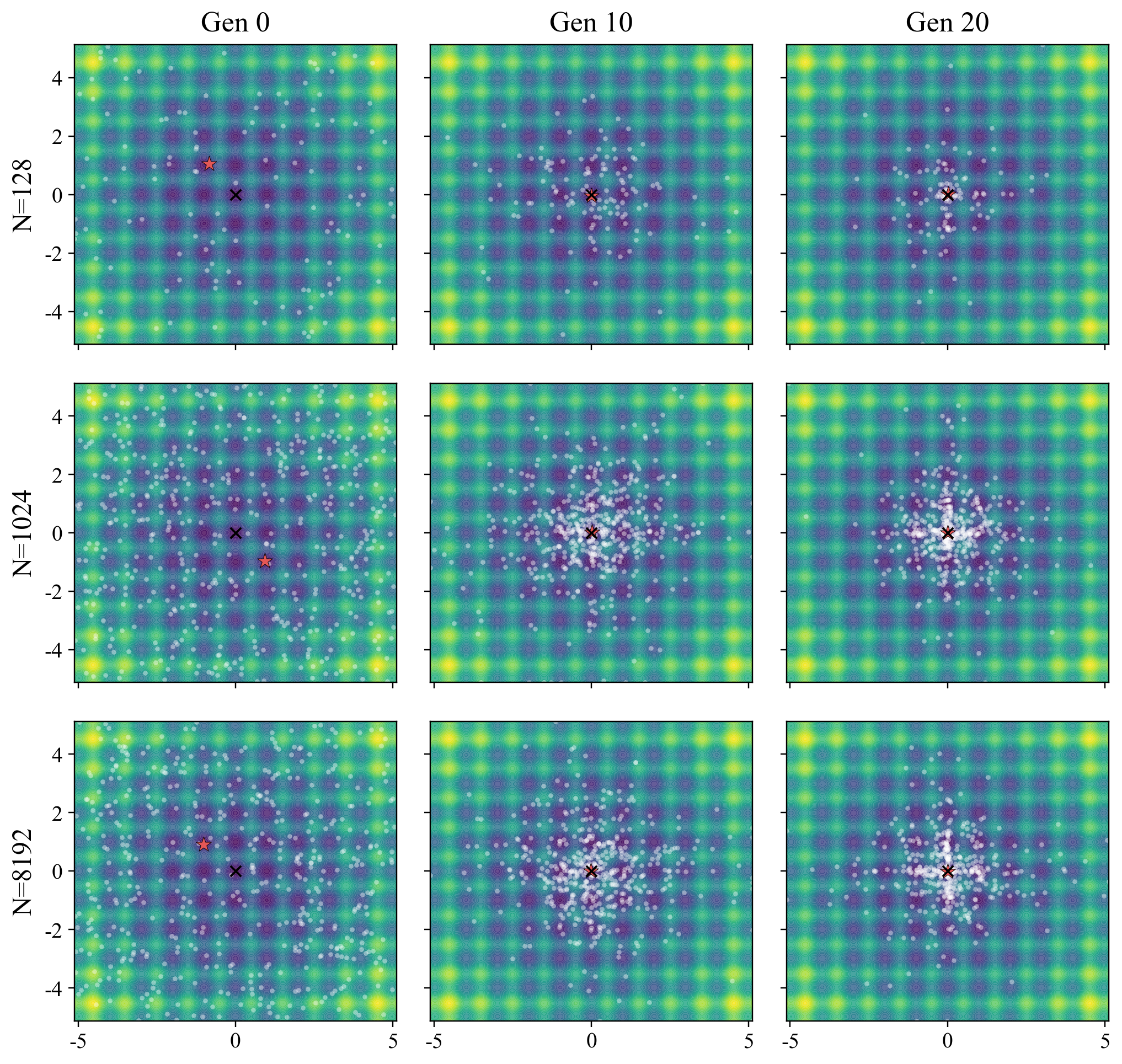}

        \centering
        \vspace{-0.1cm}
        \subcaption{PSO}
    \end{minipage}
    \vspace{0.3cm}
    \centering
    \begin{minipage}[b]{0.48\textwidth}
        \centering
        \includegraphics[width=\textwidth]{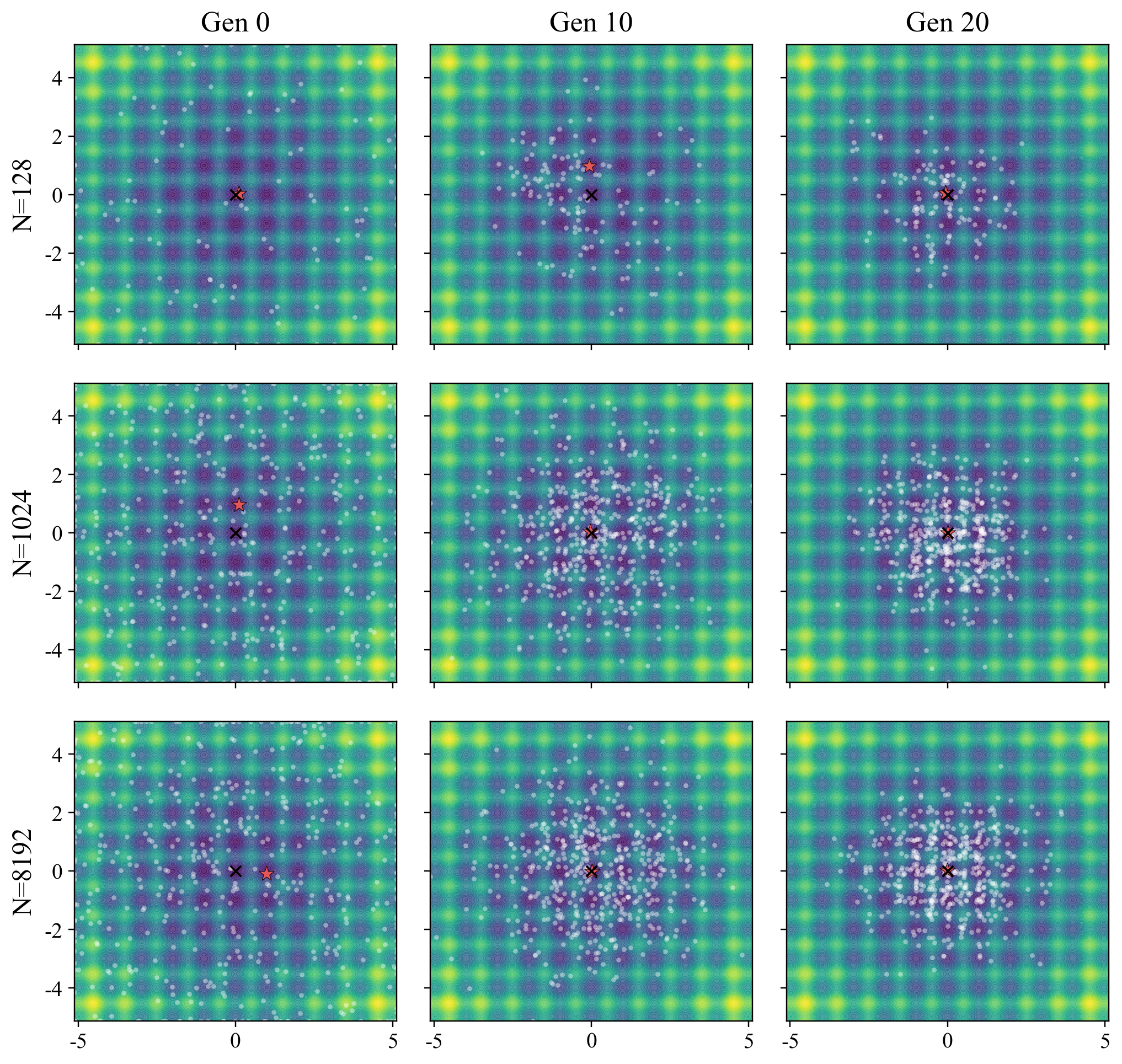}

        \centering
        \vspace{-0.1cm}
        \subcaption{DE}
    \end{minipage}
    \vspace{-0.5cm}
    \caption{Population snapshots on 2D Rastrigin with population sizes of 128, 1024, and 8192. Four representative EAs are shown at generations 0, 10, and 20.}
    \label{fig:2D-case}
\end{figure*}

To further examine the underlying mechanisms, we analyze four representative algorithms: PSO, CMA-ES, GA-SBX/PM, and DE (rand/1/bin), which correspond to swarm-based attraction, distribution-based adaptation, recombination-based search, and differential variation, respectively.
Fig.~\ref{fig:2D-case} visualizes the population dynamics of the four algorithms on the 2D Rastrigin function using population sizes of 128, 1024, and 8192, with snapshots recorded at generations 0, 10, and 20.

The snapshots show that larger populations provide denser initial coverage, but this coverage is transformed differently by different search operators.
CMA-ES contracts most rapidly, suggesting that the additional samples improve distribution estimation and accelerate concentration around the global basin. GA-SBX/PM also benefits from larger populations, but its search remains more structured due to crossover and polynomial mutation; larger populations reduce the risk of committing to a local basin by supplying more recombinable material. PSO exhibits attraction toward the best regions while retaining moderate spread through personal-best and velocity updates. In contrast, DE preserves the widest population cloud, since differential mutation continuously generates directions from population-level differences. This improves exploration, but also weakens short-term concentration.

We then extend the analysis to 200D Ackley, Rastrigin, and Schwefel functions. The same four algorithms are tested with population sizes of 128, 1024, and 8192. Each configuration is run for 100 generations over 15 independent trials.
At each generation, we record the best-so-far fitness and the decision-space diversity, defined as
\begin{equation}
\mathrm{Div}(t)=
\frac{1}{N}\sum_{i=1}^{N}
\frac{\|\mathbf{x}_i^{(t)}-\bar{\mathbf{x}}^{(t)}\|_2}
{\|\mathbf{u}-\mathbf{l}\|_2},
\end{equation}
where $\bar{\mathbf{x}}^{(t)}$ is the population center at generation $t$, and $\mathbf{u}$ and $\mathbf{l}$ denote the upper and lower bounds of the search space. We further compute the diversity retention ratio as
\begin{equation}
R_{\mathrm{div}}=\frac{\mathrm{Div}(T)}{\mathrm{Div}(0)}.
\end{equation}
This metric measures the proportion of decision-space spread retained after evolution.
A high value of $R_{\mathrm{div}}$, however, does not necessarily imply better optimization performance, since retained diversity is beneficial only when it can be converted into effective search progress.

\begin{figure}[htbp]
    \centering
    \includegraphics[width=0.46\textwidth]{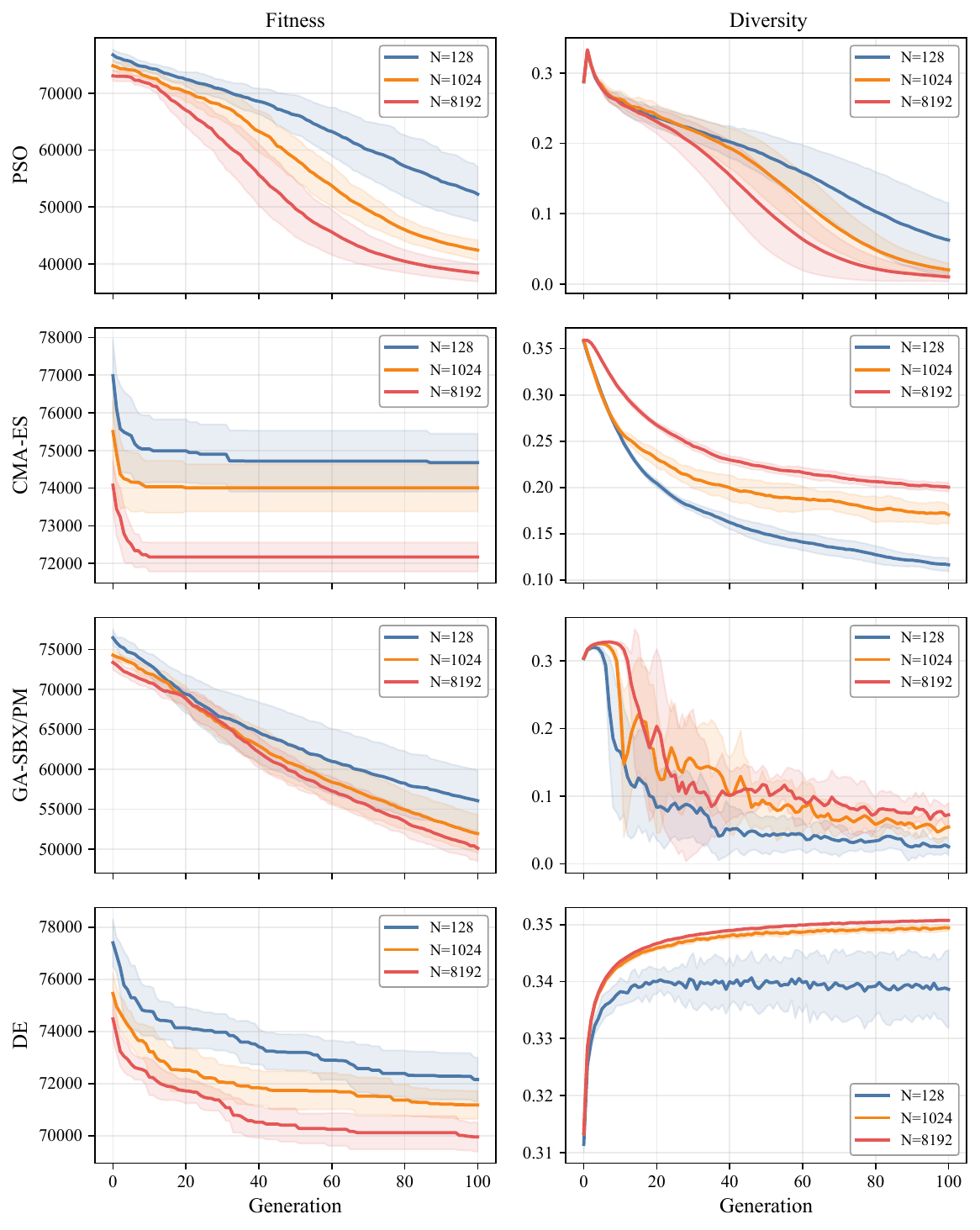}
    \caption{Convergence and decision-space diversity on 200D Schwefel. Four representative algorithms are tested with population sizes of 128, 1024, and 8192 for 100 generations. Results are averaged over 15 runs. }
    \label{fig:large-pop-dynamics-200d}
\end{figure}

Fig.~\ref{fig:large-pop-dynamics-200d} reports the convergence and diversity dynamics on 200D Schwefel. Larger populations consistently improve the final fitness of all four algorithms, but the diversity curves show that this benefit is realized through different search mechanisms.
Specifically, PSO exhibits the strongest contraction: diversity decays faster with larger populations and eventually collapses to a very low level. This indicates that PSO mainly uses the enlarged population for early basin identification, and then rapidly converts the discovered elite information into exploitation through personal and global best attraction.
CMA-ES follows a more conservative trajectory. Larger populations retain higher diversity and show a gradually slowing decay, indicating that additional samples support more stable distribution adaptation rather than immediate contraction.
GA-SBX/PM presents an intermediate behavior. Its diversity drops sharply at the early stage and then fluctuates at a low but nonzero level, with larger populations retaining more diversity. This implies that selection quickly filters out poor regions, while crossover and mutation continue to provide residual variation. Hence, a large population improves GA-SBX/PM by increasing the supply of useful genetic material and delaying complete diversity loss.
DE maintains the highest diversity among the four algorithms. Its diversity remains high and even increases for medium and large populations, which is consistent with the use of population-difference-based mutation. A larger population provides more diverse difference vectors and therefore sustains exploration. However, compared with PSO and GA-SBX/PM, this exploration is converted into convergence more slowly during the 100 iterations.

\begin{figure}[htbp]
    \centering

    \begin{minipage}[b]{0.48\textwidth}
        \centering
        \includegraphics[width=\textwidth]{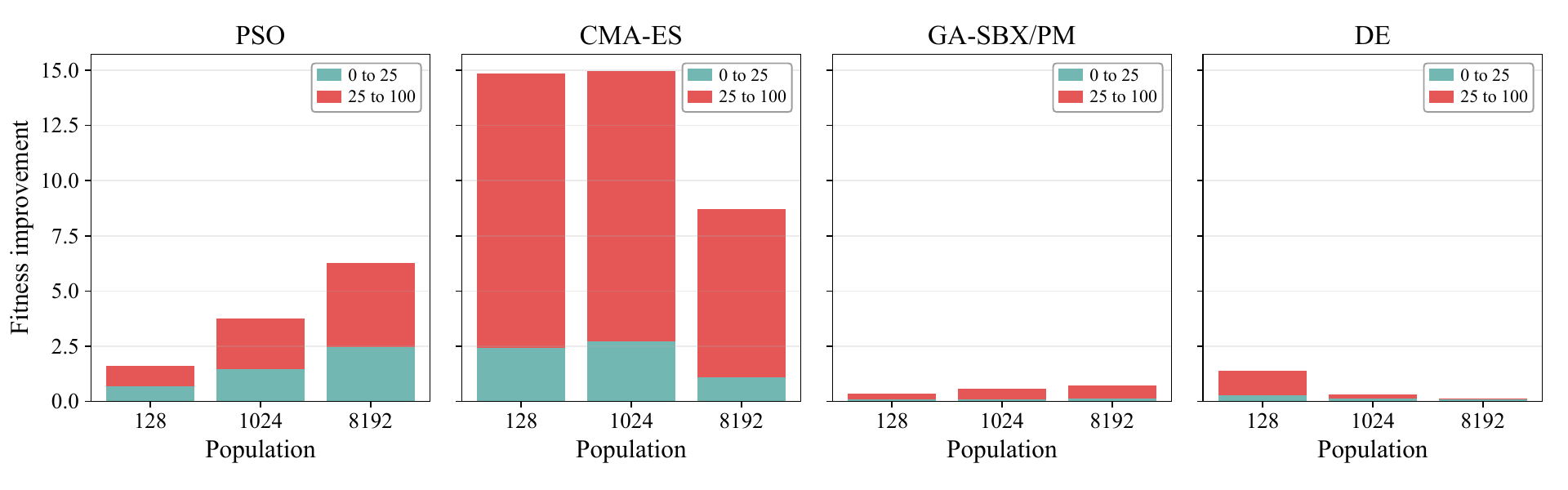}
        \centering
        \vspace{-0.6cm}
        \subcaption{Ackley}
    \end{minipage}

    \vspace{0.1cm}

    \begin{minipage}[b]{0.48\textwidth}
       \centering
        \includegraphics[width=\textwidth]{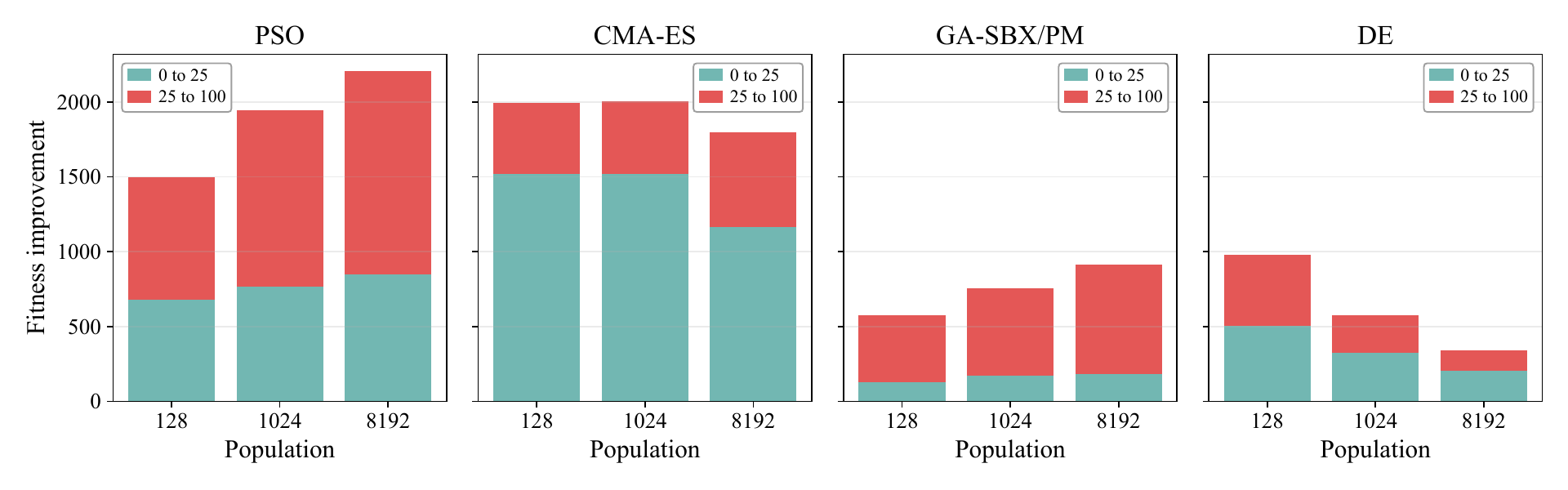}
        \centering
        \vspace{-0.6cm}
        \subcaption{Rastrigin}
    \end{minipage}

    \vspace{0.1cm}

    \begin{minipage}[b]{0.48\textwidth}
       \centering
        \includegraphics[width=\textwidth]{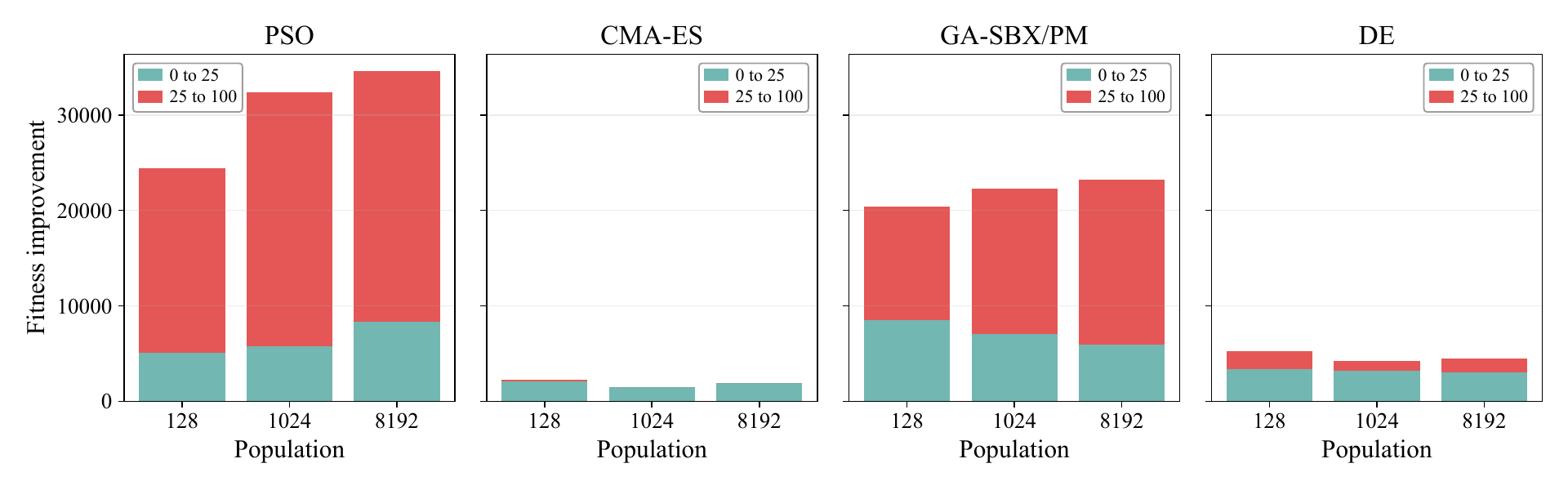}
        \centering
        \vspace{-0.6cm}
        \subcaption{Schwefel}
    \end{minipage}

    \caption{Improvement decomposition on 200D numerical problems. Total improvement over 100 generations is split into early improvement, $Q_0-Q_{25}$, and late improvement, $Q_{25}-Q_{100}$. Results are averaged over 15 runs. }
    \label{fig:improvement-decomposition-200d}
\end{figure}

To separate early and late contributions, Fig.~\ref{fig:improvement-decomposition-200d} decomposes the total improvement into $Q_0-Q_{25}$ and $Q_{25}-Q_{100}$, where $Q_0$, $Q_{25}$, and $Q_{100}$ denote the best-so-far fitness values at generations 0, 25, and 100, respectively. Unlike the 2D case, the 200D results show that large populations can also contribute substantially after the early stage. This effect is particularly evident for PSO and GA-SBX/PM on multimodal or deceptive landscapes, where larger populations not only improve initial basin coverage but also sustain later progress. CMA-ES demonstrates a trade-off between sampling reliability and adaptation speed. For DE, late-stage improvement remains limited despite high diversity retention, confirming that additional diversity alone is insufficient.

Overall, the experiments show that the role of large populations is mechanism-dependent. Their main value is not simply to increase diversity, but to improve the availability of useful search information. Whether this information becomes optimization progress depends on the algorithm's ability to transform population coverage into selection pressure, distribution adaptation, recombination material, or effective exploratory directions.

\subsubsection{Effects on Neuroevolution Tasks}
\begin{figure}[htbp]
    \centering
    \includegraphics[width=0.45\textwidth]{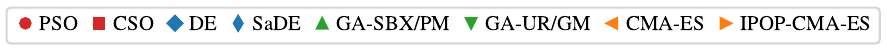}

    \vspace{0.2cm}

   \begin{minipage}[b]{0.24\textwidth}
        \centering
        \includegraphics[width=\textwidth]{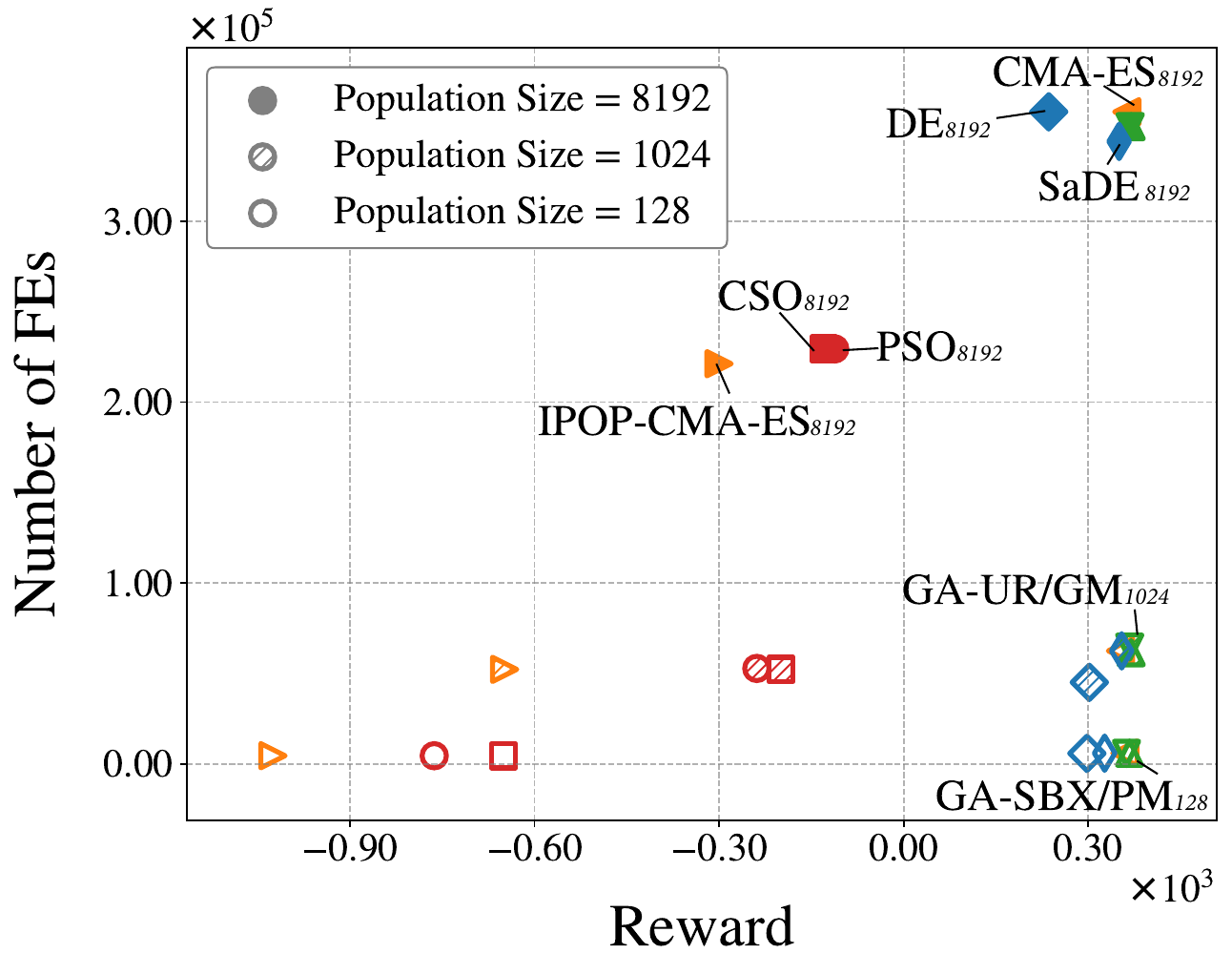}
        \centering
        \vspace{-0.5cm}
        \subcaption{ $f_{b_{4}}$: Swimmer}
    \end{minipage}
   \hfill
    \begin{minipage}[b]{0.24\textwidth}
        \centering
        \includegraphics[width=\textwidth]{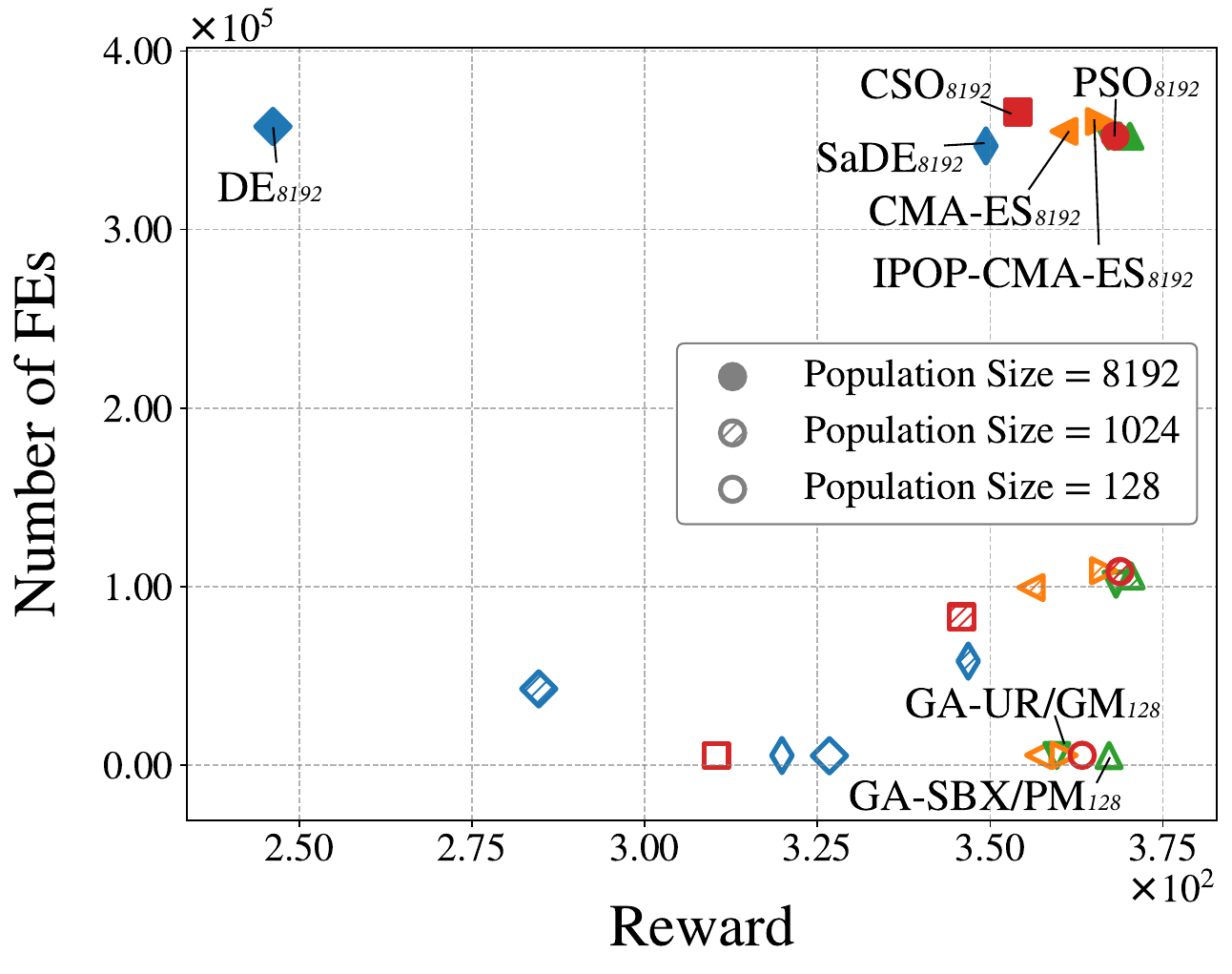}
        \centering
        \vspace{-0.5cm}
        \subcaption{ $f_{b_{5}}$: Walker2d}
    \end{minipage}

    \vspace{0.2cm}

    \includegraphics[width=0.45\textwidth]{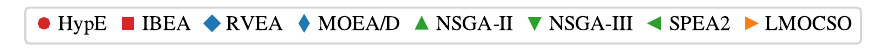}

    \vspace{0.2cm}

   \begin{minipage}[b]{0.24\textwidth}
        \centering
        \includegraphics[width=\textwidth]{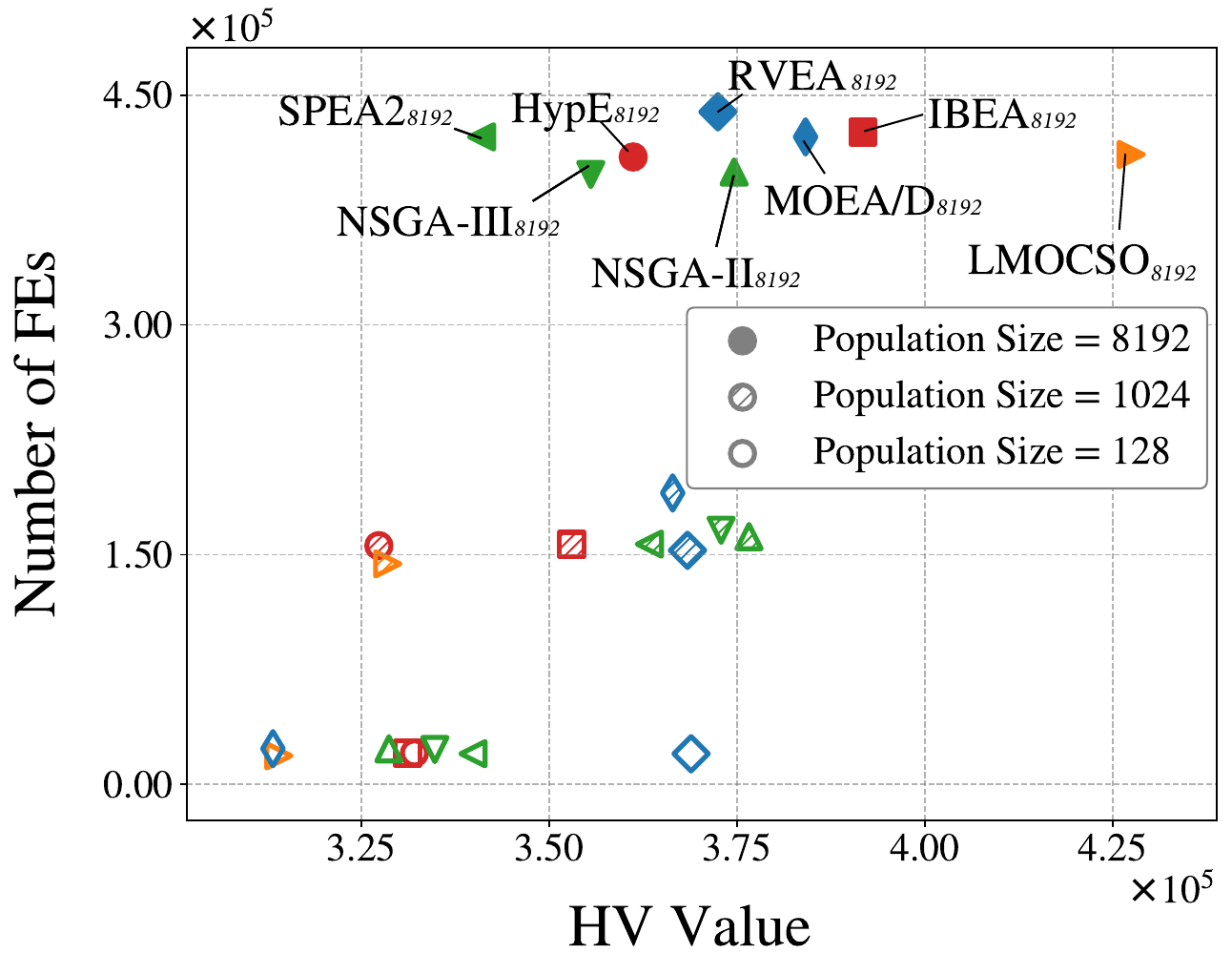}
        \centering
        \vspace{-0.5cm}
        \subcaption{ $f_{b_{6}}$: MoHopper-m2}
    \end{minipage}
   \hfill
    \begin{minipage}[b]{0.24\textwidth}
        \centering
        \includegraphics[width=\textwidth]{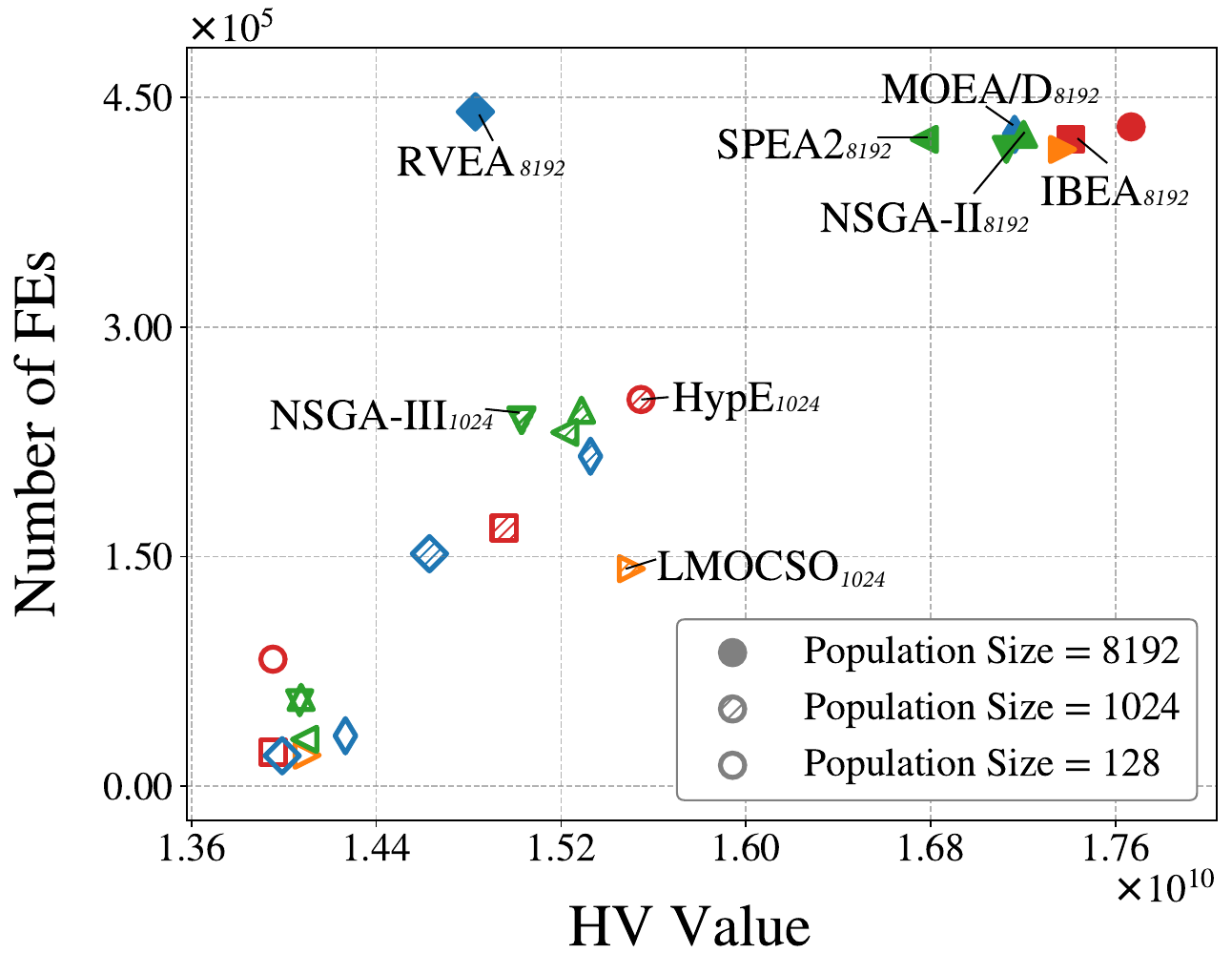}
        \centering
        \vspace{-0.5cm}
        \subcaption{ $f_{b_{7}}$: MoHopper-m3}
    \end{minipage}
    \caption{Performance of EAs tested on neuroevolution tasks under varying population sizes, evaluated in terms of solution quality and number of FEs completed within 600 seconds. Higher reward/HV values denote better performance. Results represent averaged performance values across 10 independent runs. Marker styles indicate population scales: hollow symbols for small populations (128), forward-slash-filled symbols for medium populations (1024), and solid symbols for large populations (8192). Different marker shapes distinguish between algorithms.
    }
    \label{fig:popsize}
\end{figure}

We further examine whether the large-population effects observed on numerical benchmarks remain relevant in high-cost optimization scenarios. Neuroevolution tasks involve high-dimensional policy parameters and simulation-based evaluations, making them substantially more expensive than analytic benchmark functions. This setting therefore serves as an application-oriented validation of GPU-enabled large-population EAs.
All algorithms are executed on an NVIDIA GeForce RTX 3090 GPU with three population sizes, 128, 1024, and 8192. Each configuration is run for 600 seconds over 10 independent trials. Unlike the numerical case study, where the number of generations is fixed to isolate evolutionary dynamics, the neuroevolution experiments adopt a fixed-time protocol. This setting reflects practical deployment, where the objective is to obtain high-quality policies within a limited wall-clock budget. Fig.~\ref{fig:popsize} reports achieved reward on single-objective tasks and HV value for multi-objective optimization conditions, together with the NFEs achieved on four neuroevolution tasks. Complete experimental data is available in Supplementary Material, Section VI-C.

For both single-objective tasks, the data points shift from the bottom-left to the top-right as the population grows. This trajectory indicates a dual gain in reward and evaluations completed. CMA-ES benefits the most from this scaling due to its reliance on sample size for covariance matrix estimation. With a large population, the algorithm efficiently collects sufficient statistics in parallel on the GPU, which leads to the most significant improvement in reward. Similarly, larger populations allow PSO and CSO to maintain a more diverse set of candidate solutions, thereby improving convergence to the global optimum. In contrast, classical GAs with SBX/PM or UR/GM operators exhibit only marginal progress. Their static crossover and mutation repertoire cannot maintain diversity in very high-dimensional weight spaces. Consequently, additional samples merely reiterate local patterns. Furthermore, these methods experience diminishing returns due to the overhead of global communication in large populations, which limits their scalability. SaDE employs on-line strategy adaptation to overcome this limitation and thus decisively outperforms standard DE.

For the bi-objective ($f_{b_{6}}$) and tri-objective ($f_{b_{7}}$) cases, HV rises monotonically with population size. However, the magnitude of this rise depends on the algorithm. HypE shows significant HV gain in the tri-objective setting because its hypersphere-based extreme-value sampling naturally exploits large batches to probe the high-dimensional tails of the Pareto front. Since the bi-objective surface offers fewer such extremes, the extra samples are less valuable in that context. RVEA also gains markedly, which confirms that its angular-regulation mechanism preserves exploration at scale. Conversely, NSGA-II/III and SPEA2 suffer from excessive selection pressure. Non-dominated sorting and density estimators promote early convergence around incumbent elitists. As a result, additional individuals are quickly culled rather than used to diversify the front, which leads to HV saturation at a population size of 8192. Collectively, these results indicate that enlarging the parallel evaluation budget is more cost-effective than pursuing additional generations under a fixed wall-time. GPU parallelism amplifies the strengths of algorithms that are sample-hungry or sampling-efficient, such as CMA-ES and HypE.

\section{Implications for Hardware-Aware EA Design}
The empirical results suggest that GPU parallelism should be treated as part of the design and evaluation assumptions of scalable EAs. Three implications follow from the preceding analyses.

First, EA mechanisms should be designed with hardware compatibility in mind. The acceleration results show that GPUs can provide substantial speedups, but these gains depend on the computational form of the algorithm. Dense sampling, vectorized variation, and batched state updates map well to GPU execution. In contrast, irregular sorting, neighborhood replacement, archive maintenance, and frequent synchronization can limit scalability. Therefore, future EAs should not treat GPU implementation as a post hoc acceleration step. Algorithmic mechanisms should be formulated to expose batched computation, reduce sequential dependencies, regularize memory access, and control synchronization costs.

Second, GPU-based EA benchmarking should report both sample efficiency and time-to-solution. FE-budgeted evaluation remains necessary because it compares algorithms under the same number of objective evaluations. However, CPU-oriented FE budgets may provide only a short observation window on GPUs. They may terminate the run before delayed adaptation, sustained exploration, or late-stage refinement becomes visible. Time-budgeted evaluation provides a complementary view because it measures the solution quality achieved under the same wall-clock budget. Thus, GPU-enabled EA studies should combine FE-budgeted and time-budgeted protocols rather than relying on a single evaluation unit.

Third, population size should be treated as a central design dimension rather than only as a computational burden. Under CPU-oriented assumptions, large populations are often avoided because they increase runtime and reduce the number of generations under a fixed FE budget. GPU parallelism relaxes this constraint by making large populations affordable within practical time budgets. The experiments show that large populations can improve search-space coverage and diversity, but their final benefit depends on the mechanism that converts additional samples into progress. This observation motivates large-population EAs whose operators are explicitly designed to exploit broad sampling.

Taken together, these implications define a hardware-aware view of EA scalability. A scalable EA should not only execute faster on GPUs, but should also use the additional parallel computation to improve search behavior. This requires aligning evolutionary operators, evaluation protocols, and population-size configurations with the execution model of modern parallel hardware.

\section{Conclusion}

This paper studied the scaling behavior of EAs on GPUs from a hardware-aware evaluation perspective. The results show that GPU execution should not be viewed only as an acceleration technique for existing algorithms. Instead, it changes the resource model under which EAs are evaluated and designed.

The empirical study leads to four main conclusions. First, GPU acceleration is heterogeneous across algorithms because different evolutionary mechanisms expose different levels of batched computation, memory regularity, and synchronization. Second, FE-budgeted evaluation remains useful for sample-efficiency comparison, but it can be insufficient under GPU execution because the same FE budget may provide only a limited observation window. Time-budgeted evaluation is therefore needed to assess practical time-to-solution and long-horizon search behavior. Third, GPU benefit depends on scaling regimes induced by problem dimension and population size. Fourth, large populations enabled by GPUs can improve optimization performance when the algorithm can convert additional samples into useful search information.

Overall, these findings suggest that future EA research should move from post hoc GPU acceleration toward hardware-aware evaluation and design. By developing parallel-friendly mechanisms, exploiting large-population regimes, and evaluating algorithms under both FE and time budgets, EAs can better use modern hardware to improve computational efficiency and optimization performance.

\bibliographystyle{IEEEtran}
\begin{footnotesize}
\bibliography{reference}
\end{footnotesize}

\clearpage

\onecolumn
\begin{@twocolumnfalse}
\begin{center}
    \fontsize{24}{29}\selectfont  Beyond Speedups: Hardware-Aware Evaluation of Evolutionary Algorithms on GPUs \\
    (Supplementary Document)
    \vspace{0.5em}
\end{center}
\end{@twocolumnfalse}


\section{Introduction} \label{app:su-introduction}

This document is organized as follows: Section II provides the downloadable resources, including the testing platforms, software libraries, and benchmark functions utilized in the experiments. Section III lists the abbreviations used throughout the paper. Section IV introduces the neuroevolution tasks employed in the study. Section V details the experimental setup. Specifically, it outlines the device configurations and the parameter settings for each evaluated evolutionary algorithm. Finally, Section VI analyzes the experimental results. It presents visualizations and statistical summaries to illustrate the algorithmic performance across various hardware platforms based on the hardware-aware evaluation indicators.

\section{Downloadable Material} \label{app:su-downloadable}
All experiments in this paper are performed on the EvoX platform \cite{Huang2024}, a comprehensive framework tailored for evolutionary computation research. EvoX streamlines the execution and management of algorithms across a range of hardware configurations.
The experiments include both single-objective and multi-objective optimization tasks, which are categorized into two primary categories: numerical optimization and neuroevolution. The numerical optimization problems are primarily sourced from the CEC 2022 \cite{Abdelatti2021}, DTLZ \cite{Deb2002a}, and ZDT \cite{Zitzler2000} benchmark suites, while the neuroevolution tasks make use of environments from the Brax engine. These benchmark functions are extensively employed in evolutionary computation to evaluate the performance of algorithms across a range of optimization problems. Detailed descriptions and datasets are available through the links provided below.
\begin{itemize}[noitemsep,topsep=0pt]
     \item \textbf{EvoX}: \url{https://github.com/EMI-Group/evox}

\item \textbf{CEC2022 Benchmark}: \url{https://github.com/P-N-Suganthan/2022-SO-BO}

\item \textbf{DTLZ Test Suite}: \url{https://github.com/EMI-Group/evox/blob/main/src/evox/problems/numerical/dtlz.py}

\item \textbf{ZDT Test Suite}: \url{https://github.com/EMI-Group/evox/blob/main/src/evox/problems/numerical/zdt.py}

\item \textbf{Brax-Based Robotic Control Tasks}: \url{https://github.com/google/brax}
\end{itemize}

\section{Abbreviations} \label{app:su-abbreviations}
Table \ref{table:abbreviations} provides a list of the abbreviations employed throughout this paper, serving as a reference to maintain clarity and consistency in the presentation of terms and concepts.

\begin{table}[H]
\centering
\caption{List of abbreviations used in this article}
\label{table:abbreviations}
\begin{tabular}{lp{0.8\textwidth}}
\toprule
\textbf{Abbreviations} & \textbf{Descriptions} \\
\midrule
EA & Evolutionary algorithm. \\

Popsize & Population size of an EA.\\
FE & Fitness evaluation of evolutionary algorithm. \\
$\mathcal{P}_e$ & The proposed evaluation metric that comprehensively considers power consumption and fitness evaluations. \\
$\mathcal{P}_t$ & A simplified version of $\mathcal{P}_e$ that replaces power consumption with the algorithm's runtime.\\
$\mathcal{O}_{norm}$ & The normalized optimization performance achieved by algorithms.\\
$\mathcal{S}$ & State space of robot control tasks, encapsulates all possible configurations and statuses that the robotic system can
attain. \\
$\mathcal{A}$ & Action space of robot control tasks, comprises the set of all actionable controls or decisions that the robot can execute to transition from one state to another within the environment. \\
ANOVA & Analysis of variance. \\

\midrule
PSO \cite{Kennedy1995} & Particle swarm optimization \\
CSO \cite{Cheng2015} &  Competitive Swarm Optimizer \\
DE \cite{Storn1997} & Differential Evolution \\
SaDE \cite{Qin2009} & Differential Evolution with Strategy adaptation \\
CMA-ES \cite{Hansen2003} &  Evolution Strategy with Covariance Matrix Adaptation \\
IPOP-CMA-ES \cite{Auger2005a} &  A Restart CMA Evolution Strategy With Increasing Population Size \\
GA-UR/GM \cite{Syswerda1989} &  Genetic algorithm with uniform random crossover and gaussian mutation \\
GA-SBX/PM \cite{Goldberg1987}& Genetic algorithm with simulated binary crossover and polynomial mutation\\
NSGA-II \cite{Deb2002} & Non-dominated Sorting Genetic Algorithm II\\
NSGA-III \cite{Deb2014} &  Non-dominated Sorting Genetic Algorithm III \\
RVEA \cite{Cheng2016} & Reference Vector Guided Evolutionary Algorithm \\
MOEA/D \cite{Zhang2007} &  Multi-objective Evolutionary Algorithm Based on Decomposition \\
HypE \cite{Bader2011} & Hypervolume Estimation Algorithm for Multi-objective Optimization  \\
LMOCSO \cite{Tian2020} & Efficient Large-Scale Multiobjective Optimization Based on Competitive Swarm Optimizer \\
SPEA2 \cite{Kim2004} & Strength Pareto Evolutionary Algorithm II \\
IBEA \cite{Zitzler2004} & Indicator-Based Evolutionary Algorithm \\

\bottomrule
\end{tabular}
\end{table}

\section{Neuroevolution Benchmarks}
\label{app:neuroevolution}
\subsubsection{Robot Control Task based on Brax Engine}
Brax provides a collection of robotic control environments that are widely used in reinforcement learning (RL) and evolutionary algorithm research. These environments simulate physically realistic dynamics and serve as benchmarks for evaluating various control strategies. Each task involves controlling an articulated agent with continuous actions in a simulated 3D environment.

\begin{figure}[H]
    \centering
    \begin{minipage}[b]{0.24\textwidth}
        \centering
        \includegraphics[width=\textwidth]{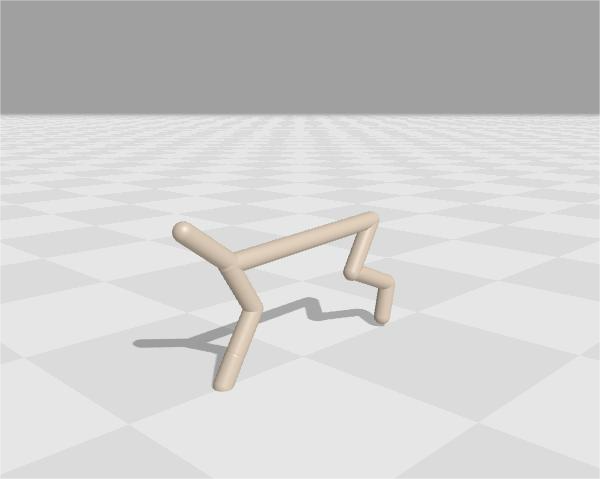}
        \centering
        \subcaption{Halfcheetah}
    \end{minipage}
    \begin{minipage}[b]{0.24\textwidth}
        \centering
        \includegraphics[width=\textwidth]{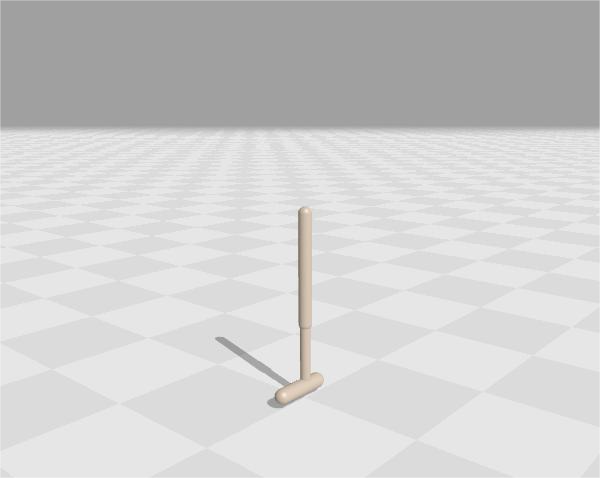}
        \centering
        \subcaption{Hopper}
    \end{minipage}

  \begin{minipage}[b]{0.24\textwidth}
        \centering
        \includegraphics[width=\textwidth]{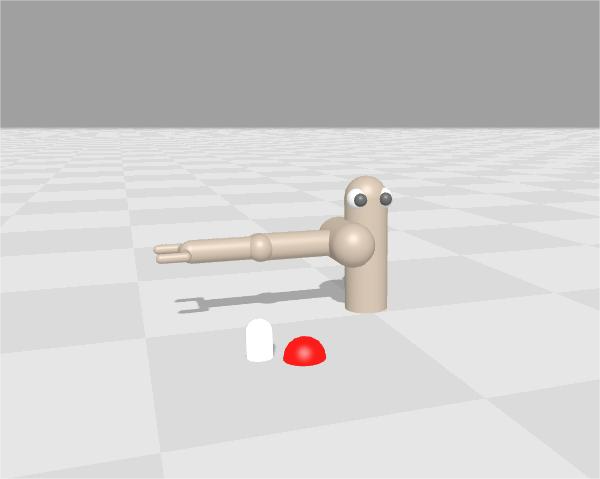}
        \centering
        \subcaption{Pusher}
    \end{minipage}
    \begin{minipage}[b]{0.24\textwidth}
        \centering
        \includegraphics[width=\textwidth]{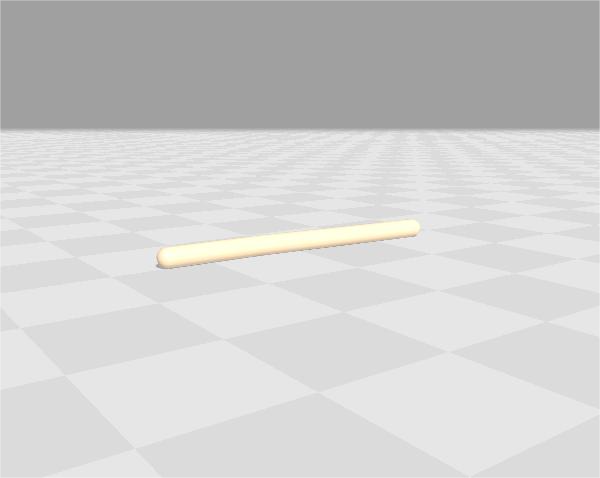}
        \centering
        \subcaption{Swimmer}
    \end{minipage}

    \caption{Robotic control tasks based on Brax engine.}
    \label{fig:s-brax}
\end{figure}

Fig.~\ref{fig:s-brax} illustrates several robot control environments based on the Brax engine. These environments leverage a differentiable physics engine built on JAX, which allows for highly efficient parallel simulation across hardware accelerators such as GPUs and TPUs. This architecture enables fast gradient-based learning and facilitates large-scale experimentation. Furthermore, Brax is compatible with various Reinforcement Learning (RL) algorithms, including Proximal Policy Optimization (PPO), Augmented Random Search (ARS), and Evolutionary Strategies (ES). Consequently, these environments serve as versatile benchmarks for both policy gradient methods and evolutionary strategies.

\subsubsection{Multi-objective Neuroevolution}
In the Brax environment, robot control tasks are conventionally defined as single-objective optimization problems. To align with the objectives of our experimental analysis, we reformulate five selected environments into multi-objective optimization tasks, thereby enhancing their suitability for comprehensive performance evaluation. The specific descriptions of the reformulated multi-objective tasks are as follows:

\begin{itemize}
\item \textbf{MoHopper-m2:} The observation and action spaces are defined as \( S \in \mathbb{R}^{11} \) and \( A \in \mathbb{R}^{3} \), respectively. Each episode comprises 1000 steps. The first objective is to maximize the forward reward, given by:\begin{equation}
f_1 = w_1 \cdot v_x - \sum_i a_i^2 + C,
\end{equation}where \( v_x \) denotes the velocity along the x-axis, \( w_1 \) represents the corresponding weight, \( a_i \) denotes the action of each actuator, and \( C = 1 \) is a constant survival reward that ensures the agent remains active.

The second objective is to maintain or increase the height of the hopper, which is formulated as:\begin{equation}
f_2 = 10 \cdot (h_{\text{curr}} - h_{\text{init}}) - \sum_i a_i^2 + C,
\end{equation}where \( h_{\text{curr}} \) denotes the current height of the hopper, and \( h_{\text{init}} \) represents the initial height. The term \( \sum_i a_i^2 \) penalizes excessive actuator actions, while \( C \) corresponds to the survival reward.

\item \textbf{MoHopper-m3:} The observation and action spaces are \( S \in \mathbb{R}^{11} \) and \( \mathcal{A} \in \mathbb{R}^3 \), respectively. Each episode spans 1000 steps. The first objective is the forward reward, defined as:\begin{equation}
    f_1 = w_1 \cdot v_x + C,
\end{equation}where \( v_x \) denotes the velocity in the \( x \)-direction, \( w_1 \) represents the corresponding weight, and \( C = 1 \) serves as a survival reward, thereby indicating the agent's continued activity. The second objective, i.e., height, is given by:\begin{equation}
    f_2 = 10 \cdot (h_{\text{curr}} - h_{\text{init}}) + C,
\end{equation}where \( h_{\text{curr}} \) and \( h_{\text{init}} \) denote the current and initial heights of the hopper, respectively. The third objective is the control cost, which is expressed as:\begin{equation}
    f_3 = -\sum_i a_i^2 + C,
\end{equation}where $a_{i}$ denotes the action of the corresponding actuator.

\item \textbf{MoReacher:} The observation and action spaces are defined as $\mathcal{S} \in \mathbb{R}^{11}$ and $\mathcal{A} \in \mathbb{R}^{2}$, respectively. Each episode consists of 1000 steps. The first objective is the distance reward:\begin{equation}
    f_1 = -d,
\end{equation}where $d$ denotes the Euclidean distance between the fingertip of the Reacher and the target. The second objective represents the control cost:\begin{equation}
    f_2 = - \sum_i a_i^2,
\end{equation}where $a_i$ denotes the action of each actuator.

\item \textbf{MoSwimmer:} The observation and action spaces are defined as $\mathcal{S} \in \mathbb{R}^{8}$ and $\mathcal{A} \in \mathbb{R}^{2}$, respectively. Each episode consists of 1000 steps. The first objective is the forward reward:\begin{equation}
    f_1 = w_1 \cdot v_x,
\end{equation}where $v_x$ denotes the velocity in the $x$-direction and $w_1$ represents the weight assigned to the velocity. The second objective is the control cost:\begin{equation}
    f_2 = -w_2 \cdot \sum_i a_i^2,
\end{equation}where $a_i$ denotes the action of the $i$-th actuator and $w_2$ represents the weight of the control cost.

\item \textbf{MoWalker2d:} The observation space and action space are defined as $\mathcal{S} \in \mathbb{R}^{17}$ and $\mathcal{A} \in \mathbb{R}^{6}$, respectively. Each episode consists of 1000 steps. The first objective is the forward reward:\begin{equation}
    f_1 = w_1 \cdot v_x,
\end{equation}where $v_x$ denotes the velocity along the $x$-axis and $w_1$ represents the velocity weight. The second objective is the control cost:\begin{equation}
    f_2 = -w_2 \cdot \sum_i a_i^2,
\end{equation}where $a_i$ denotes the action of each actuator, and $w_2$ represents the weight for the control cost.
\end{itemize}

\section{Experiments Setups}\label{app:experiments}

\subsection{Algorithm Parameters Settings}

Table \ref{table:s-parameters} details the parameter settings for the sixteen EAs evaluated in the experiments.
\begin{table}[H]
\centering
\caption{Parameter settings of the tested algorithms.}
\label{table:s-parameters}
\begin{tabular}{ll}
\toprule
\textbf{Algorithm} & \textbf{Key Parameter Settings} \\ \midrule
PSO & inertia weight = 0.6, cognitive coefficient = 2.5, social coefficient = 0.8 \\
CSO & phi = 0 \\
DE & random selection, differential weight = 0.5, cross probability = 0.9, batch size = population size \\
SaDE &  number of differential vectors = 9, learning period = 50 \\
CMA-ES & cm = 1, recombination weights = None\\
IPOP-CMA-ES & cm = 1, stagnation threshold = 50, recombination weights = None\\
GA-UR/GM & gaussian mutation, uniform random crossover  \\
GA-SBX/PM & polynomial mutation, simulated binary crossover  \\
NSGA-II &  uniform random selection, polynomial mutation, simulated binary crossover\\
NSGA-III & uniform random selection, polynomial mutation, simulated binary crossover \\
RVEA  & alpha = 2, fr = 0.1, reference vector guided selection, polynomial mutation, simulated binary crossover \\
MOEA/D  & penalty-based boundary intersection, polynomial mutation, simulated binary crossover  \\
HypE &  number of samples = 10000, polynomial mutation, simulated binary crossover\\
LMOCSO &  alpha = 2,  reference vector guided selection, polynomial mutation\\
SPEA2 & polynomial mutation, simulated binary crossover \\
IBEA & kappa = 0.05, polynomial mutation, simulated binary crossover\\
\bottomrule
\end{tabular}
\end{table}

\subsection{Benchmarking Function Settings}
Table~\ref{tab:s-numerical} summarizes the configuration of the benchmark functions used in our numerical optimization tasks, including their dimensionality, search space, and minimum. These functions are widely adopted in evolutionary computation research and span a range of landscapes from unimodal to highly multimodal.

\begin{table}[H]
\centering
\caption{Numerical Benchmark functions used in the experiment}
\begin{tabular}{l l c c c}
\toprule
\textbf{Functions} & \textbf{Name} & \textbf{Dimension} & \textbf{Search space} & \textbf{Minimum} \\
\midrule
$f_{a_{1-5}}$ & CEC2022 F1-F5 & 20 & $[-100, 100]$    & 0 \\
$f_{a_{6}}$ & Ackley   & 50 & $[-32, 32]$      & 0 \\
$f_{a_{7}}$ & Griewank        & 50 & $[-600, 600]$      & 0 \\
$f_{a_{8}}$ & Rosenbrock     & 50 & $[-5, 10]$  & 0 \\
$f_{a_{9}}$ & Schwefel  & 50 & $[-500, 500]$      & 0 \\
$f_{a_{10}}$ & Sphere & 50 & $[-5.12, 5.12]$    & 0 \\
$f_{a_{11-17}}$ & DTLZ 1-7  & 50 & $[0, 1]$    & 0 \\
$f_{a_{18-20}}$ & ZDT 1-3   & 50 & $[0, 1]$    & 0 \\
\bottomrule
\end{tabular}
\label{tab:s-numerical}
\end{table}

All tasks employed a standardized policy architecture and evaluation protocol. Specifically, the policy was implemented as a multilayer perceptron (MLP) using Flax, which comprised a single hidden layer with 16 tanh-activated neurons. While this internal structure remained fixed, the input and output dimensions were adjusted to align with the observation and action spaces of each environment. Regarding optimization, the model parameters were constrained within a fixed range of $[-8,8]$ per dimension. Furthermore, each task was evaluated in a Brax environment, subject to a maximum of 1000 simulation steps per episode. Table~\ref{tab:s-neuro-config} summarizes these settings for all neuroevolution tasks.

\begin{table}[H]
\centering
\captionsetup{justification=centering}
\caption{Neuroevolution of Robotic control tasks}
\label{tab:s-neuro-config}

\begin{tabular}{cc c ccc}
\toprule
\multicolumn{2}{c}{\textbf{Single-objective}} & & \multicolumn{3}{c}{\textbf{Multi-objective} } \\

\cmidrule(lr){1-2} \cmidrule(lr){4-6}

$f_{b_{1}}$-$f_{b_{5}}$ & d  & & $f_{b_{6}}$-$f_{b_{10}}$ & m & d  \\

\midrule

Halfcheetah & 390  &  & MoHopper-m2 & 2 & 243 \\
Hopper & 243  &  & MoHopper-m3 & 3 & 243  \\
Pusher & 503  & & MoReacher & 2 & 226 \\
Reacher & 226  & & MoSwimmer & 2 & 178 \\
Swimmer & 178  & & MoWalker2d & 2 & 390 \\

\bottomrule
\end{tabular}
\end{table}

\section{Experimental Results}
\subsection{Time-capped Performance}\label{app:sensitivity}

We evaluate the performance of various EAs under a fixed 30-second time constraint by comparing the number of FEs completed and the quality of the final solutions. All experiments were conducted using an NVIDIA GeForce RTX 3090 GPU, with GPU-based results depicted by solid markers and CPU-based results by hollow markers.
In these plots, a horizontal shift from a hollow to a solid marker indicates improved optimization performance (lower objective or IGD value), while a vertical shift reflects increased computational efficiency (more NFEs completed). Consistent with previous findings, most EAs exhibit notable efficiency improvements when executed on the GPU. In some cases, GPU-based implementations even benefit from both dimensions, achieving superior solution quality as a result of significantly more evaluations within the same time window.

\begin{figure}[H]
    \centering
    \includegraphics[width=0.7\textwidth]{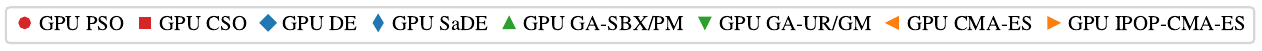}

    \vspace{0.2cm}

    \begin{minipage}[b]{0.235\textwidth}
        \centering
        \includegraphics[width=\textwidth]{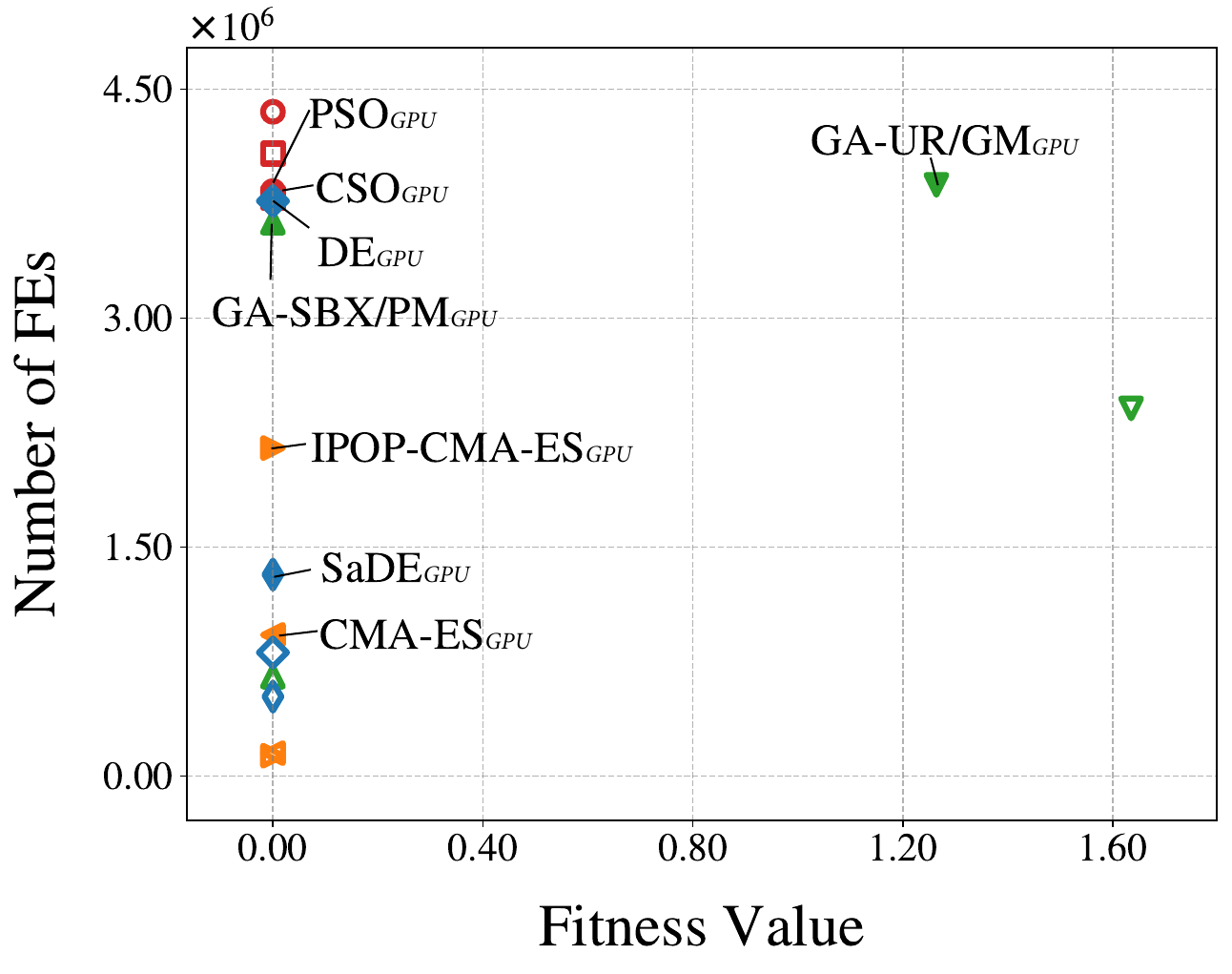}
        \centering
        \vspace{-0.5cm}
        \subcaption{ $f_{a_{1}}$: CEC2022 F1}
    \end{minipage}
   \hfill
    \begin{minipage}[b]{0.235\textwidth}
        \centering
        \includegraphics[width=\textwidth]{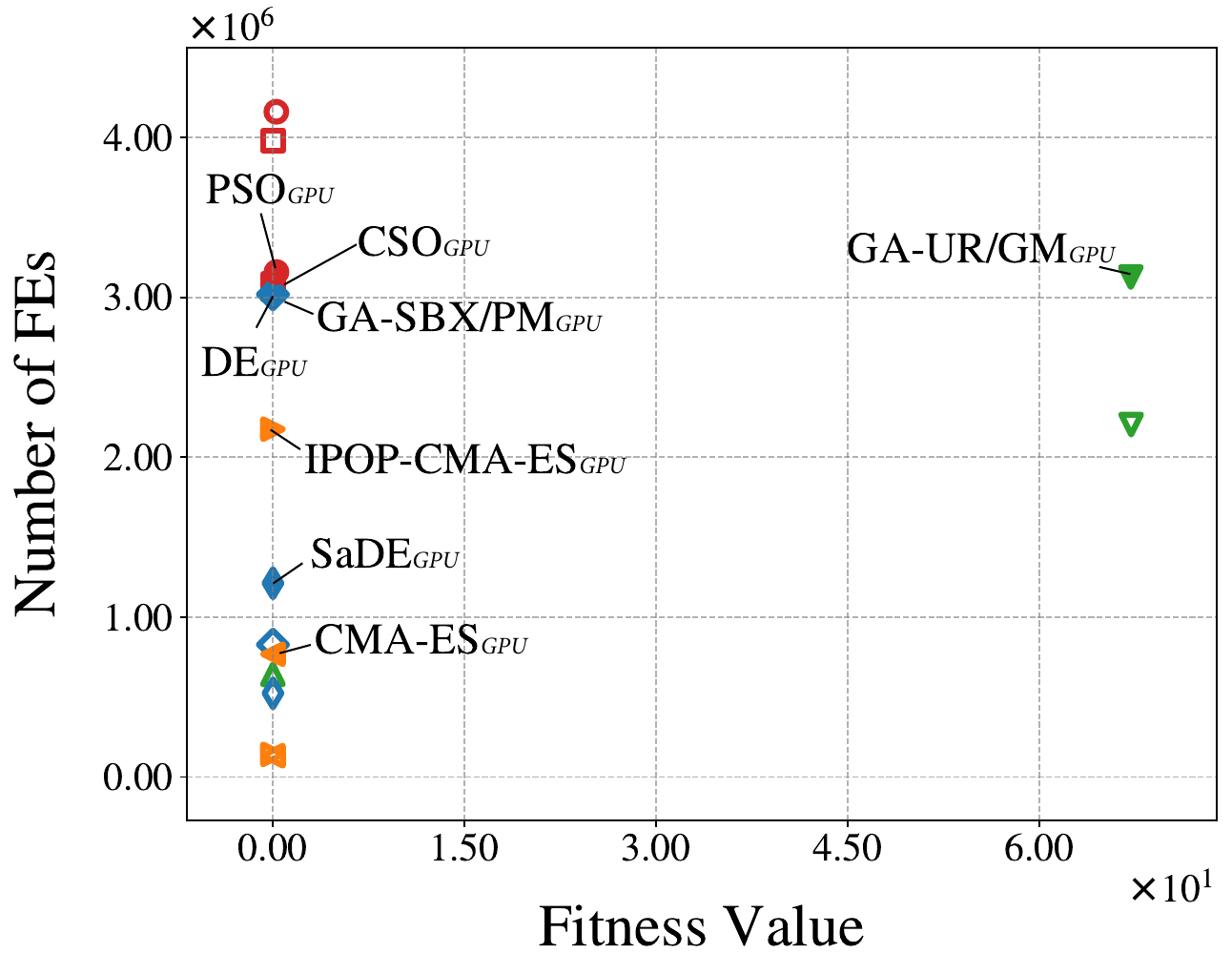}
        \centering
        \vspace{-0.5cm}
        \subcaption{ $f_{a_{3}}$: CEC2022 F3}
    \end{minipage}
    \hfill
    \begin{minipage}[b]{0.235\textwidth}
        \centering
        \includegraphics[width=\textwidth]{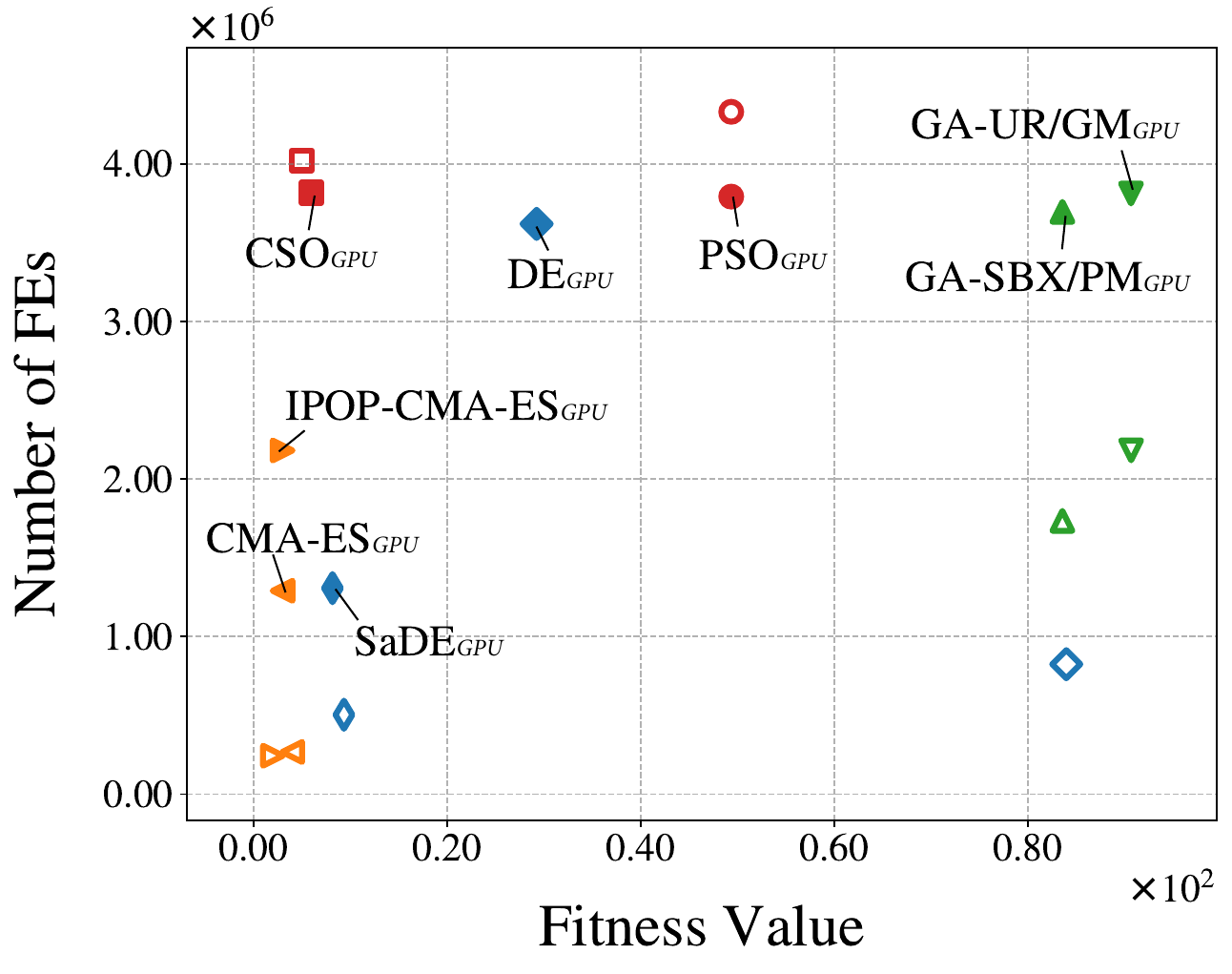}
        \centering
        \vspace{-0.5cm}
        \subcaption{ $f_{a_{4}}$: CEC2022 F4}
    \end{minipage}
   \hfill
    \begin{minipage}[b]{0.235\textwidth}
        \centering
        \includegraphics[width=\textwidth]{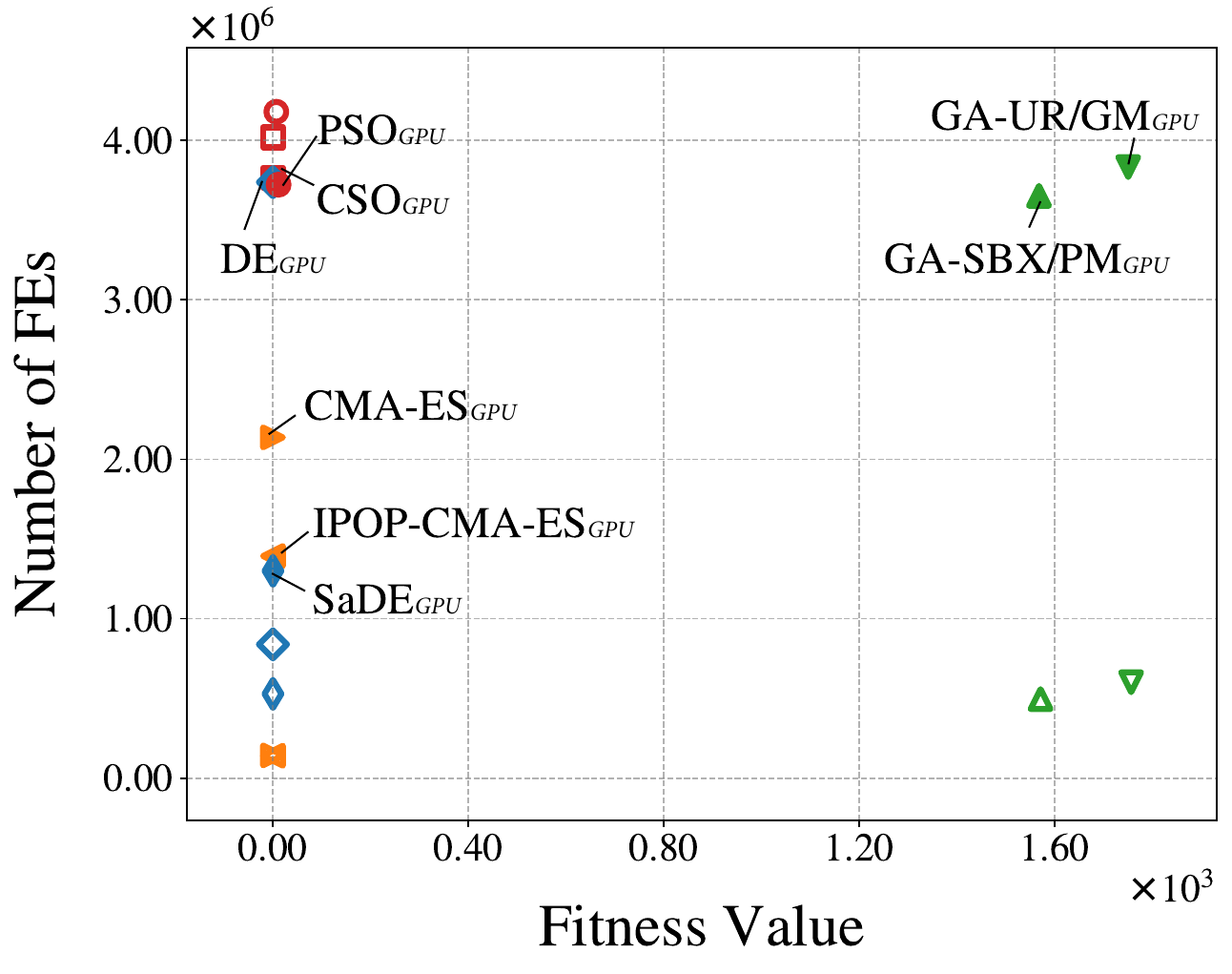}
        \centering
        \vspace{-0.5cm}
        \subcaption{ $f_{a_{5}}$: CEC2022 F5}
    \end{minipage}
    \hfill

    \vspace{0.4cm}

    \begin{minipage}[b]{0.235\textwidth}
        \centering
        \includegraphics[width=\textwidth]{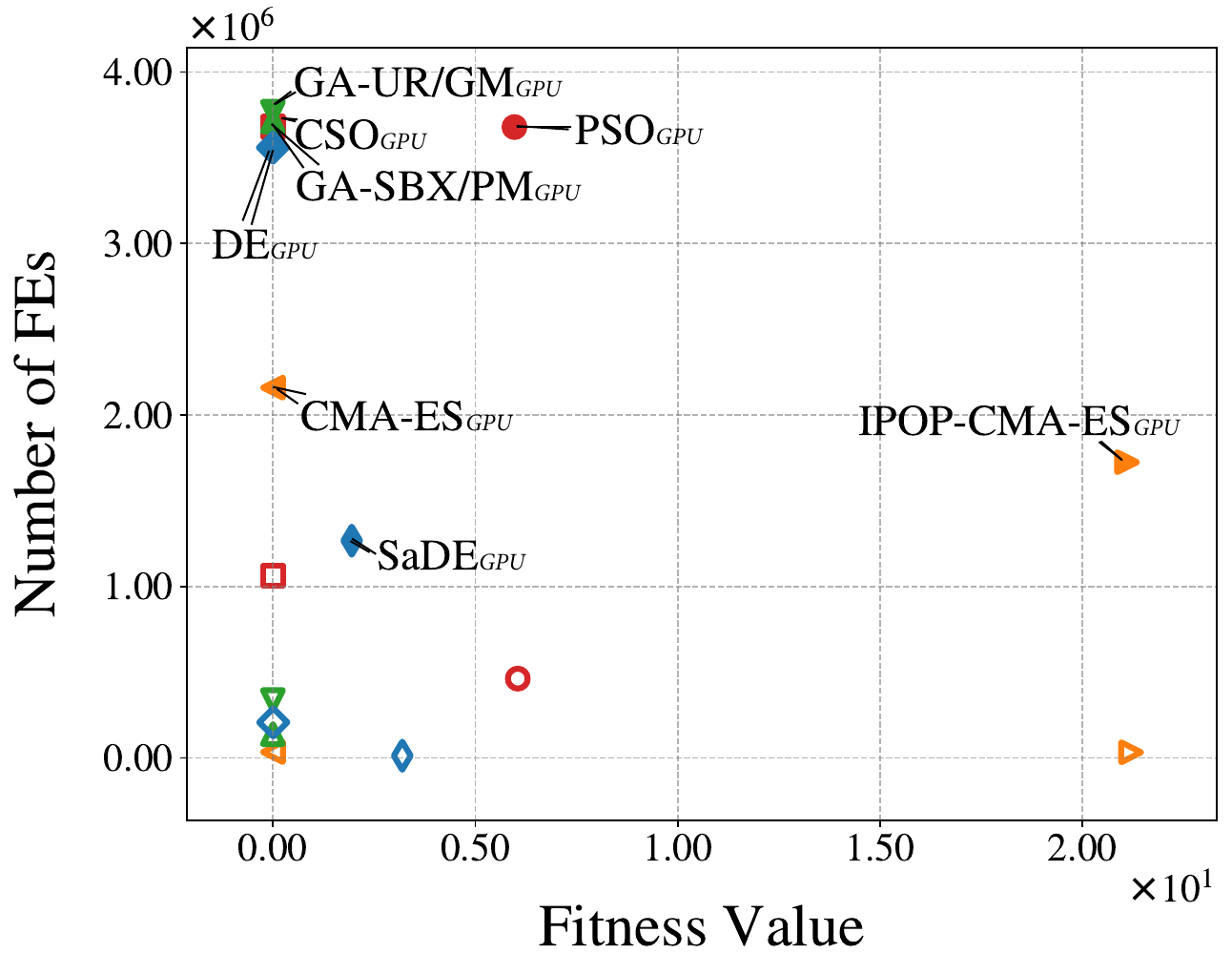}
        \centering
        \vspace{-0.5cm}
        \subcaption{ $f_{a_{6}}$: Ackley}
    \end{minipage}
   \hfill
    \begin{minipage}[b]{0.235\textwidth}
        \centering
        \includegraphics[width=\textwidth]{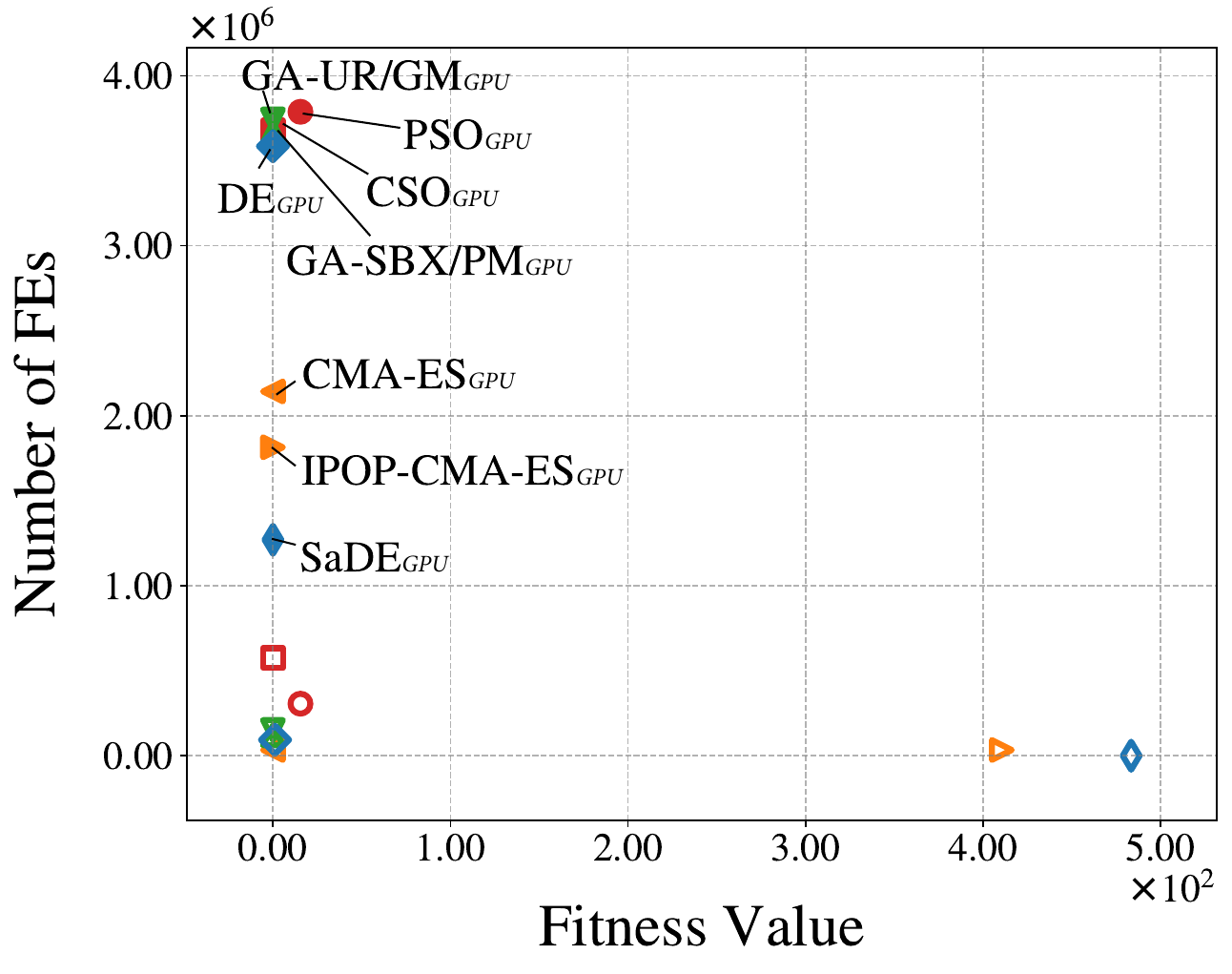}
        \centering
        \vspace{-0.5cm}
        \subcaption{ $f_{a_{7}}$: Griewank}
    \end{minipage}
    \hfill
    \begin{minipage}[b]{0.235\textwidth}
        \centering
        \includegraphics[width=\textwidth]{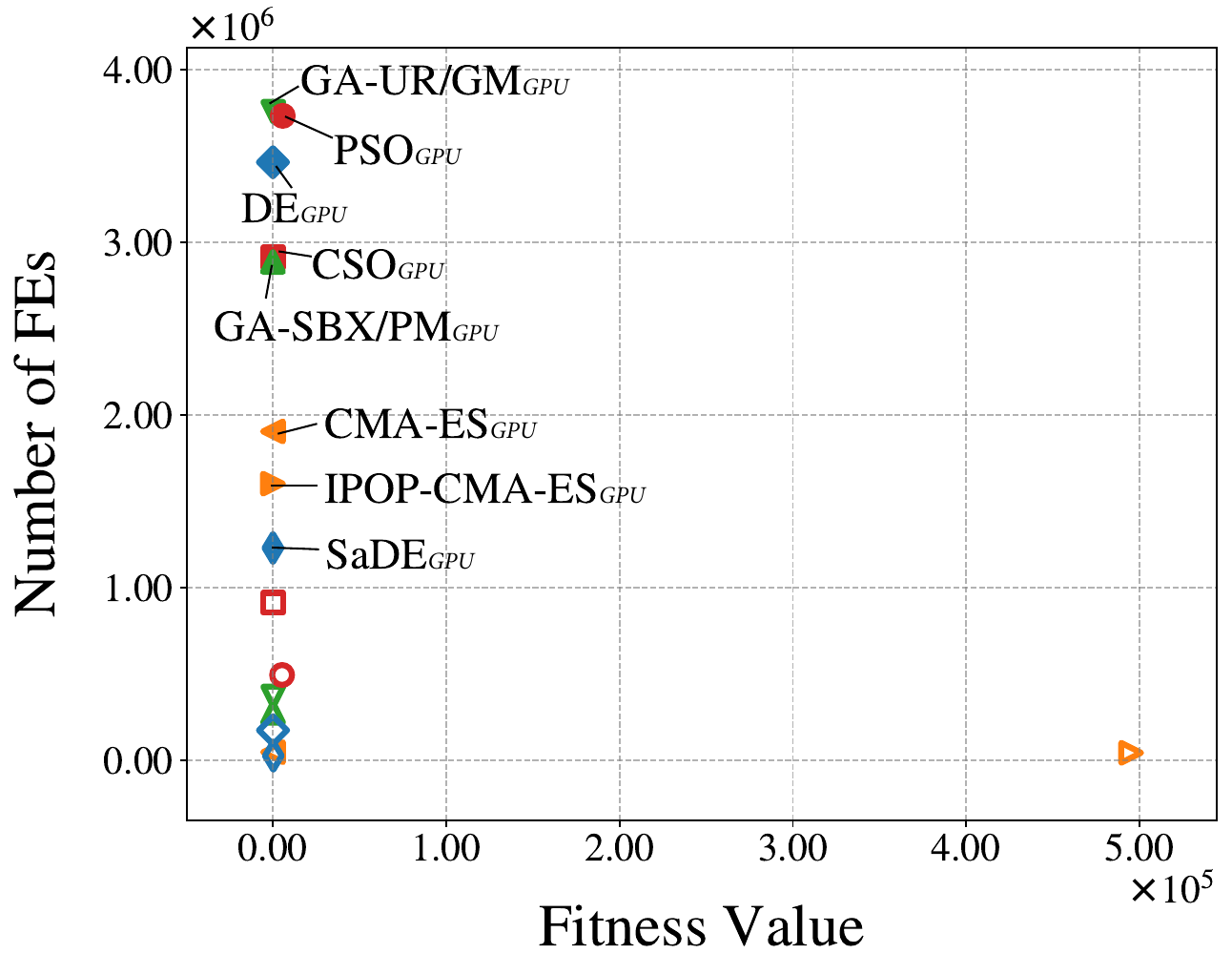}
        \centering
        \vspace{-0.5cm}
        \subcaption{ $f_{a_{8}}$: Rosenbrock}
    \end{minipage}
   \hfill
    \begin{minipage}[b]{0.235\textwidth}
        \centering
        \includegraphics[width=\textwidth]{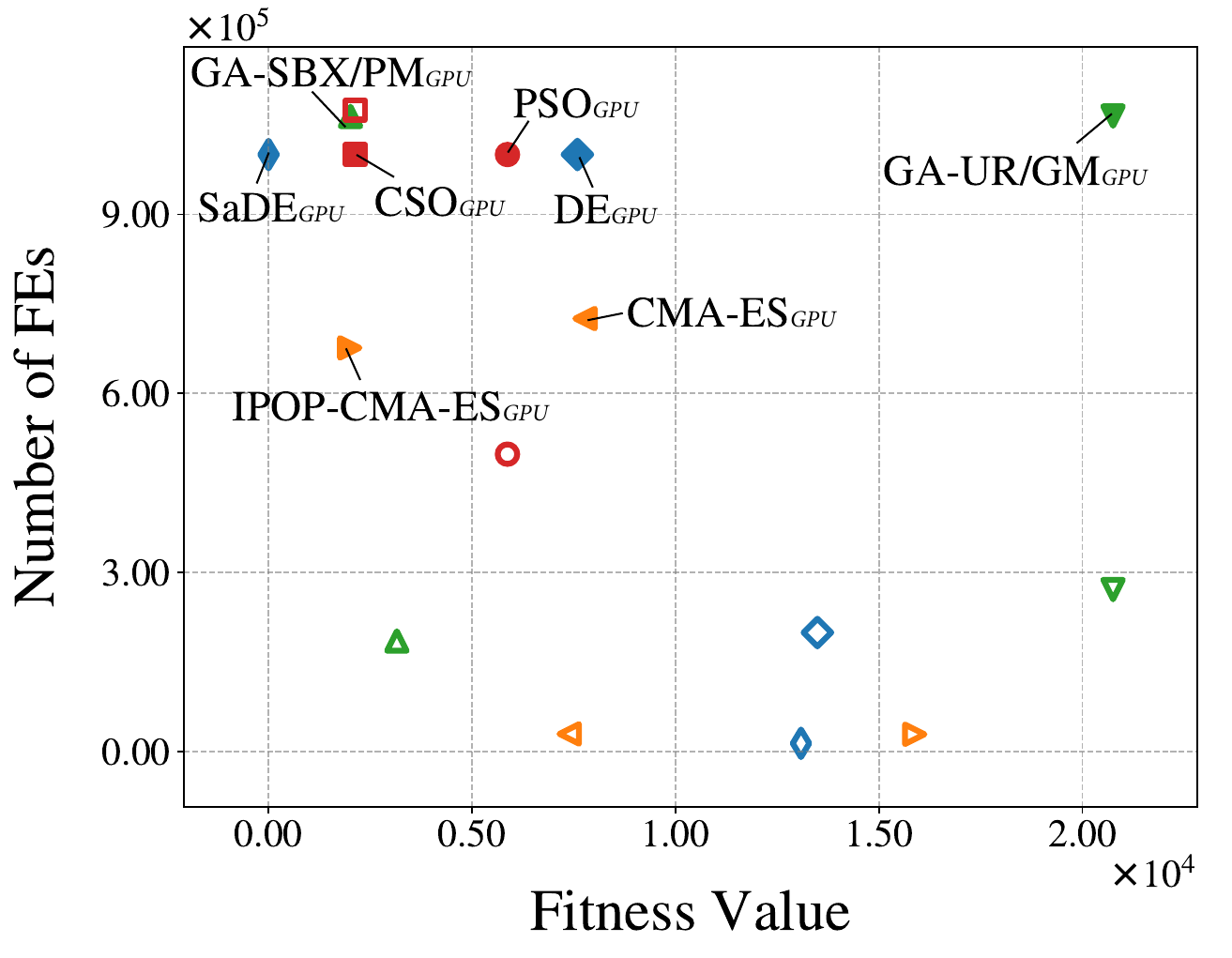}
        \centering
        \vspace{-0.5cm}
        \subcaption{ $f_{a_{9}}$: Schwefel}
    \end{minipage}
    \hfill

    \vspace{0.2cm}

    \includegraphics[width=0.7\textwidth]{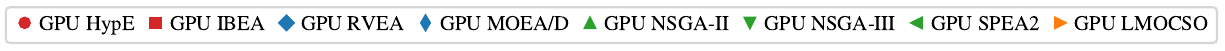}

    \vspace{0.2cm}

    \begin{minipage}[b]{0.235\textwidth}
        \centering
        \includegraphics[width=\textwidth]{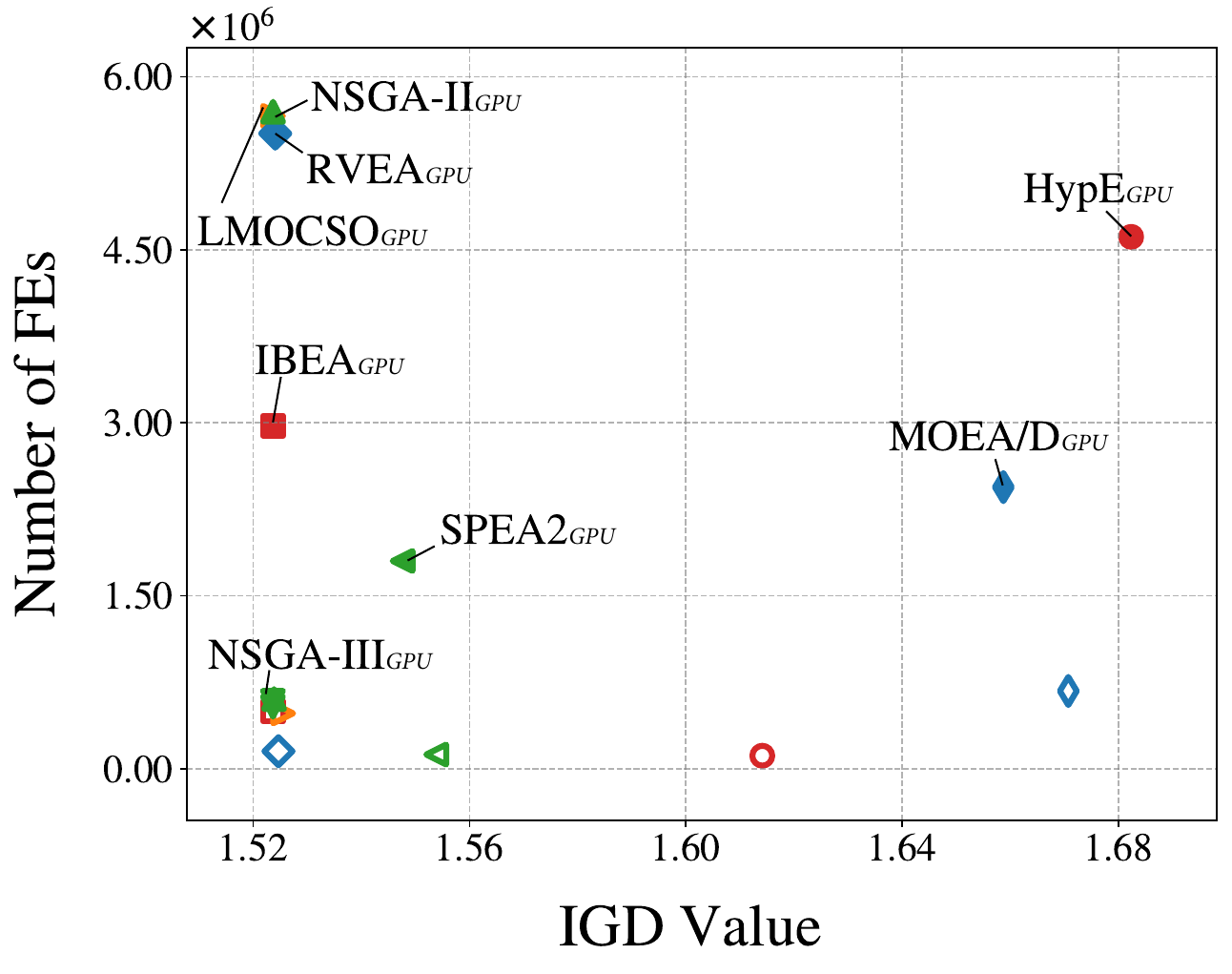}
        \centering
        \vspace{-0.5cm}
        \subcaption{ $f_{a_{12}}$: DTLZ2}
    \end{minipage}
   \hfill
    \begin{minipage}[b]{0.235\textwidth}
        \centering
        \includegraphics[width=\textwidth]{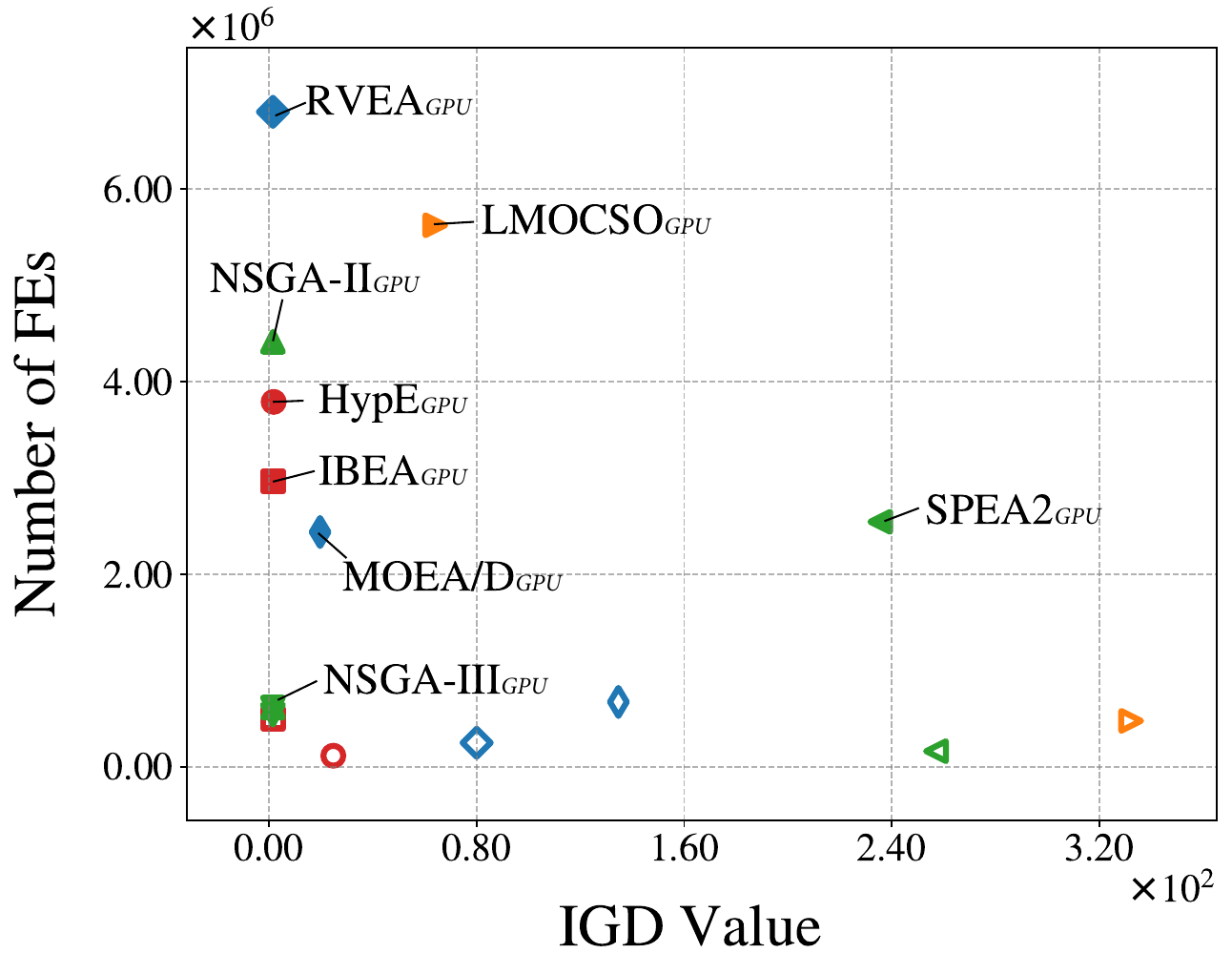}
        \centering
        \vspace{-0.5cm}
        \subcaption{ $f_{a_{13}}$: DTLZ3}
    \end{minipage}
    \hfill
    \begin{minipage}[b]{0.235\textwidth}
        \centering
        \includegraphics[width=\textwidth]{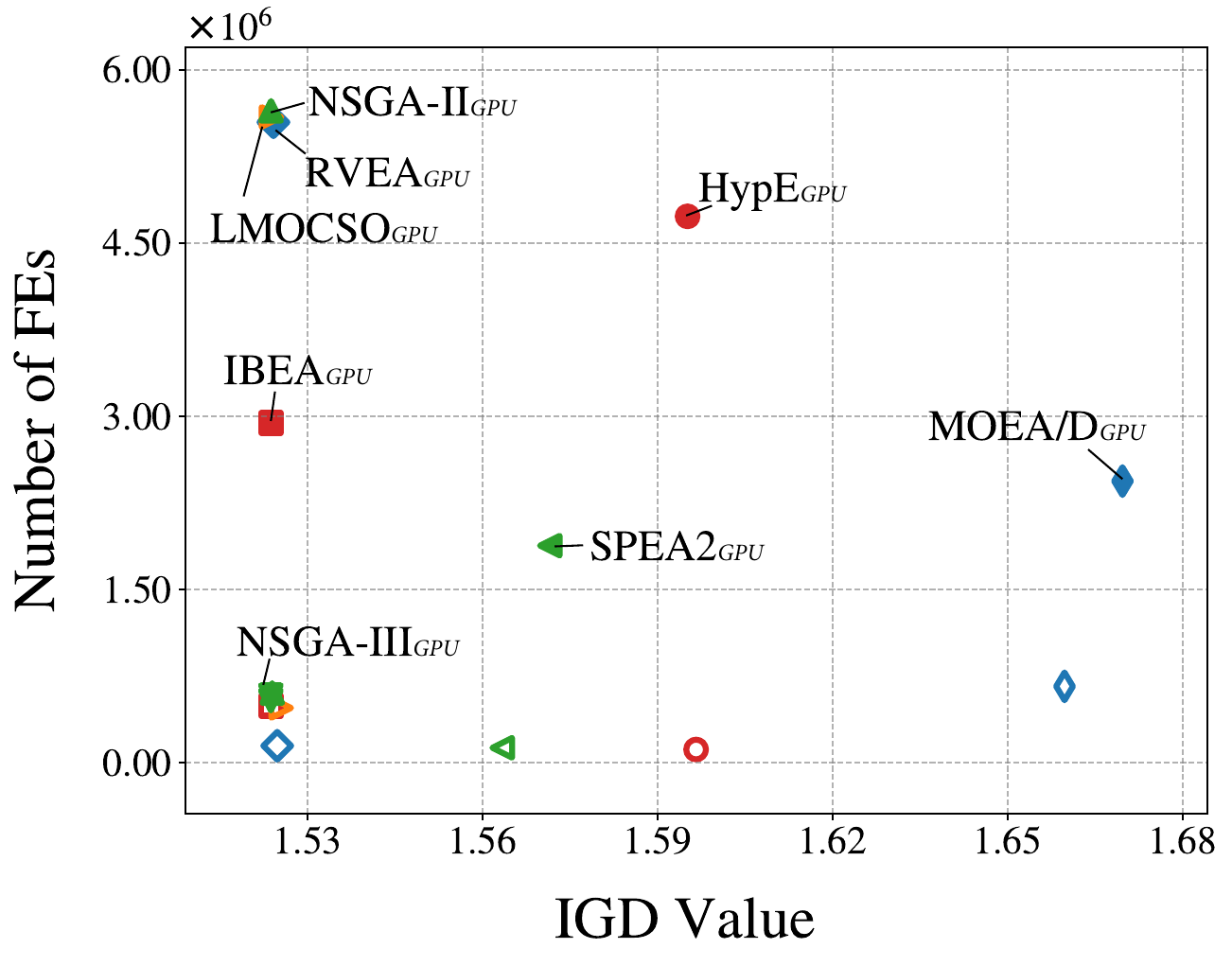}
        \centering
        \vspace{-0.5cm}
        \subcaption{ $f_{a_{14}}$: DTLZ4}
    \end{minipage}
   \hfill
    \begin{minipage}[b]{0.235\textwidth}
        \centering
        \includegraphics[width=\textwidth]{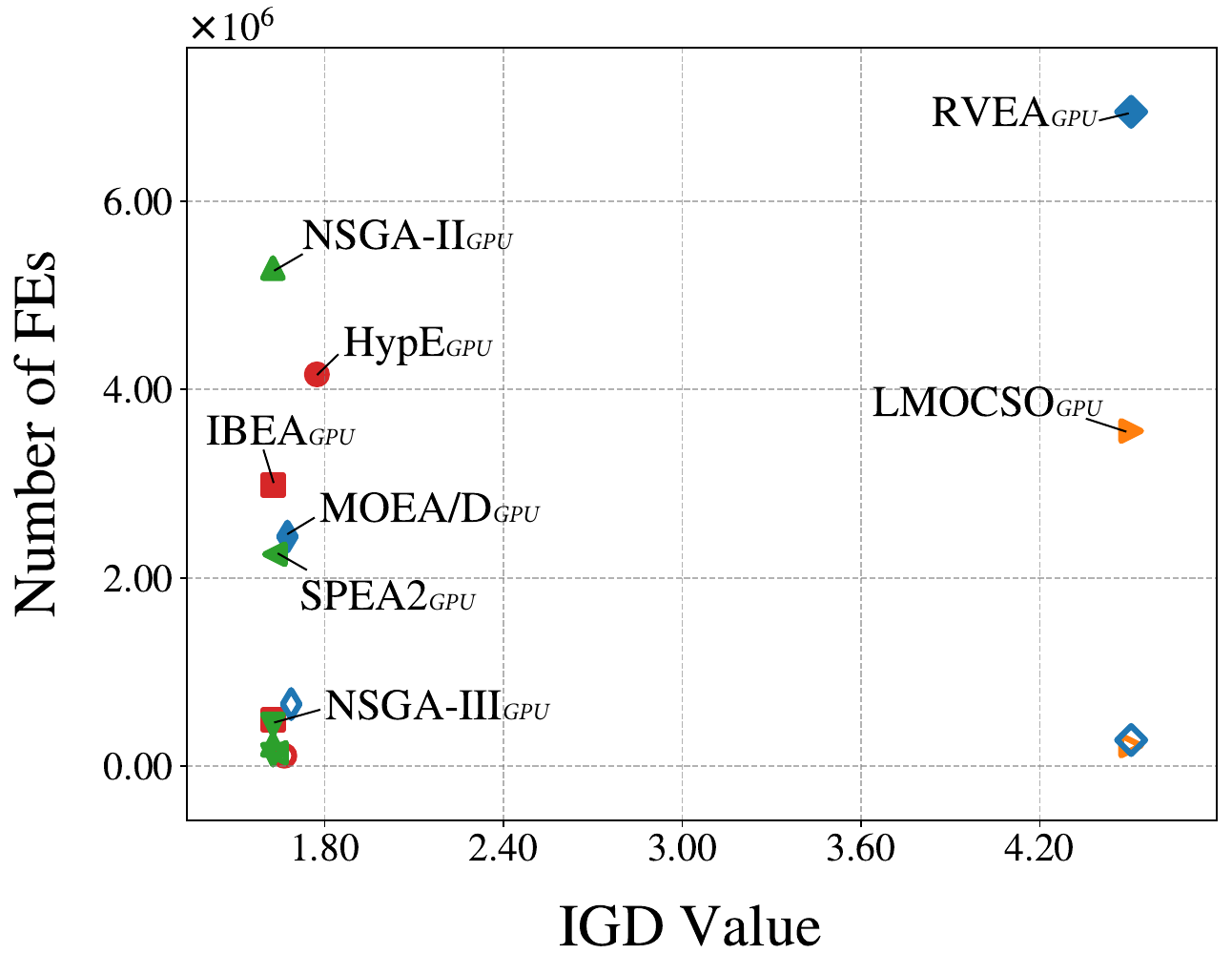}
        \centering
        \vspace{-0.5cm}
        \subcaption{ $f_{a_{15}}$: DTLZ5}
    \end{minipage}
    \vspace{0.2cm}

    \begin{minipage}[b]{0.235\textwidth}
        \centering
        \includegraphics[width=\textwidth]{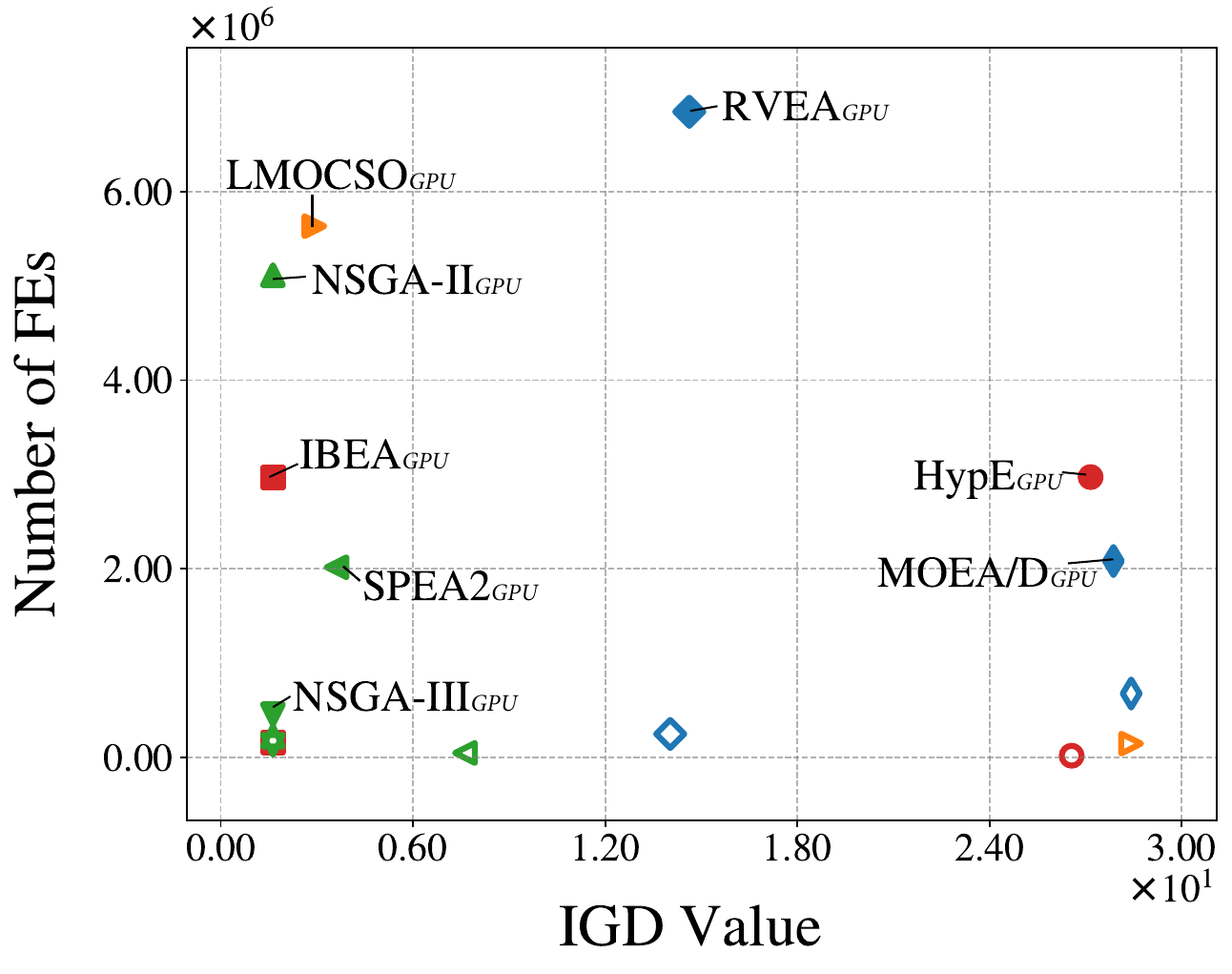}
        \centering
        \vspace{-0.5cm}
        \subcaption{ $f_{a_{16}}$: DTLZ6}
    \end{minipage}
   \hfill
    \begin{minipage}[b]{0.235\textwidth}
        \centering
        \includegraphics[width=\textwidth]{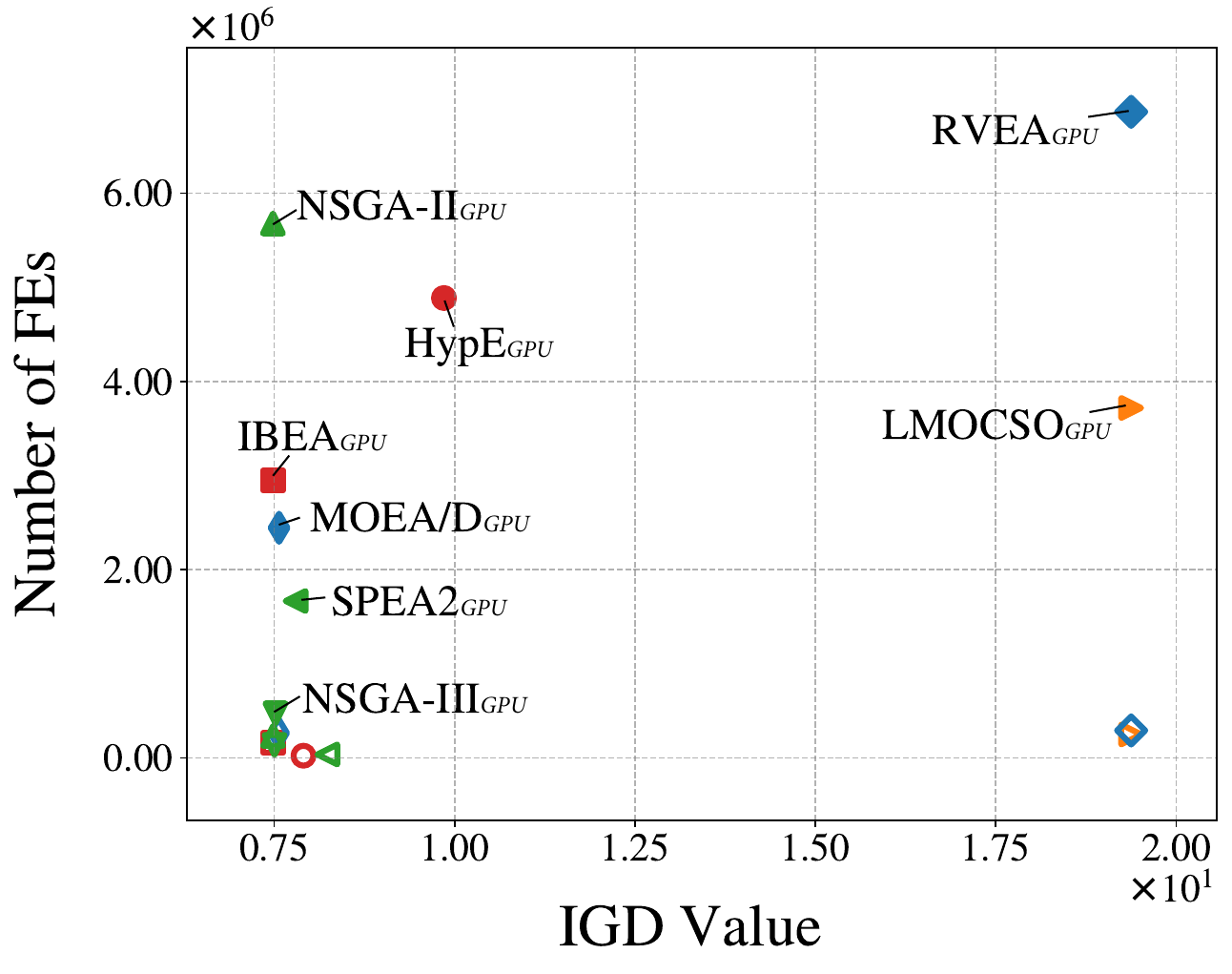}
        \centering
        \vspace{-0.5cm}
        \subcaption{ $f_{a_{17}}$: DTLZ7}
    \end{minipage}
    \hfill
    \begin{minipage}[b]{0.235\textwidth}
        \centering
        \includegraphics[width=\textwidth]{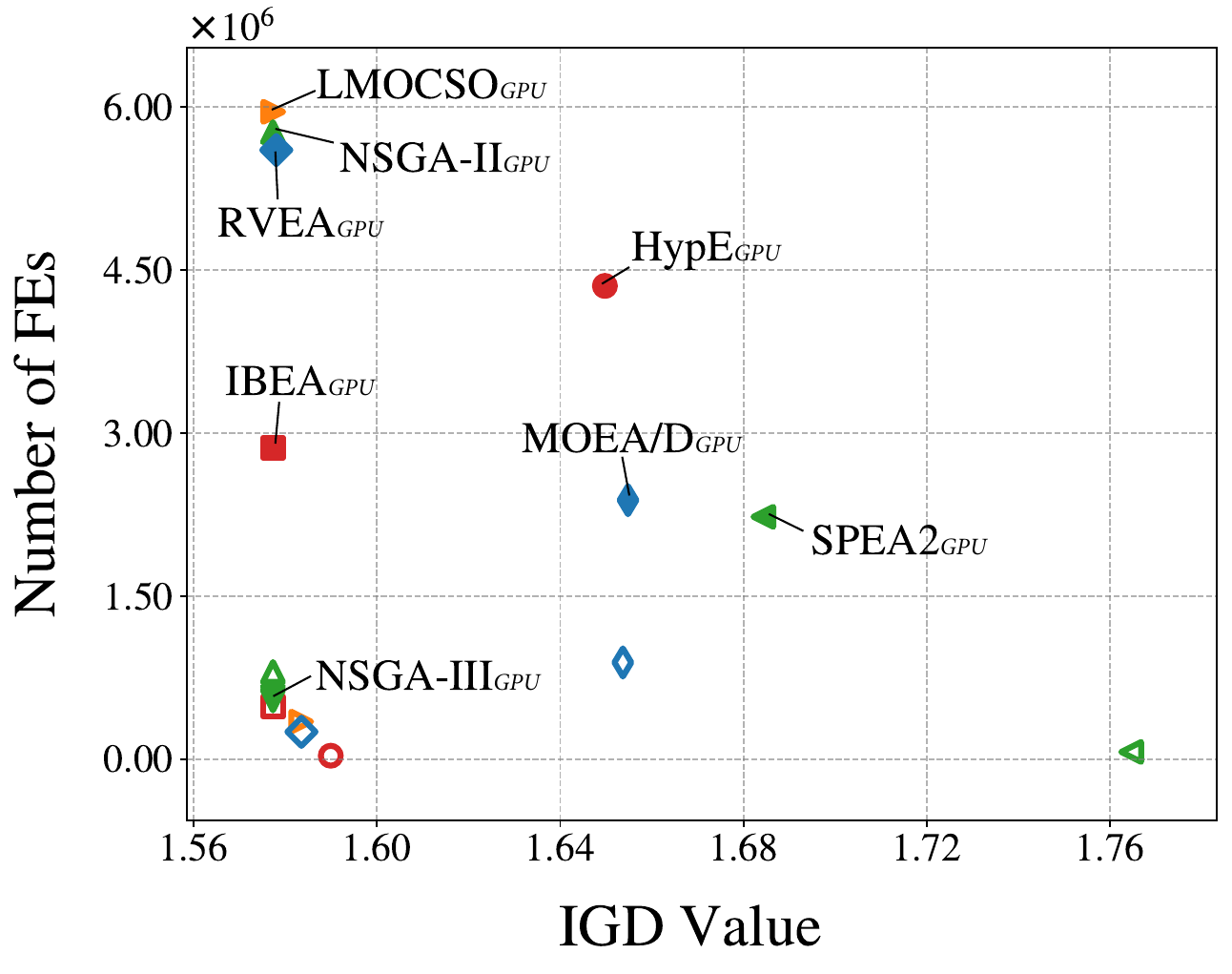}
        \centering
        \vspace{-0.5cm}
        \subcaption{ $f_{a_{19}}$: ZDT2}
    \end{minipage}
   \hfill
    \begin{minipage}[b]{0.235\textwidth}
        \centering
        \includegraphics[width=\textwidth]{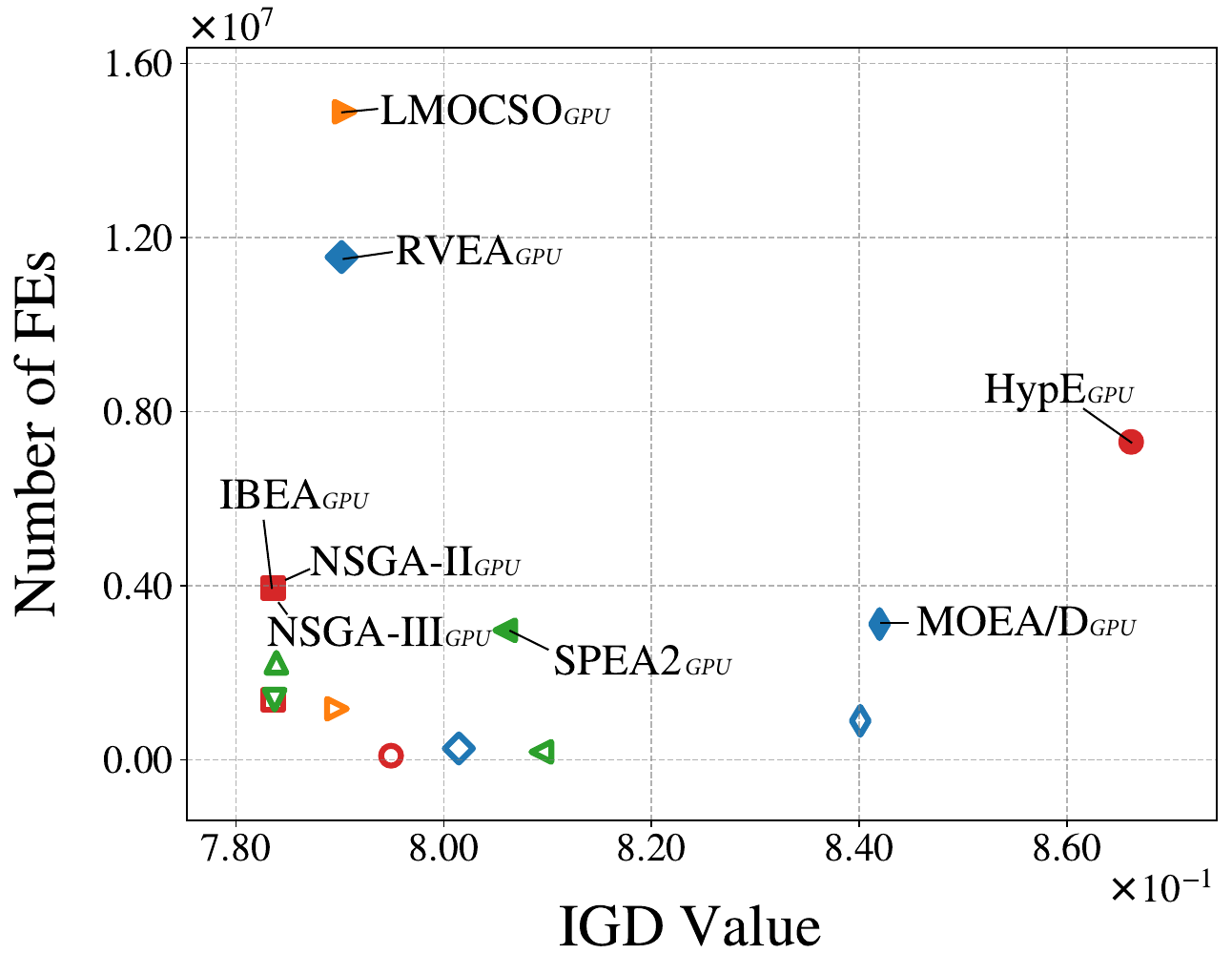}
        \centering
        \vspace{-0.5cm}
        \subcaption{ $f_{a_{20}}$: ZDT3}
    \end{minipage}
    \hfill

    \caption{Performance comparison of EAs tested on CPU and GPU (NVIDIA GeForce RTX-3090), evaluated in terms of solution quality and number of FEs completed within 30 seconds. Lower fitness/IGD values denote better performance. Results represent averaged performance values across 15 independent runs. Solid markers denote GPU-based (NVIDIA GeForce RTX-3090) implementations; hollow markers indicate CPU-based counterparts. Algorithms achieving notable improvements in both efficiency and accuracy on the GPU compared to the CPU are highlighted with connecting lines.}
\end{figure}

\subsection{Fixed-FE Truncation under GPU Execution}
Fig.~\ref{fig:s-truncation-effect} presents the convergence curves of PSO, GA-SBX/PM and GA-UR/GM on 200-dimensional Rastrigin and Schwefel functions. The vertical dashed line marks the wall-clock time at which $10^6$ FEs are completed on an NVIDIA RTX 3090 GPU. The region before this line corresponds to the conventional fixed-FE observation window, while the region after it shows the additional search trajectory made observable by GPU acceleration.

\subsection{Scaling Regimes with Problem Dimension and Population Size}\label{app:scaling}
This section analyzes the computational efficiency of evolutionary algorithms across three hardware platforms, focusing on their scaling behavior under varying experimental parameters. Specifically, we investigate the sensitivity of performance to changes in population size and problem dimensionality, both scaled exponentially from 16 to 8192. This analysis aims to reveal how different architectures respond to increasing computational demands and to identify conditions under which GPU acceleration offers the most significant advantages.

\subsubsection{Scaling Problem Dimension}
\begin{figure}[H]
  \centering
    \begin{minipage}[b]{0.47\textwidth}
        \includegraphics[width=\textwidth]{Figures/parameter/legend.pdf}
    \end{minipage}
    \centering
    \begin{minipage}[b]{0.47\textwidth}
        \includegraphics[width=\textwidth]{Figures/parameter/legend.pdf}
    \end{minipage}
    \vspace{0.2cm}
     \centering
    \begin{minipage}[b]{0.48\textwidth}
        \centering
        \includegraphics[width=0.48\textwidth]{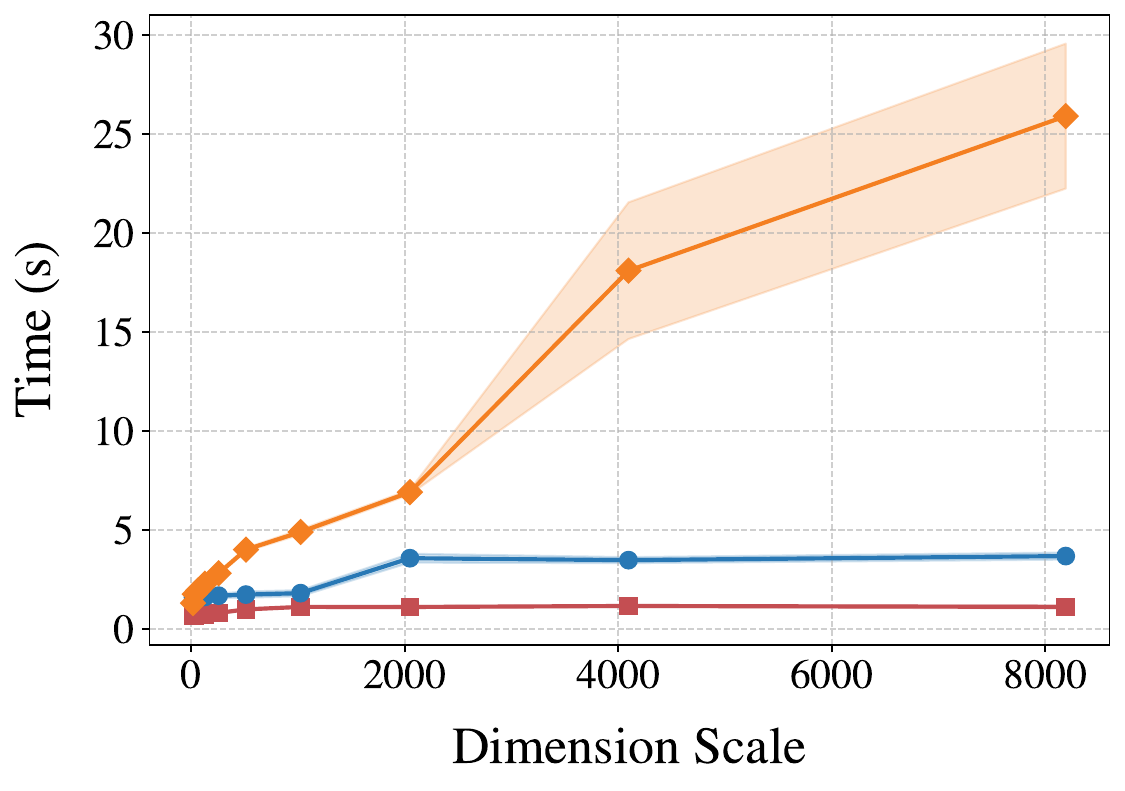}
        \centering
        \includegraphics[width=0.48\textwidth]{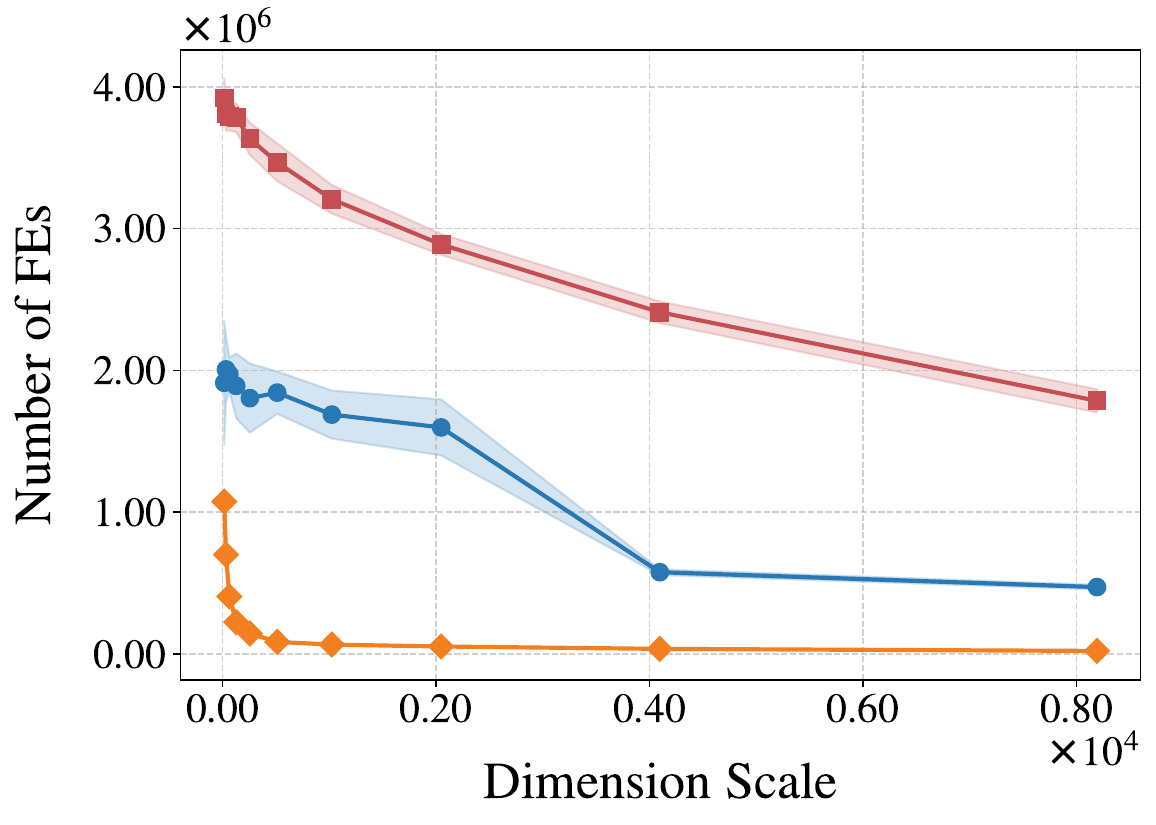}

        \centering
        \subcaption{PSO on Ackley (Popsize = 128)}
    \end{minipage}
    \vspace{0.3cm}
    \centering
    \begin{minipage}[b]{0.48\textwidth}
        \centering
        \includegraphics[width=0.48\textwidth]{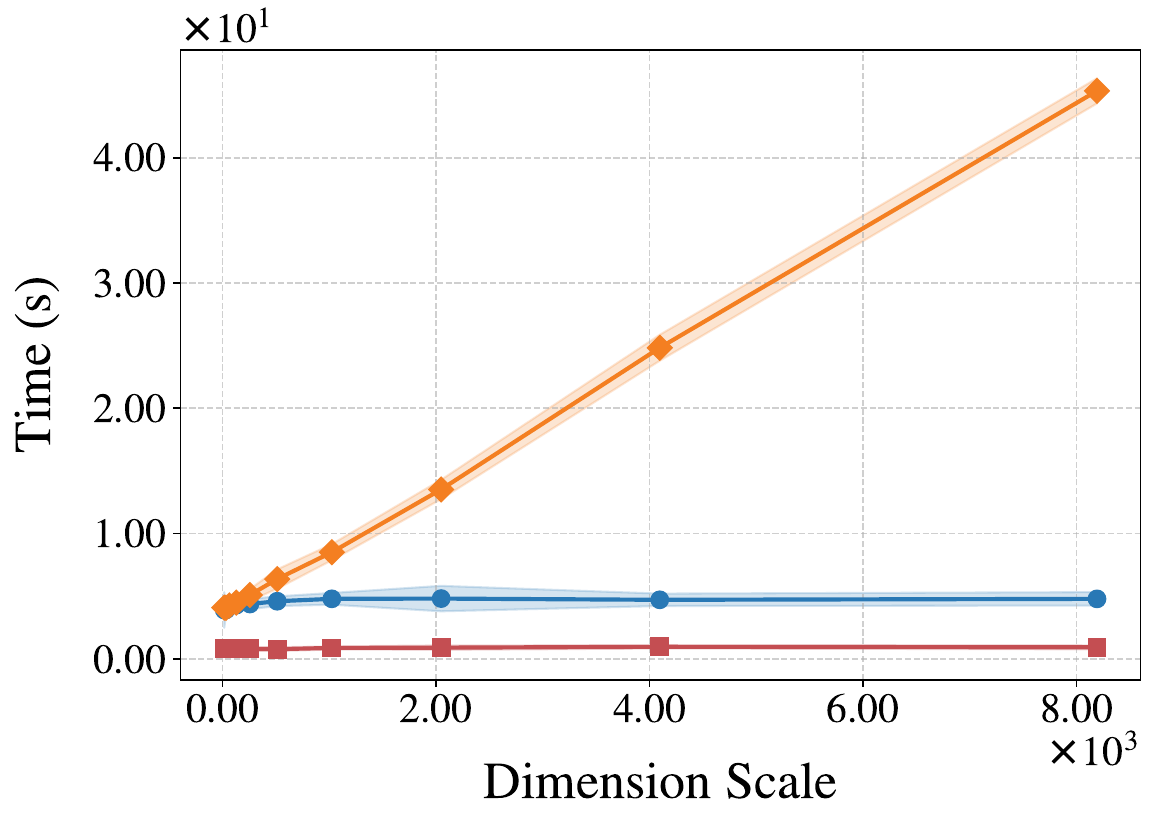}
        \centering
        \includegraphics[width=0.48\textwidth]{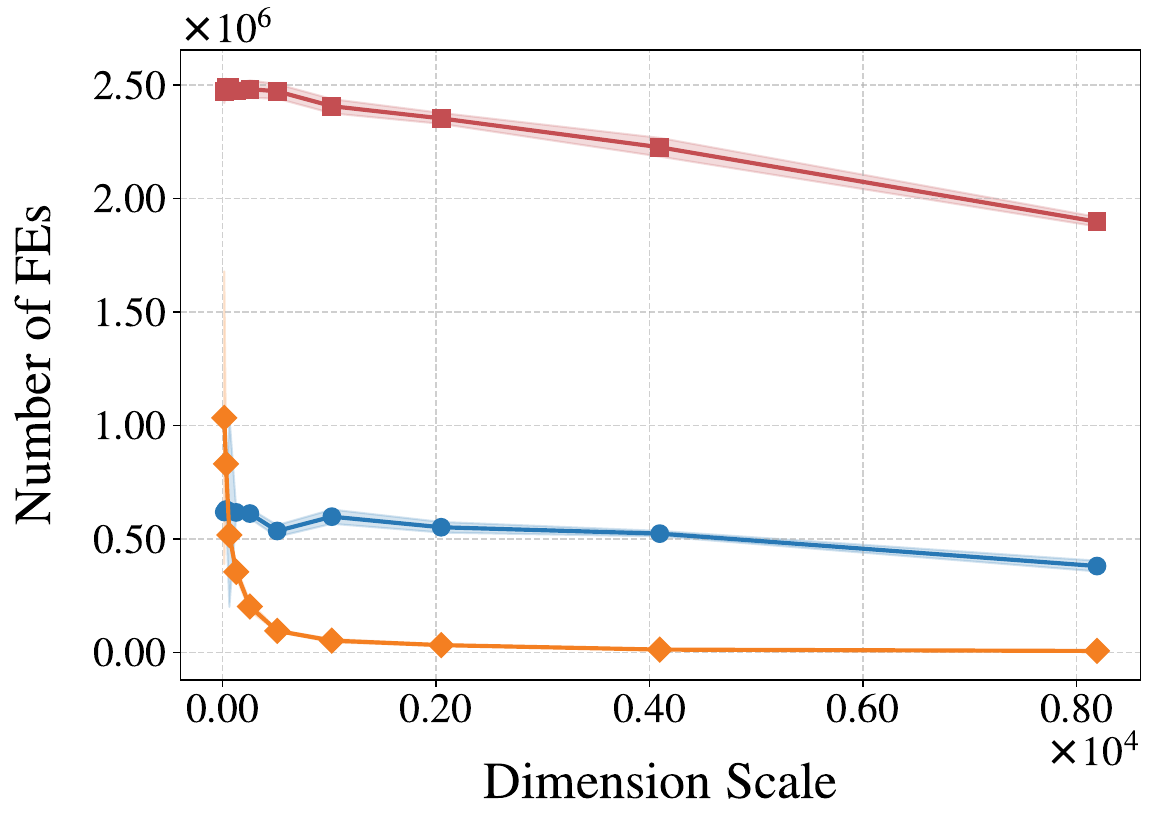}

        \centering
        \subcaption{MOEA/D on DTLZ1 (Popsize = 128)}
    \end{minipage}
    \vspace{0.3cm}
   \centering
    \begin{minipage}[b]{0.48\textwidth}
        \centering
        \includegraphics[width=0.48\textwidth]{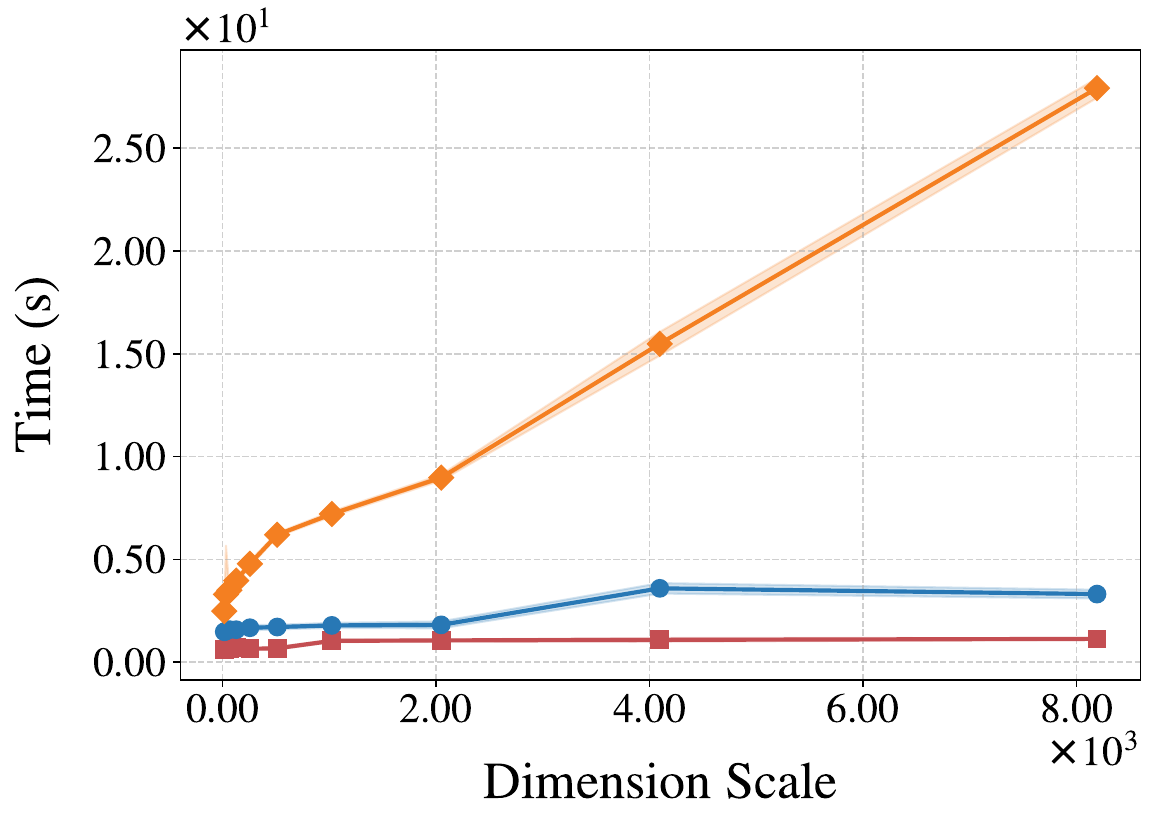}
        \centering
        \includegraphics[width=0.48\textwidth]{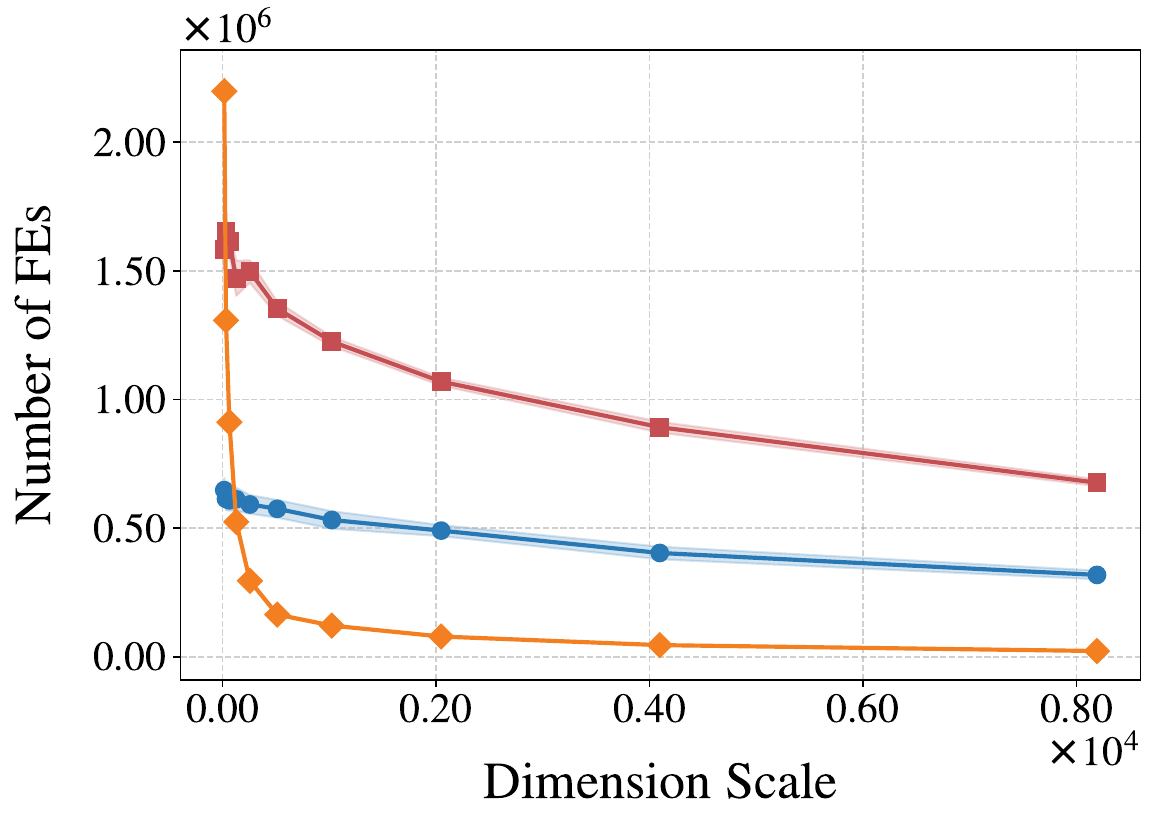}

        \centering
        \subcaption{PSO on Schwefel (Popsize = 128)}
    \end{minipage}
    \vspace{0.3cm}
    \centering
    \begin{minipage}[b]{0.48\textwidth}
        \centering
        \includegraphics[width=0.48\textwidth]{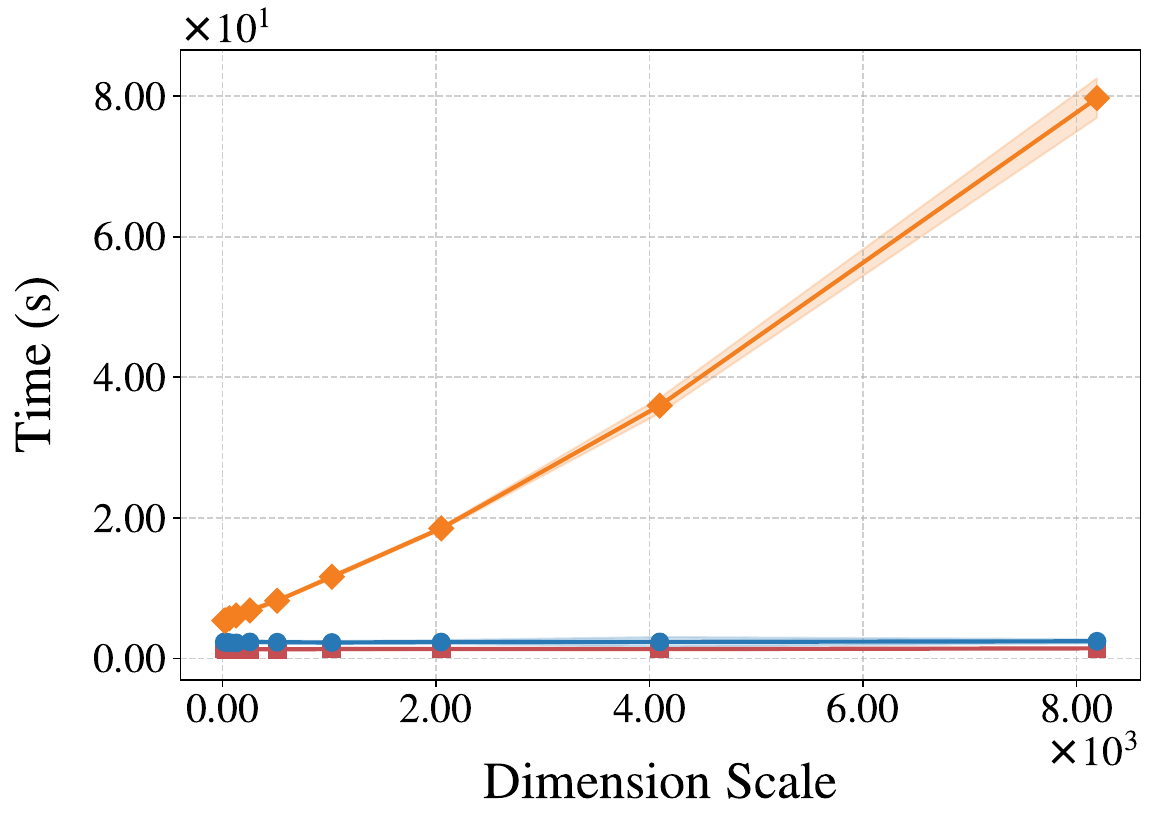}
        \centering
        \includegraphics[width=0.48\textwidth]{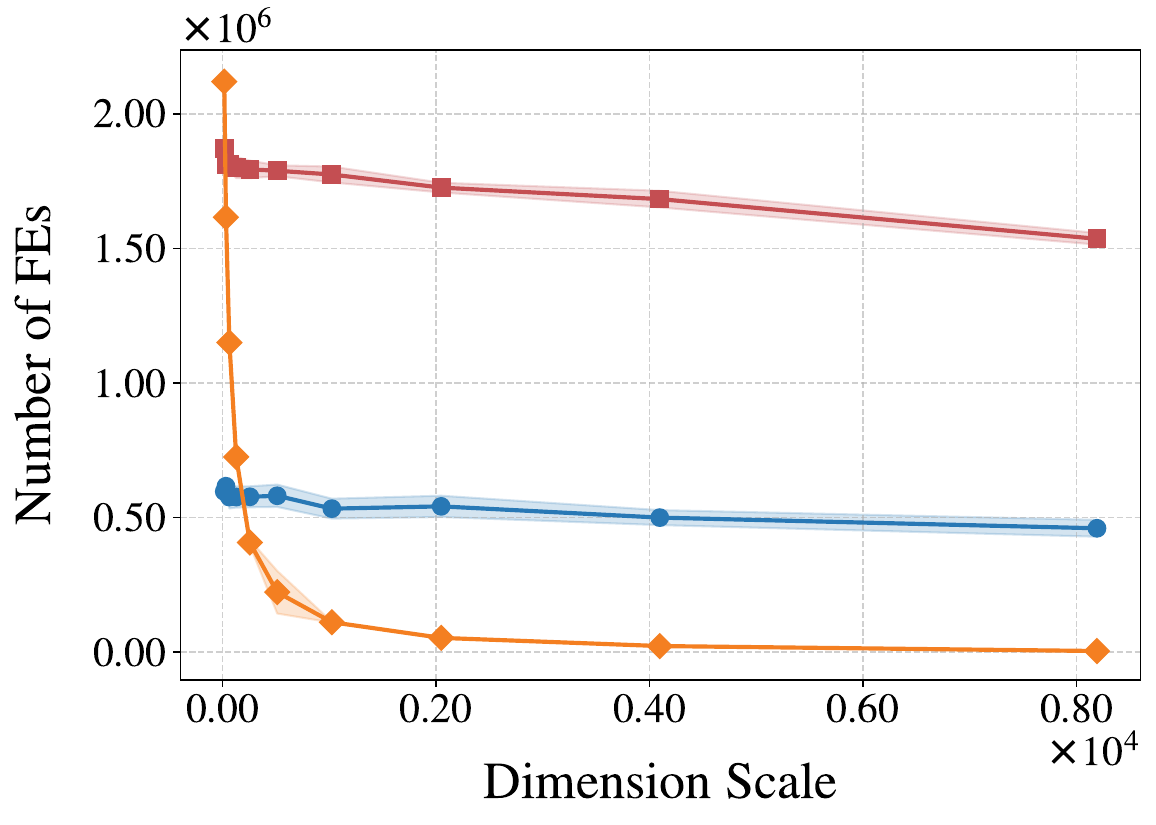}

        \centering
        \subcaption{MOEA/D on ZDT1 (Popsize = 128)}
    \end{minipage}
     \vspace{0.3cm}
    \centering
    \begin{minipage}[b]{0.48\textwidth}
        \centering
        \includegraphics[width=0.48\textwidth]{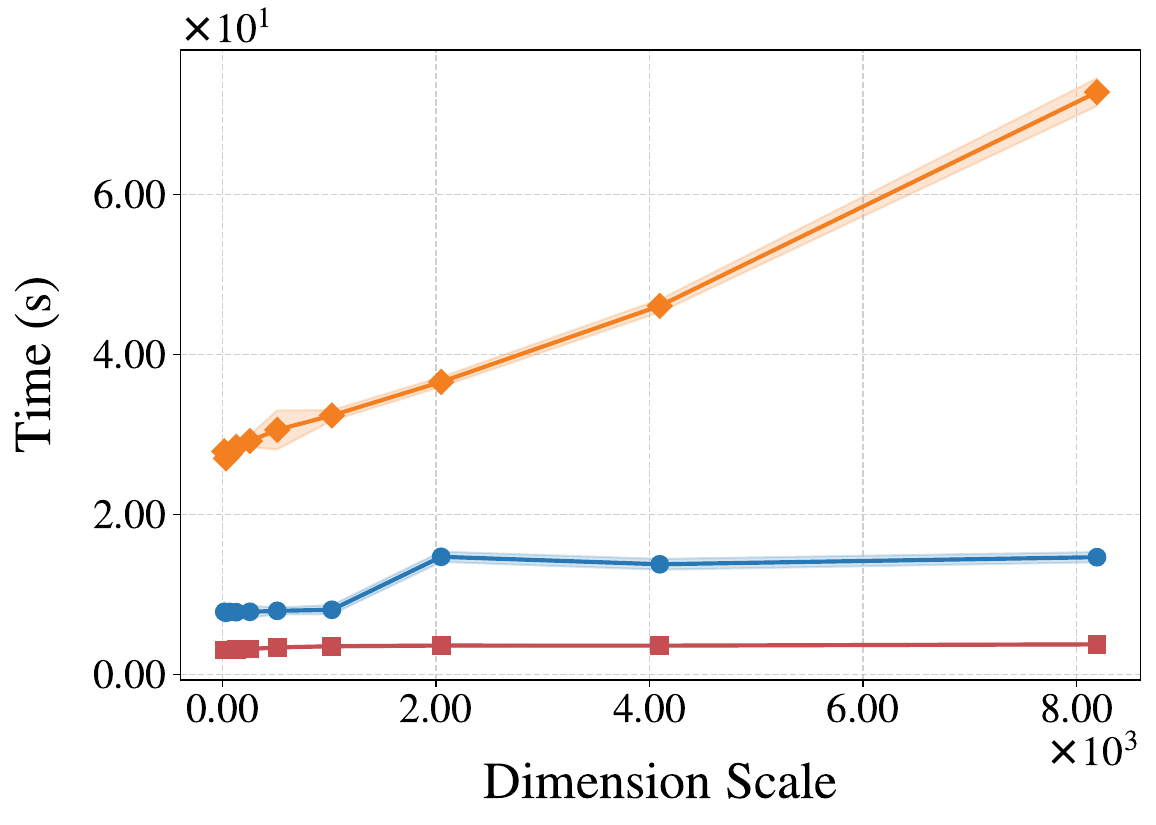}
        \centering
        \includegraphics[width=0.48\textwidth]{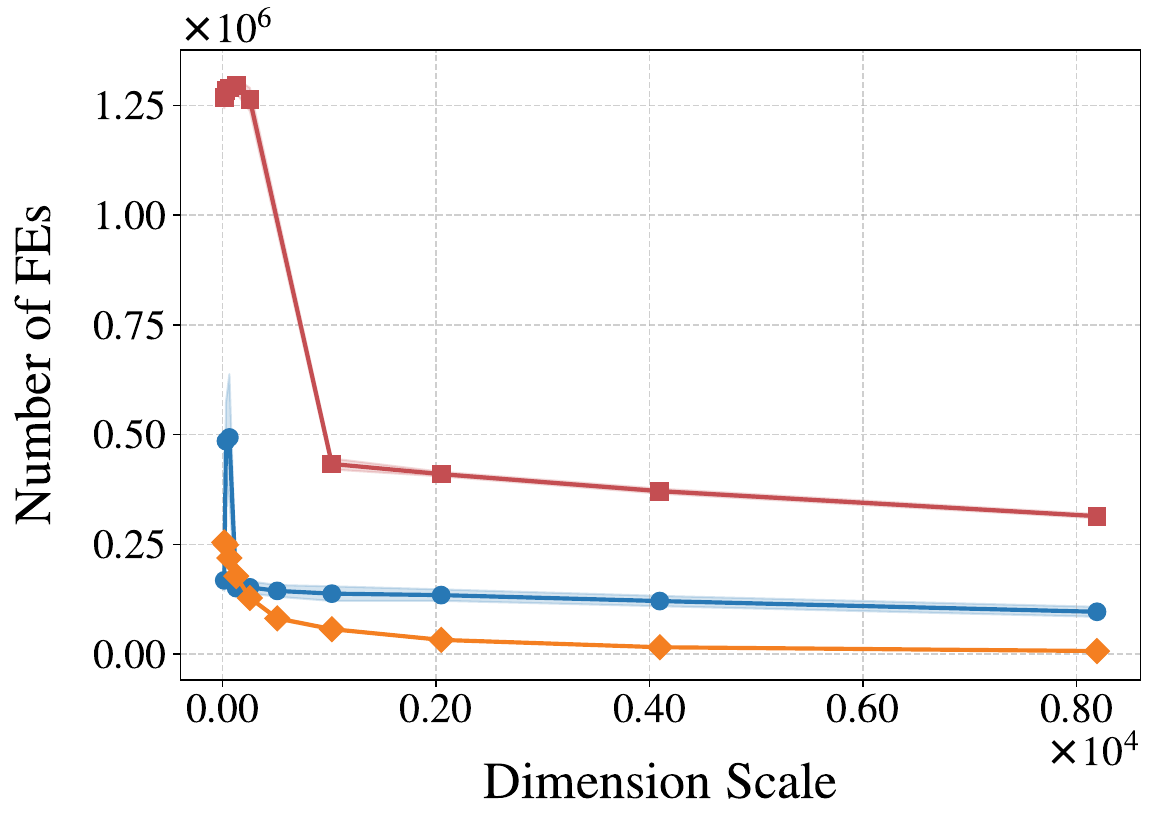}

        \centering
        \subcaption{SaDE on Ackley (Popsize = 128)}
    \end{minipage}
    \centering
    \begin{minipage}[b]{0.48\textwidth}
        \centering
        \includegraphics[width=0.48\textwidth]{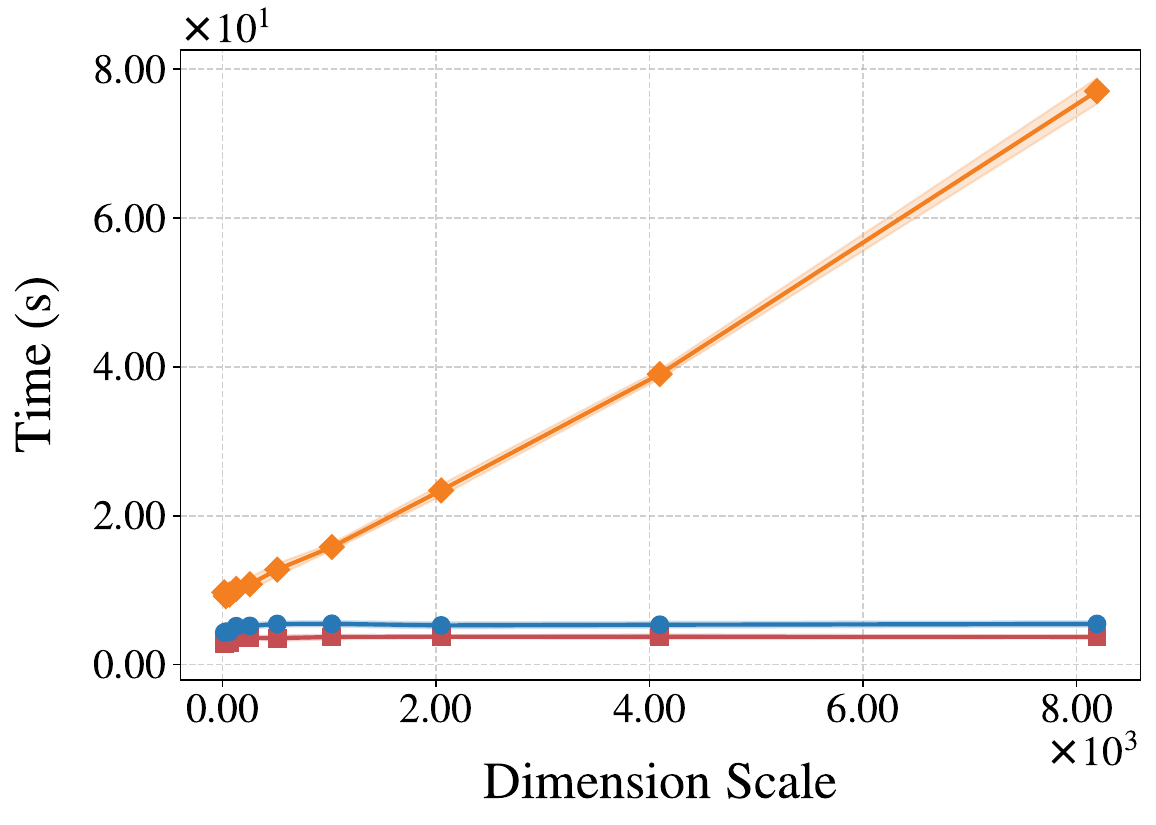}
        \centering
        \includegraphics[width=0.48\textwidth]{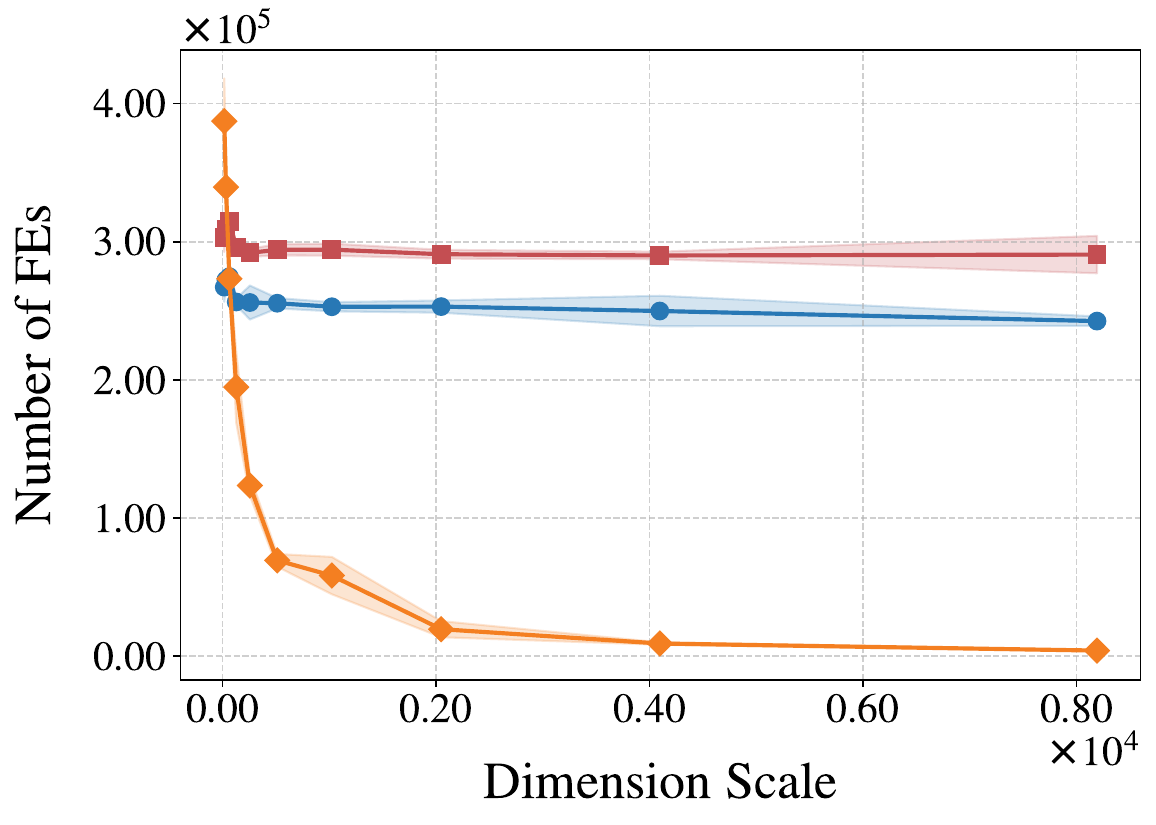}

        \centering
        \subcaption{NSGA-III on DTLZ1 (Popsize = 128)}
    \end{minipage}
    \centering
    \begin{minipage}[b]{0.48\textwidth}
        \centering
        \includegraphics[width=0.48\textwidth]{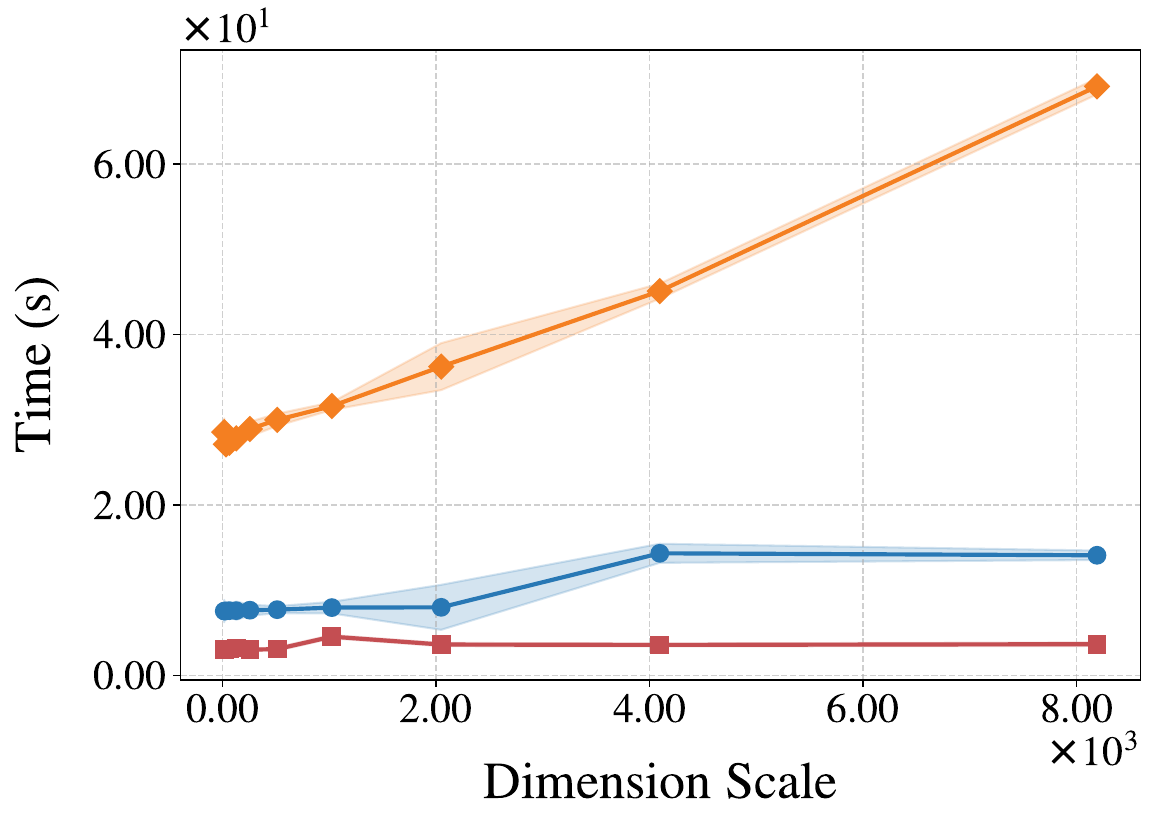}
        \centering
        \includegraphics[width=0.48\textwidth]{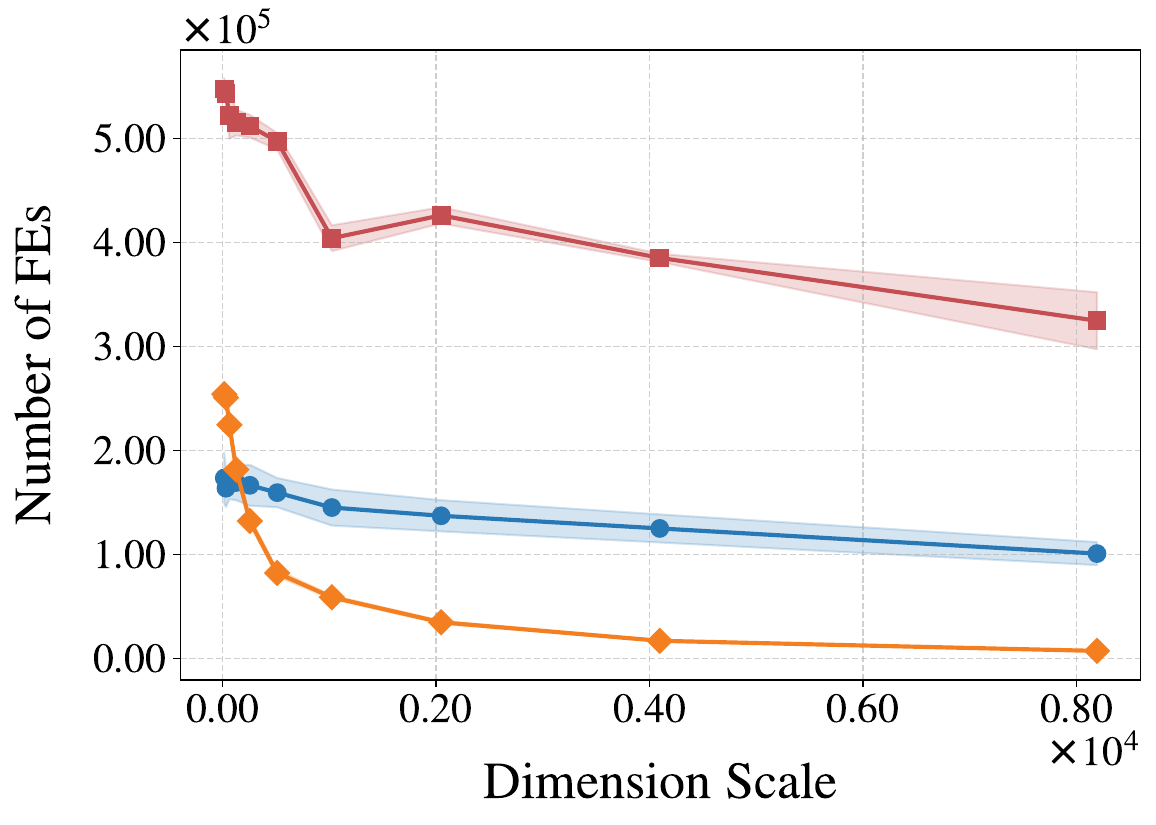}

        \centering
        \subcaption{SaDE on Schwefel (Popsize = 128)}
    \end{minipage}
    \centering
    \begin{minipage}[b]{0.48\textwidth}
        \centering
        \includegraphics[width=0.48\textwidth]{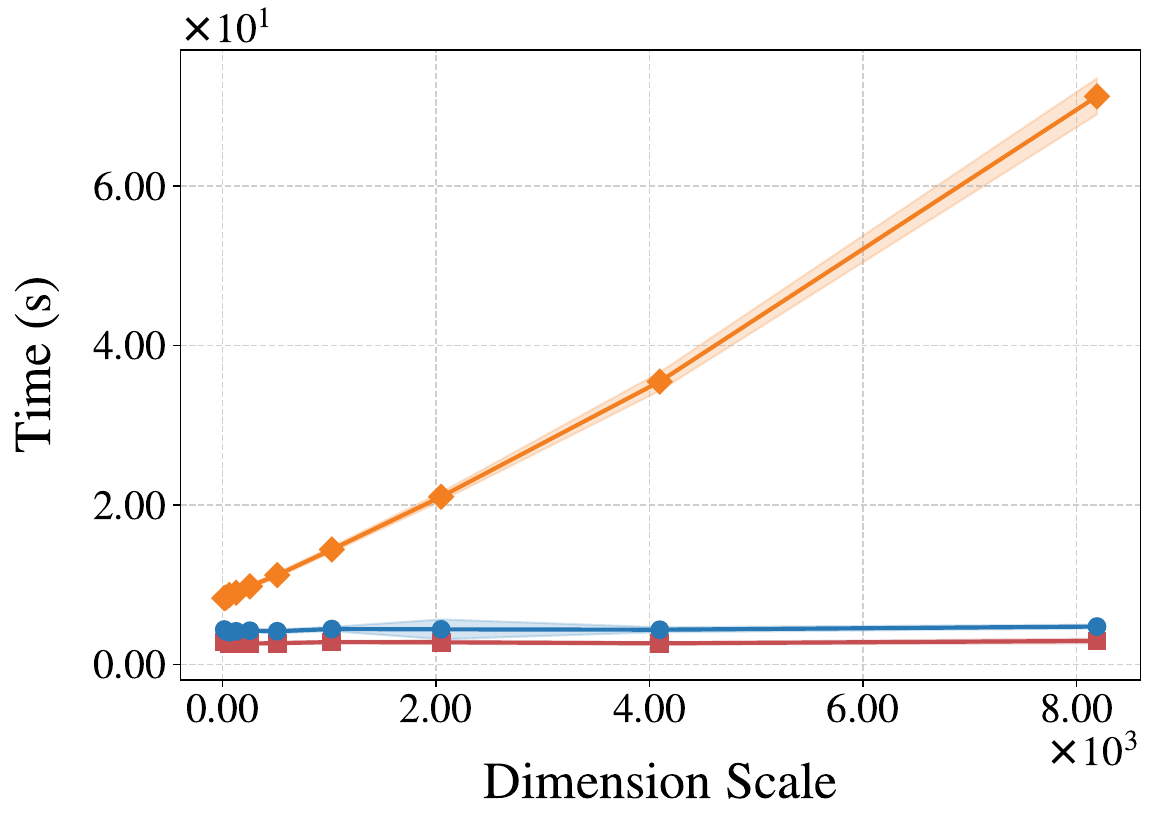}
        \centering
        \includegraphics[width=0.48\textwidth]{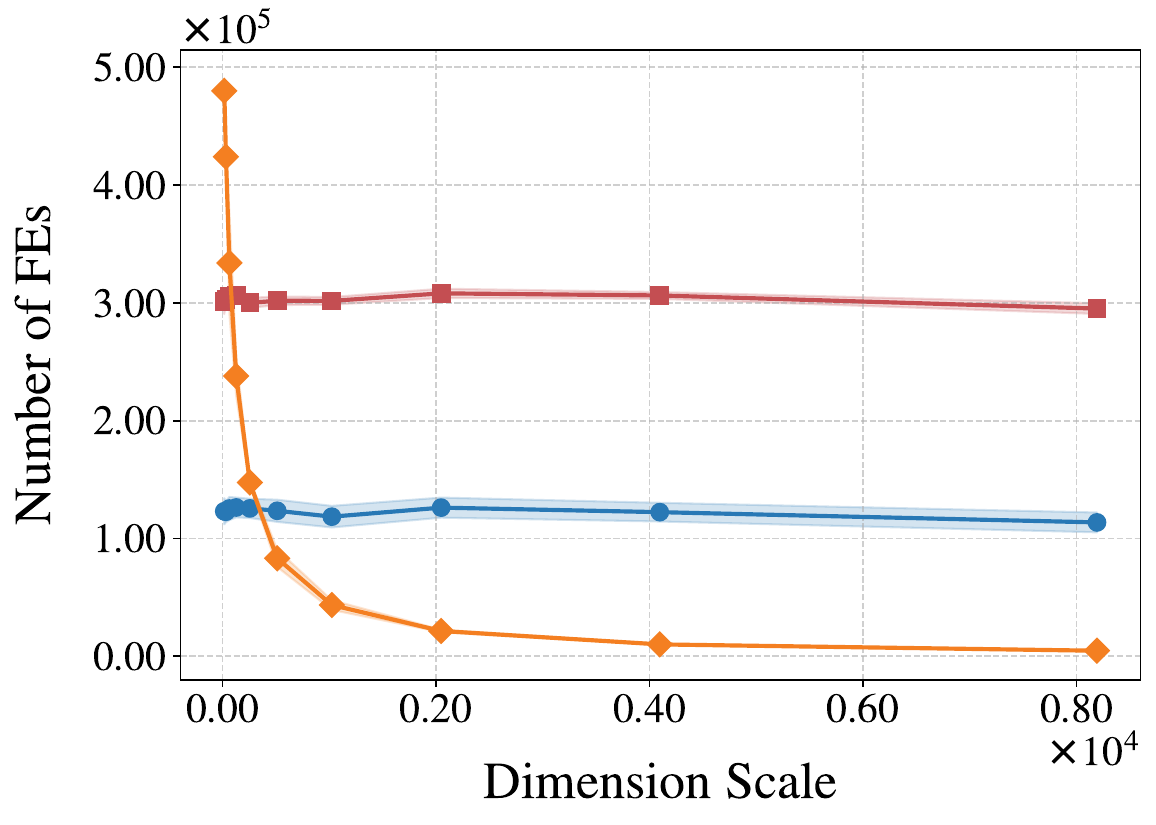}

        \centering
        \subcaption{NSGA-III on ZDT1 (Popsize = 128)}
    \end{minipage}

    \caption{Computational performance tested across varying problem dimensions on three hardware platforms. \textbf{Left}: Total runtime over 100 generations. \textbf{Right}: Number of FEs completed within a 30-second time budget. Results are averaged over 15 independent runs; solid lines indicate mean values, and shaded regions represent standard deviations.}
    \label{fig:s-varying dimension}
\end{figure}

Fig.~\ref{fig:s-varying dimension} illustrates the computational efficiency of EAs on different hardware platforms under increasing problem dimensionalities. The left panel shows the \textbf{runtime} over 100 generations, while the right panel reports the \textbf{number of FEs} completed within a fixed 30-second time window. Problem dimensions were scaled exponentially, taking values from {16, 32, 64, 128, 256, 512, 1024, 2048, 4096, 8192}. Algorithms were evaluated with a fixed population size of 128 across all configurations. Each setting was repeated 15 times, with the plotted curves representing the mean performance and the shaded bands indicating standard deviations. Different colors denote different hardware platforms. Lower runtime and higher NFE counts are indicative of greater computational efficiency.

\subsubsection{Scaling Population Size}
\begin{figure}[H]
    \centering
     \begin{minipage}[b]{0.47\textwidth}
        \includegraphics[width=\textwidth]{Figures/parameter/legend.pdf}
    \end{minipage}
    \centering
    \begin{minipage}[b]{0.47\textwidth}
        \includegraphics[width=\textwidth]{Figures/parameter/legend.pdf}
    \end{minipage}
    \vspace{0.2cm}
    \centering
    \begin{minipage}[b]{0.48\textwidth}
        \centering
        \includegraphics[width=0.48\textwidth]{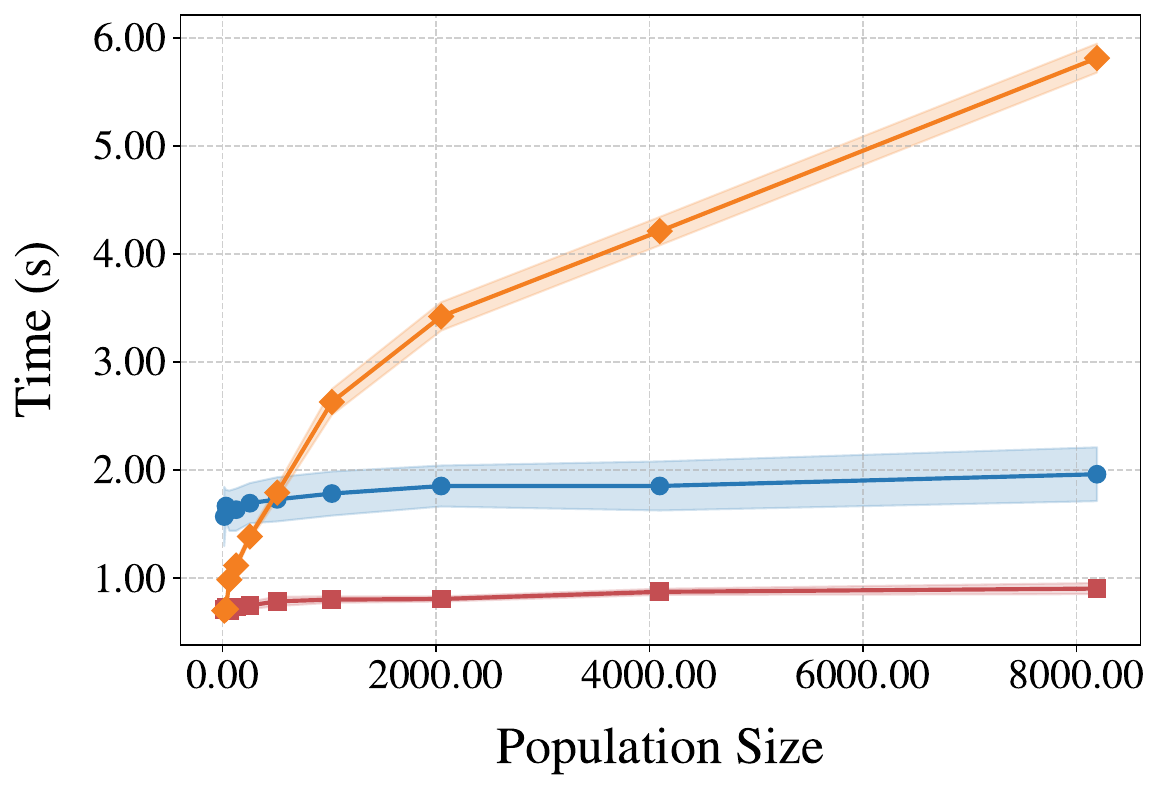}
        \centering
        \includegraphics[width=0.48\textwidth]{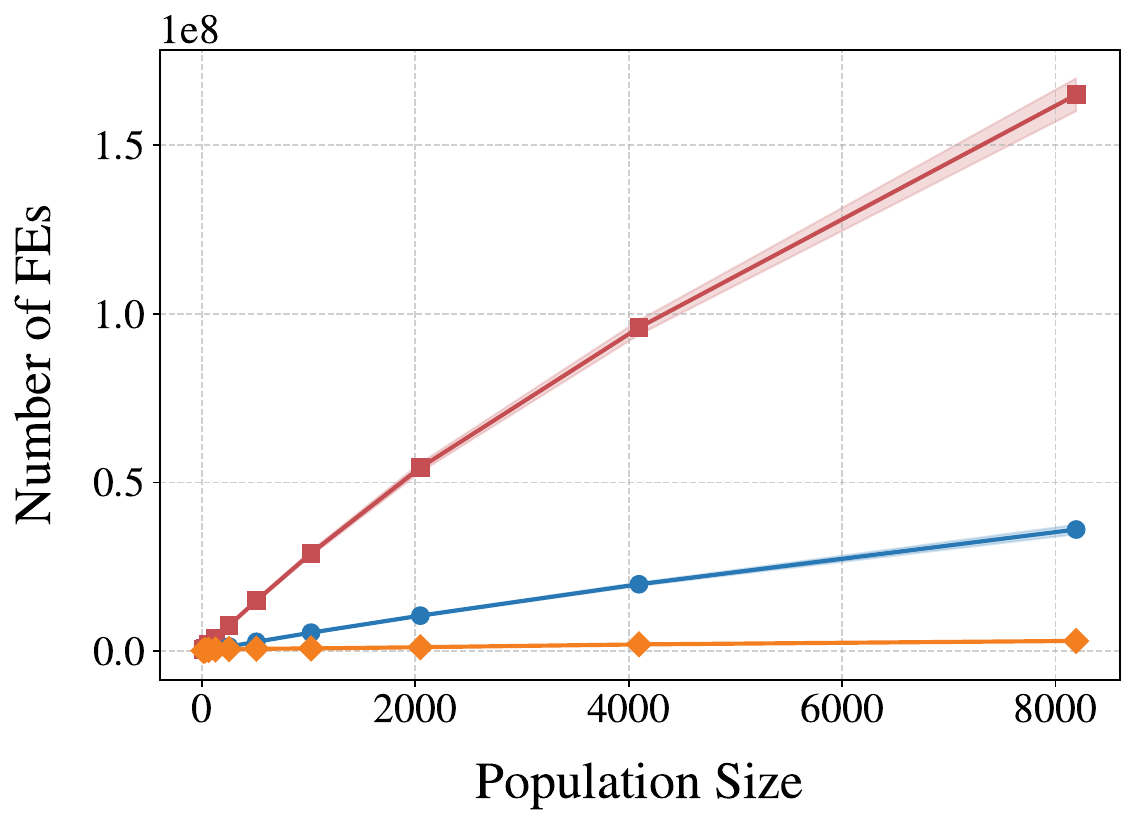}

        \centering
        \subcaption{PSO on Ackley (D = 50)}
    \end{minipage}
    \vspace{0.3cm}
    \centering
    \begin{minipage}[b]{0.48\textwidth}
        \centering
        \includegraphics[width=0.48\textwidth]{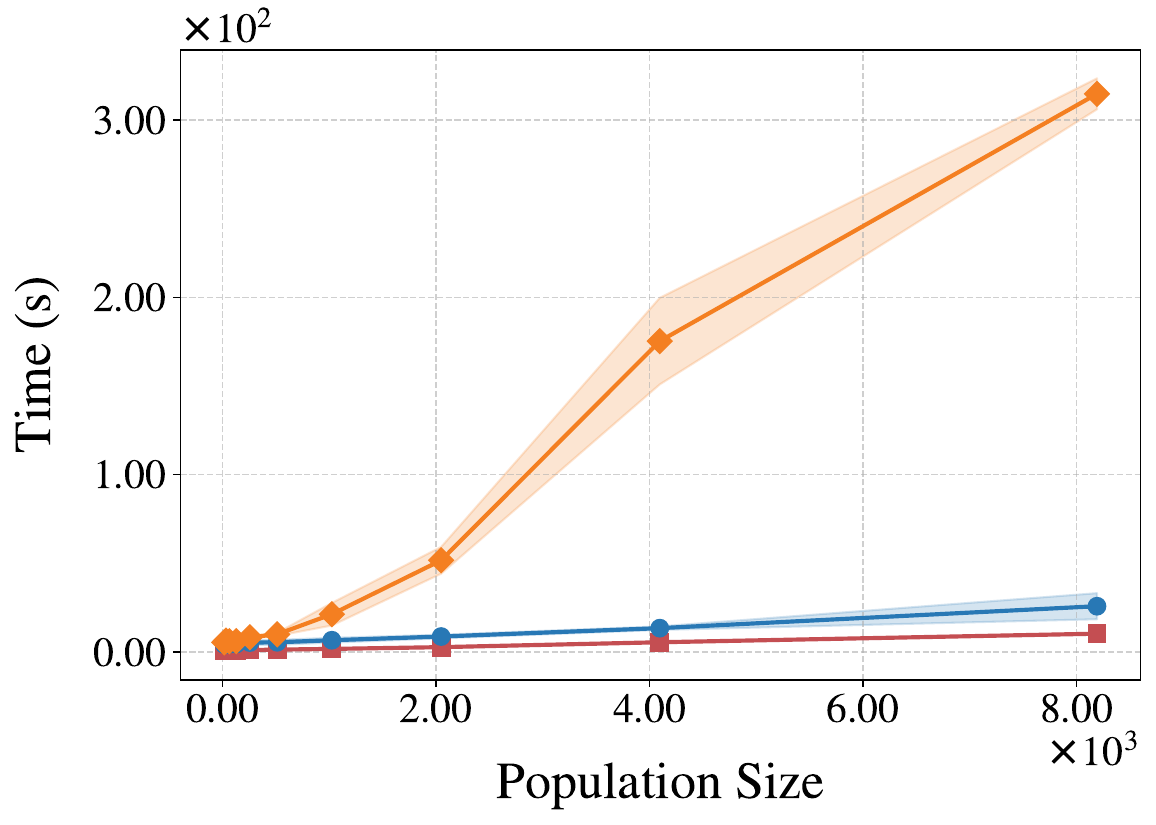}
        \centering
        \includegraphics[width=0.48\textwidth]{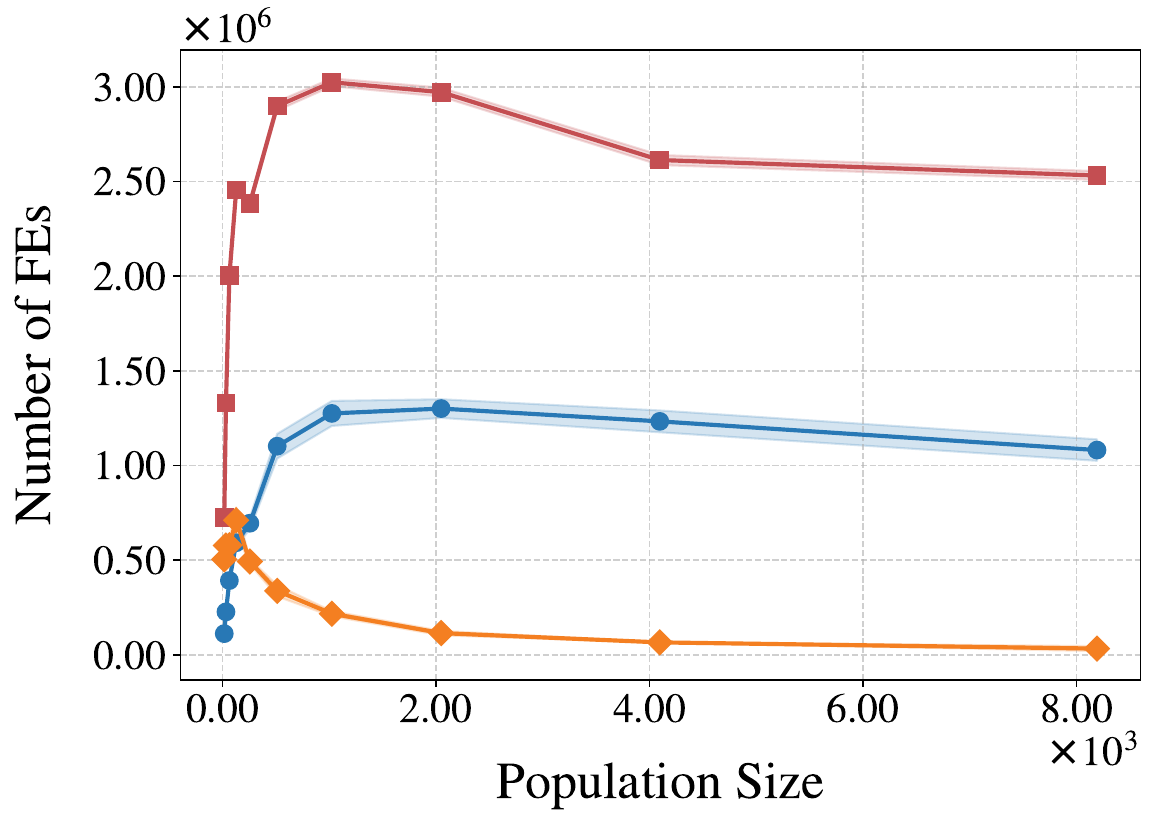}

        \centering
        \subcaption{MOEA/D on DTLZ1 (D = 50)}
    \end{minipage}
    \vspace{0.3cm}
    \centering
    \begin{minipage}[b]{0.48\textwidth}
        \centering
        \includegraphics[width=0.48\textwidth]{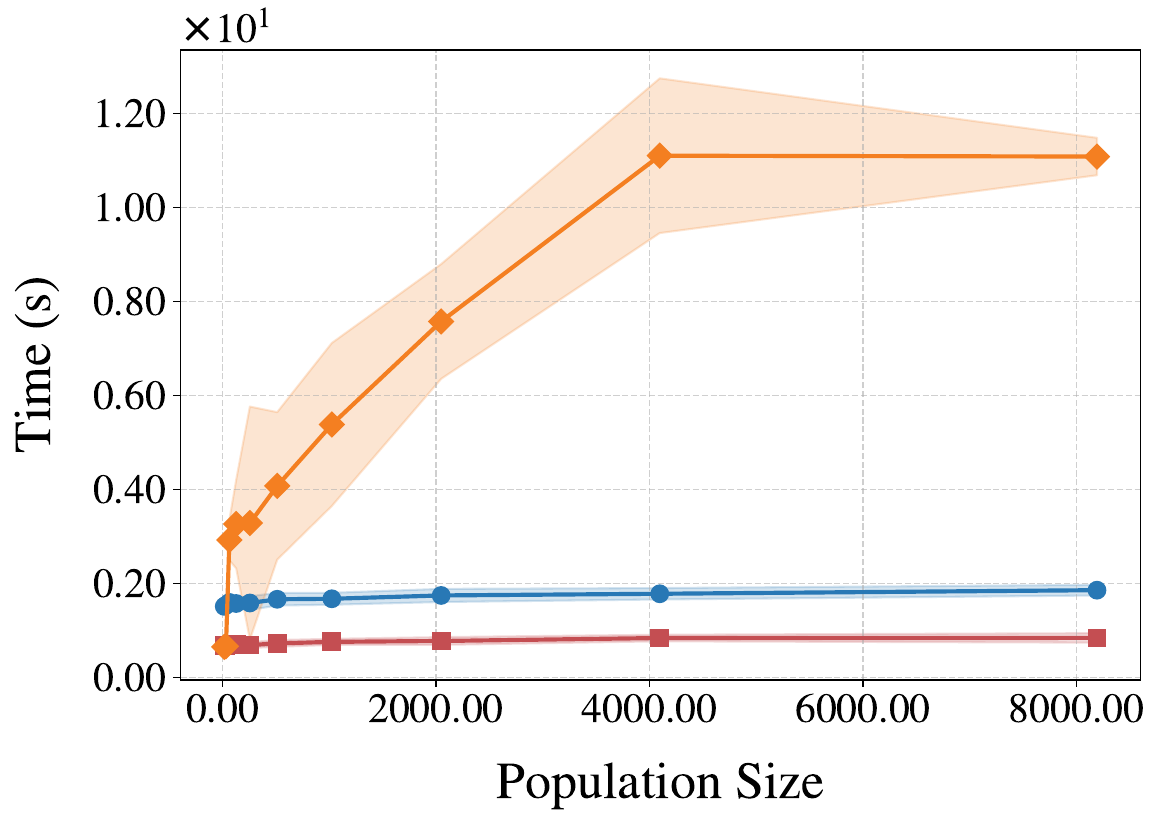}
        \centering
        \includegraphics[width=0.48\textwidth]{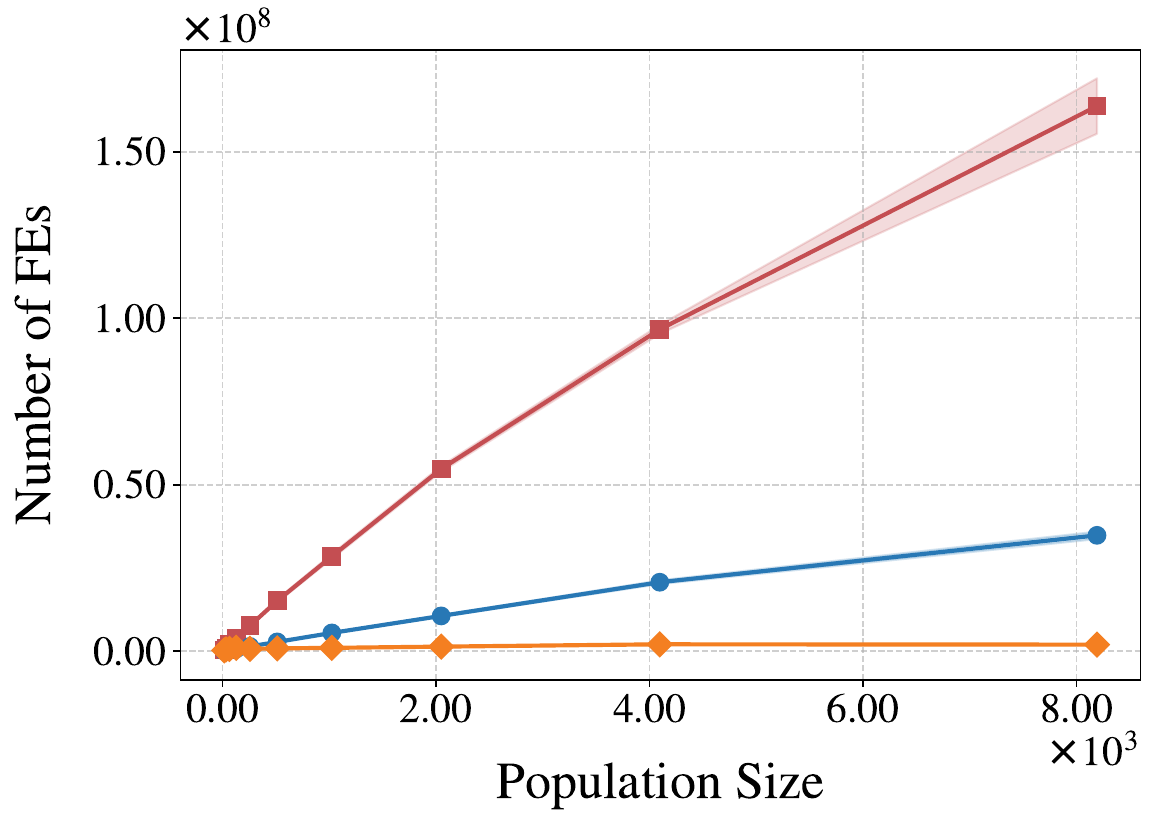}

        \centering
        \subcaption{PSO on Schwefel (D = 50)}
    \end{minipage}
    \vspace{0.3cm}
    \centering
    \begin{minipage}[b]{0.48\textwidth}
        \centering
        \includegraphics[width=0.48\textwidth]{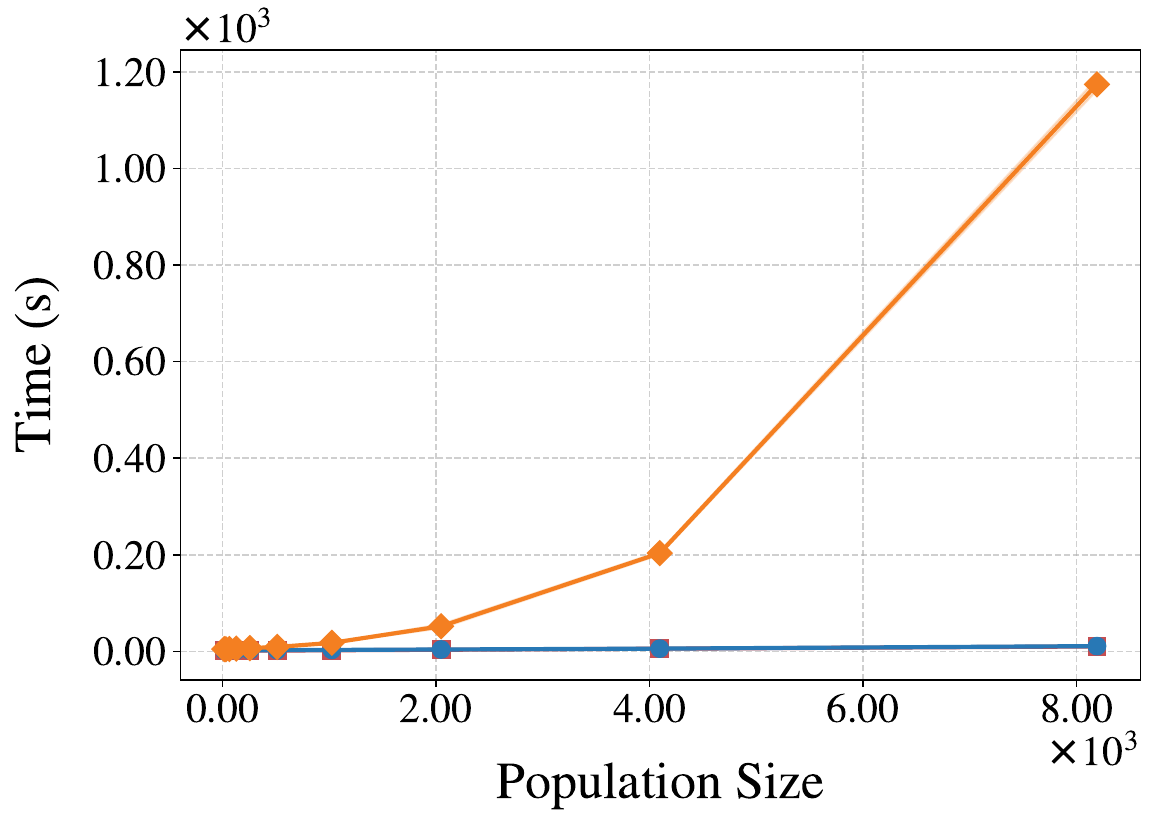}
        \centering
        \includegraphics[width=0.48\textwidth]{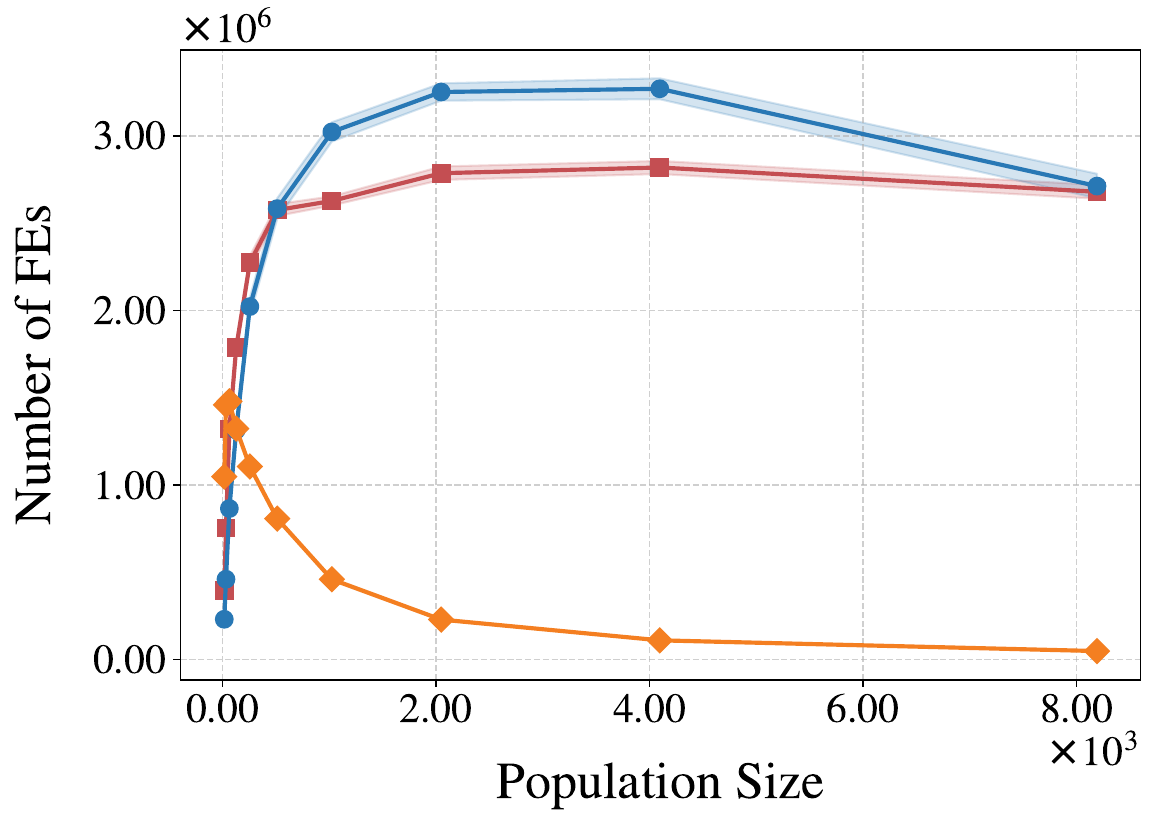}

        \centering
        \subcaption{MOEA/D on ZDT1 (D = 50)}
    \end{minipage}
     \vspace{0.3cm}
    \centering
    \begin{minipage}[b]{0.48\textwidth}
        \centering
        \includegraphics[width=0.48\textwidth]{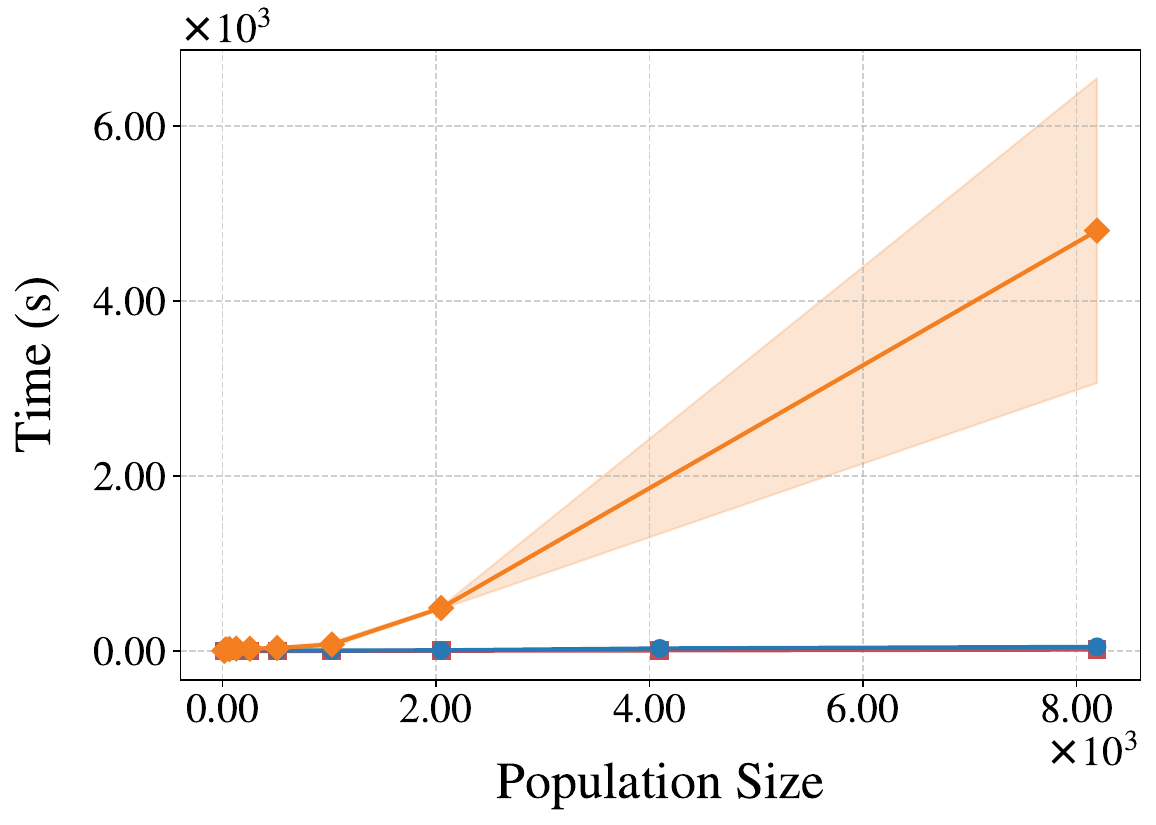}
        \centering
        \includegraphics[width=0.48\textwidth]{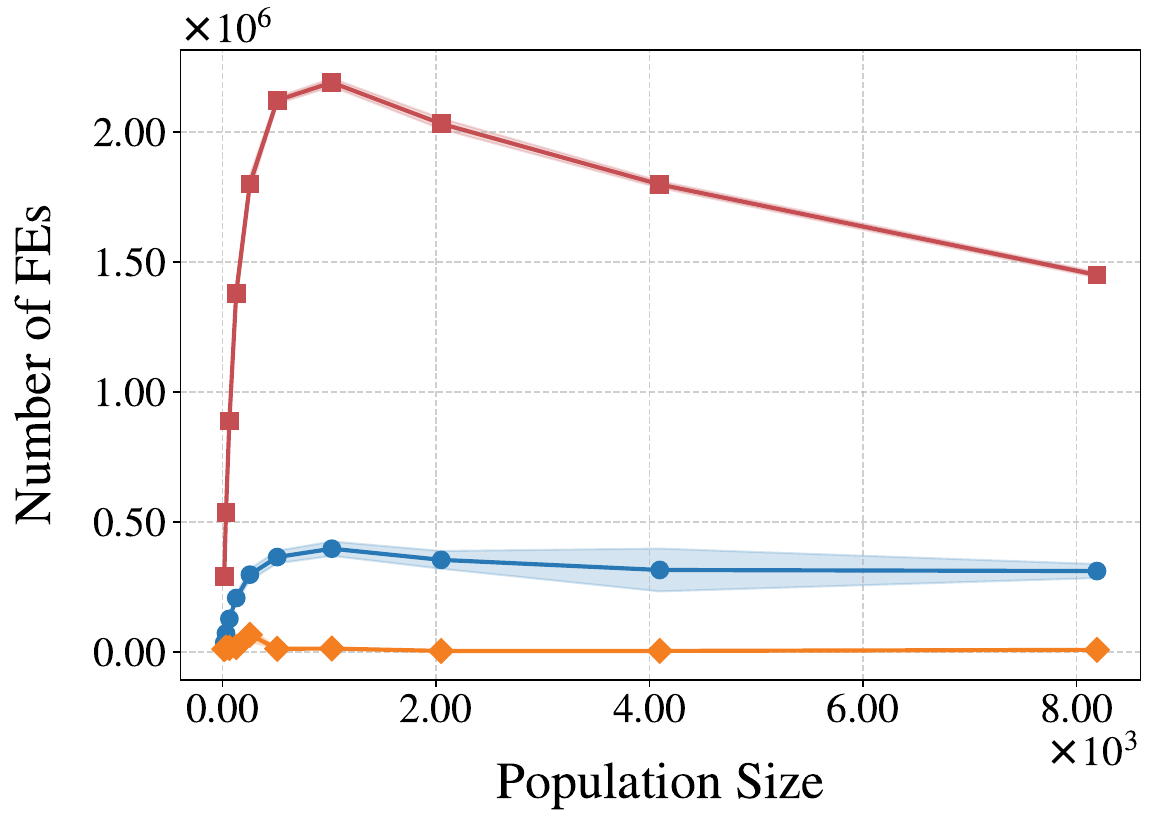}

        \centering
        \subcaption{SaDE on Ackley (D = 50)}
    \end{minipage}
    \centering
    \begin{minipage}[b]{0.48\textwidth}
        \centering
        \includegraphics[width=0.48\textwidth]{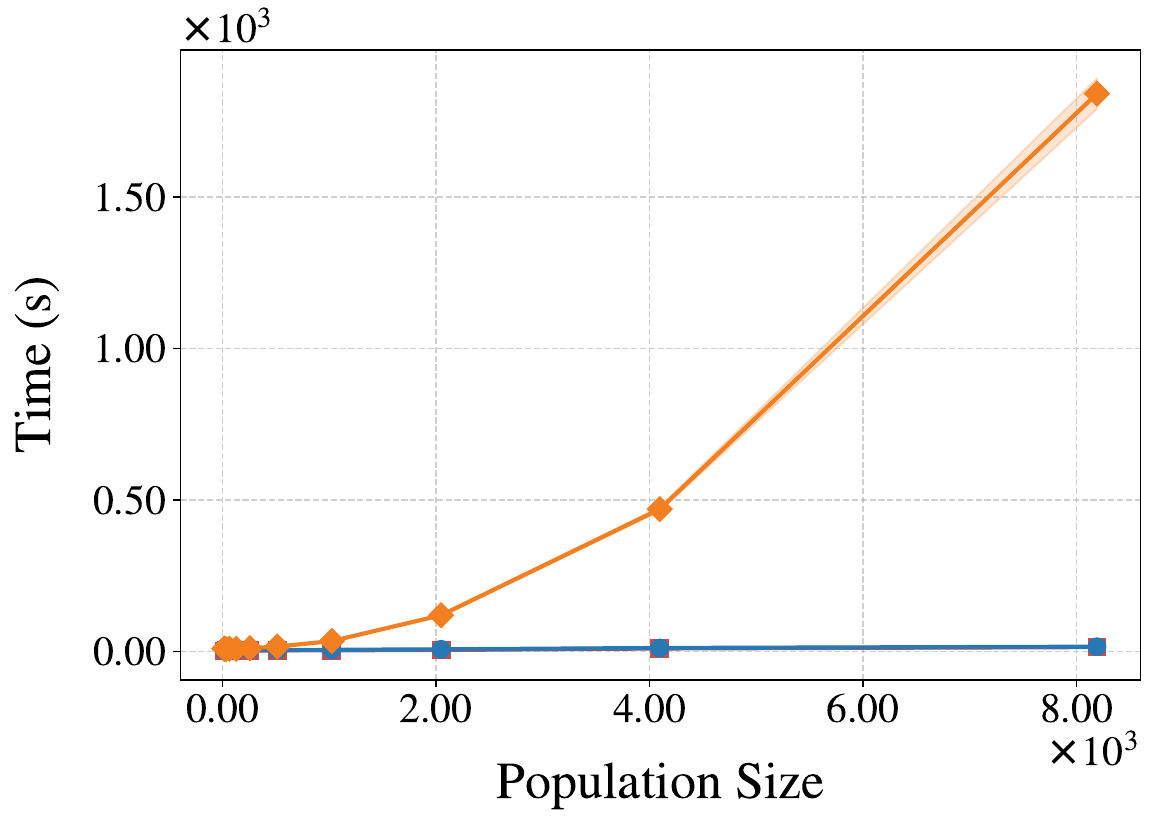}
        \centering
        \includegraphics[width=0.48\textwidth]{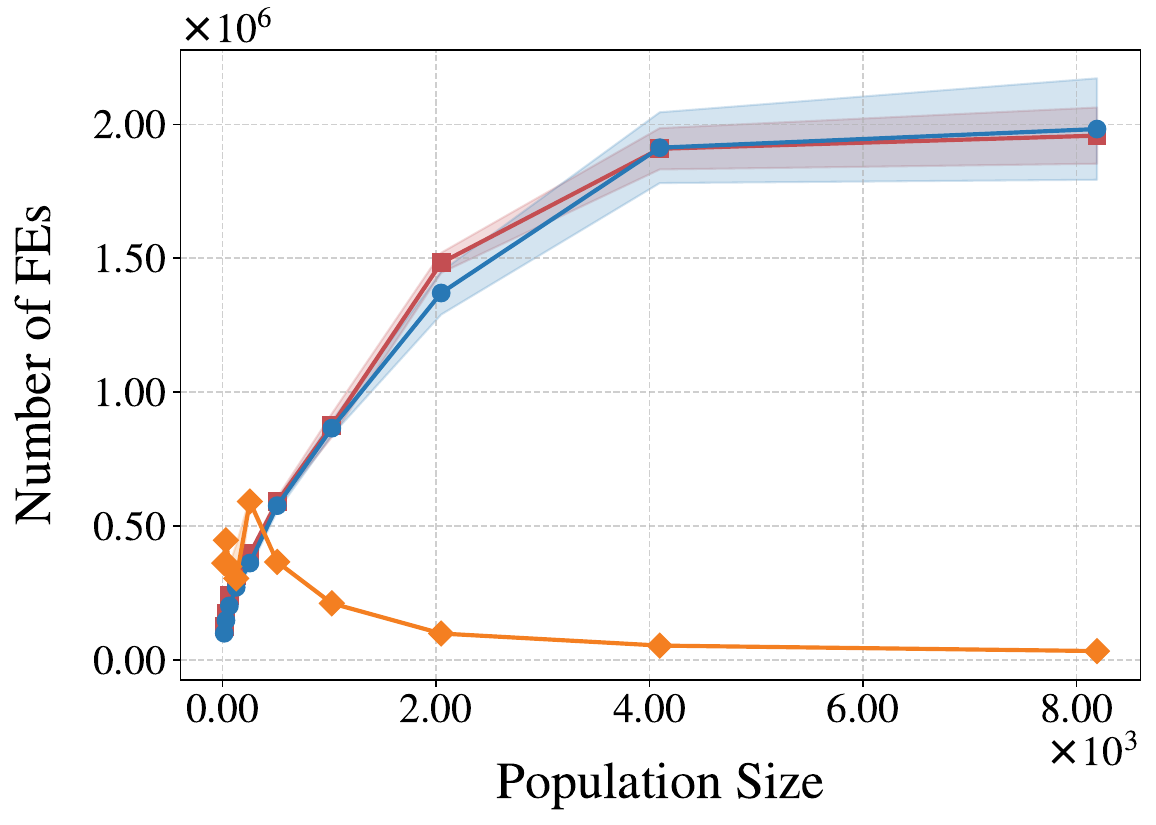}

        \centering
        \subcaption{NSGA-III on DTLZ1 (D = 50)}
    \end{minipage}
    \centering
    \begin{minipage}[b]{0.48\textwidth}
        \centering
        \includegraphics[width=0.48\textwidth]{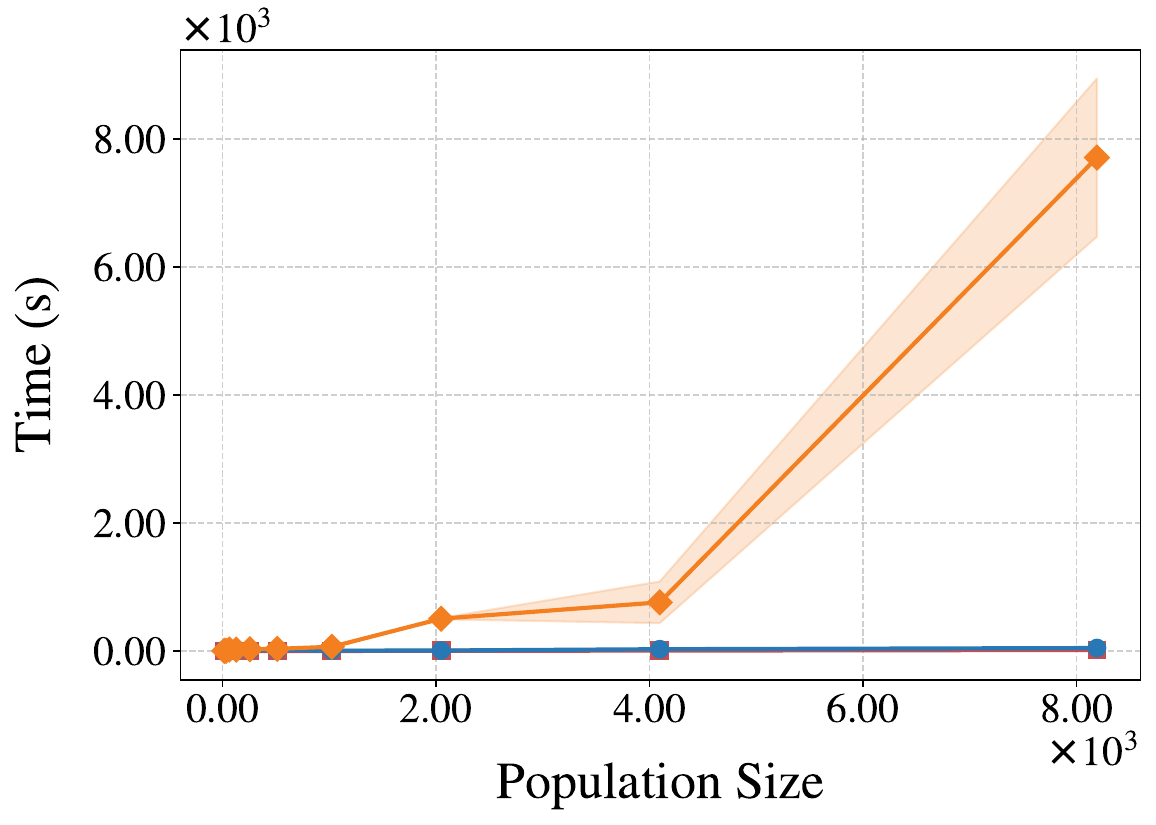}
        \centering
        \includegraphics[width=0.48\textwidth]{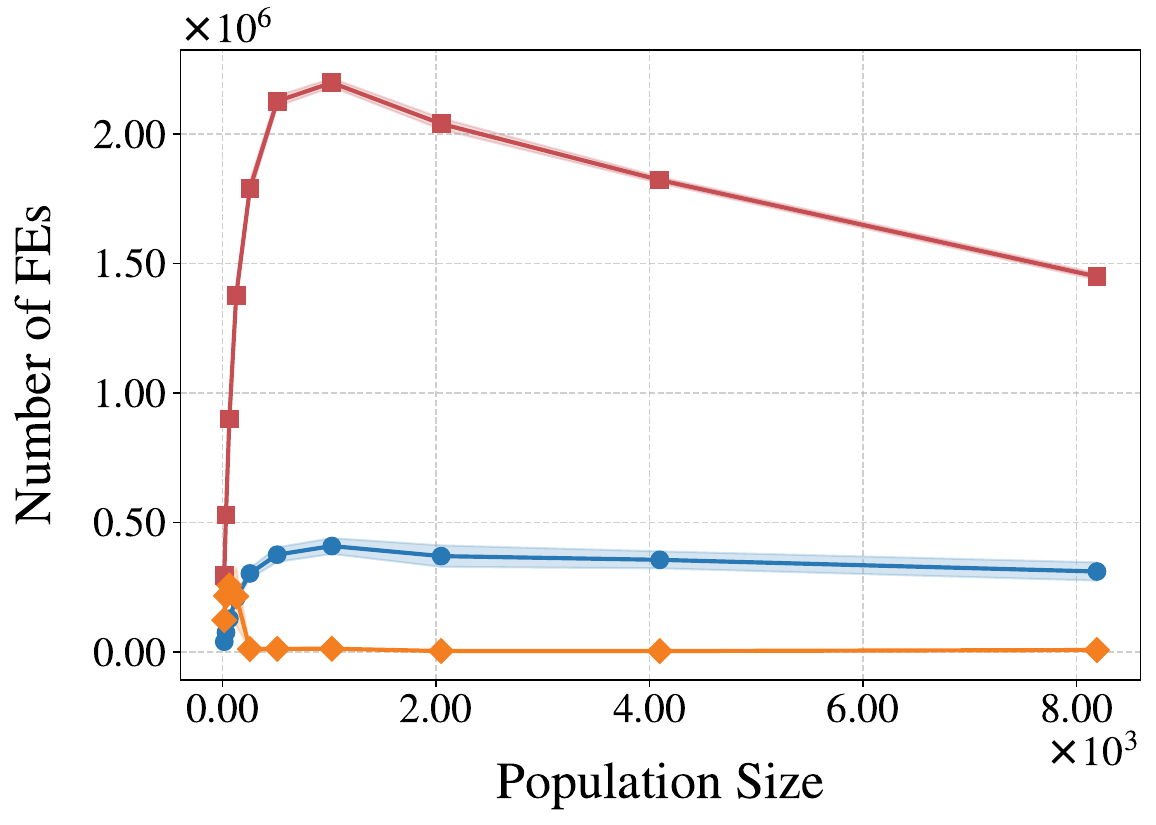}

        \centering
        \subcaption{SaDE on Schwefel (D = 50)}
    \end{minipage}
    \centering
    \begin{minipage}[b]{0.48\textwidth}
        \centering
        \includegraphics[width=0.48\textwidth]{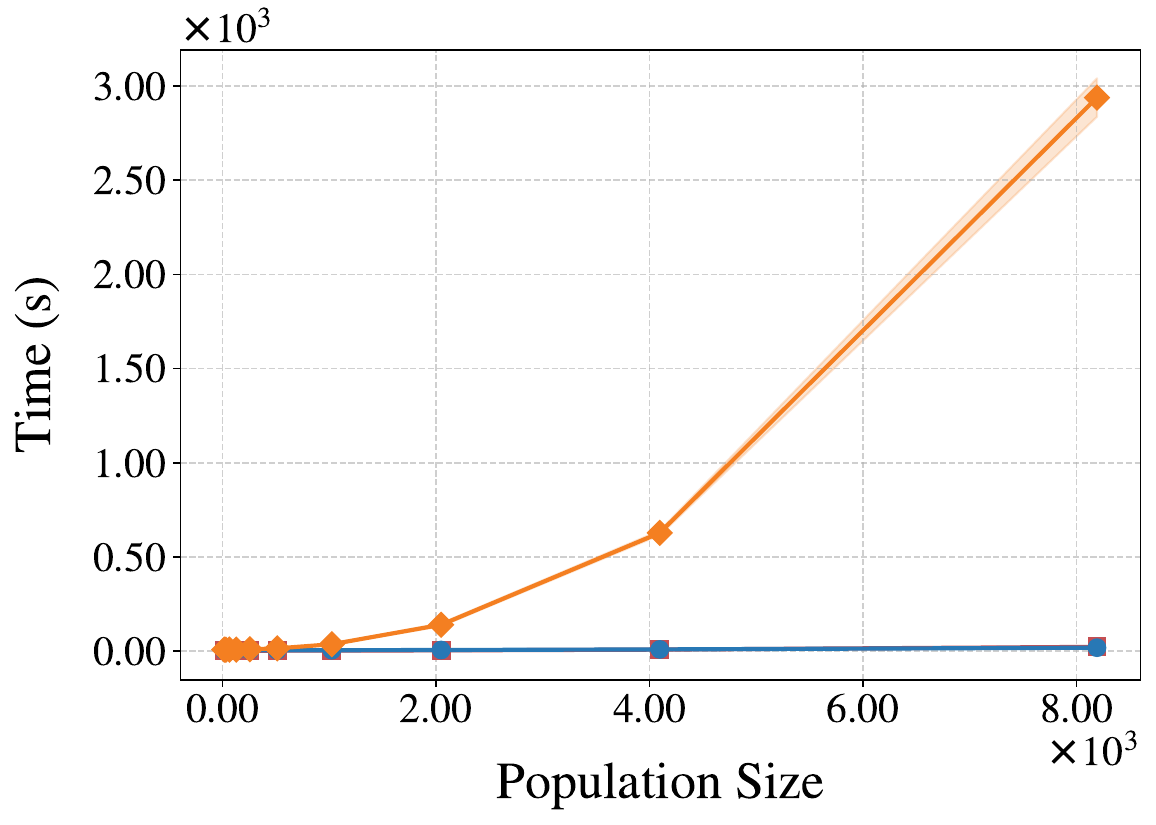}
        \centering
        \includegraphics[width=0.48\textwidth]{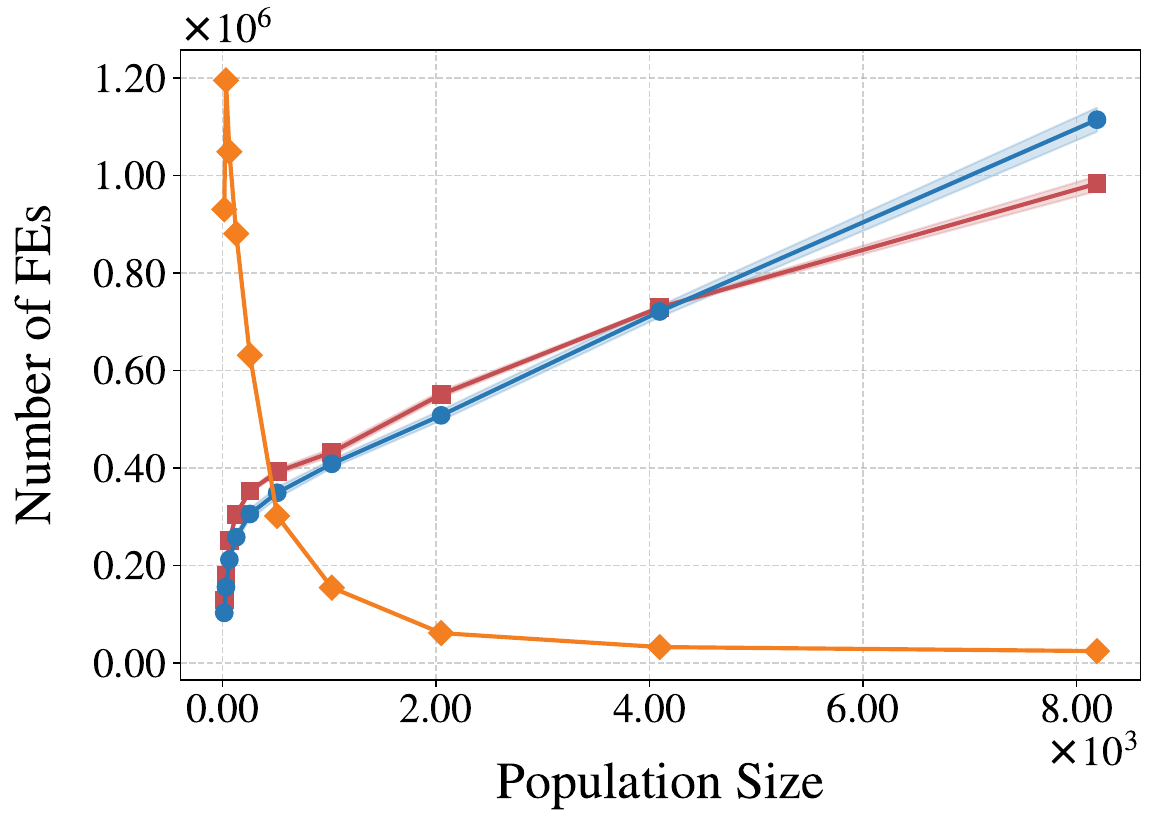}

        \centering
        \subcaption{NSGA-III on ZDT1 (D = 50)}
    \end{minipage}
    \caption{Computational performance of two EAs configured with varying population sizes across three hardware platforms. \textbf{Left}: Total runtime over 100 generations. \textbf{Right}: Number of FEs completed within a 30-second time budget. Each curve represents the average of 15 independent runs; solid lines denote mean values, and shaded areas indicate standard deviations.}
    \label{fig:s-varying popsize1}
\end{figure}

\begin{figure}[htbp]
    \centering

    \begin{minipage}[b]{0.24\textwidth}
        \centering
        \includegraphics[width=\textwidth]{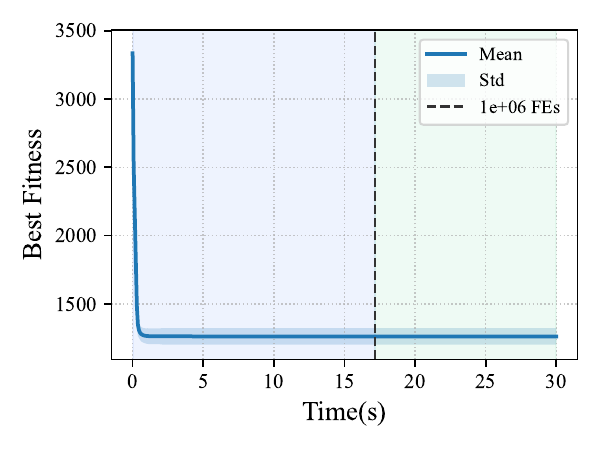}
        \centering
        \vspace{-0.6cm}
        \subcaption{PSO on 200D Rastrigin}
    \end{minipage}
   \hfill
   \begin{minipage}[b]{0.24\textwidth}
        \centering
        \includegraphics[width=\textwidth]{Figures/evaluation/DE_Rastrigin_200D.pdf}
        \centering
        \vspace{-0.6cm}
        \subcaption{DE on 200D Rastrigin}
    \end{minipage}
   \hfill
    \begin{minipage}[b]{0.24\textwidth}
        \centering
        \includegraphics[width=\textwidth]{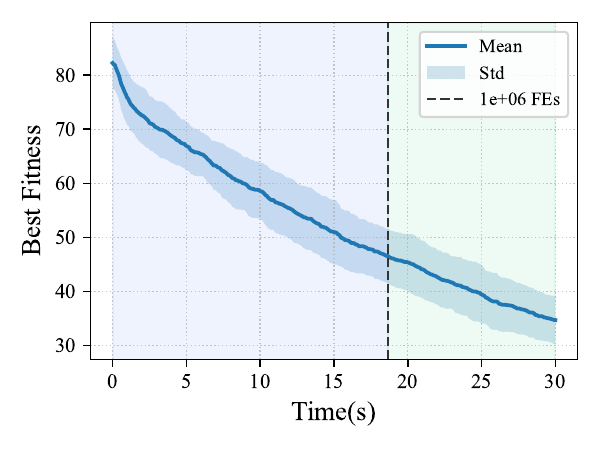}
        \centering
        \vspace{-0.6cm}
        \subcaption{GA-SBX/PM on 200D Rastrigin}
    \end{minipage}
    \hfill
    \begin{minipage}[b]{0.24\textwidth}
        \centering
        \includegraphics[width=\textwidth]{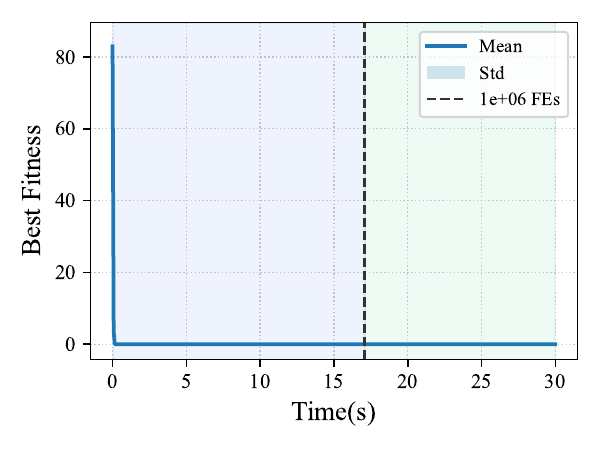}
        \centering
        \vspace{-0.6cm}
        \subcaption{GA-UR/GM on 200D Rastrigin}
    \end{minipage}

    \vspace{0.2cm}

    \begin{minipage}[b]{0.24\textwidth}
        \centering
        \includegraphics[width=\textwidth]{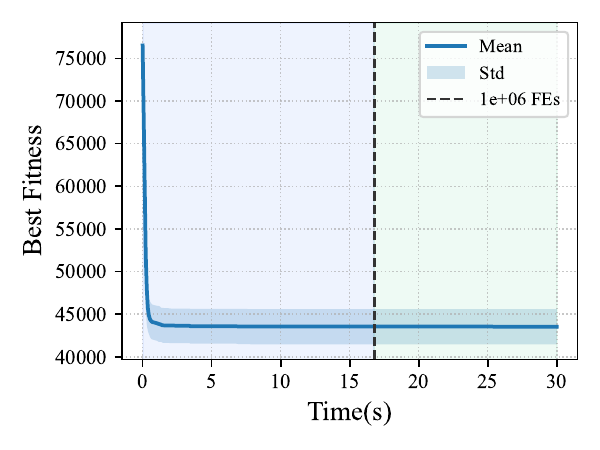}
        \centering
        \vspace{-0.6cm}
        \subcaption{PSO on 200D Schwefel}
    \end{minipage}
   \hfill
   \begin{minipage}[b]{0.24\textwidth}
        \centering
        \includegraphics[width=\textwidth]{Figures/evaluation/DE_Schwefel_200D.pdf}
        \centering
        \vspace{-0.6cm}
        \subcaption{DE on 200D Schwefel}
    \end{minipage}
   \hfill
    \begin{minipage}[b]{0.24\textwidth}
        \centering
        \includegraphics[width=\textwidth]{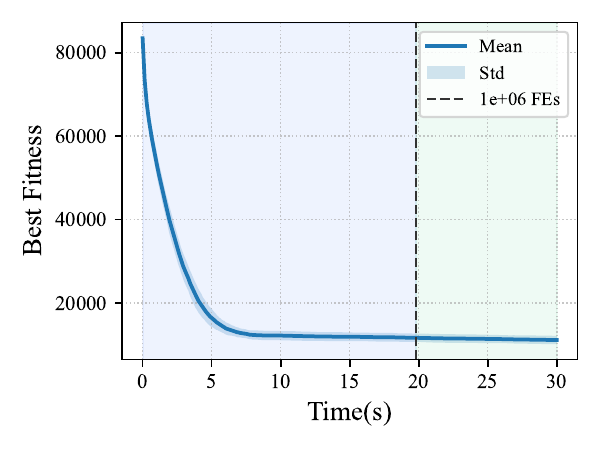}
        \centering
        \vspace{-0.6cm}
        \subcaption{GA-SBX/PM on 200D Schwefel}
    \end{minipage}
    \hfill
    \begin{minipage}[b]{0.24\textwidth}
        \centering
        \includegraphics[width=\textwidth]{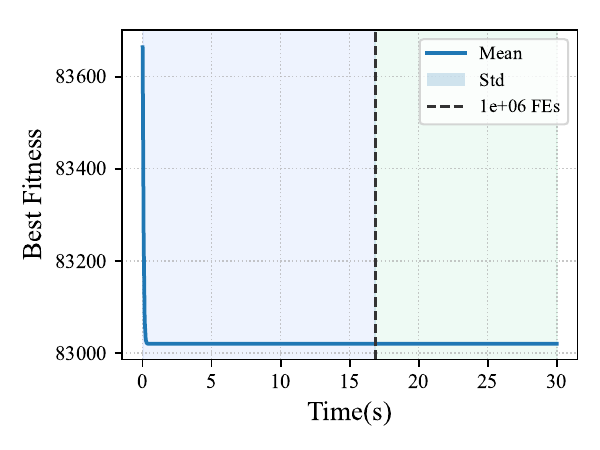}
        \centering
        \vspace{-0.6cm}
        \subcaption{GA-UR/GM on 200D Schwefel}
    \end{minipage}

    \caption{Truncation effect of fixed-FE evaluation under GPU execution. Each curve shows the mean best fitness over 15 independent runs, with shaded regions indicating the standard deviation. The vertical dashed line marks the wall-clock time at which $10^6$ FEs are completed.
    }
 \label{fig:s-truncation-effect}
\end{figure}

Fig.~\ref{fig:s-varying popsize1} presents the impact of different population sizes on computational efficiency of EAs, with values set to {16, 32, 64, 128, 256, 512, 1024, 2048, 4096, 8192}. The problem dimensionality is fixed at 50 across all tests. The left panel shows the \textbf{runtime} over 100 generations, while the right panel reports the \textbf{number of FEs} completed within a fixed 30-second time window. Problem dimensions were fixed at 50 across all configurations. Each setting was repeated 15 times, with the plotted curves representing the mean performance and the shaded bands indicating standard deviations. Different colors denote different hardware platforms. Lower runtime and higher NFE counts are indicative of greater computational efficiency.

\subsection{Exploiting Large Population Size on GPU-based EAs}
\subsubsection{Results on Numerical Problems}\label{app:population-n}
\begin{figure}[H]
    \centering
    \begin{minipage}[b]{0.28\textwidth}
        \centering
        \includegraphics[width=\textwidth]{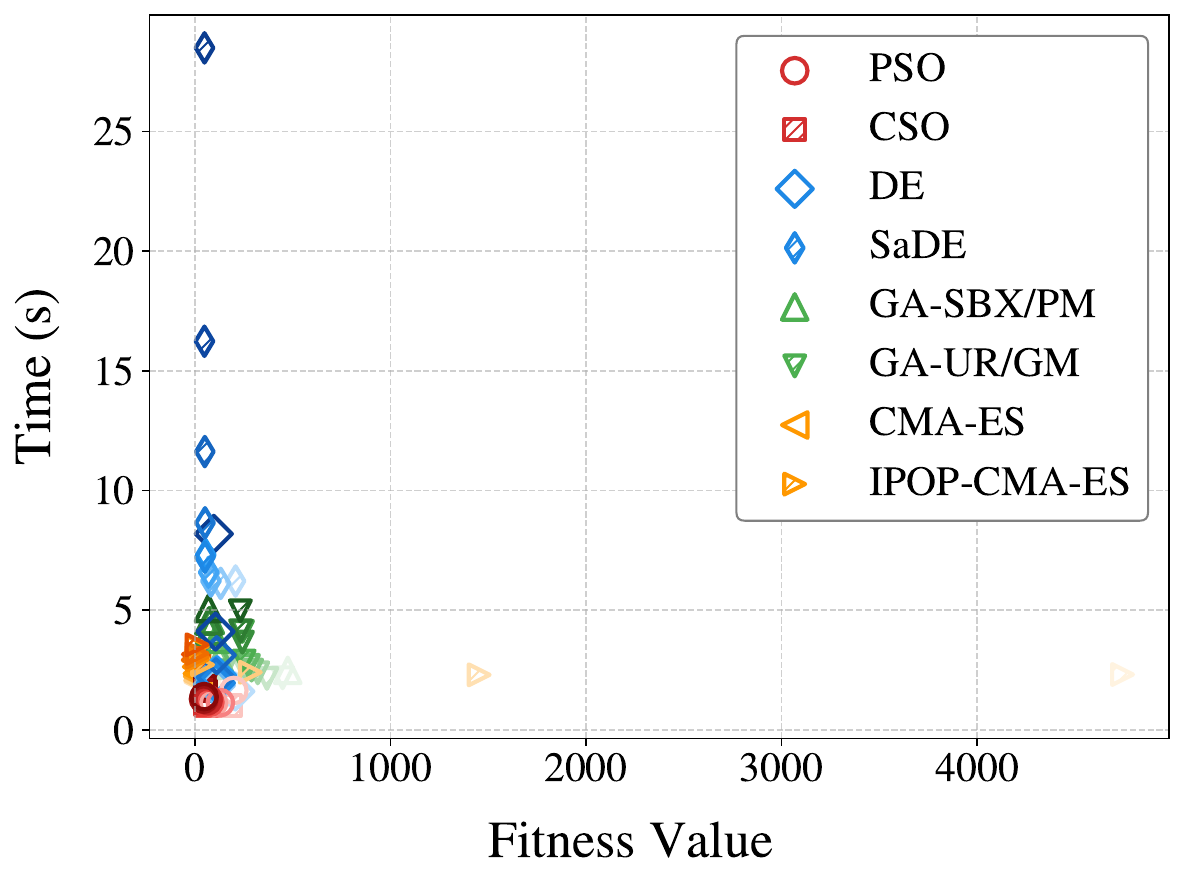}
        \centering
        \vspace{-0.4cm}
        \subcaption{$f_{a_{2}}$}
    \end{minipage}
  \hspace{0.2cm}
  \centering
    \begin{minipage}[b]{0.28\textwidth}
        \centering
        \includegraphics[width=\textwidth]{Figures/sub/3/3090/numerical/scatter/Ackley-all.pdf}
        \centering
        \vspace{-0.4cm}
        \subcaption{$f_{a_{6}}$}
    \end{minipage}
   \hspace{0.2cm}
  \centering
    \begin{minipage}[b]{0.28\textwidth}
        \centering
        \includegraphics[width=\textwidth]{Figures/sub/3/3090/numerical/scatter/Sphere-all.pdf}
        \centering
        \vspace{-0.4cm}
        \subcaption{$f_{a_{10}}$}
    \end{minipage}

  \vspace{0.4cm}

    \begin{minipage}[b]{0.28\textwidth}
        \centering
        \includegraphics[width=\textwidth]{Figures/sub/3/3090/numerical/scatter/DTLZ1-all.pdf}
        \centering
        \vspace{-0.4cm}
        \subcaption{$f_{a_{11}}$}
    \end{minipage}
    \hspace{0.2cm}
  \centering
    \begin{minipage}[b]{0.28\textwidth}
        \centering
        \includegraphics[width=\textwidth]{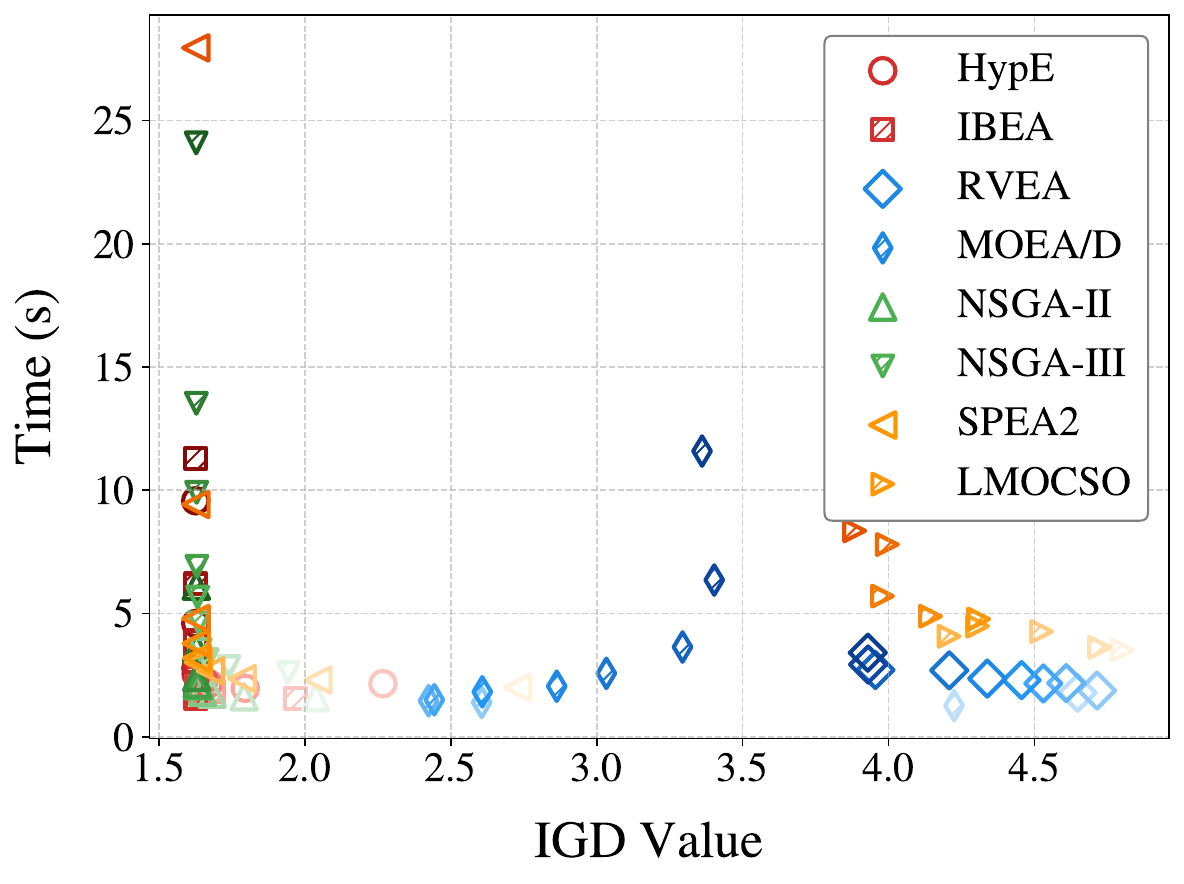}
        \centering
        \vspace{-0.4cm}
        \subcaption{$f_{a_{15}}$}
    \end{minipage}
   \hspace{0.2cm}
  \centering
    \begin{minipage}[b]{0.28\textwidth}
        \centering
        \includegraphics[width=\textwidth]{Figures/sub/3/3090/numerical/scatter/ZDT2-all.pdf}
        \centering
        \vspace{-0.4cm}
        \subcaption{$f_{a_{19}}$}
    \end{minipage}
    \caption{Performance comparison of EAs under varying population sizes, evaluated in terms of solution quality and time consumed over 100 iterations. Lower fitness/IGD values denote better performance. Results represent averaged performance values across 15 independent runs. Markers represent individual algorithms, color-coded from light to dark to indicate increasing population sizes (from 16 to 8192).}

    \label{fig:s-pop-curve}
\end{figure}

We evaluate EA performance under two distinct computational constraints: (i) fixed-generation (100 iterations) and (ii) fixed-time (30-second) conditions, Each algorithm was tested across ten population sizes (range from 16 to 8192) over 100 iterations, conducting 15 independent repetitions on an NVIDIA GeForce RTX-3090 GPU, with solution quality, runtime, and NFEs completed serving as key performance metrics. Table~\ref{tab:s-pop-pv} presents the mean of best fitness/IGD value obtained over 100 iterations with varying population size on NVIDIA GeForce RTX-3090. Fig.~\ref{fig:s-pop-curve} complements Table~\ref{tab:s-pop-pv} with Pareto front visualizations.

\begin{table}[htbp]
\centering
\caption{Mean of Best Fitness / IGD value Obtained under 100 Iterations with 10 Population Size on NVIDIA GeForce RTX-3090.}
\small
\resizebox{0.7\textwidth}{!}{
\begin{tabular}{llcccccccc}
\toprule
Func. & \makecell{Population\\Size} & \makecell{PSO} & \makecell{CSO} & \makecell{DE} & \makecell{SaDE} & \makecell{CMA-\\ES} & \makecell{IPOP-\\CMA-ES} & \makecell{GA-\\SBX/PM} & \makecell{GA-\\UR/GM} \\
\midrule
\addlinespace[5pt]
\multirow{10}{*}{$f_{a_2}$}
& 16 & 174.97 & 200.27 & 192.59 & 195.33 & 470.65 & 453.65 & 19.48 & 3892.91 \\
& 32 & 118.67 & 71.97 & 84.99 & 124.98 & 177.67 & 372.37 & 18.21 & 1386.11 \\
& 64 & 119.68 & 56.98 & 93.49 & 78.19 & 221.95 & 316.64 & 17.34 & 308.84 \\
& 128 & 49.14 & 59.27 & 99.33 & 66.35 & 112.93 & 289.93 & 16.36 & 40.26 \\
& 256 & 49.10 & 55.38 & 113.14 & 56.54 & 93.57 & 273.92 & 15.75 & 16.71 \\
& 512 & 73.81 & 54.50 & 112.86 & 52.44 & 105.47 & 255.14 & 13.60 & 13.33 \\
& 1024 & 49.10 & 54.17 & 107.47 & 54.10 & 79.43 & 244.68 & 9.91 & 12.38 \\
& 2048 & 49.08 & 53.71 & 114.48 & 50.09 & 75.97 & 243.47 & 5.23 & 13.12 \\
& 4096 & 49.08 & 53.54 & 102.78 & 49.85 & 75.71 & 234.48 & 5.39 & 12.36 \\
& 8192 & 49.08 & 52.81 & 97.04 & 49.99 & 73.51 & 233.31 & 5.28 & 12.00 \\ \hline
\addlinespace[5pt]
\multirow{10}{*}{$f_{a_6}$}
& 16 & 17.60 & 14.71 & 15.45 & 10.84 & 2.20 & 0.78 & 21.31 & 21.30 \\
& 32 & 12.46 & 10.63 & 12.68 & 6.62 & 2.19 & 0.00 & 4.75 & 21.27 \\
& 64 & 10.54 & 9.34 & 14.81 & 4.50 & 1.93 & 0.00 & 2.54 & 21.28 \\
& 128 & 7.59 & 9.66 & 18.53 & 3.23 & 1.54 & 0.00 & 1.31 & 21.22 \\
& 256 & 5.94 & 9.31 & 19.88 & 2.42 & 1.45 & 0.00 & 1.56 & 21.02 \\
& 512 & 3.93 & 9.32 & 20.09 & 1.73 & 1.32 & 0.00 & 3.04 & 18.98 \\
& 1024 & 3.22 & 8.82 & 20.37 & 0.63 & 0.95 & 0.00 & 3.21 & 11.72 \\
& 2048 & 2.41 & 8.94 & 20.31 & 0.42 & 0.24 & 0.00 & 0.76 & 3.37 \\
& 4096 & 2.05 & 8.79 & 20.19 & 0.31 & 0.15 & 0.00 & 0.03 & 0.09 \\
& 8192 & 1.51 & 8.62 & 20.29 & 0.24 & 0.10 & 0.00 & 0.00 & 0.00 \\ \hline
\addlinespace[5pt]
\multirow{10}{*}{$f_{a_{10}}$}
& 16   & 51.31 & 28.19 & 44.67 & 16.08 & 12.77 & 2.00 & 2.25 & 843.69 \\
& 32   & 32.00 & 13.02 & 19.08 & 3.13  & 8.53  & 0.00 & 1.08 & 655.66 \\
& 64   & 13.90 & 8.36  & 27.43 & 0.90  & 4.04  & 0.00 & 0.03 & 449.58 \\
& 128  & 4.58  & 7.08  & 41.04 & 0.20  & 1.46  & 0.00 & 0.01 & 351.15 \\
& 256  & 1.21  & 7.28  & 79.10 & 0.08  & 0.99  & 0.00 & 0.01 & 179.21 \\
& 512  & 0.45  & 6.82  & 116.50 & 0.02 & 0.64  & 0.00 & 0.10 & 56.78 \\
& 1024 & 0.18  & 6.97  & 117.63 & 0.01 & 0.55  & 0.00 & 0.05 & 6.54 \\
& 2048 & 0.03  & 5.99  & 117.95 & 0.01 & 0.51  & 0.00 & 0.00 & 0.11 \\
& 4096 & 0.00  & 6.34  & 137.86 & 0.00 & 0.52  & 0.00 & 0.00 & 0.00 \\
& 8192 & 0.00  & 5.93  & 133.91 & 0.00 & 0.38  & 0.00 & 0.00 & 0.00 \\
\bottomrule
\toprule
Func. & \makecell{Population\\Size} & HypE & IBEA & RVEA & MOEA/D & NSGA-II & NSGA-III & SPEA2 & LMOCSO \\
\midrule
\addlinespace[5pt]
\multirow{10}{*}{$f_{a_{11}}$}
& 16      & 314.18 & 235.91 & 586.23  & 1116.02 & 365.79  & 299.74   & 497.86 & 871.80 \\
& 32      & 188.39 & 172.11 & 740.06  & 794.18  & 281.19  & 228.85   & 318.81 & 851.12 \\
& 64      & 148.97 & 122.81 & 305.75  & 824.91  & 217.35  & 287.93   & 202.87 & 817.60 \\
& 128     & 147.71 & 110.14 & 298.39  & 918.94  & 261.36  & 237.25   & 186.41 & 753.89 \\
& 256     & 123.85 & 100.59 & 323.10  & 1011.99 & 232.98  & 242.56   & 142.39 & 716.83 \\
& 512     & 133.94 & 110.15 & 334.38  & 1036.51 & 227.82  & 224.67   & 119.91 & 672.64 \\
& 1024    & 140.91 & 92.78  & 332.80  & 1051.77 & 220.06  & 196.49   & 140.50 & 687.58 \\
& 2048    & 144.61 & 110.62 & 362.61  & 1046.40 & 212.60  & 197.79   & 111.06 & 616.33 \\
& 4096    & 139.63 & 110.71 & 1038.03 & 1029.58 & 193.53  & 192.94   & 111.44 & 635.26 \\
& 8192    & 137.87 & 105.74 & 1068.18 & 1014.46 & 180.24  & 177.24   & 111.22 & 618.74 \\ \hline
\addlinespace[5pt]
\multirow{10}{*}{$f_{a_{15}}$}
& 16      & 2.2677 & 1.9680 & 4.6477 & 4.2245 & 2.0371  & 1.9434   & 2.7285 & 4.8017 \\
& 32      & 1.7963 & 1.6865 & 4.7141 & 2.6062 & 1.7921  & 1.7401   & 2.0449 & 4.7233 \\
& 64      & 1.6712 & 1.6412 & 4.6089 & 2.4241 & 1.6825  & 1.6658   & 1.7853 & 4.5242 \\
& 128     & 1.6537 & 1.6310 & 4.5304 & 2.4438 & 1.6496  & 1.6419   & 1.6769 & 4.2075 \\
& 256     & 1.6364 & 1.6281 & 4.4557 & 2.6082 & 1.6390  & 1.6373   & 1.6361 & 4.3052 \\
& 512     & 1.6302 & 1.6269 & 4.3382 & 2.8632 & 1.6333  & 1.6327   & 1.6307 & 4.3086 \\
& 1024    & 1.6282 & 1.6266 & 4.2081 & 3.0332 & 1.6313  & 1.6304   & 1.6283 & 4.1439 \\
& 2048    & 1.6273 & 1.6264 & 3.9551 & 3.2942 & 1.6298  & 1.6294   & 1.6272 & 3.9793 \\
& 4096    & 1.6266 & 1.6264 & 3.9300 & 3.4027 & 1.6288  & 1.6284   & 1.6267 & 3.9960 \\
& 8192    & 1.6264 & 1.6263 & 3.9288 & 3.3610 & 1.6283  & 1.6279   & 1.6265 & 3.8841 \\\hline
\addlinespace[5pt]
\multirow{10}{*}{$f_{a_{19}}$}
& 16      & 3.1467 & 3.4636 & 3.6157 & 5.1363 & 2.7801  & 3.0238   & 4.5059 & 3.4256 \\
& 32      & 2.3067 & 2.7873 & 2.9867 & 3.7253 & 2.0274  & 2.2809   & 3.0948 & 3.0772 \\
& 64      & 1.805  & 1.9318 & 2.4781 & 3.2443 & 1.7804  & 1.7887   & 2.5331 & 2.4471 \\
& 128     & 1.6302 & 1.6861 & 5.1294 & 3.2506 & 1.6315  & 1.6316   & 1.8225 & 5.1412 \\
& 256     & 1.591  & 1.5907 & 5.1482 & 3.0809 & 1.6003  & 1.6034   & 1.665  & 5.1944 \\
& 512     & 1.5825 & 1.5801 & 5.0399 & 3.1803 & 1.5901  & 1.5897   & 1.5906 & 5.191  \\
& 1024    & 1.5791 & 1.5785 & 5.0725 & 3.4082 & 1.5862  & 1.585    & 1.5793 & 5.0912 \\
& 2048    & 1.5784 & 1.5779 & 4.9281 & 3.5546 & 1.5841  & 1.5833   & 1.5784 & 4.974  \\
& 4096    & 1.5781 & 1.5777 & 4.9346 & 3.8128 & 1.5835  & 1.5822   & 1.5779 & 4.9351 \\
& 8192    & 1.578  & 1.5776 & 4.7815 & 4.0614 & 1.583   & 1.5819   & 1.5777 & 4.8466 \\
\bottomrule
\end{tabular}
}
\label{tab:s-pop-pv}
\end{table}

\begin{figure}[H]
    \centering
    \begin{minipage}[b]{0.7\textwidth}
        \centering
        \includegraphics[width=\textwidth]{Figures/sub/3/legend.pdf}
    \end{minipage}
    \vspace{0.3cm}

    \centering
    \begin{minipage}[b]{0.48\textwidth}
        \centering
        \includegraphics[width=0.48\textwidth]{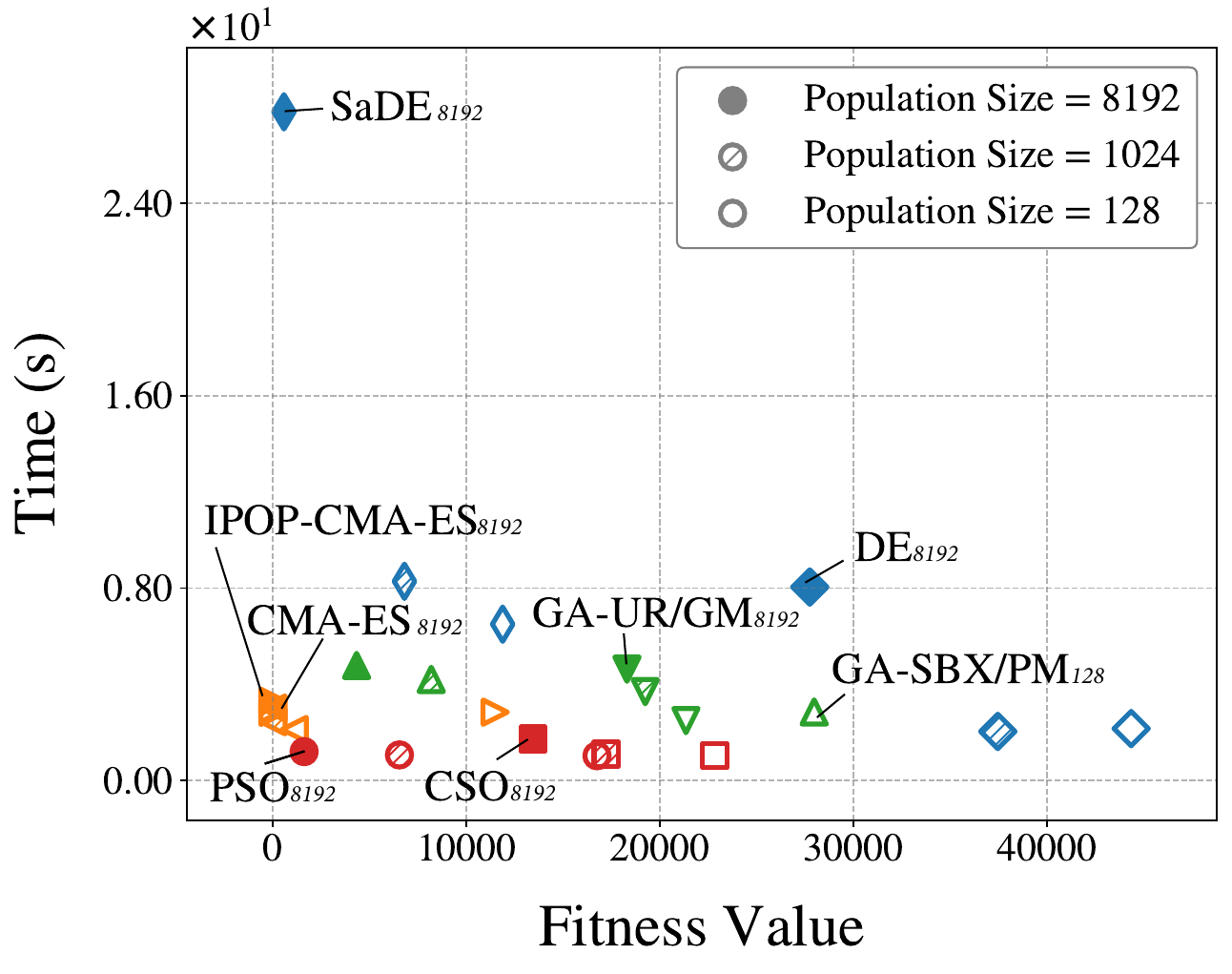}
        \centering
        \includegraphics[width=0.48\textwidth]{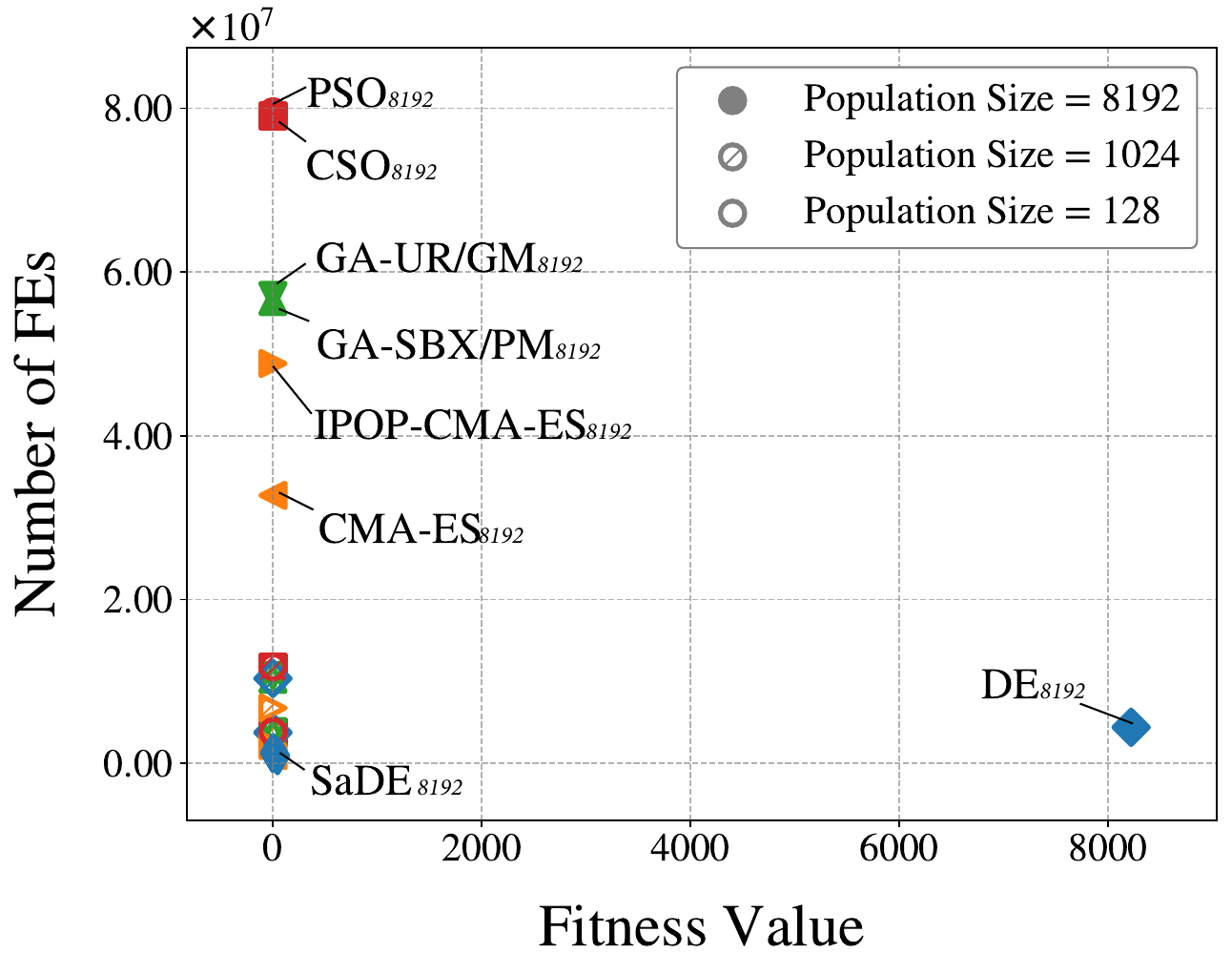}

        \centering
        \subcaption{CEC2022 F1}
    \end{minipage}
    \vspace{0.3cm}
    \begin{minipage}[b]{0.48\textwidth}
        \centering
        \includegraphics[width=0.48\textwidth]{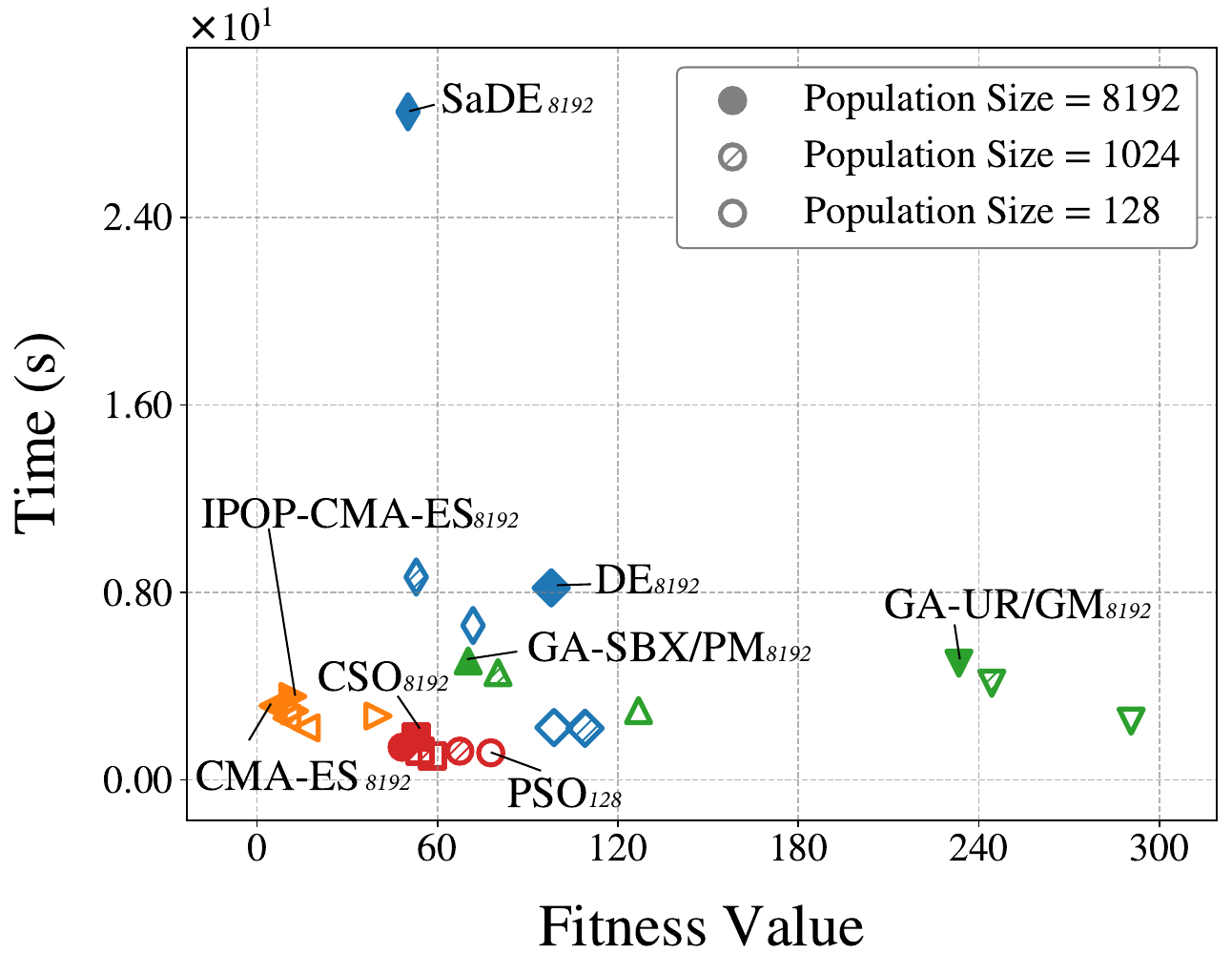}
        \centering
        \includegraphics[width=0.48\textwidth]{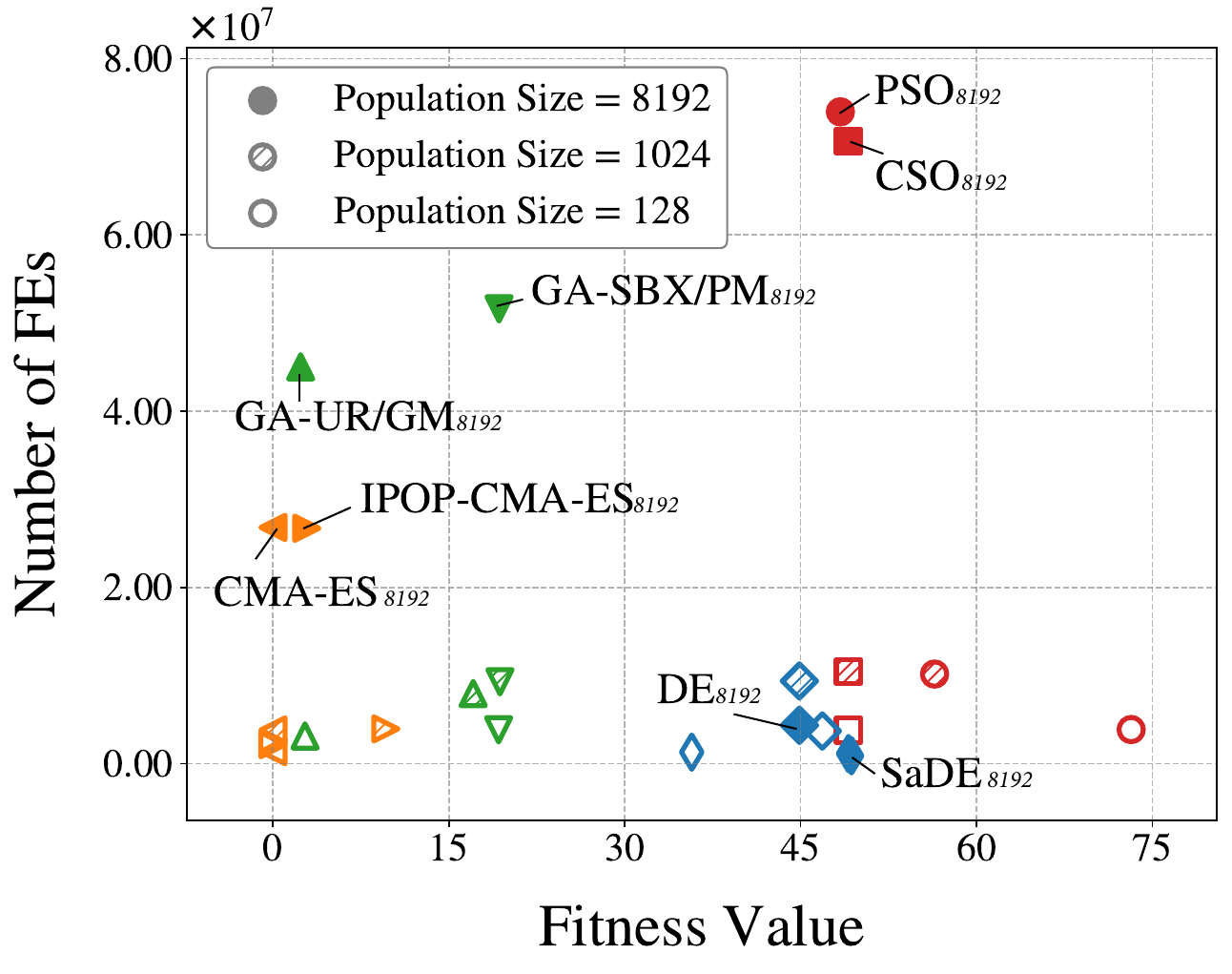}

        \centering
        \subcaption{CEC2022 F2}
    \end{minipage}
    \vspace{0.3cm}
    \begin{minipage}[b]{0.48\textwidth}
        \centering
        \includegraphics[width=0.48\textwidth]{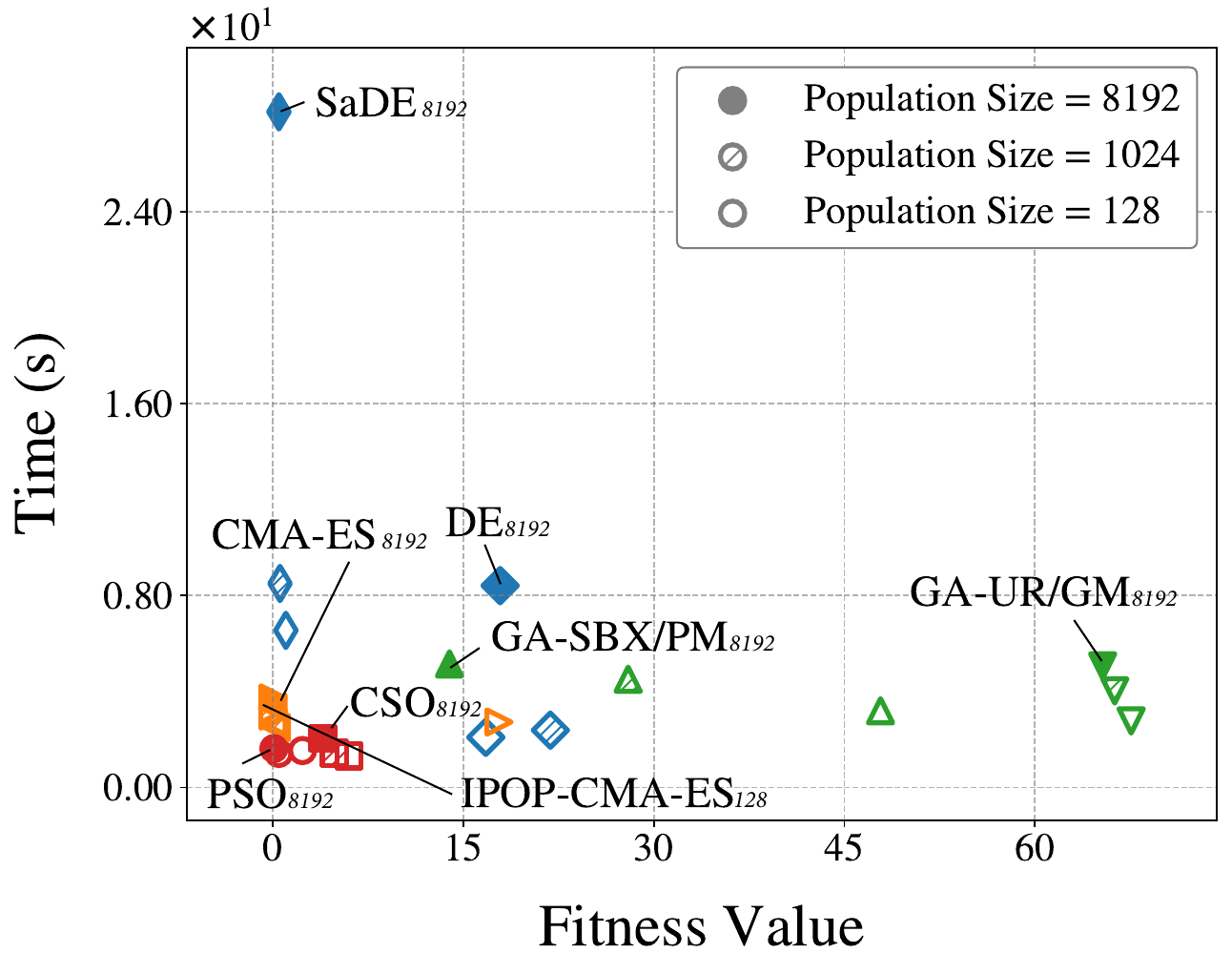}
        \centering
        \includegraphics[width=0.48\textwidth]{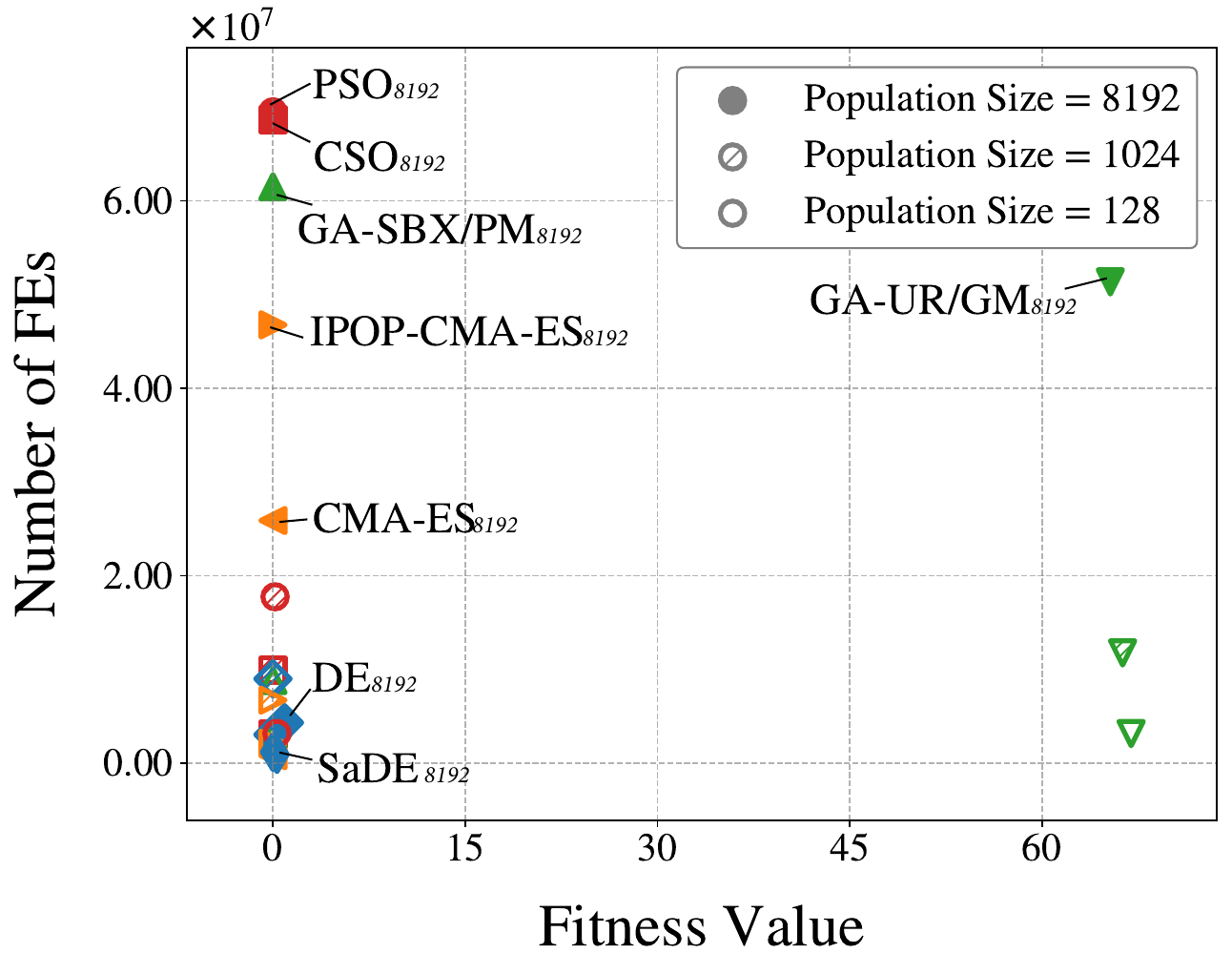}

        \centering
        \subcaption{CEC2022 F3}
    \end{minipage}
    \vspace{0.3cm}
    \begin{minipage}[b]{0.48\textwidth}
        \centering
        \includegraphics[width=0.48\textwidth]{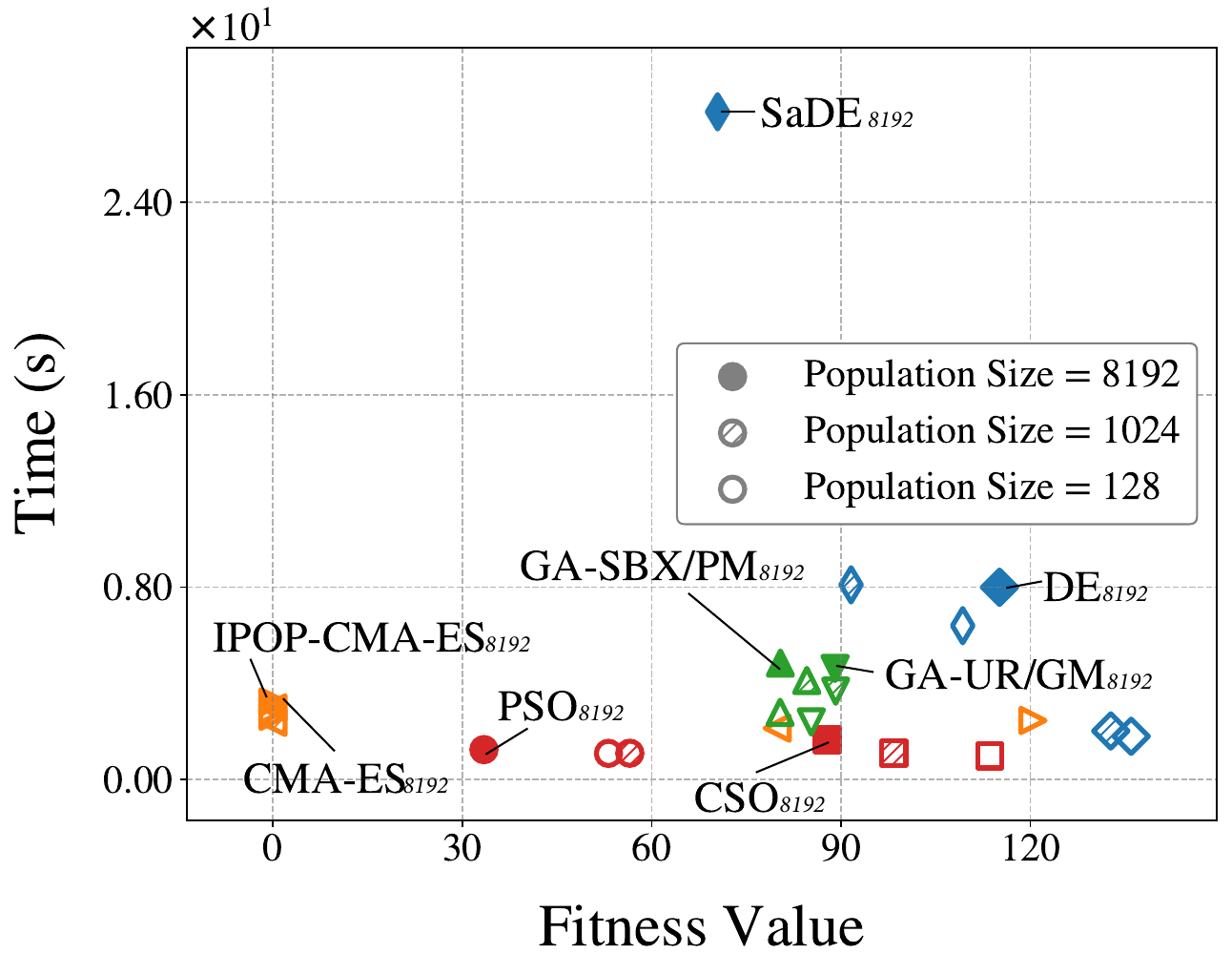}
        \centering
        \includegraphics[width=0.48\textwidth]{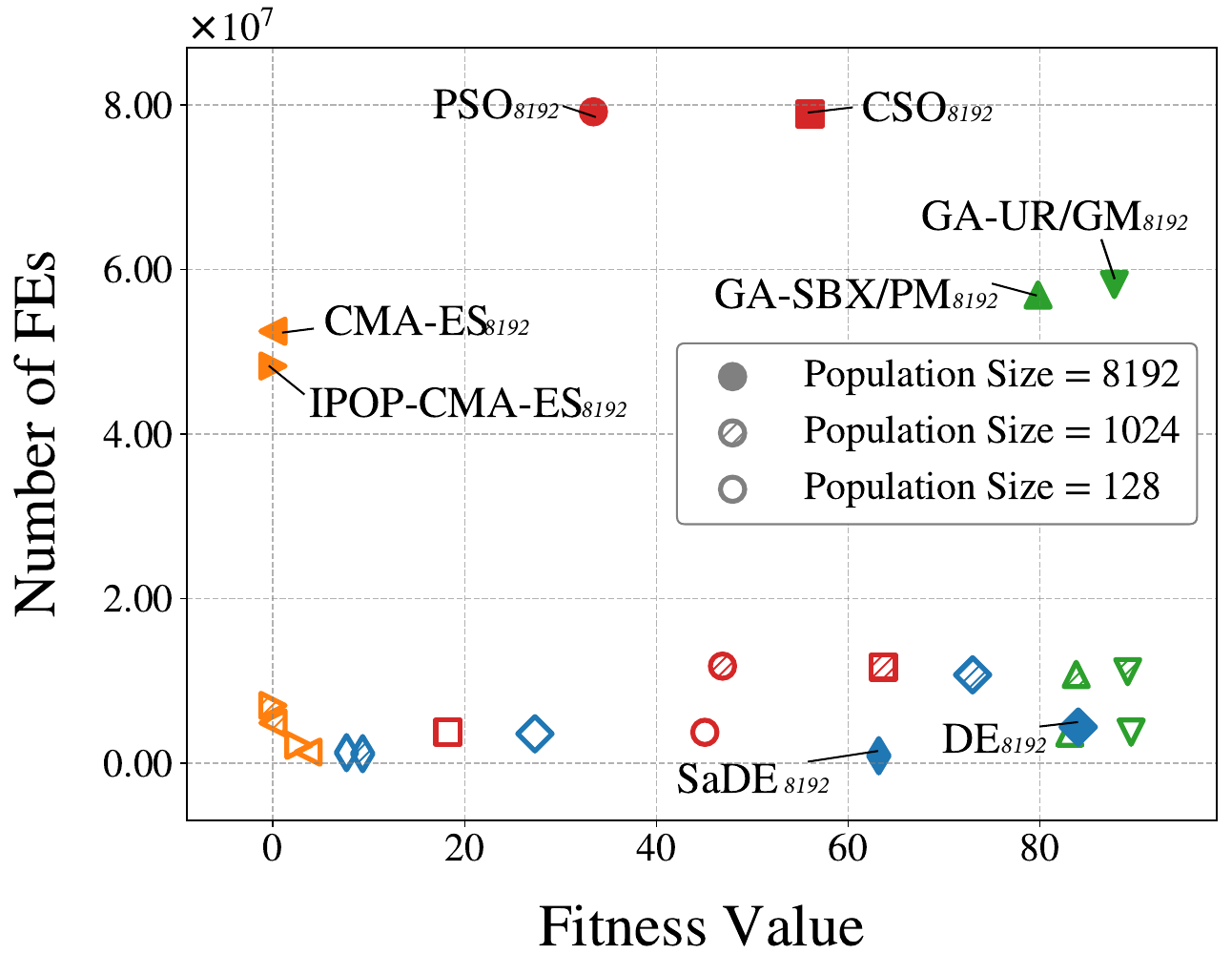}
        \centering
        \subcaption{CEC2022 F4}
    \end{minipage}
    \vspace{0.3cm}
    \begin{minipage}[b]{0.48\textwidth}
        \centering
        \includegraphics[width=0.48\textwidth]{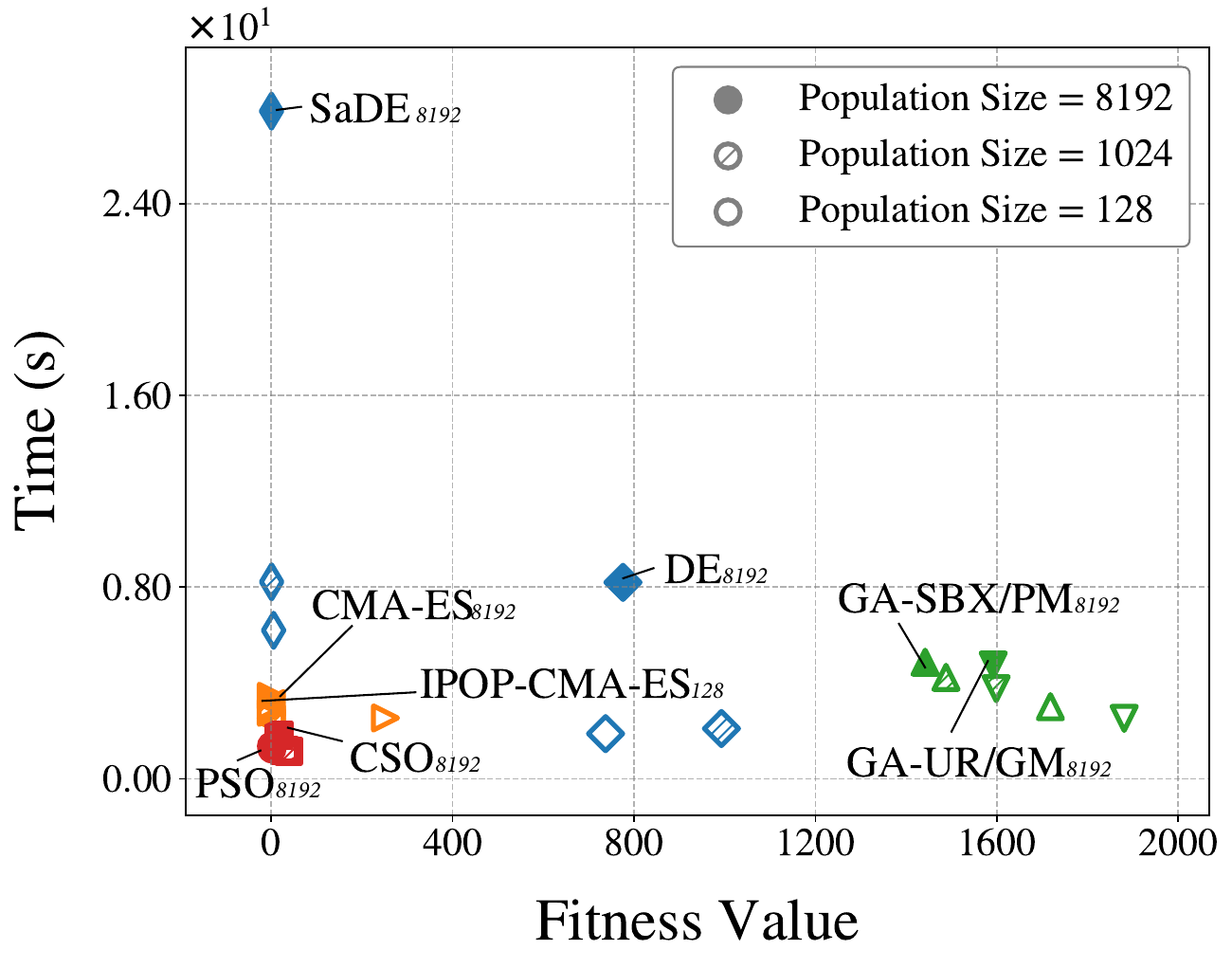}
        \centering
        \includegraphics[width=0.48\textwidth]{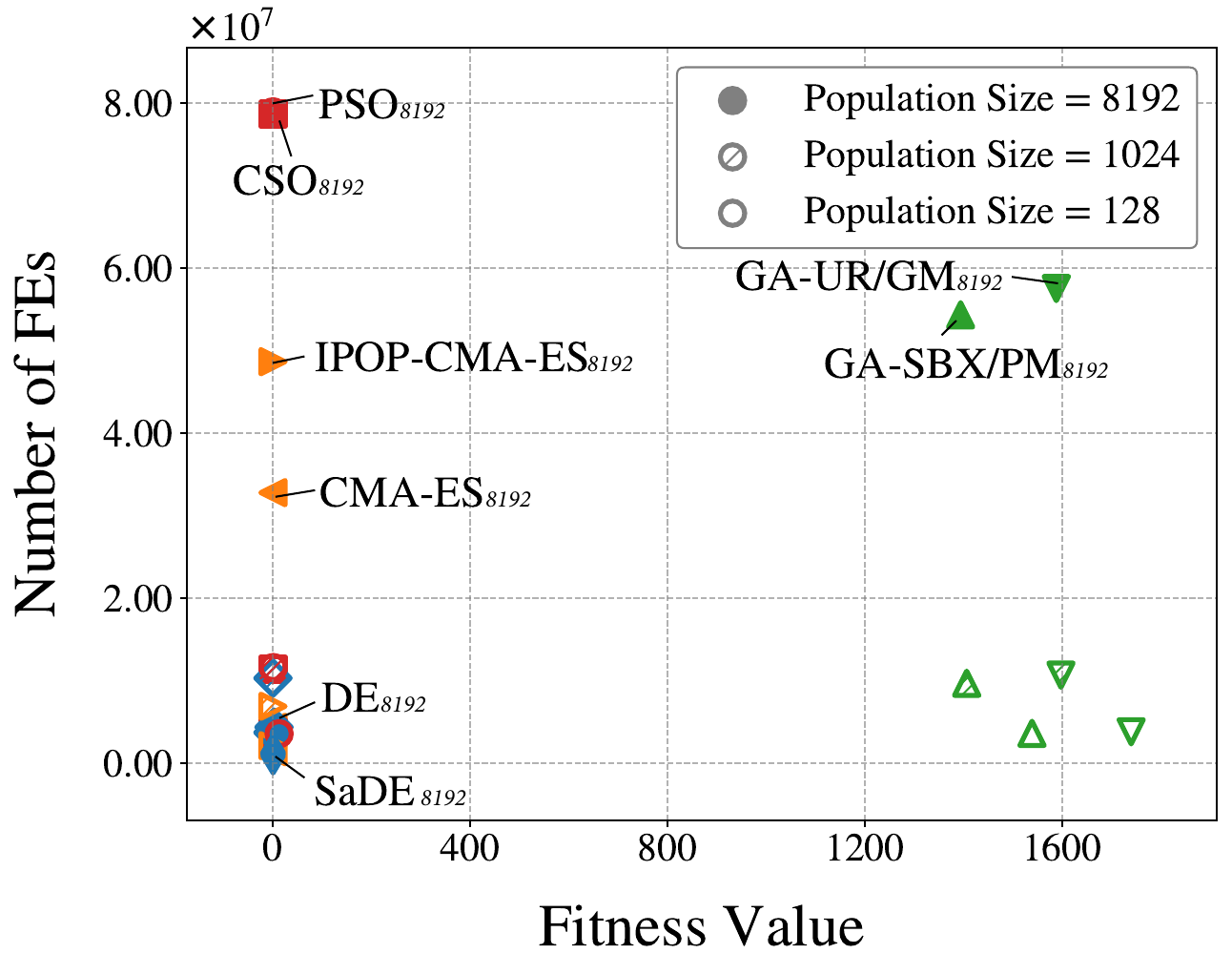}
        \centering
        \subcaption{CEC2022 F5}
    \end{minipage}
    \vspace{0.3cm}
    \centering
    \begin{minipage}[b]{0.48\textwidth}
        \centering
        \includegraphics[width=0.48\textwidth]{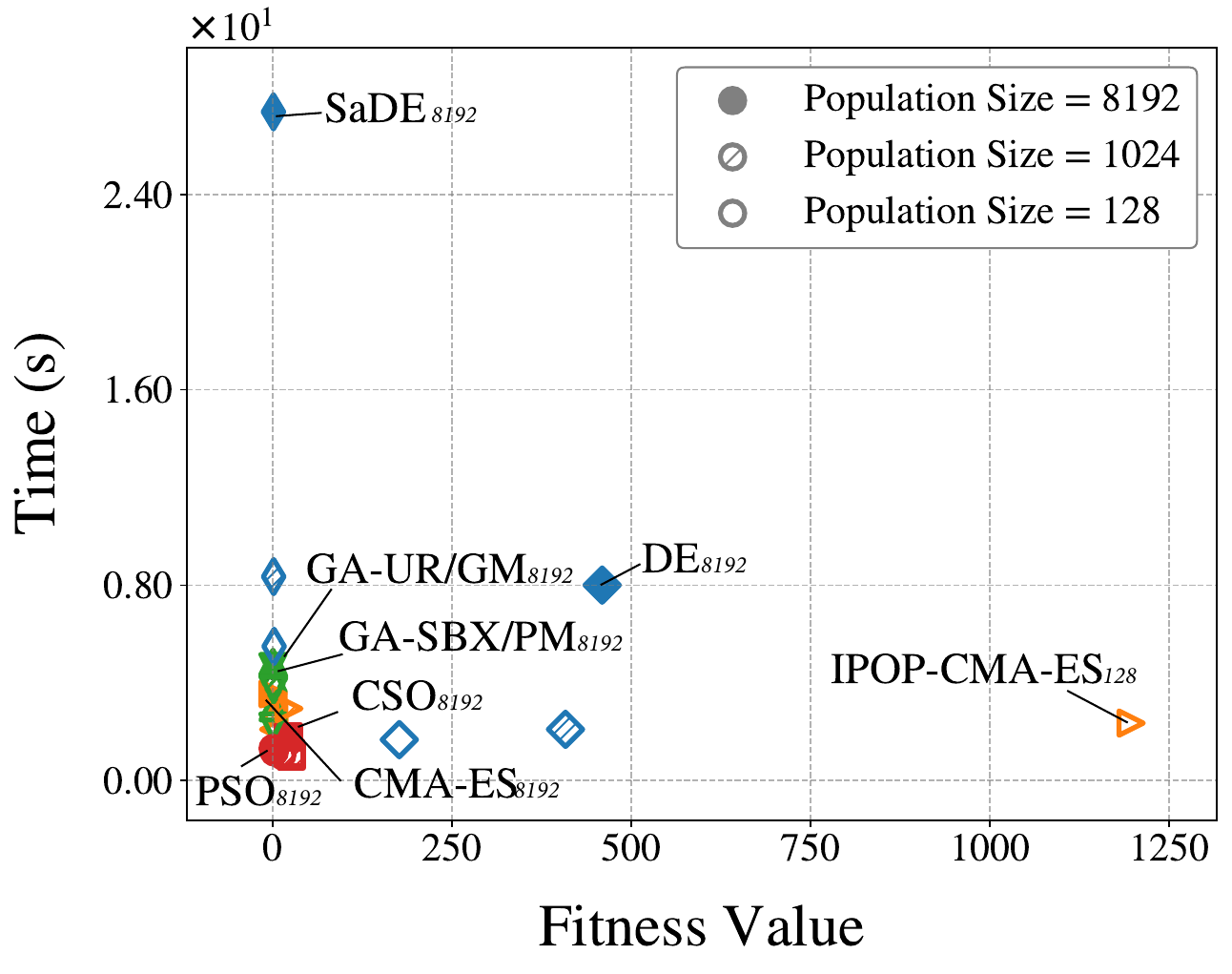}
        \centering
        \includegraphics[width=0.48\textwidth]{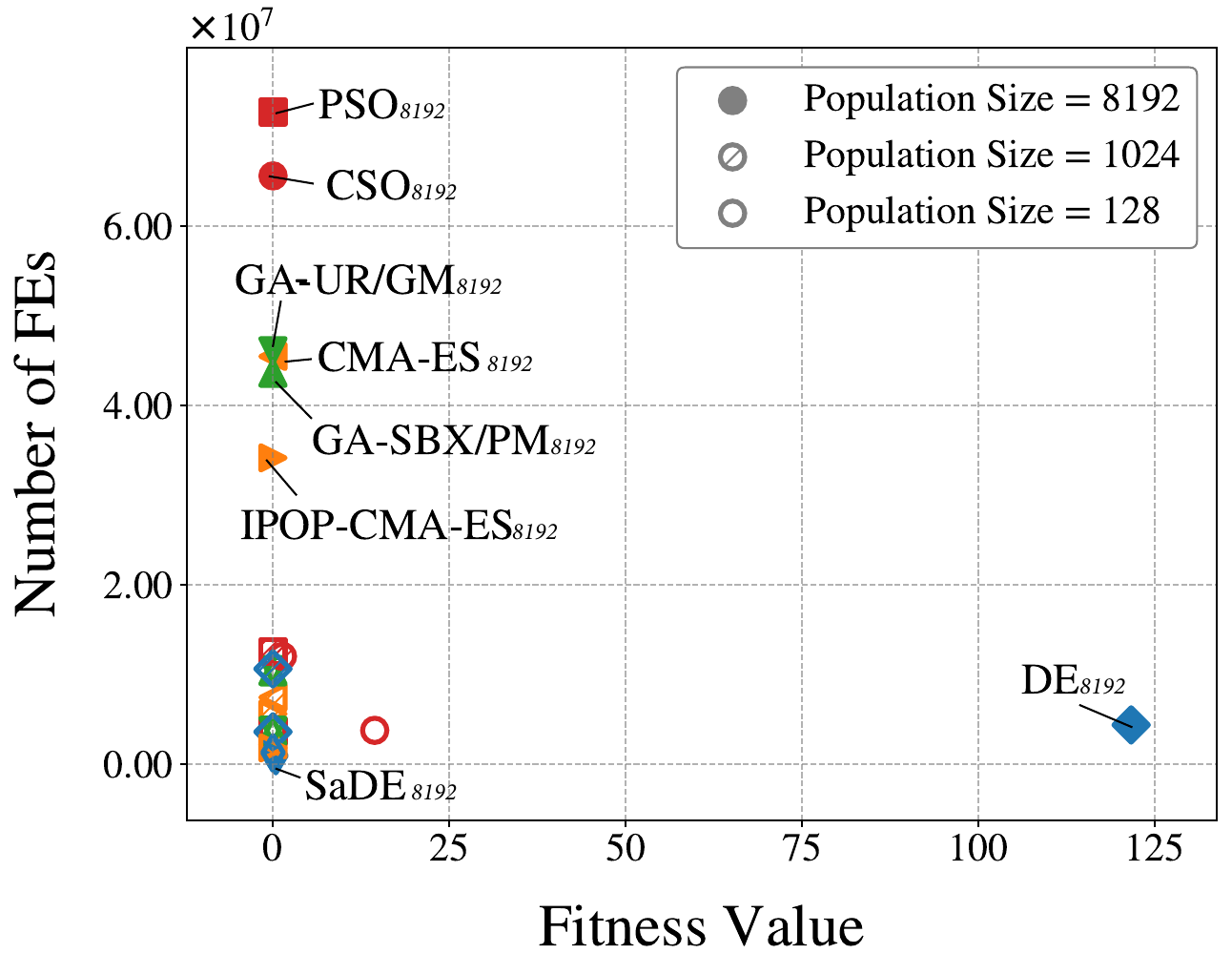}
        \centering
        \subcaption{Griewank}
    \end{minipage}
    \vspace{0.3cm}
    \centering
    \begin{minipage}[b]{0.48\textwidth}
        \centering
        \includegraphics[width=0.48\textwidth]{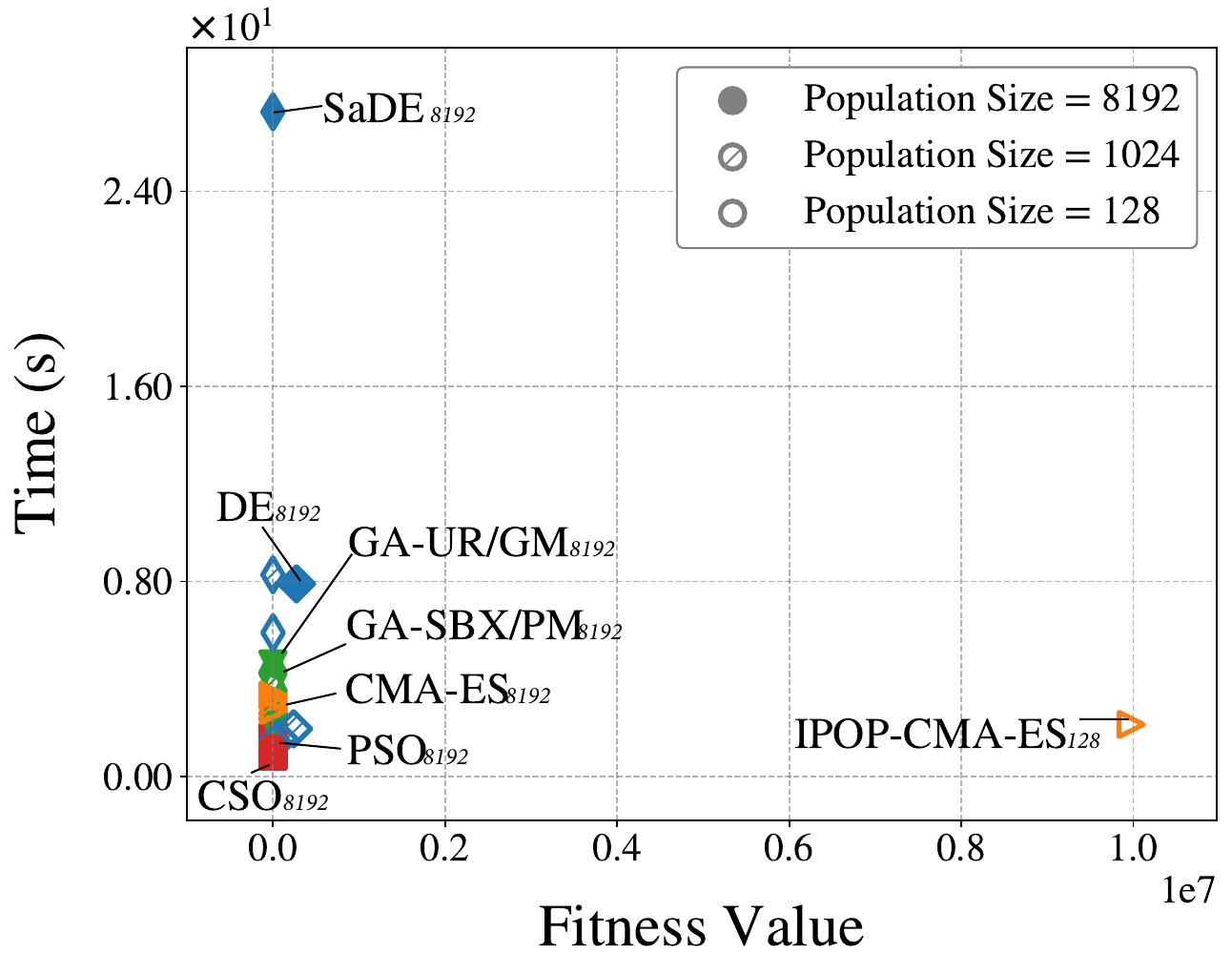}
        \centering
        \includegraphics[width=0.48\textwidth]{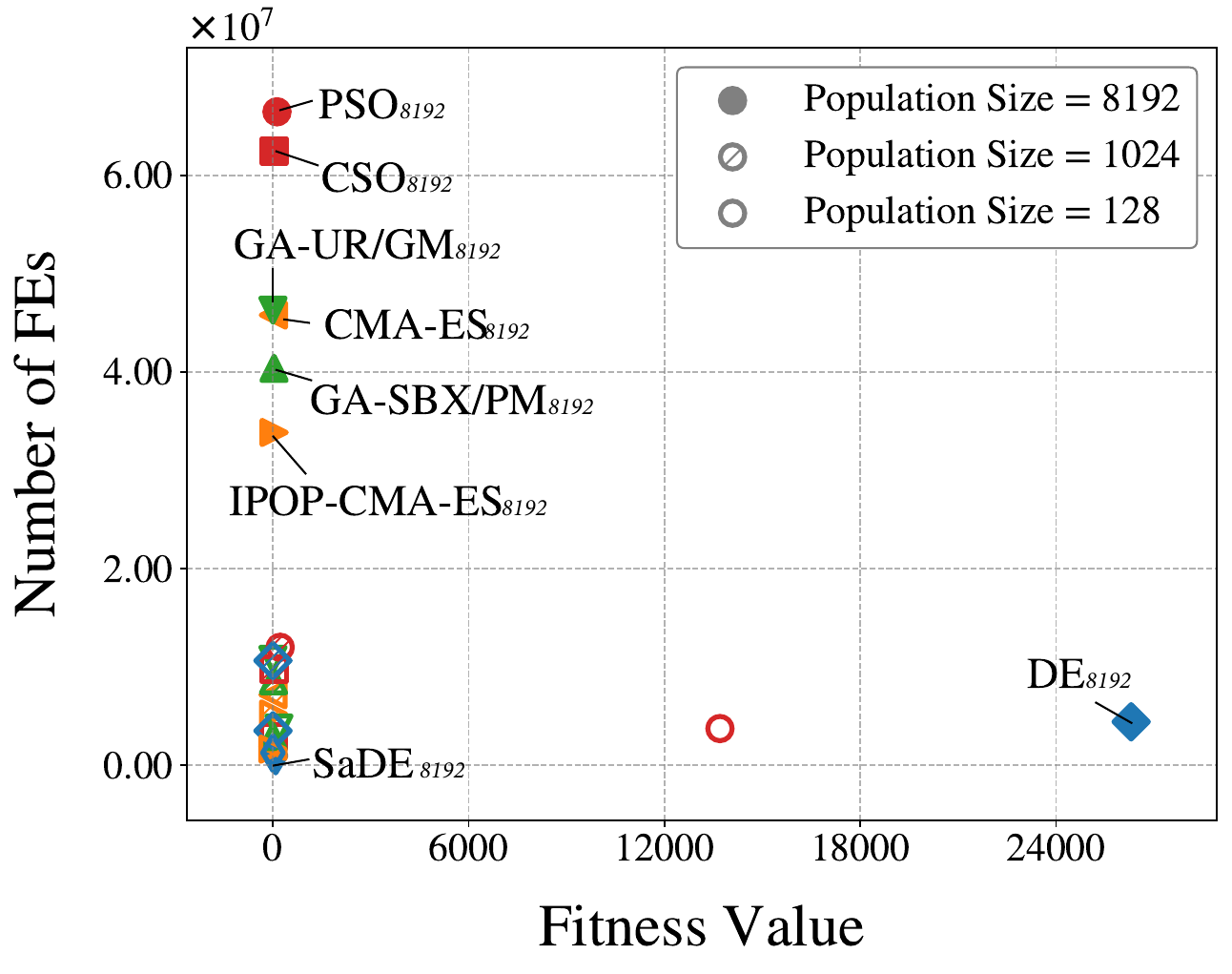}
        \centering
        \subcaption{Rosenbrock}
    \end{minipage}
    \vspace{0.3cm}
    \centering
    \begin{minipage}[b]{0.48\textwidth}
        \centering
        \includegraphics[width=0.48\textwidth]{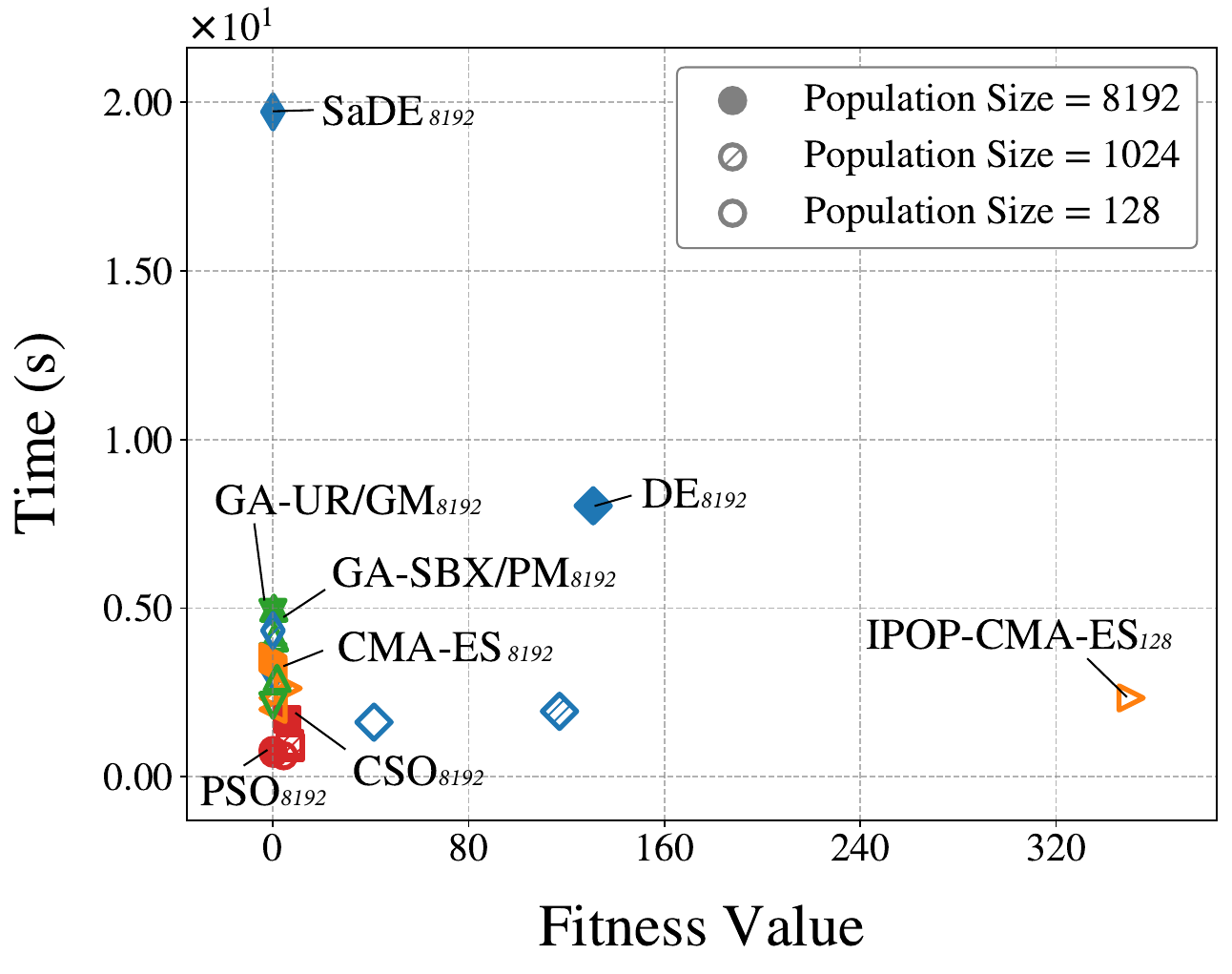}
        \centering
        \includegraphics[width=0.48\textwidth]{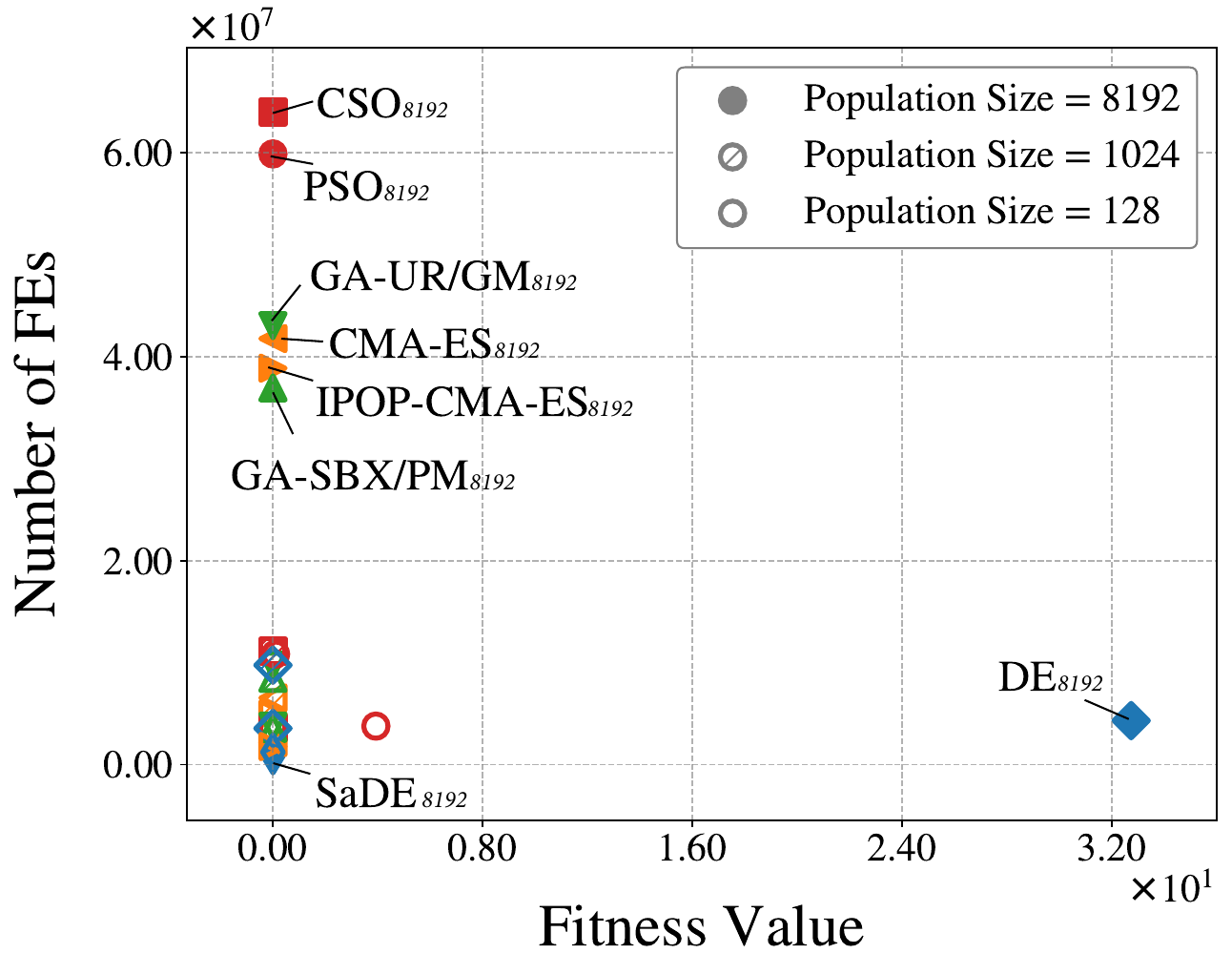}

        \centering
        \subcaption{Sphere}
    \end{minipage}
    \vspace{0.3cm}

    \caption{Performance comparison of SOEAs tested on numerical problems under varying population sizes, evaluated in terms of solution quality and computational efficiency under fixed-generation (100 iterations) versus fixed-time (30-second) constraints for EAs on an NVIDIA RTX-3090 GPU. Lower fitness/IGD values denote better performance. Results represent averaged performance values across 15 independent runs. Marker styles indicate population scales: hollow symbols for small populations (128), forward-slash-filled symbols for medium populations (1024), and solid symbols for large populations (8192). Different marker shapes distinguish between algorithms.}
    \label{fig:s-varying-popsize-fixed-generation}
\end{figure}

\begin{figure}[H]
    \centering
    \begin{minipage}[b]{0.7\textwidth}
        \centering
        \includegraphics[width=\textwidth]{Figures/sub/3/legend-m.pdf}
    \end{minipage}
    \vspace{0.3cm}

    \centering
    \begin{minipage}[b]{0.48\textwidth}
        \centering
        \includegraphics[width=0.48\textwidth]{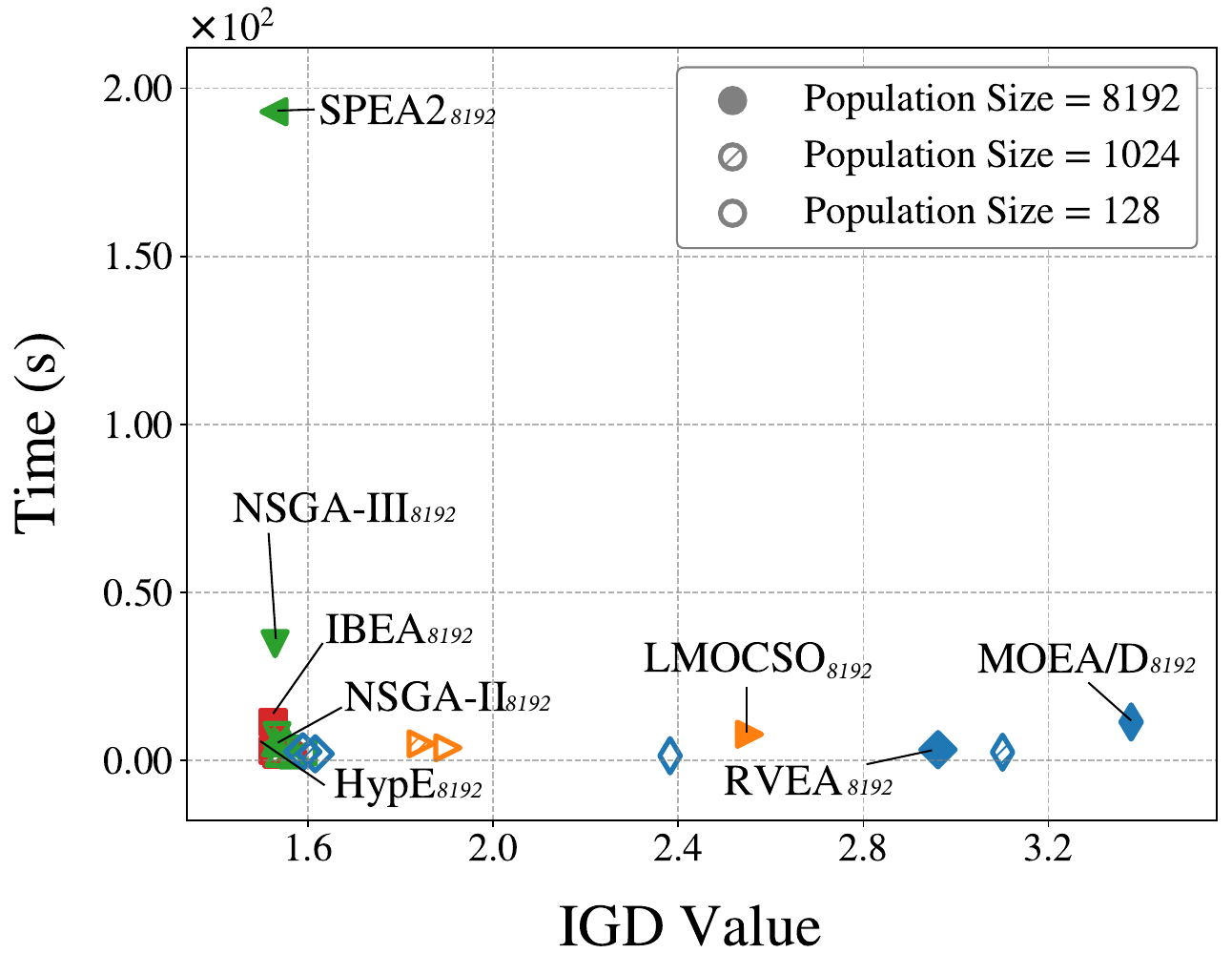}
        \centering
        \includegraphics[width=0.48\textwidth]{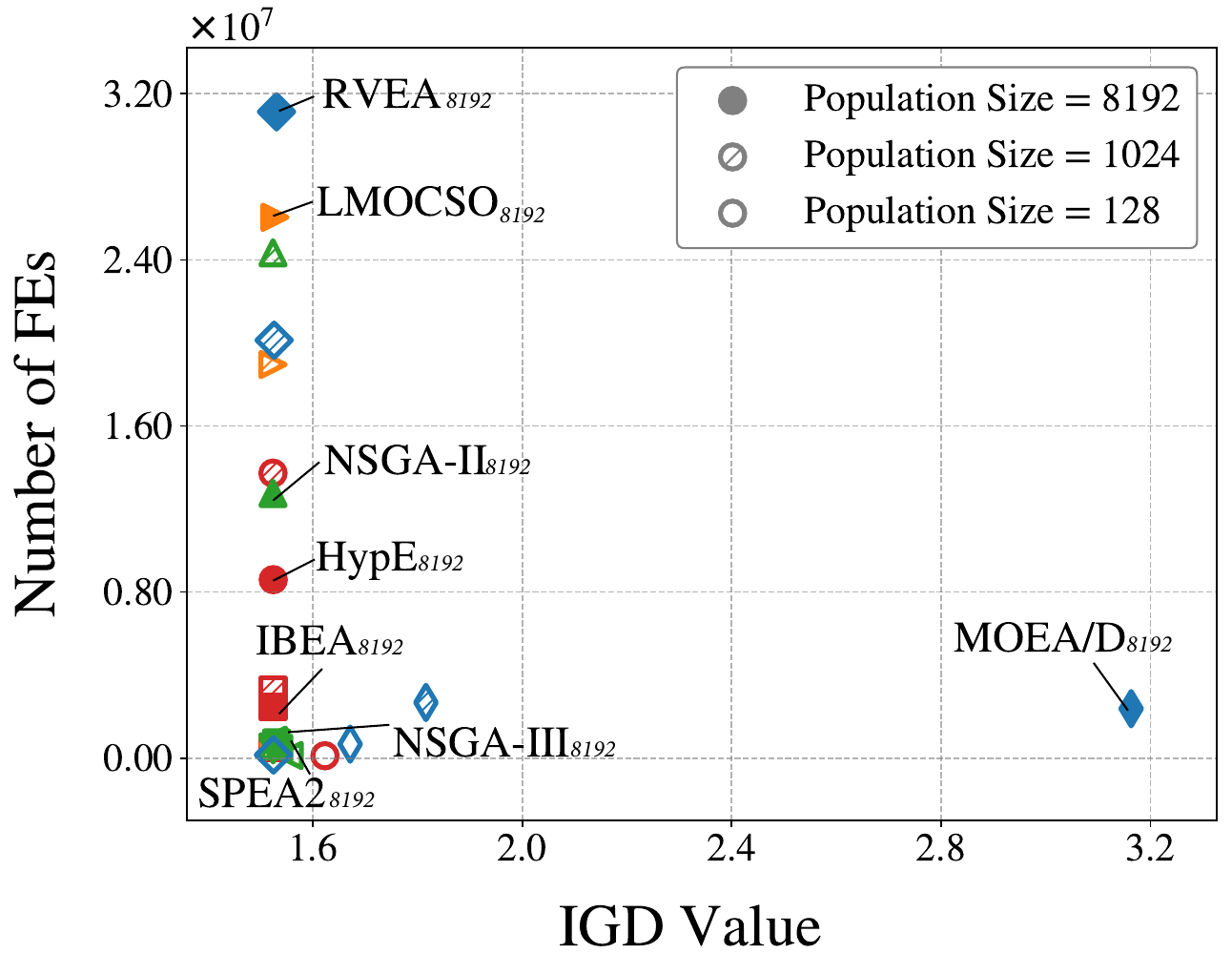}

        \centering
        \subcaption{DTLZ2}
    \end{minipage}
    \vspace{0.3cm}
    \centering
    \begin{minipage}[b]{0.48\textwidth}
        \centering
        \includegraphics[width=0.48\textwidth]{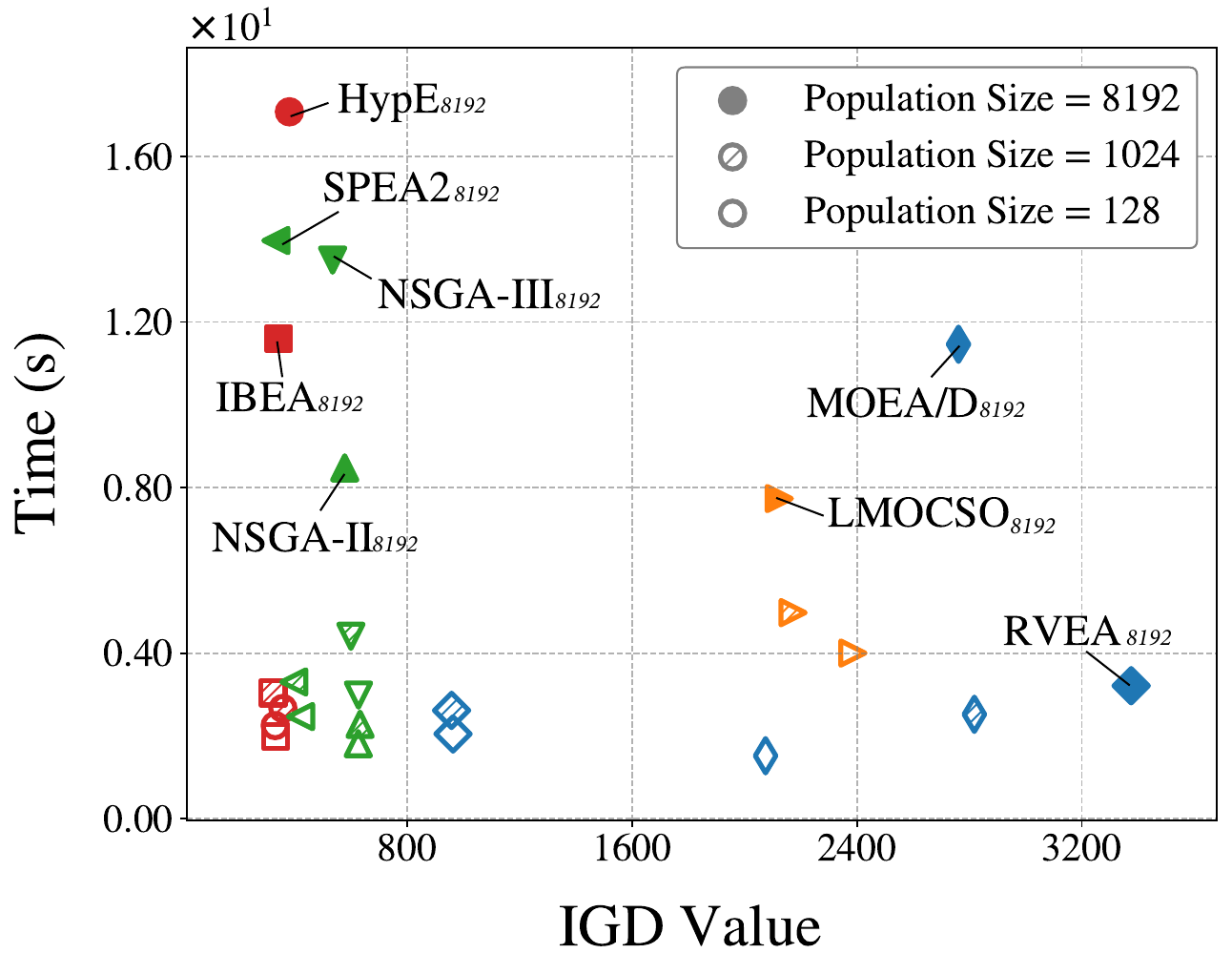}
        \centering
        \includegraphics[width=0.48\textwidth]{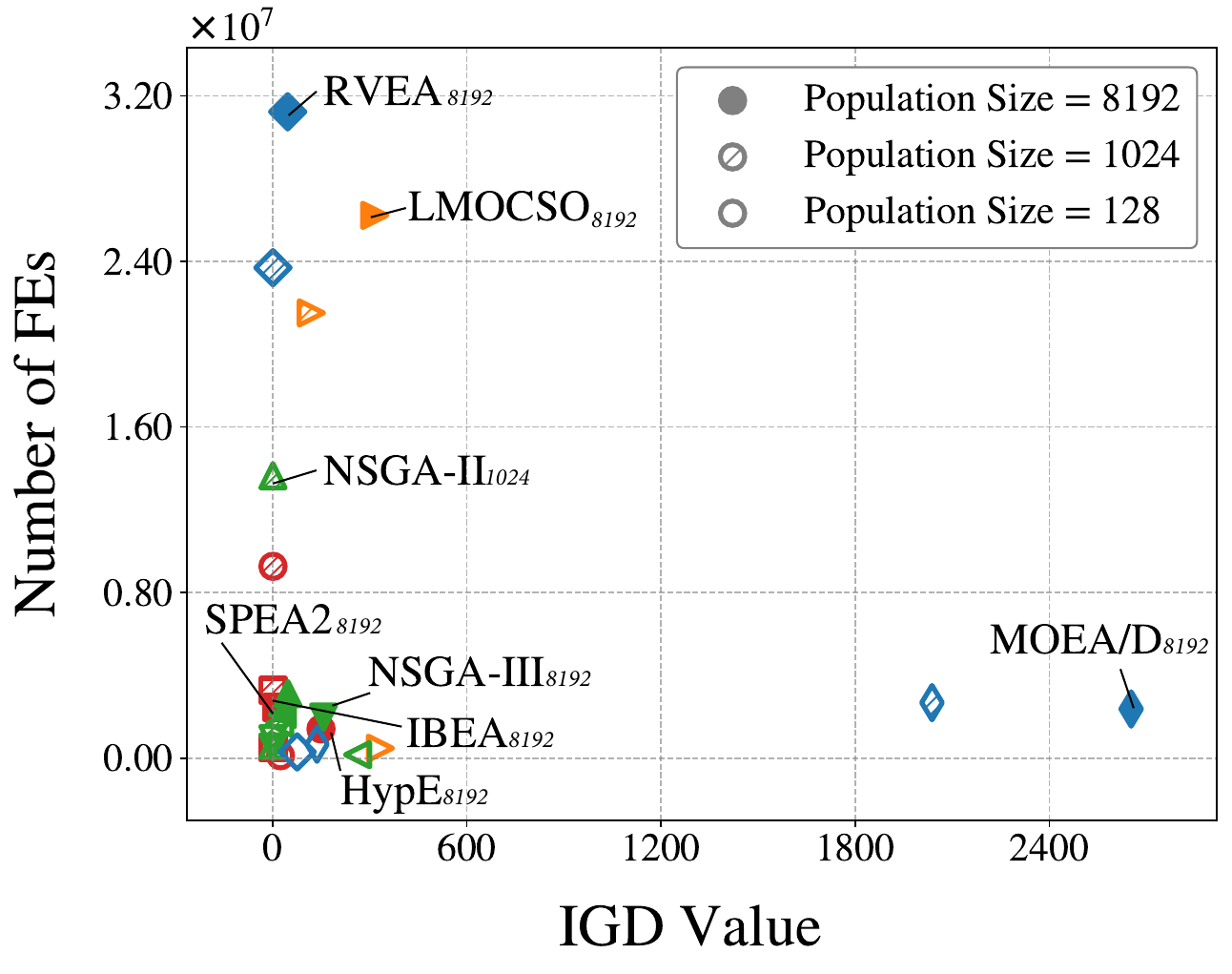}

        \centering
        \subcaption{DTLZ3}
    \end{minipage}
    \vspace{0.3cm}
    \centering
    \begin{minipage}[b]{0.48\textwidth}
        \centering
        \includegraphics[width=0.48\textwidth]{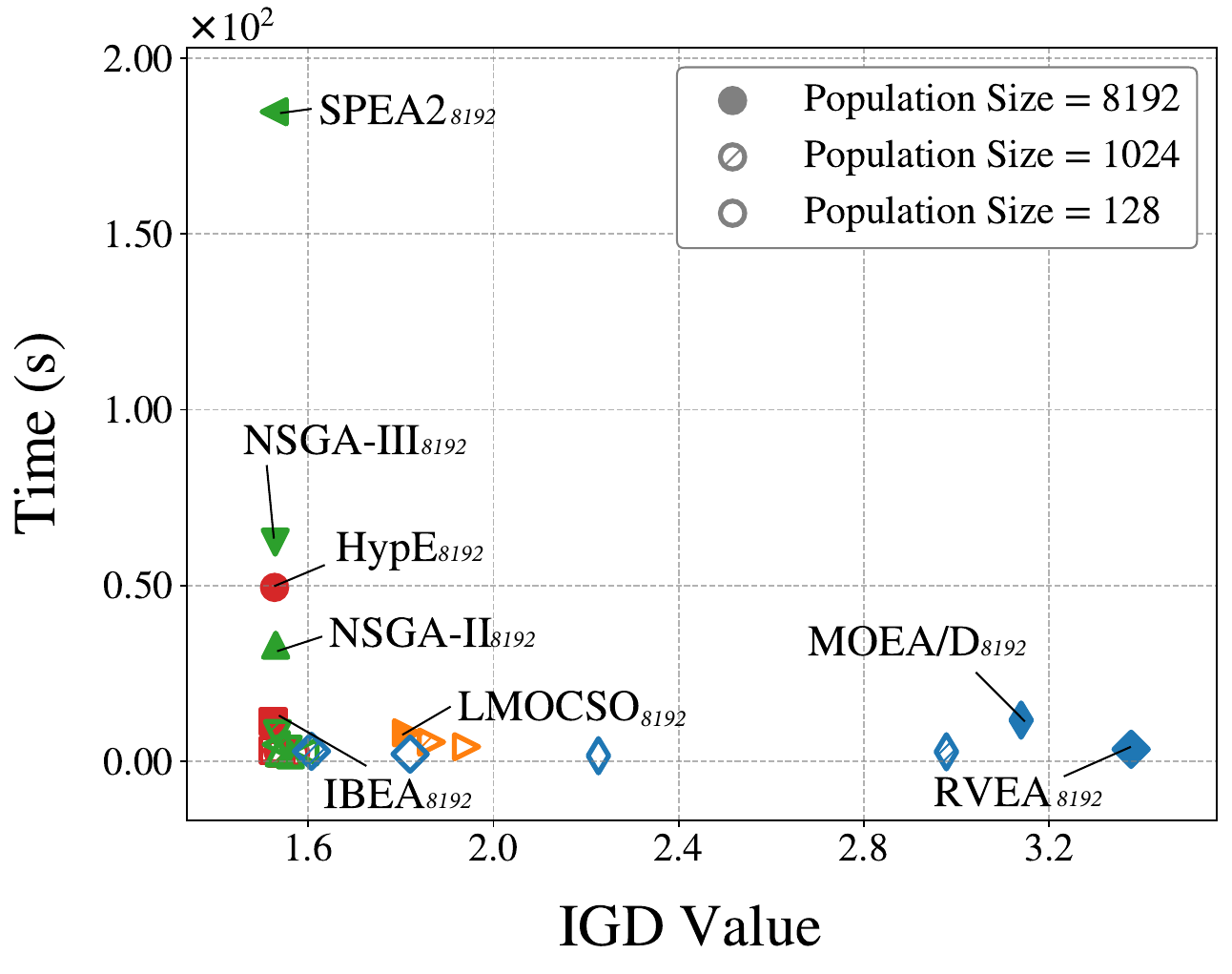}
        \centering
        \includegraphics[width=0.48\textwidth]{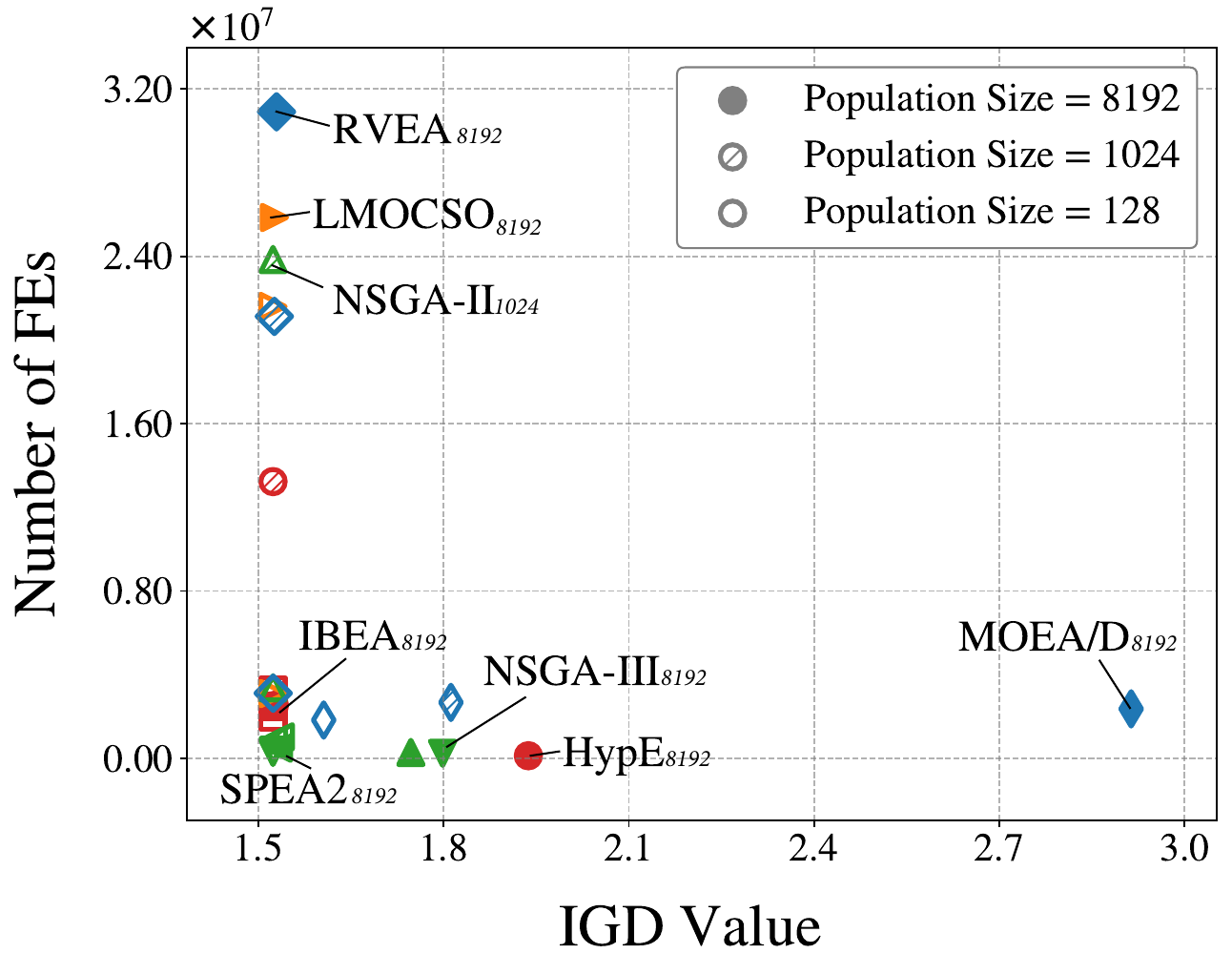}

        \centering
        \subcaption{DTLZ4}
    \end{minipage}
    \vspace{0.3cm}
    \centering
    \begin{minipage}[b]{0.48\textwidth}
        \centering
        \includegraphics[width=0.48\textwidth]{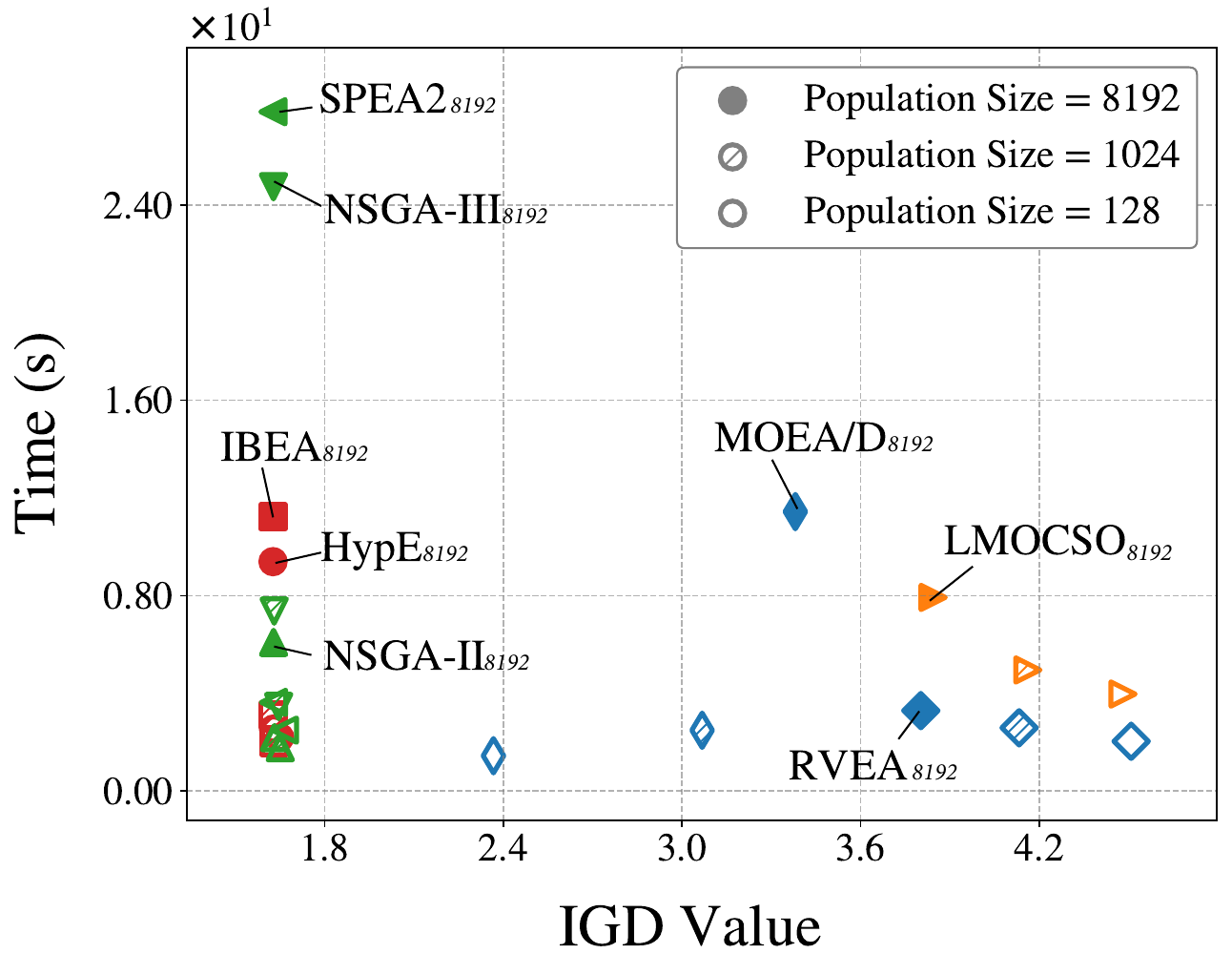}
        \centering
        \includegraphics[width=0.48\textwidth]{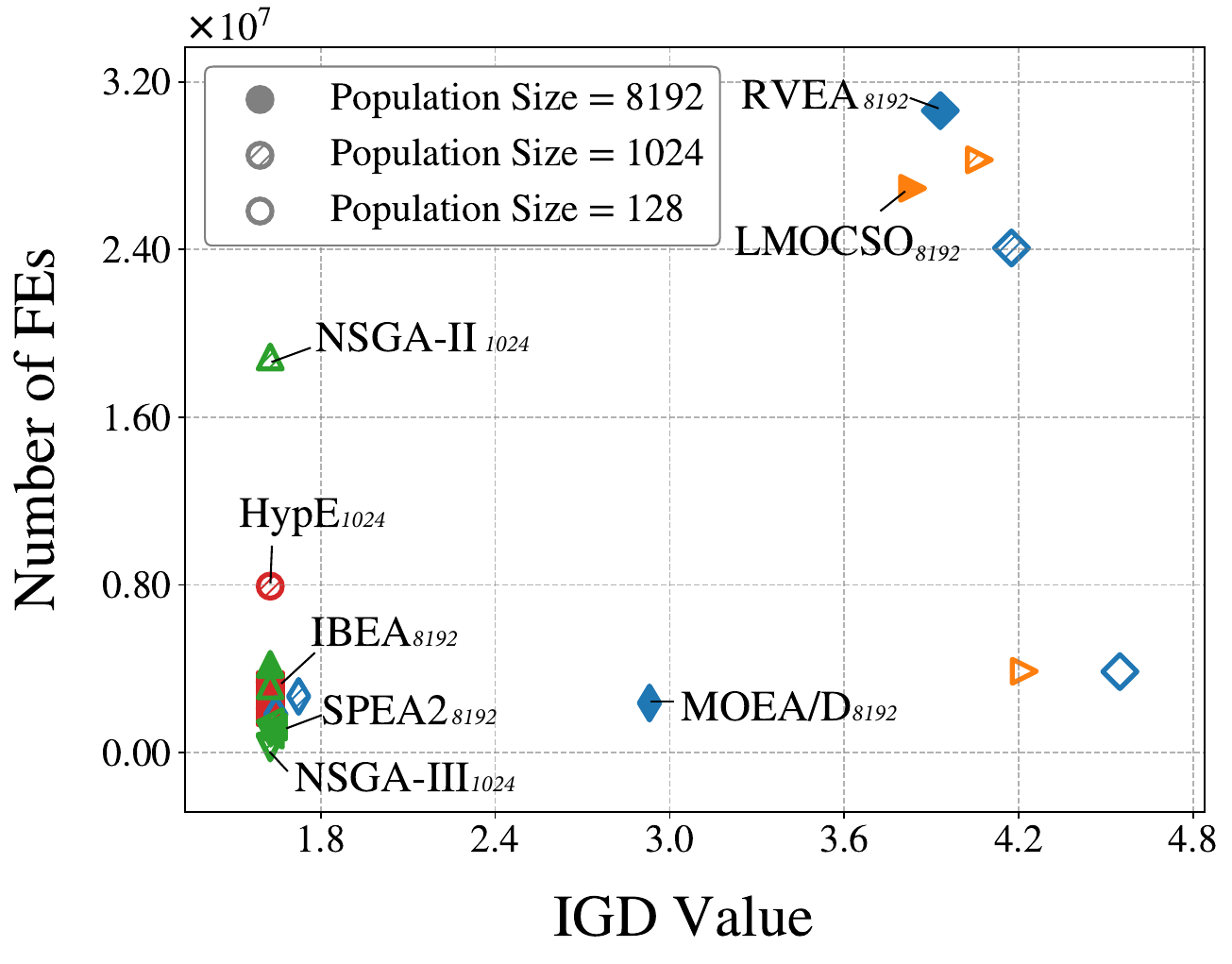}

        \centering
        \subcaption{DTLZ5}
    \end{minipage}
    \vspace{0.3cm}
    \centering
    \begin{minipage}[b]{0.48\textwidth}
        \centering
        \includegraphics[width=0.48\textwidth]{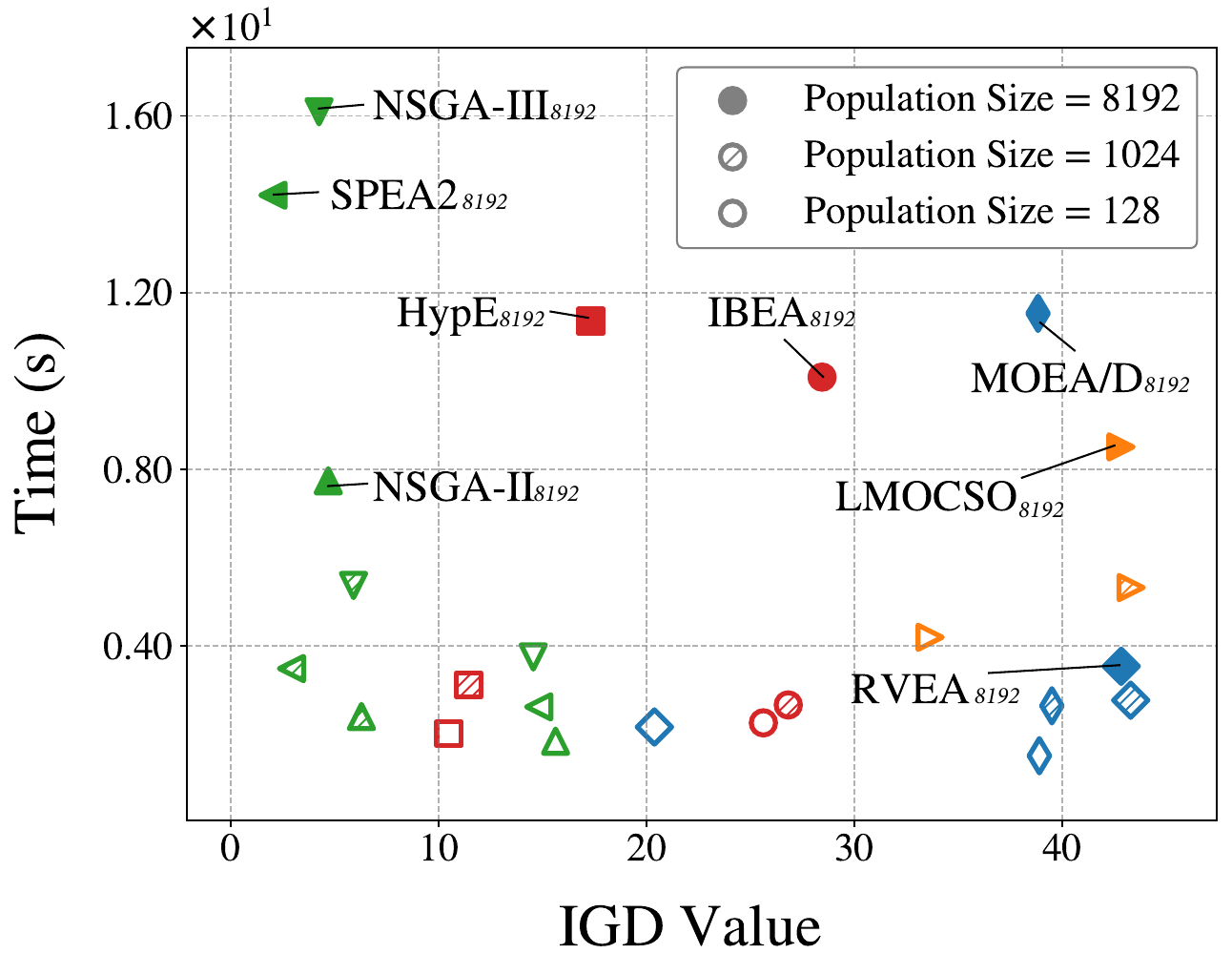}
        \centering
        \includegraphics[width=0.48\textwidth]{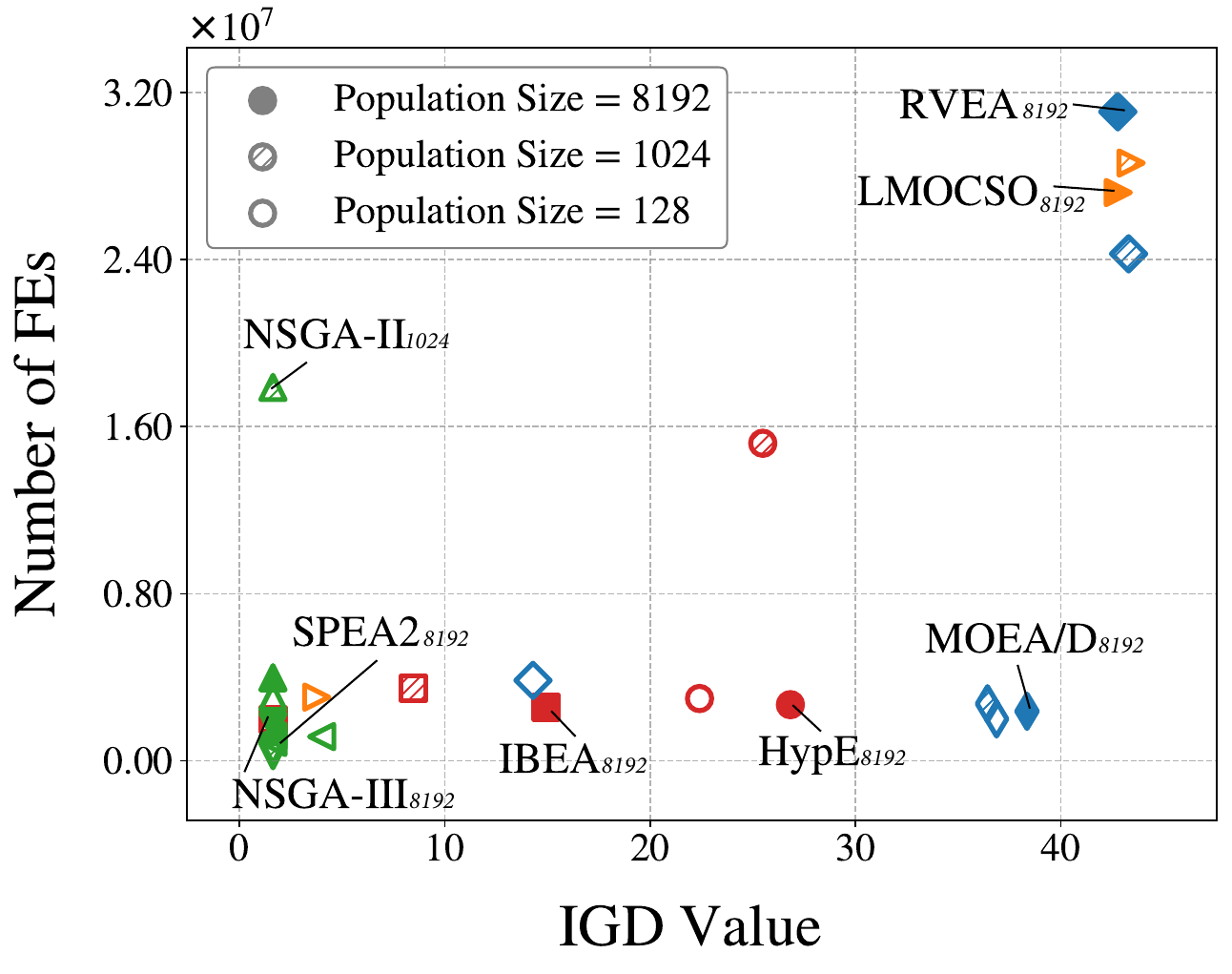}

        \centering
        \subcaption{DTLZ6}
    \end{minipage}
    \vspace{0.3cm}
    \centering
    \begin{minipage}[b]{0.48\textwidth}
        \centering
        \includegraphics[width=0.48\textwidth]{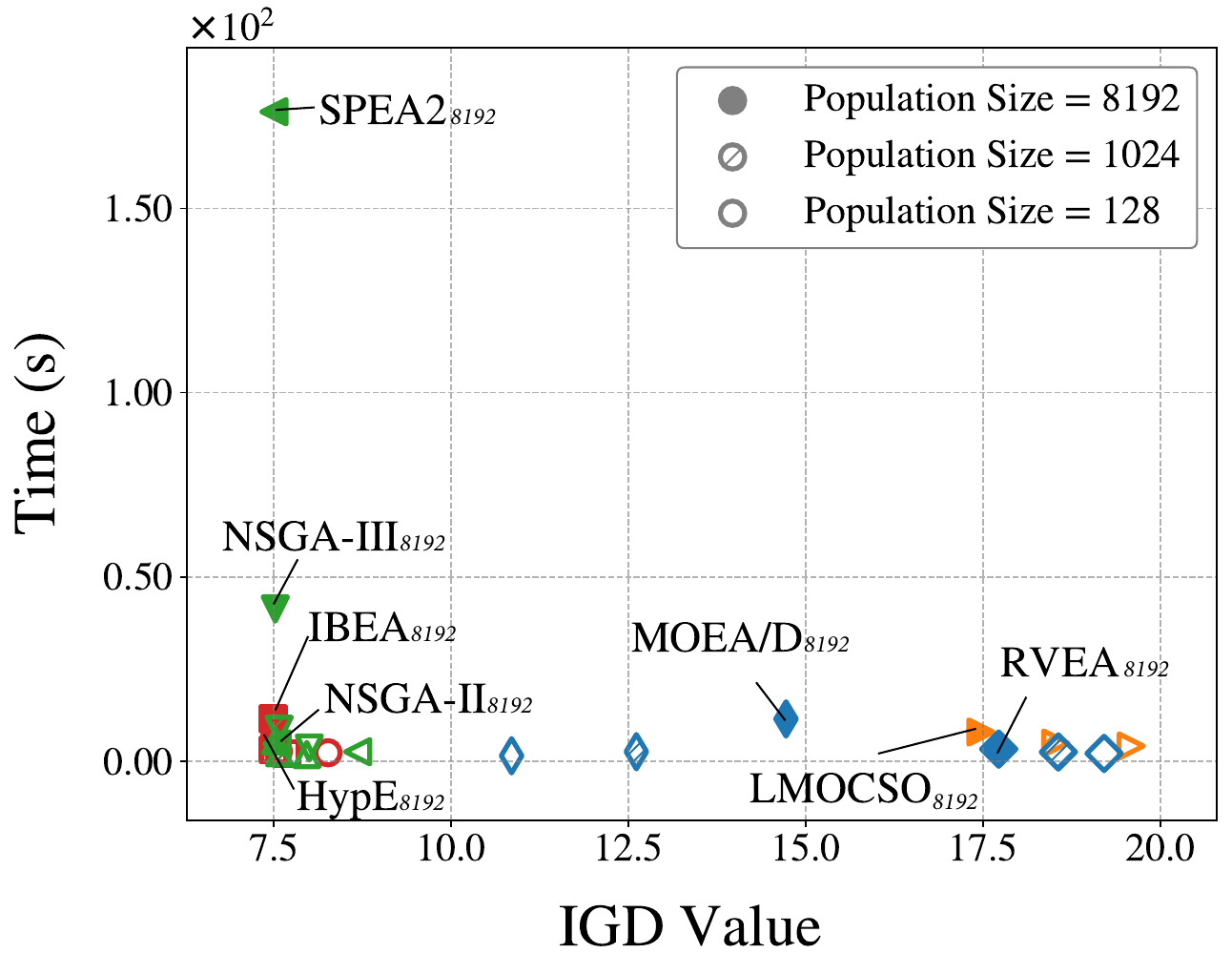}
        \centering
        \includegraphics[width=0.48\textwidth]{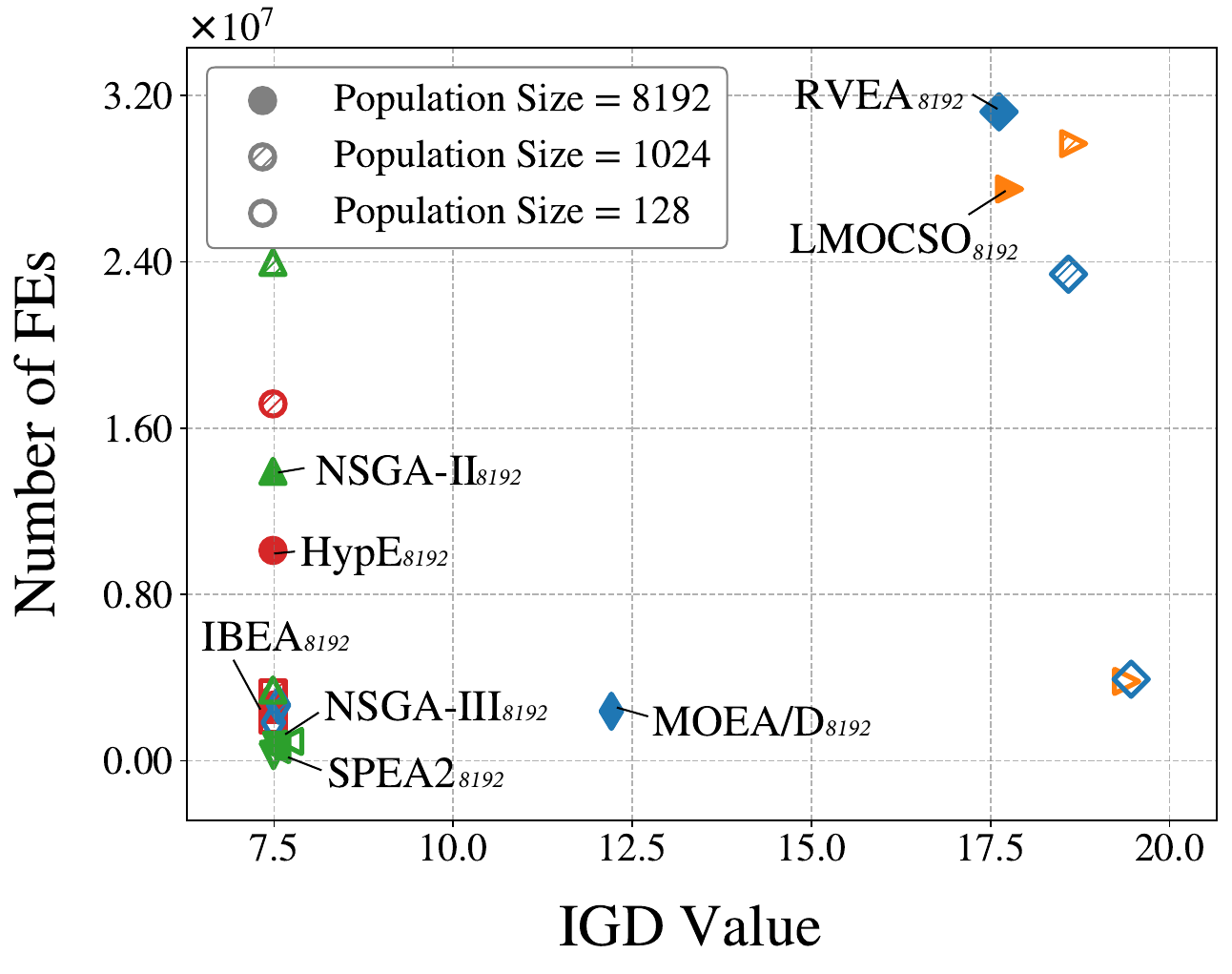}

        \centering
        \subcaption{DTLZ7}
    \end{minipage}
    \vspace{0.3cm}
    \centering
    \begin{minipage}[b]{0.48\textwidth}
        \centering
        \includegraphics[width=0.48\textwidth]{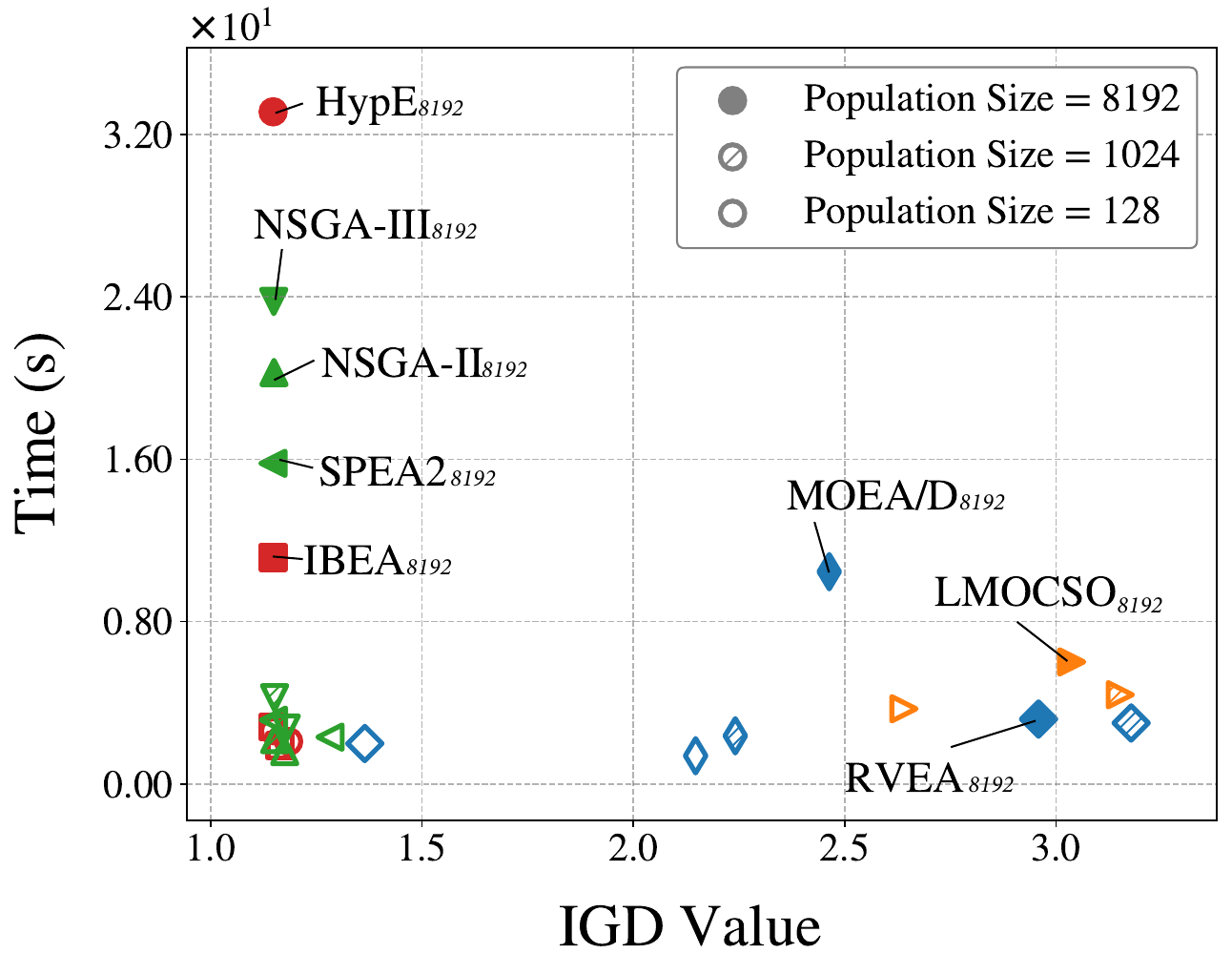}
        \centering
        \includegraphics[width=0.48\textwidth]{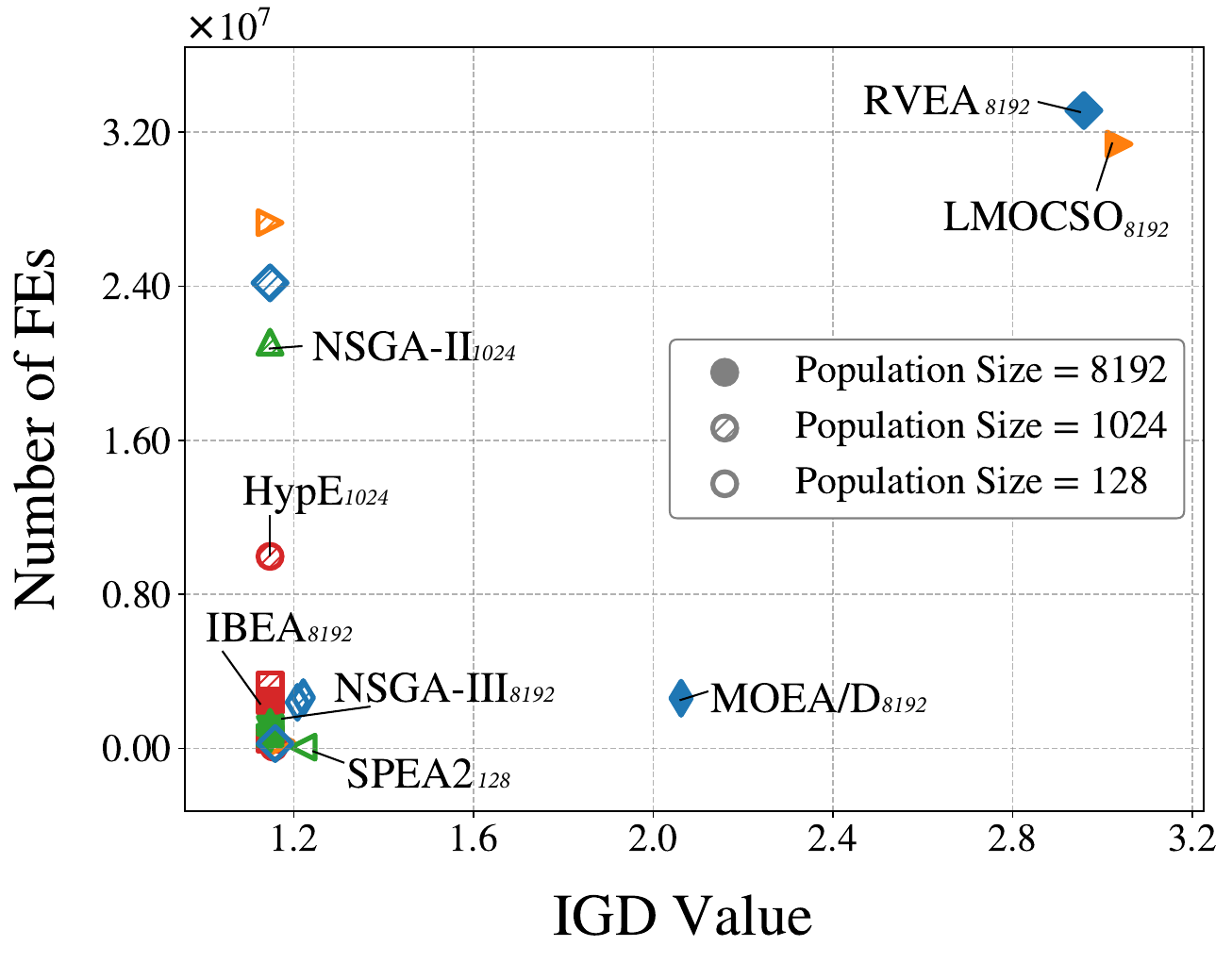}

        \centering
        \subcaption{ZDT1}
    \end{minipage}
    \vspace{0.3cm}
    \centering
    \begin{minipage}[b]{0.48\textwidth}
        \centering
        \includegraphics[width=0.48\textwidth]{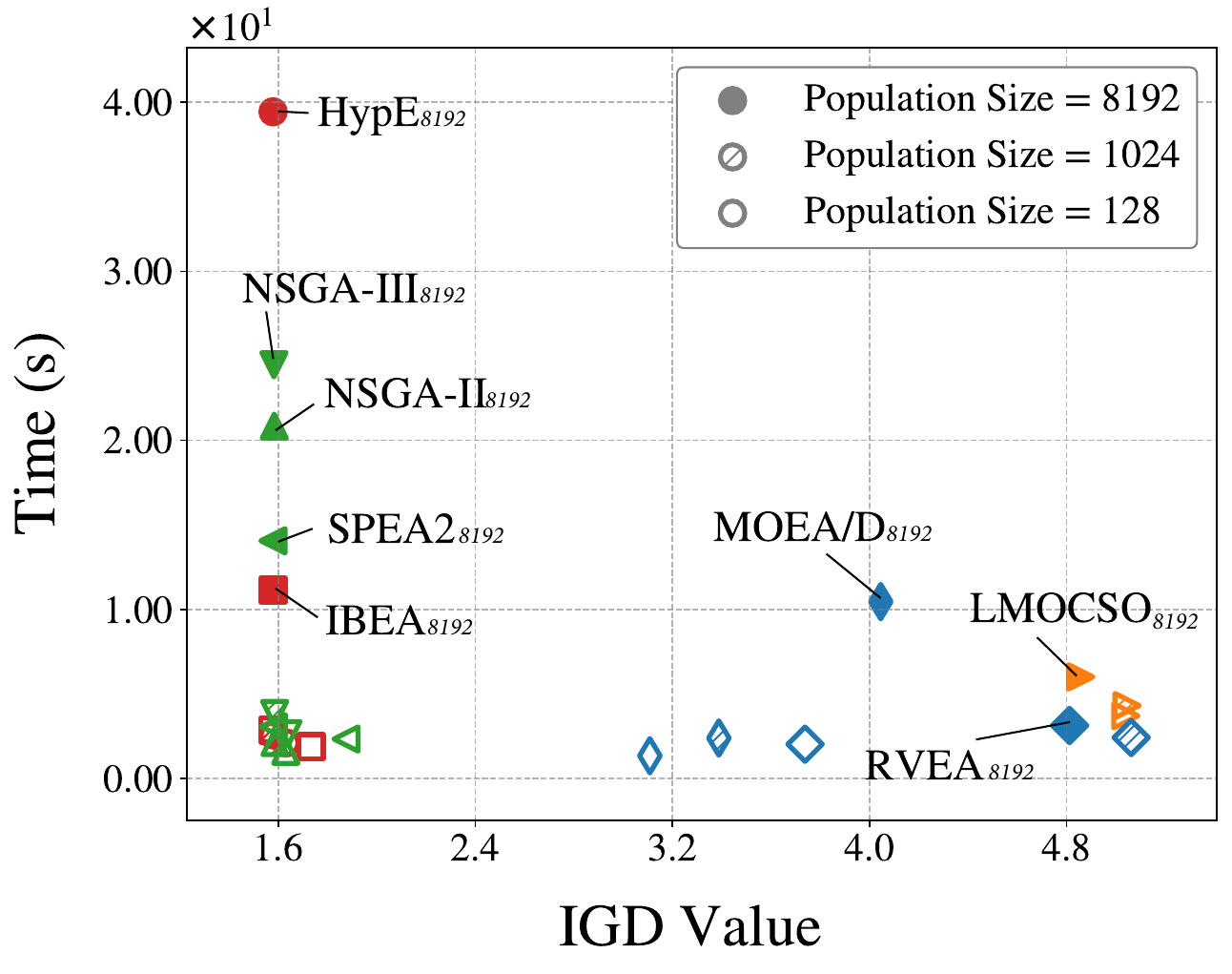}
        \centering
        \includegraphics[width=0.48\textwidth]{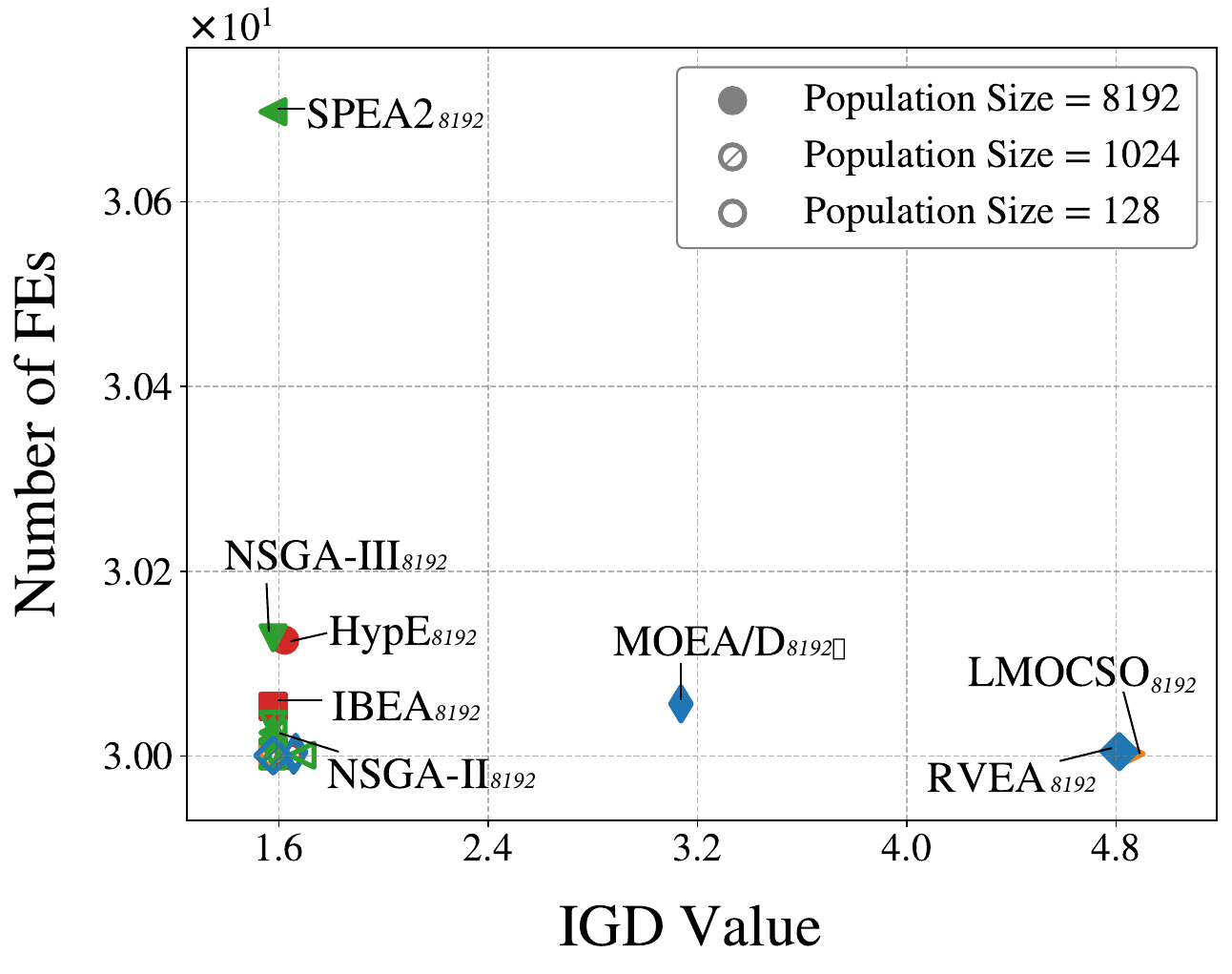}

        \centering
        \subcaption{ZDT2}
    \end{minipage}
    \caption{Performance comparison of MOEAs tested on numerical problems under varying population sizes, evaluated in terms of solution quality and computational efficiency under fixed-generation (100 iterations) versus fixed-time (30-second) constraints for EAs on an NVIDIA RTX-3090 GPU. Lower fitness/IGD values denote better performance. Results represent averaged performance values across 15 independent runs. Marker styles indicate population scales: hollow symbols for small populations (128), forward-slash-filled symbols for medium populations (1024), and solid symbols for large populations (8192). Different marker shapes distinguish between algorithms.}
    \label{fig:s-varying-popsize-numerical}
\end{figure}

\begin{figure*}[htbp]
    \centering
    \begin{minipage}[b]{0.48\textwidth}
        \centering
        \includegraphics[width=\textwidth]{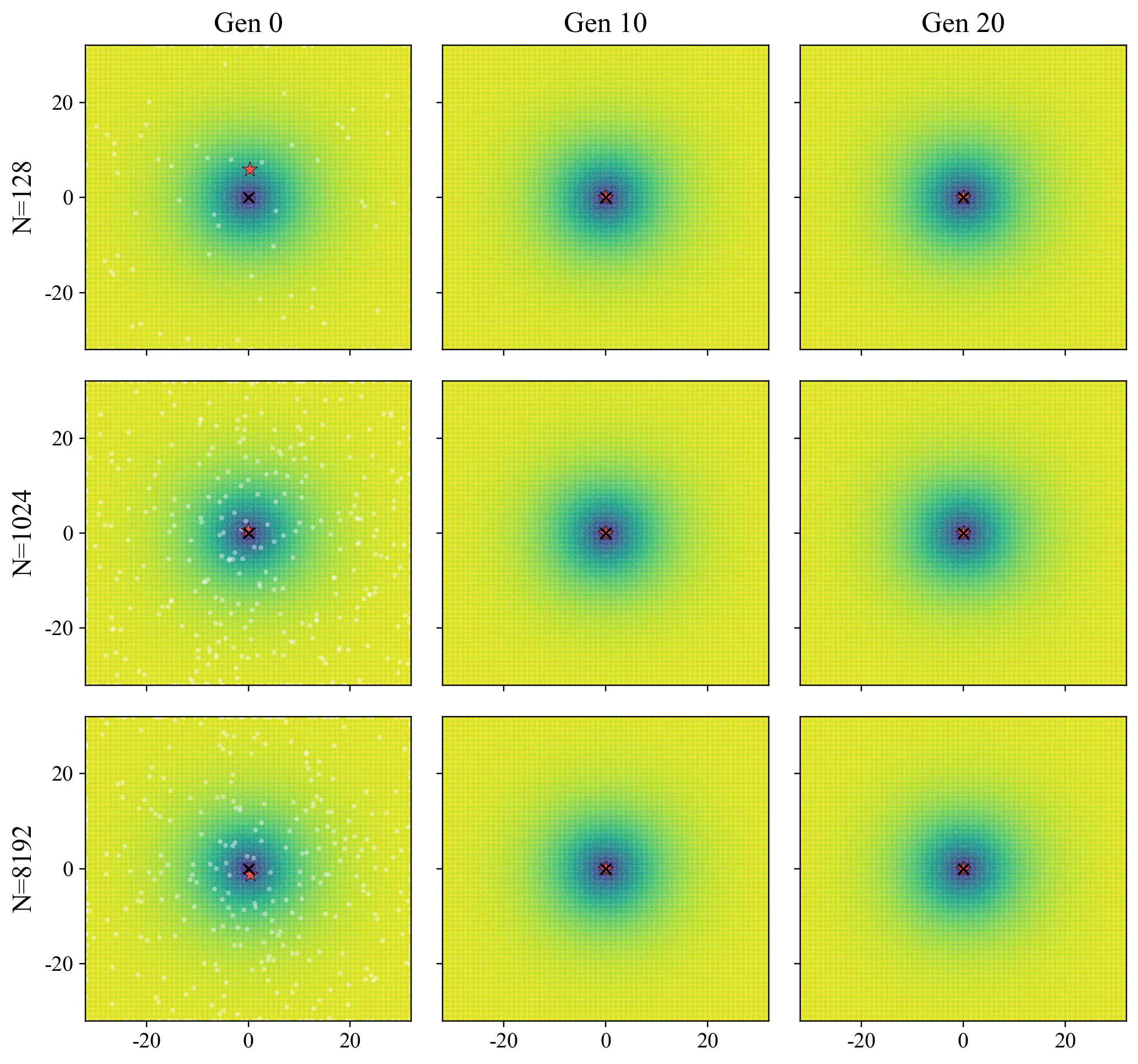}

        \centering
        \vspace{-0.1cm}
        \subcaption{CMA-ES}
    \end{minipage}
    \vspace{0.3cm}
    \centering
    \begin{minipage}[b]{0.48\textwidth}
        \centering
        \includegraphics[width=\textwidth]{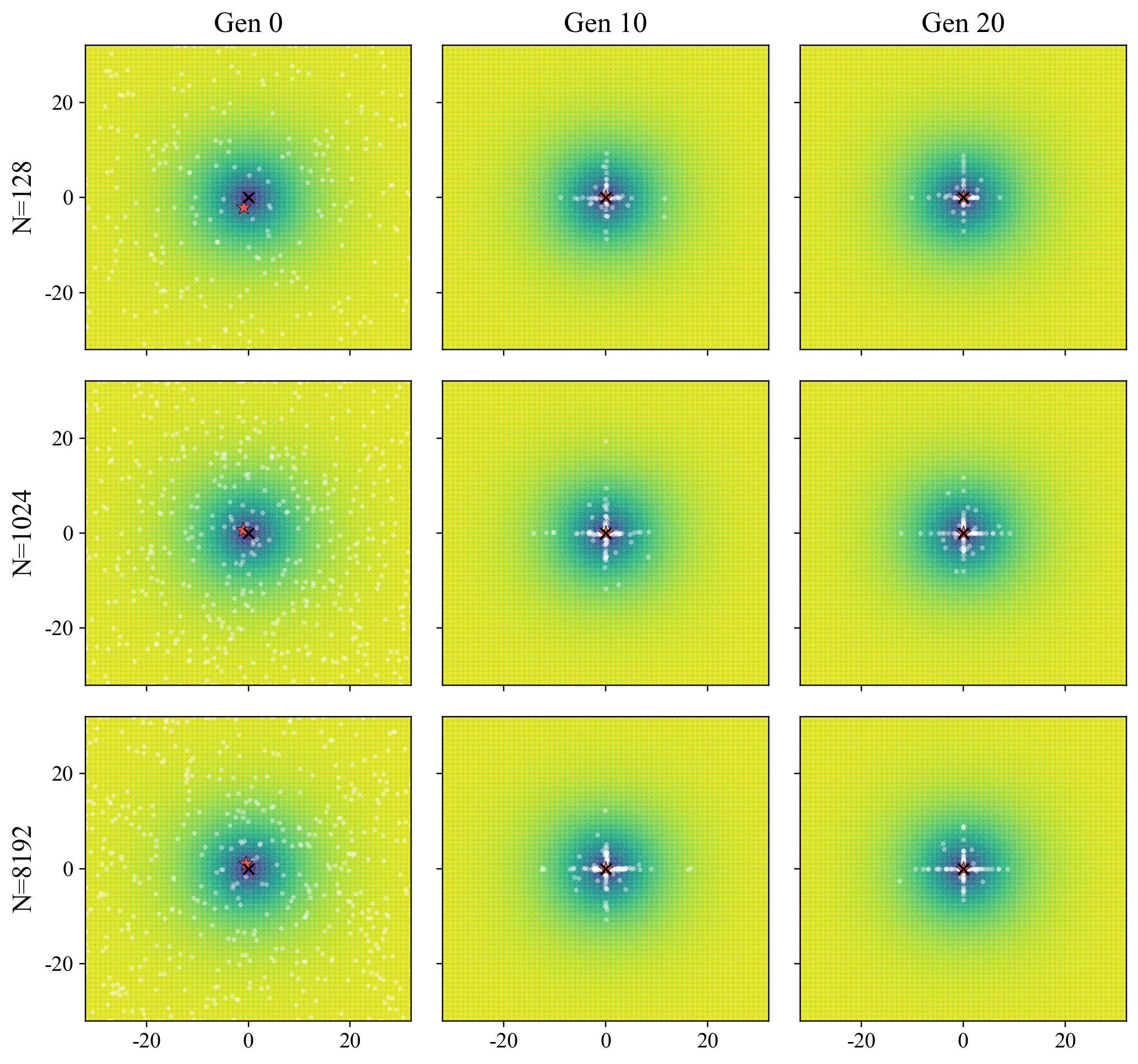}

        \centering
        \vspace{-0.1cm}
        \subcaption{GA-SBX/PM}
    \end{minipage}
     \vspace{0.3cm}
    \centering
    \begin{minipage}[b]{0.48\textwidth}
        \centering
        \includegraphics[width=\textwidth]{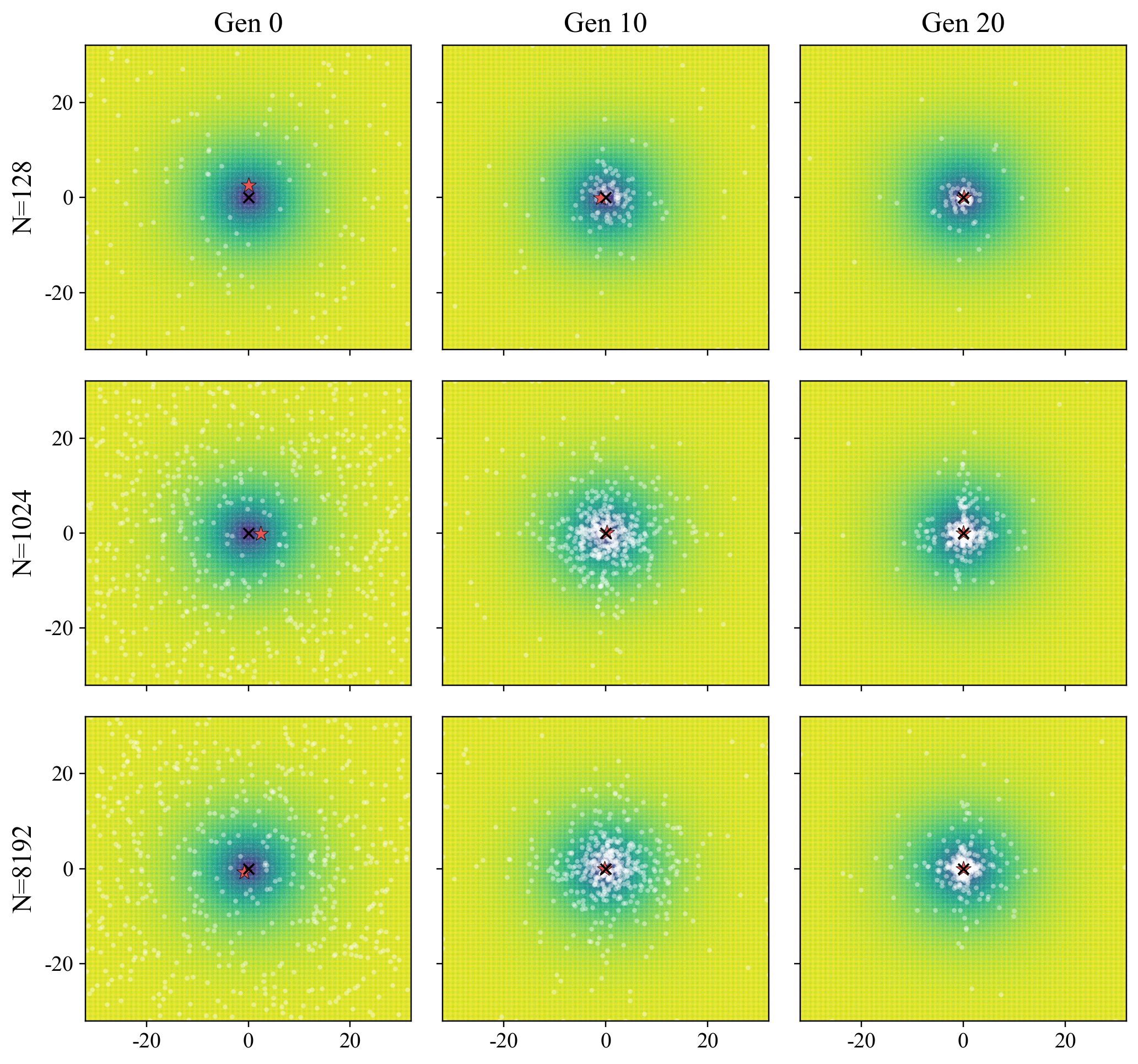}

        \centering
        \vspace{-0.1cm}
        \subcaption{PSO}
    \end{minipage}
    \vspace{0.3cm}
    \centering
    \begin{minipage}[b]{0.48\textwidth}
        \centering
        \includegraphics[width=\textwidth]{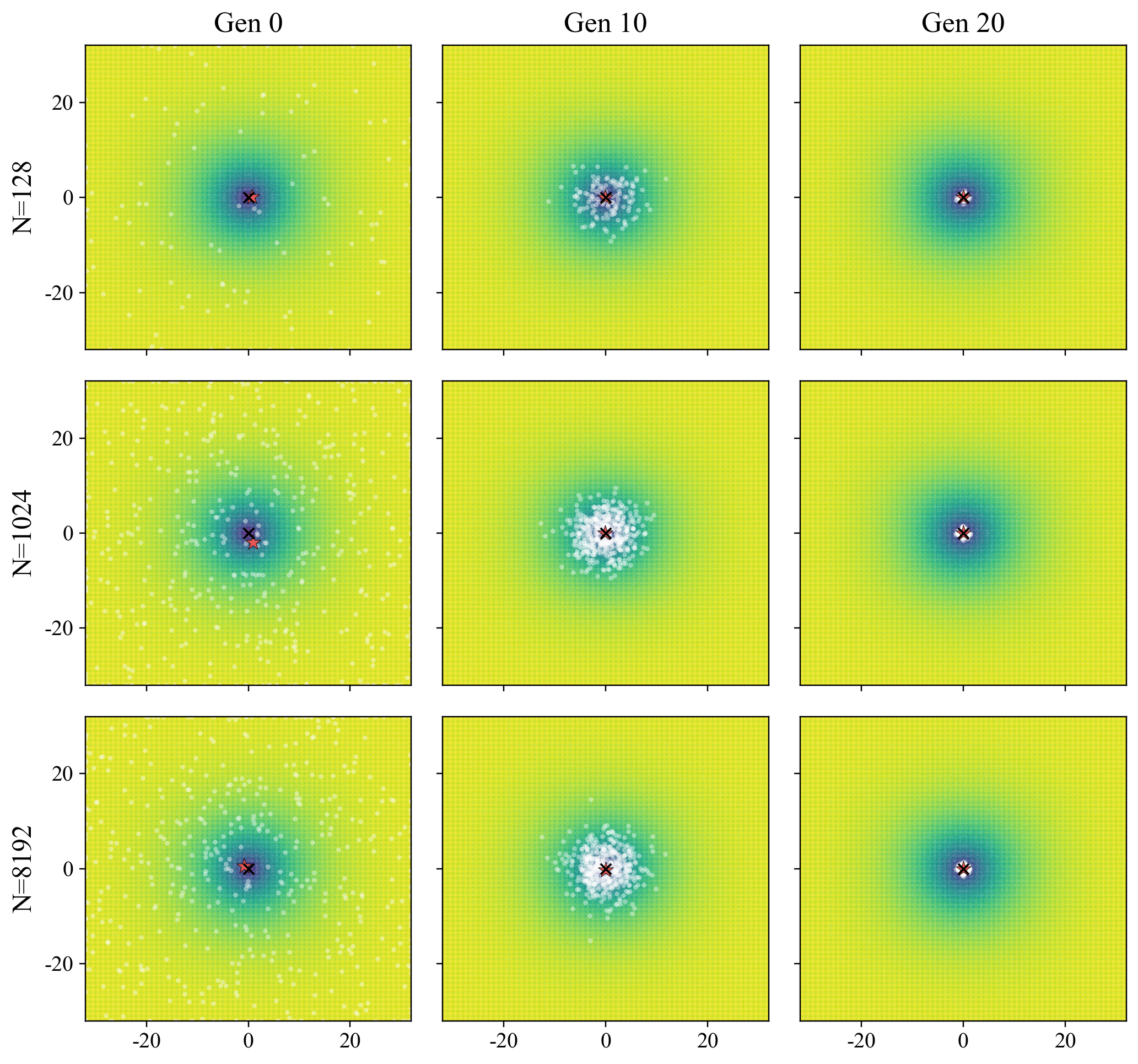}

        \centering
        \vspace{-0.1cm}
        \subcaption{DE}
    \end{minipage}
    \vspace{-0.5cm}
    \caption{Population snapshots on 2D Ackley with population sizes of 128, 1024, and 8192. Four representative EAs are shown at generations 0, 10, and 20.}
    \label{fig:2D}
\end{figure*}

To further examine the underlying mechanisms, we analyze four representative algorithms: PSO, CMA-ES, GA-SBX/PM, and DE (rand/1/bin), which correspond to swarm-based attraction, distribution-based adaptation, recombination-based search, and differential variation, respectively.
Fig.~\ref{fig:2D} visualizes the population dynamics of the four algorithms on the 2D Ackley function using population sizes of 128, 1024, and 8192, with snapshots recorded at generations 0, 10, and 20.

\setlength{\tabcolsep}{3pt}
\begin{table*}[htbp]
\centering
\begin{threeparttable}
\centering
\caption{Observed effects of scaling population in tested EAs and underlying mechanisms. (16–8192 individuals, 100 generations)}
\label{tab:trend}\begin{tabular}{p{2cm}p{6cm}p{8cm}}
\toprule
\textbf{Algorithm} & \textbf{Key mechanism} & \textbf{Effect of larger population} \\
\midrule
\makecell[tl]{Swarm-based\\{\smaller[1.5] (PSO, CSO)}}  &
Positions updated toward personal/global elites; CSO adds winner–loser competition mechanism. & \textbf{+} More particles cover a wider search radius and sustain diversity; CSO’s competitive sampling amplifies this benefit and accelerates escape from local optima.  \\
\\[-0.05em]
\makecell[tl]{Diff. Evo. \\{\smaller[1.5] (DE, SaDE)}} &
Trial vectors are built from scaled parent differences and crossover; SaDE self-adapts scaling factor ($F$) and crossover rate ($CR$) during the iterative process. &
\textbf{-} Larger parent spacing inflates step sizes, causing frequent overshoot and stagnation; SaDE regains stability via self-adaptation.  \\
\\[-0.05em]
\makecell[tl]{GA{\smaller[1.5] (GA-UR/GM,} \\ {\smaller[1.5]  GA-SBX/PM)}} & Stochastic recombination (uniform / simulated binary crossover) plus mutation pool gene statistics. &
\textbf{+} A bigger gene pool yields more reliable allele-frequency estimates, curbing genetic drift and delaying premature convergence while still broadening exploration. \\
\\[-0.05em]
\makecell[tl]{ES {\smaller[1.5] (CMA-ES, } \\ {\smaller[1.5] IPOP-CMA-ES)}} &
CMA updates full covariance and global step size $\sigma$; IPOP restarts with doubled $\lambda$. &
\textbf{+} Extra offspring sharpen covariance estimates, revealing clearer curvature directions; IPOP leverages large $\lambda$ restarts for aggressive global search. \\ \hline
\\[-0.05em]
\makecell[tl]{Indicator-based\\{\smaller[1.5] (HypE, IBEA)}}  &
Fitness assigned via hypervolume or $\varepsilon$-indicator. &
\textbf{+} Denser Pareto set improves trade-off resolution; however, indicator evaluation scales and quickly dominates runtime without GPU kernels or pruning. \\
\\[-0.05em]
\makecell[tl]{Pareto-based\\{\smaller[1.5] (SPEA2,} \\ {\smaller[1.5] NSGA-II/III)}} &
Non-dominated sorting and crowding-distance density estimation. &
\textbf{+} More individuals fill gaps and extremes, producing a smoother front; sorting overhead grows significantly. \\
\\[-0.05em]
\makecell[tl]{Decomposition\\ {\smaller[1.5] (RVEA, MOEA/D)}} & Problem split into weight-vector subproblems optimized cooperatively. & \textbf{–} An overly dense weight grid yields redundant neighborhoods, dilutes search pressure, and leads to stalled improvement despite higher cost. \\
\\[-0.05em]
\makecell[tl]{Swarm-based\\{\smaller[1.5] (LMOCSO)}}  & Pairwise winner–loser moves guided by an external archive. & \textbf{+} A massive candidate set intensifies selection pressure, accelerating convergence while maintaining diversity across objectives. \\

\bottomrule
\end{tabular}\begin{tablenotes}
    \footnotesize
    \item The symbols ``+'' and ``--'' denote beneficial and detrimental effects, respectively.
\end{tablenotes}
\end{threeparttable}
\end{table*}

\subsubsection{Results on Neuroevolution Tasks}\label{app:population-r}
We evaluated EA performance on ten neuroevolution tasks under 600 seconds, testing population sizes of 128, 1024 and 8192. Each configuration is tested for 10 times. All experiments were conducted on an NVIDIA RTX-3090 GPU and a RTX-2080-Ti, with solution quality and NFEs completed serving as key performance metrics.

\begin{figure}[htbp]
    \centering
    \begin{minipage}[b]{\textwidth}
        \centering
        \includegraphics[width=0.47\textwidth]{Figures/sub/3/legend.pdf}
         \hspace{0.5cm}
        \includegraphics[width=0.47\textwidth]{Figures/sub/3/legend-m.pdf}
    \end{minipage}

    \vspace{0.2cm}

    \begin{minipage}[b]{0.235\textwidth}
        \centering
        \includegraphics[width=\textwidth]{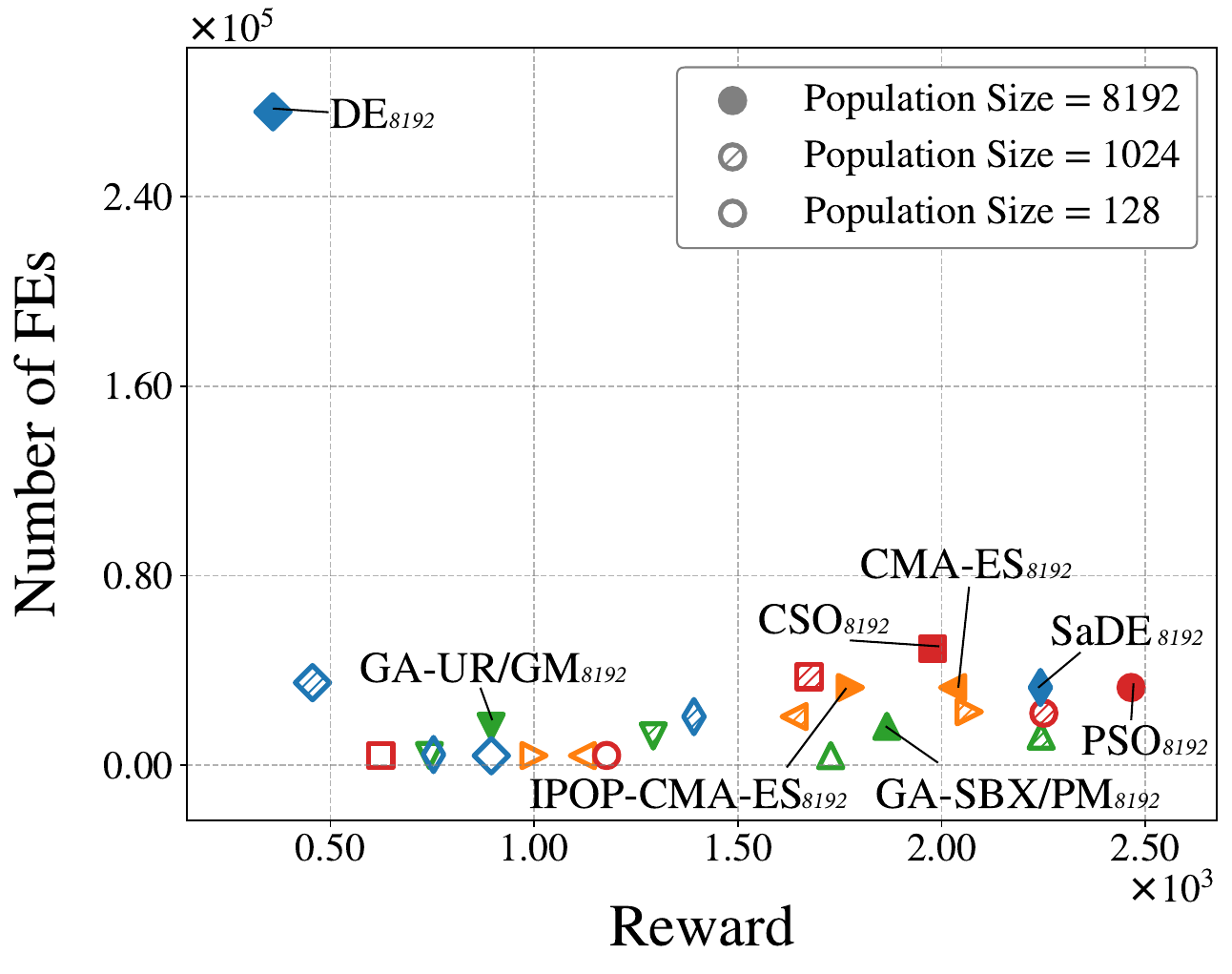}
        \centering
        \vspace{-0.5cm}
        \subcaption{Halfcheetah}
    \end{minipage}
   \hfill
    \begin{minipage}[b]{0.235\textwidth}
        \centering
        \includegraphics[width=\textwidth]{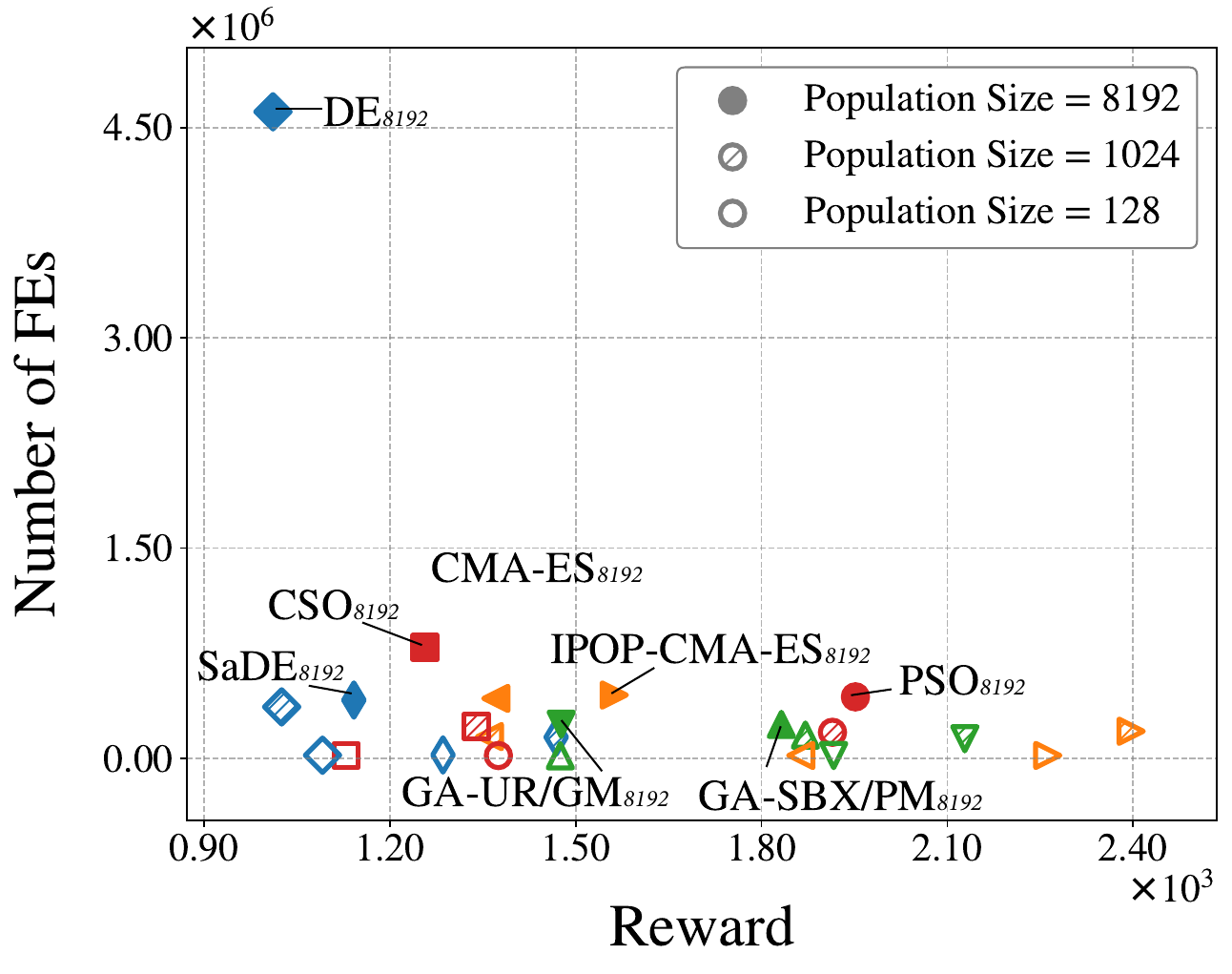}
        \centering
        \vspace{-0.5cm}
        \subcaption{Hopper}
    \end{minipage}
    \hfill
    \begin{minipage}[b]{0.235\textwidth}
        \centering
        \includegraphics[width=\textwidth]{Figures/sub/3/3090/neuro/fix_time/MoHopper-m2.pdf}
        \centering
        \vspace{-0.5cm}
        \subcaption{MoHopper-m2}
    \end{minipage}
   \hfill
    \begin{minipage}[b]{0.235\textwidth}
        \centering
        \includegraphics[width=\textwidth]{Figures/sub/3/3090/neuro/fix_time/MoHopper-m3.pdf}
        \centering
        \vspace{-0.5cm}
        \subcaption{MoHopper-m3}
    \end{minipage}
    \hfill

    \vspace{0.2cm}

    \begin{minipage}[b]{0.235\textwidth}
        \centering
        \includegraphics[width=\textwidth]{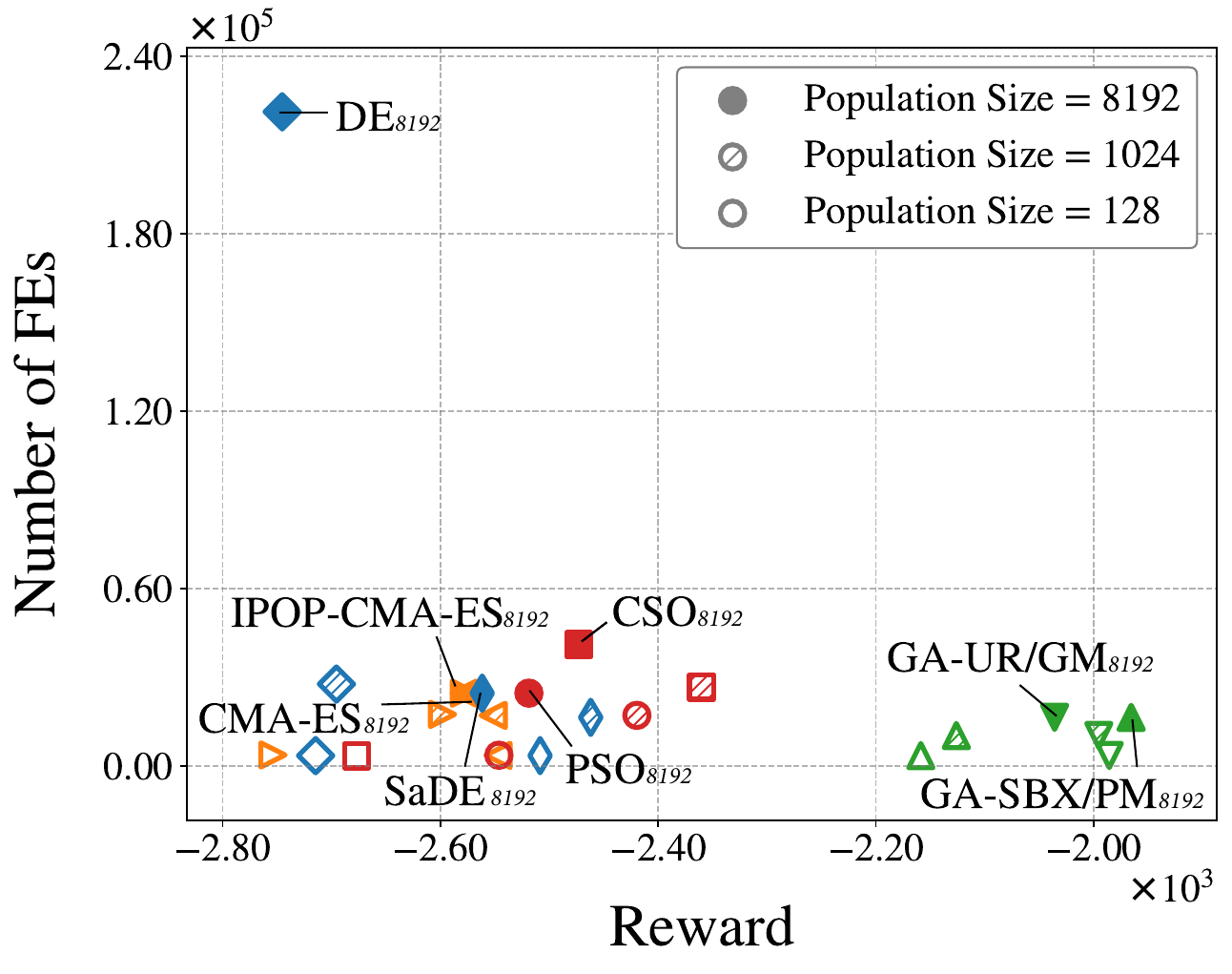}
        \centering
        \vspace{-0.5cm}
        \subcaption{Pusher}
    \end{minipage}
   \hfill
    \begin{minipage}[b]{0.235\textwidth}
        \centering
        \includegraphics[width=\textwidth]{Figures/sub/3/3090/neuro/fix_time/Reacher.pdf}
        \centering
        \vspace{-0.5cm}
        \subcaption{Reacher}
    \end{minipage}
    \hfill
    \begin{minipage}[b]{0.235\textwidth}
        \centering
        \includegraphics[width=\textwidth]{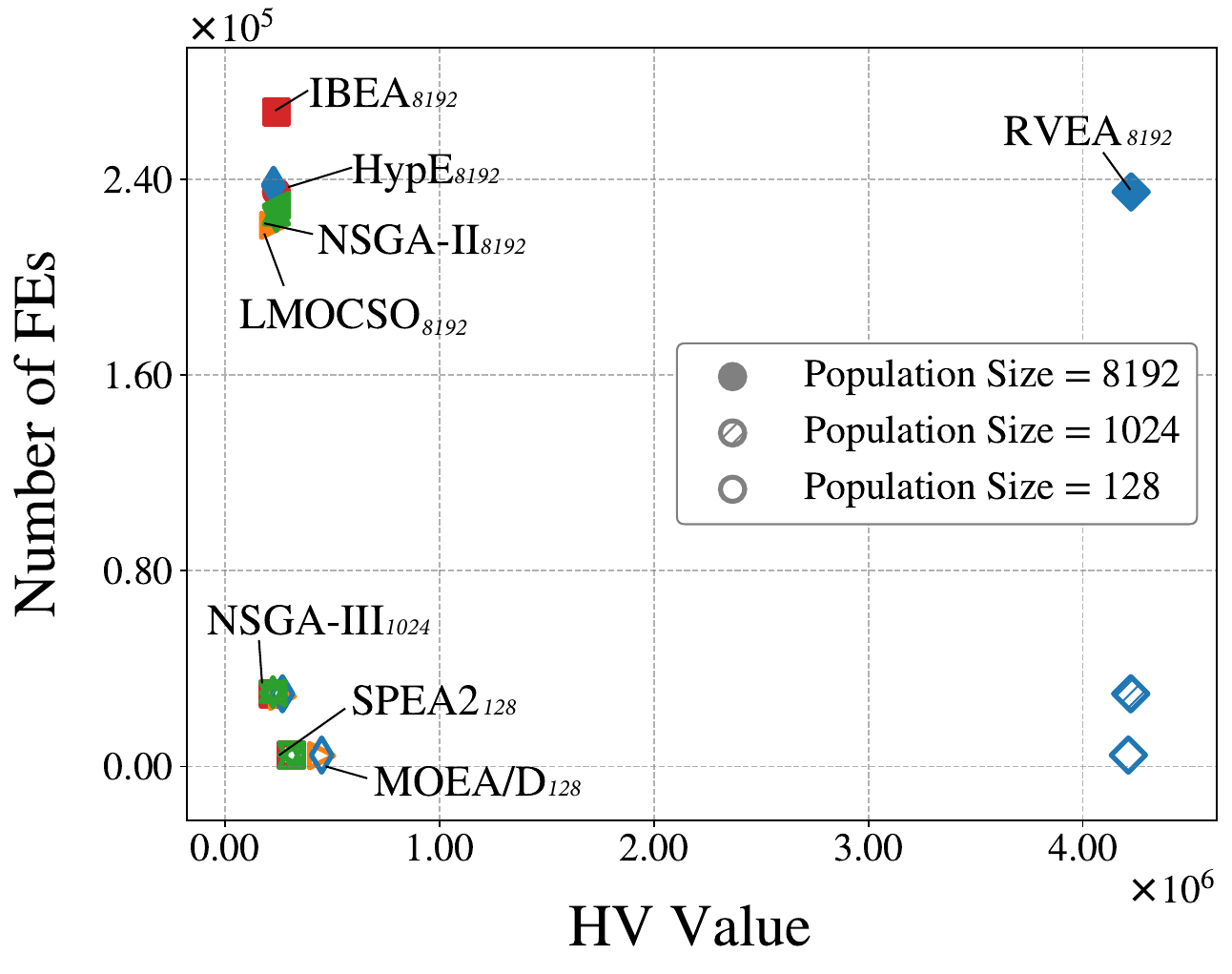}
        \centering
        \vspace{-0.5cm}
        \subcaption{MoReacher}
    \end{minipage}
   \hfill
    \begin{minipage}[b]{0.235\textwidth}
        \centering
        \includegraphics[width=\textwidth]{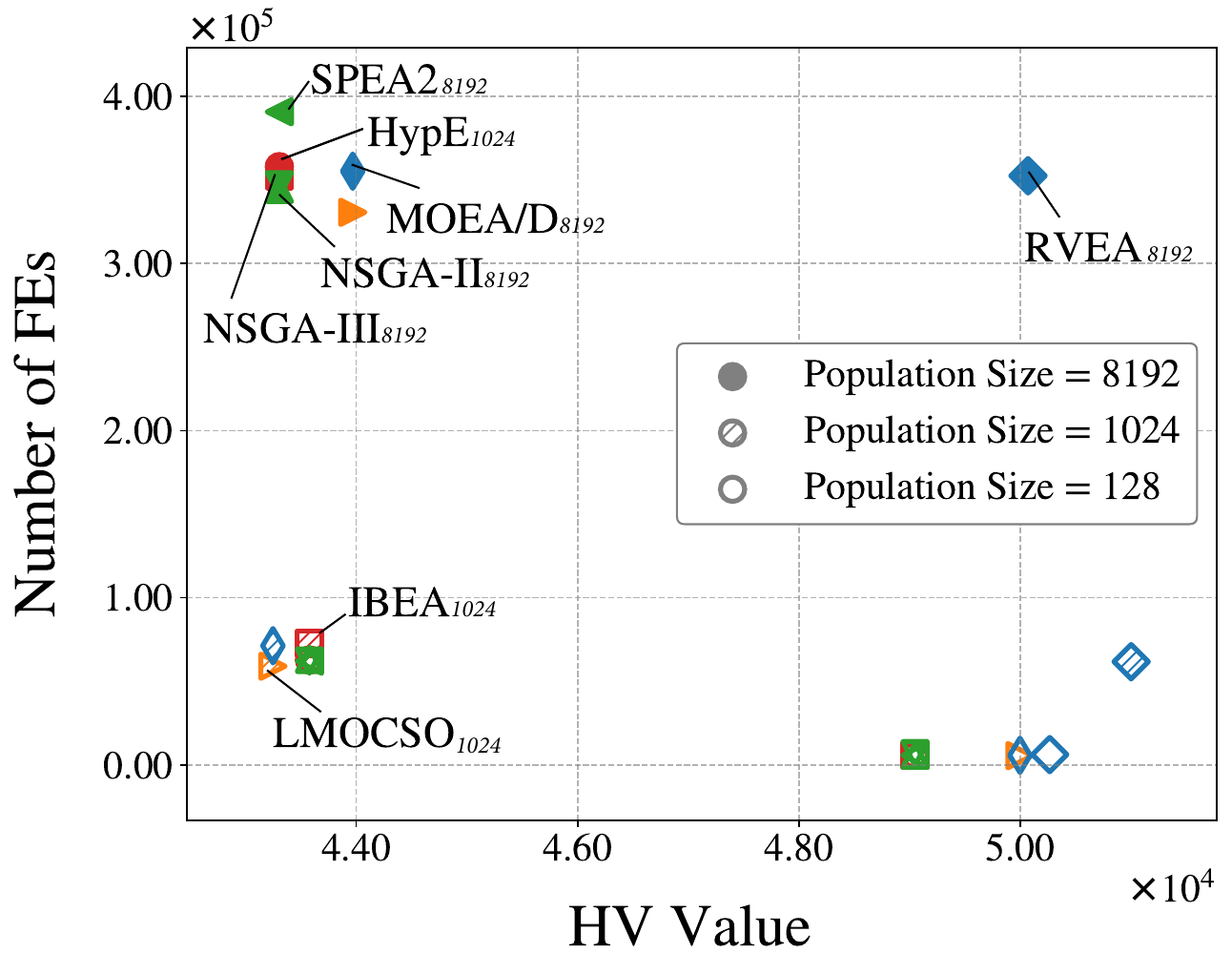}
        \centering
        \vspace{-0.5cm}
        \subcaption{MoSwimmer}
    \end{minipage}
    \hfill

    \vspace{0.2cm}

    \begin{minipage}[b]{0.235\textwidth}
        \centering
        \includegraphics[width=\textwidth]{Figures/sub/3/3090/neuro/fix_time/Swimmer.pdf}
        \centering
        \vspace{-0.5cm}
        \subcaption{Swimmer}
    \end{minipage}
    \hspace{2cm}
    \begin{minipage}[b]{0.235\textwidth}
        \centering
        \includegraphics[width=\textwidth]{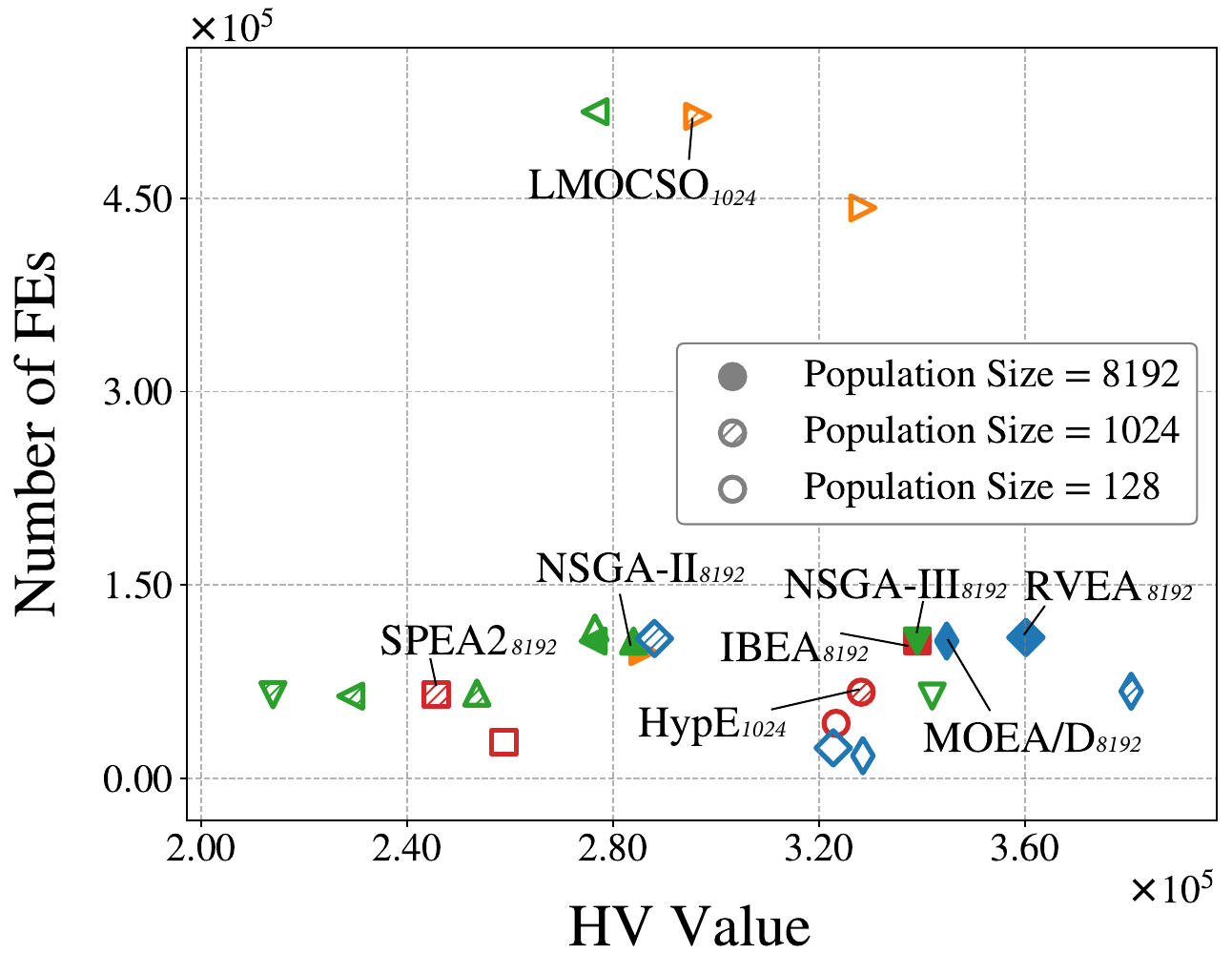}
        \centering
        \vspace{-0.5cm}
        \subcaption{MoWalker2d}
    \end{minipage}

    \caption{Performance comparison of EAs tested on neuroevolution tasks under varying population sizes with an NVIDIA RTX-3090 GPU, evaluated in terms of solution quality and number of FEs completed within 600 seconds. Higher reward/HV values denote better performance. Results represent averaged performance values across 10 independent runs. Marker styles indicate population scales: hollow symbols for small populations (128), forward-slash-filled symbols for medium populations (1024), and solid symbols for large populations (8192). Different marker shapes distinguish between algorithms.}
    \label{fig:s-varying-popsize-neuro-so}
\end{figure}

\begin{figure}[htbp]
    \centering
    \begin{minipage}[b]{\textwidth}
        \centering
        \includegraphics[width=0.47\textwidth]{Figures/sub/3/legend.pdf}
         \hspace{0.5cm}
        \includegraphics[width=0.47\textwidth]{Figures/sub/3/legend-m.pdf}
    \end{minipage}

    \vspace{0.2cm}

    \begin{minipage}[b]{0.235\textwidth}
        \centering
        \includegraphics[width=\textwidth]{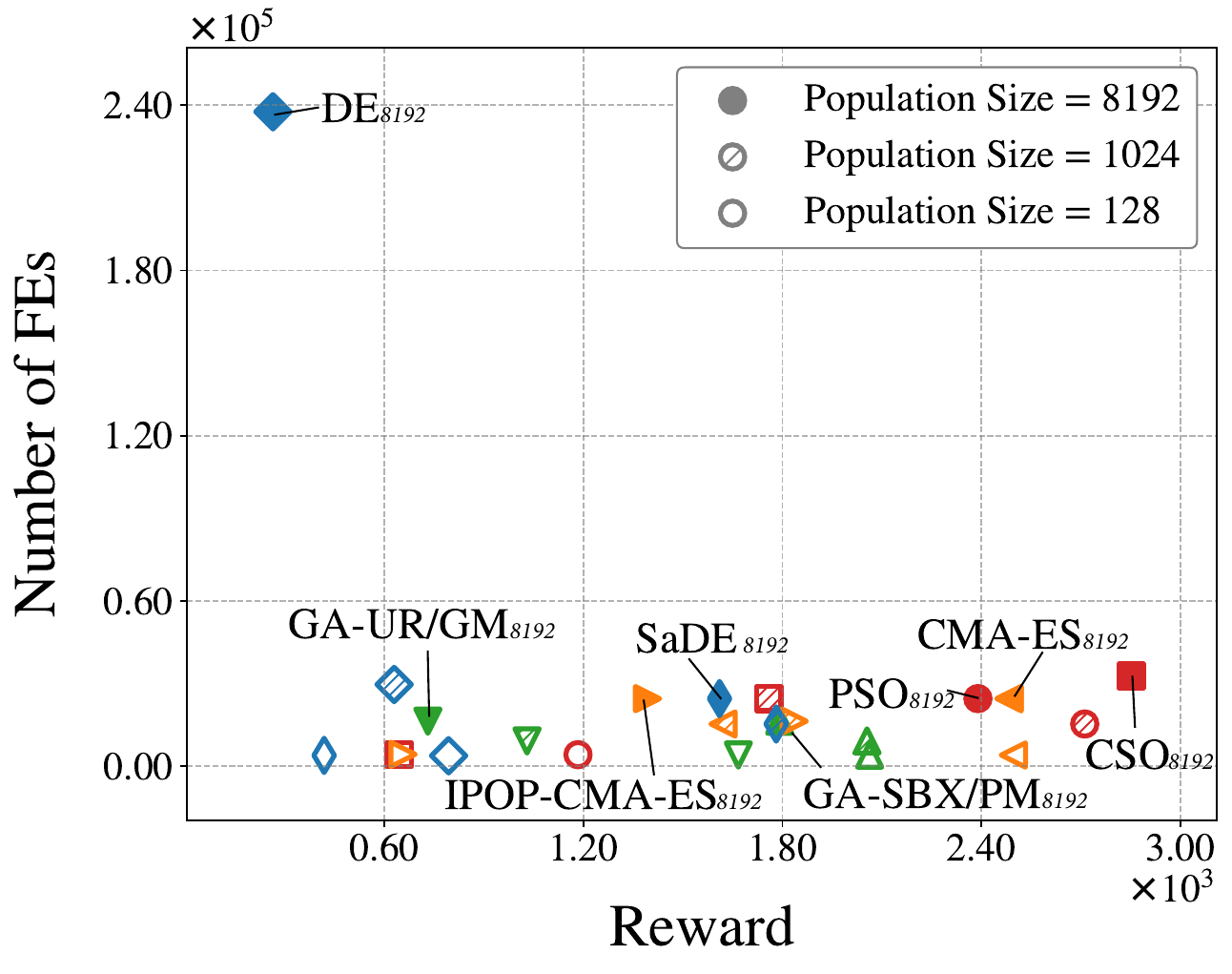}
        \centering
        \vspace{-0.5cm}
        \subcaption{Halfcheetah}
    \end{minipage}
   \hfill
    \begin{minipage}[b]{0.235\textwidth}
        \centering
        \includegraphics[width=\textwidth]{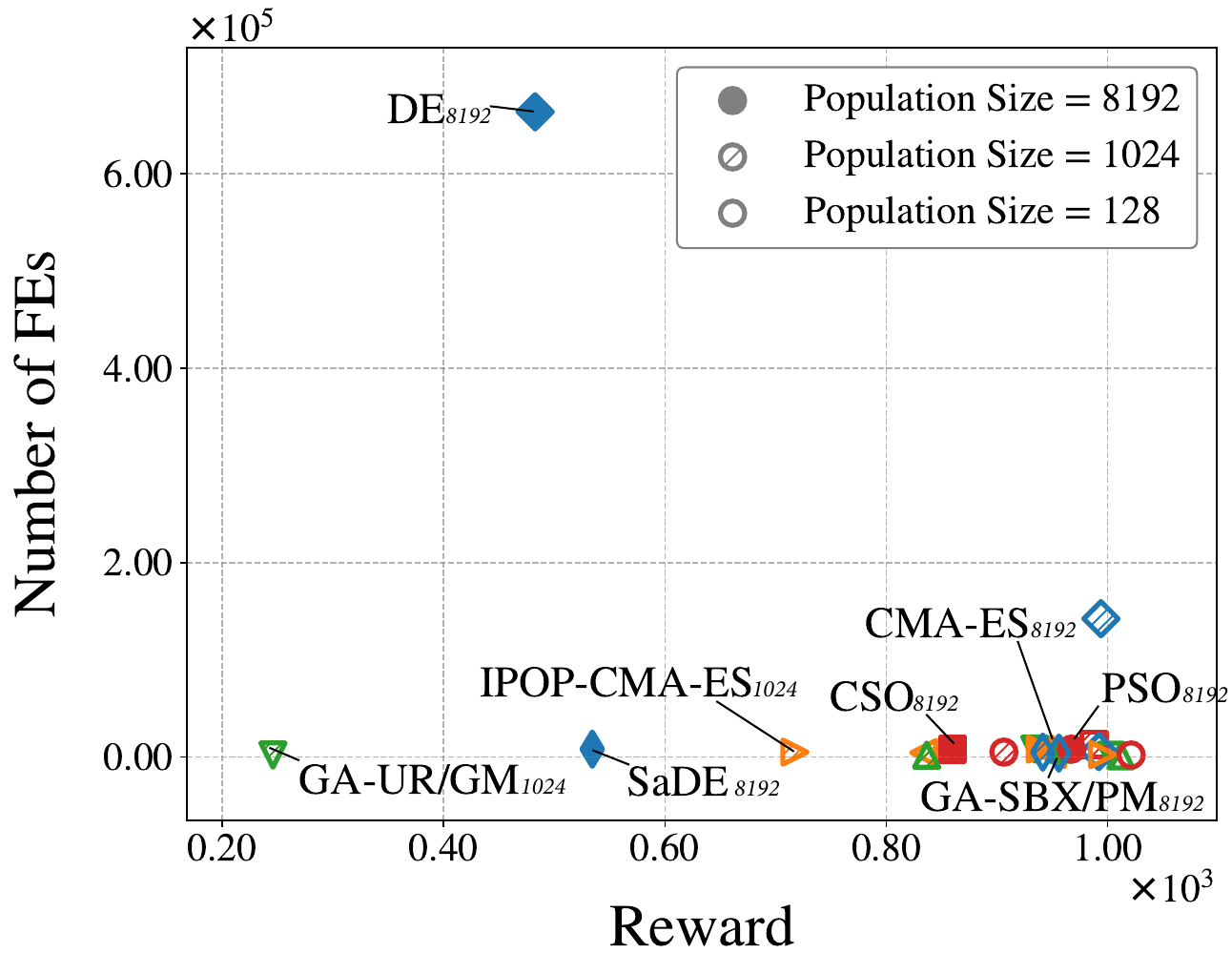}
        \centering
        \vspace{-0.5cm}
        \subcaption{Hopper}
    \end{minipage}
    \hfill
    \begin{minipage}[b]{0.235\textwidth}
        \centering
        \includegraphics[width=\textwidth]{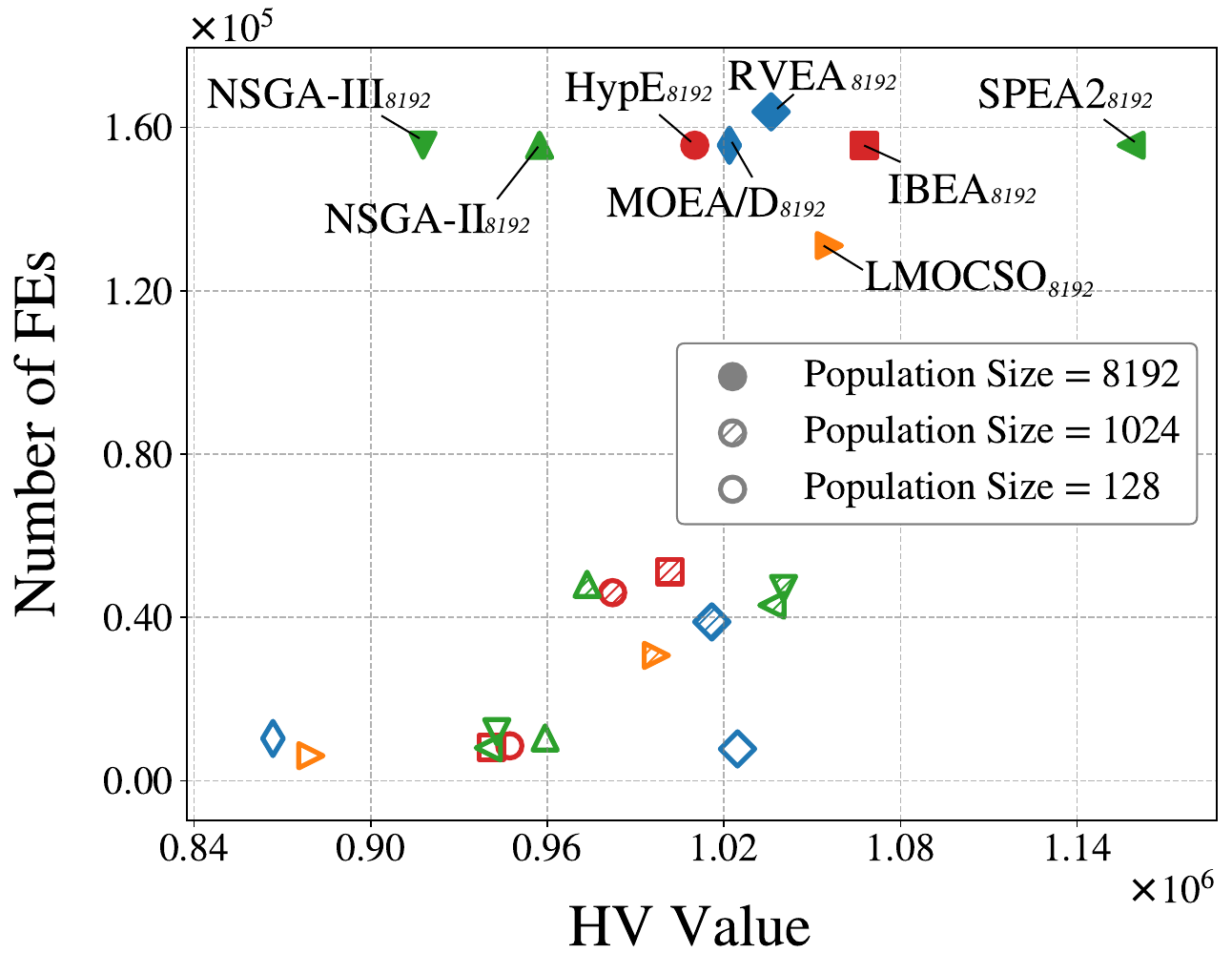}
        \centering
        \vspace{-0.5cm}
        \subcaption{MoHopper-m2}
    \end{minipage}
   \hfill
    \begin{minipage}[b]{0.235\textwidth}
        \centering
        \includegraphics[width=\textwidth]{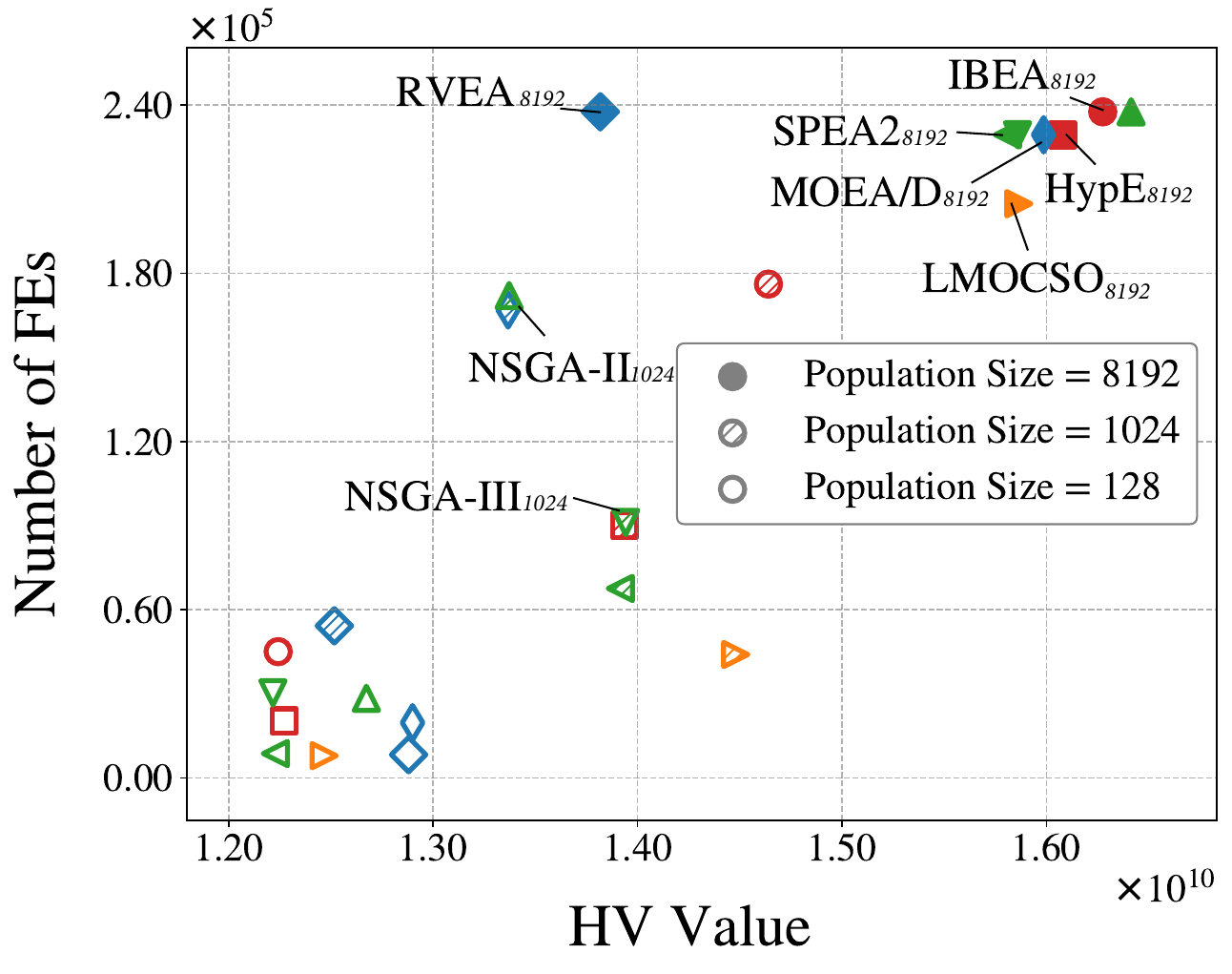}
        \centering
        \vspace{-0.5cm}
        \subcaption{MoHopper-m3}
    \end{minipage}
    \hfill

    \vspace{0.2cm}

    \begin{minipage}[b]{0.235\textwidth}
        \centering
        \includegraphics[width=\textwidth]{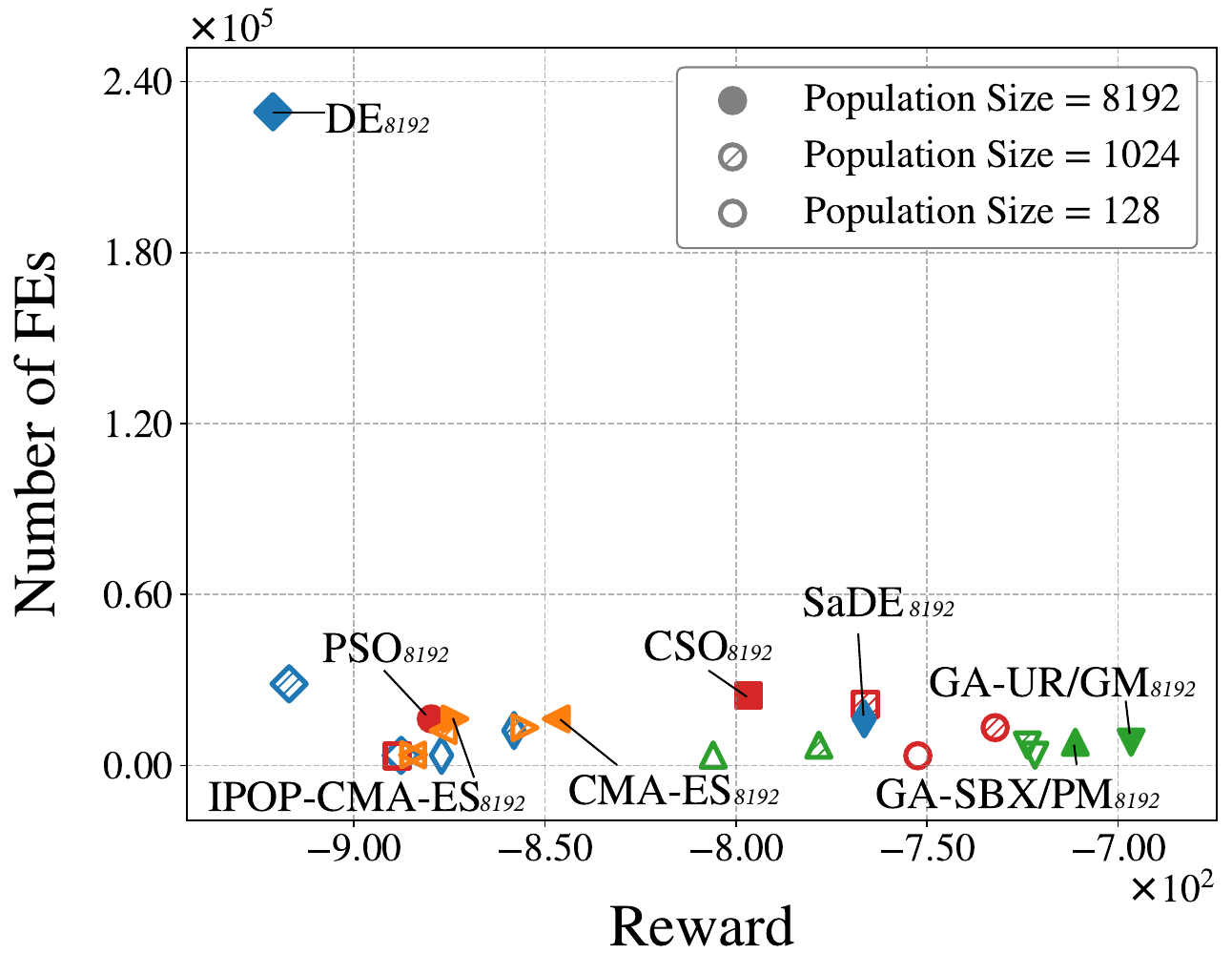}
        \centering
        \vspace{-0.5cm}
        \subcaption{Pusher}
    \end{minipage}
   \hfill
    \begin{minipage}[b]{0.235\textwidth}
        \centering
        \includegraphics[width=\textwidth]{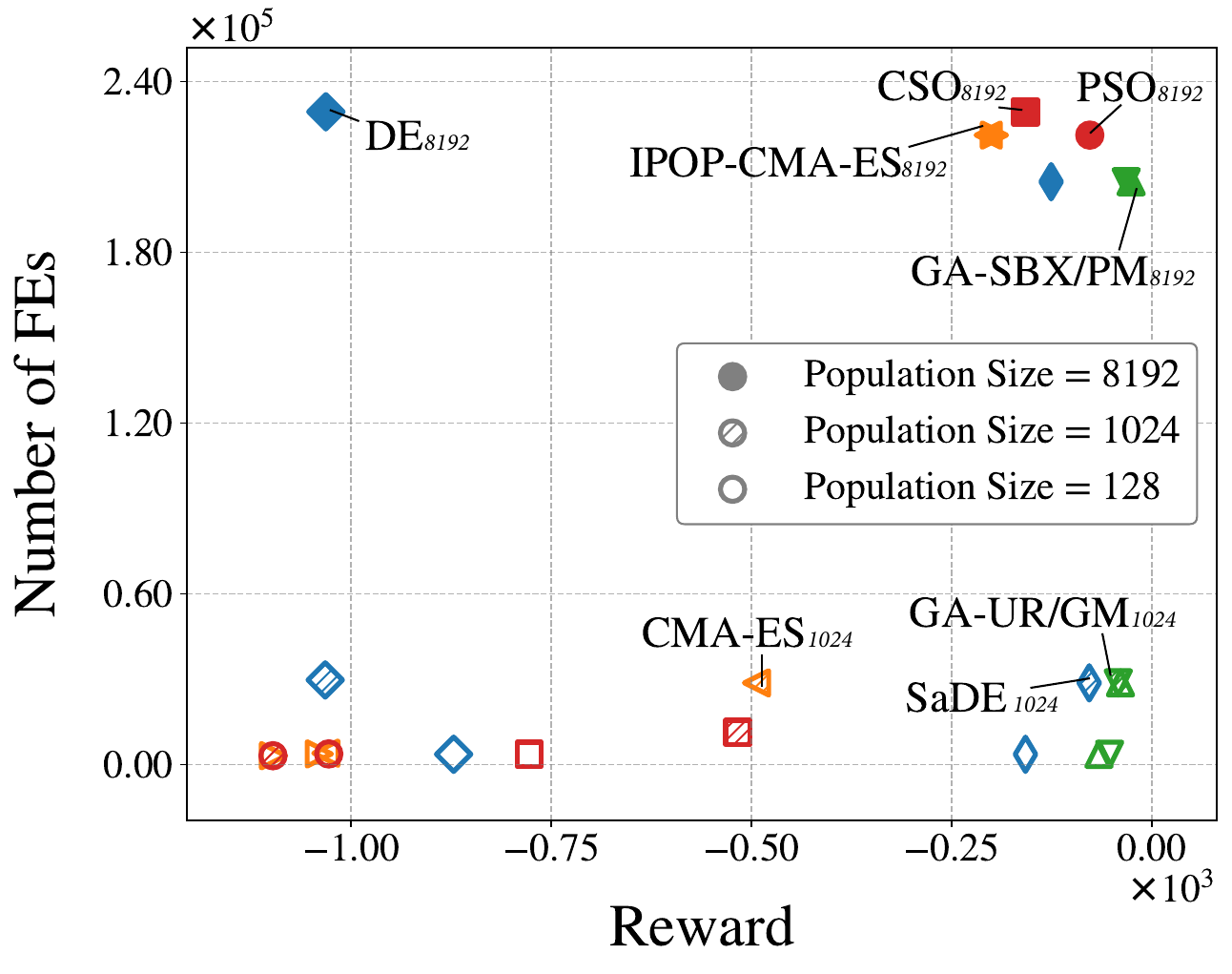}
        \centering
        \vspace{-0.5cm}
        \subcaption{Reacher}
    \end{minipage}
    \hfill
    \begin{minipage}[b]{0.235\textwidth}
        \centering
        \includegraphics[width=\textwidth]{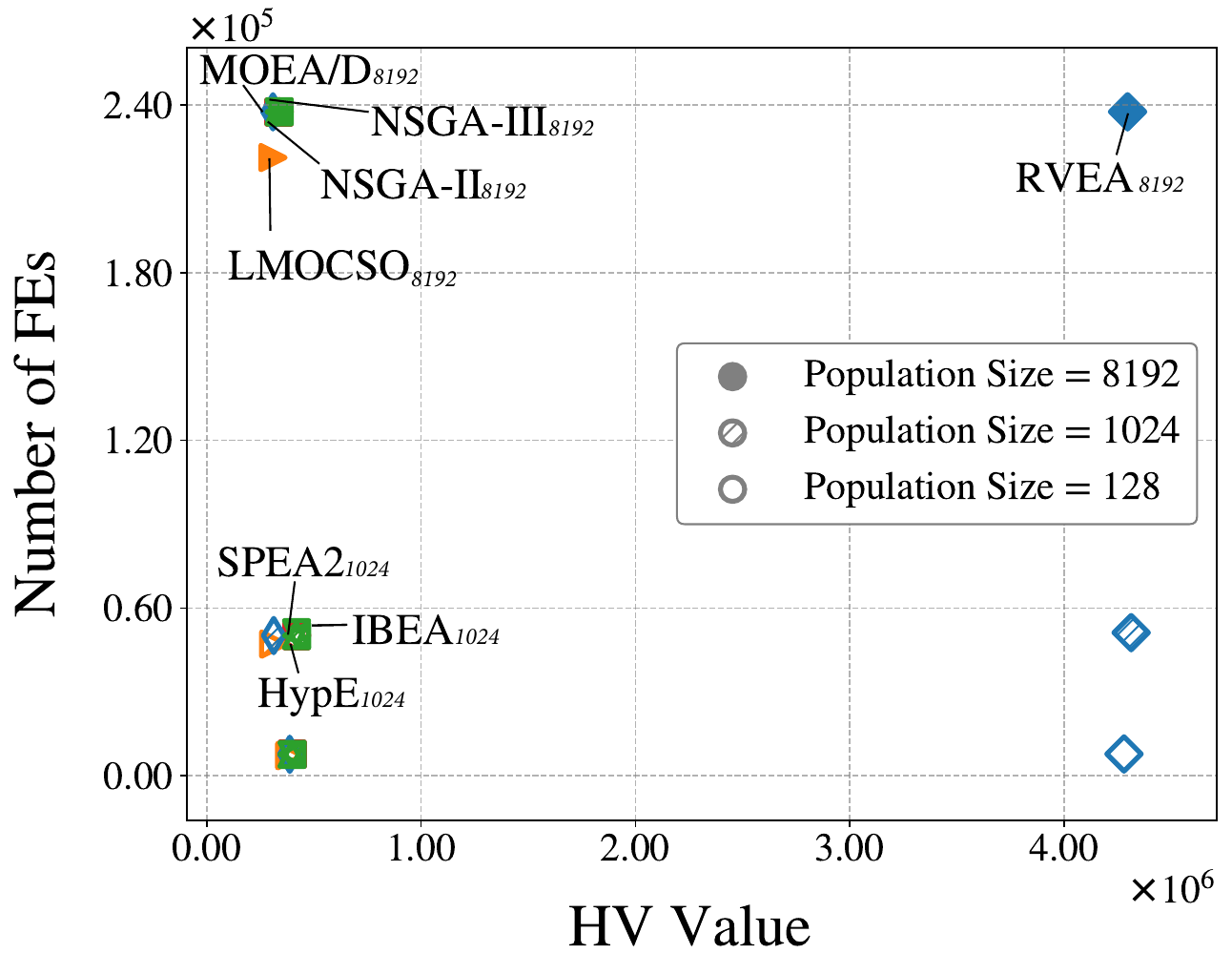}
        \centering
        \vspace{-0.5cm}
        \subcaption{MoReacher}
    \end{minipage}
   \hfill
    \begin{minipage}[b]{0.235\textwidth}
        \centering
        \includegraphics[width=\textwidth]{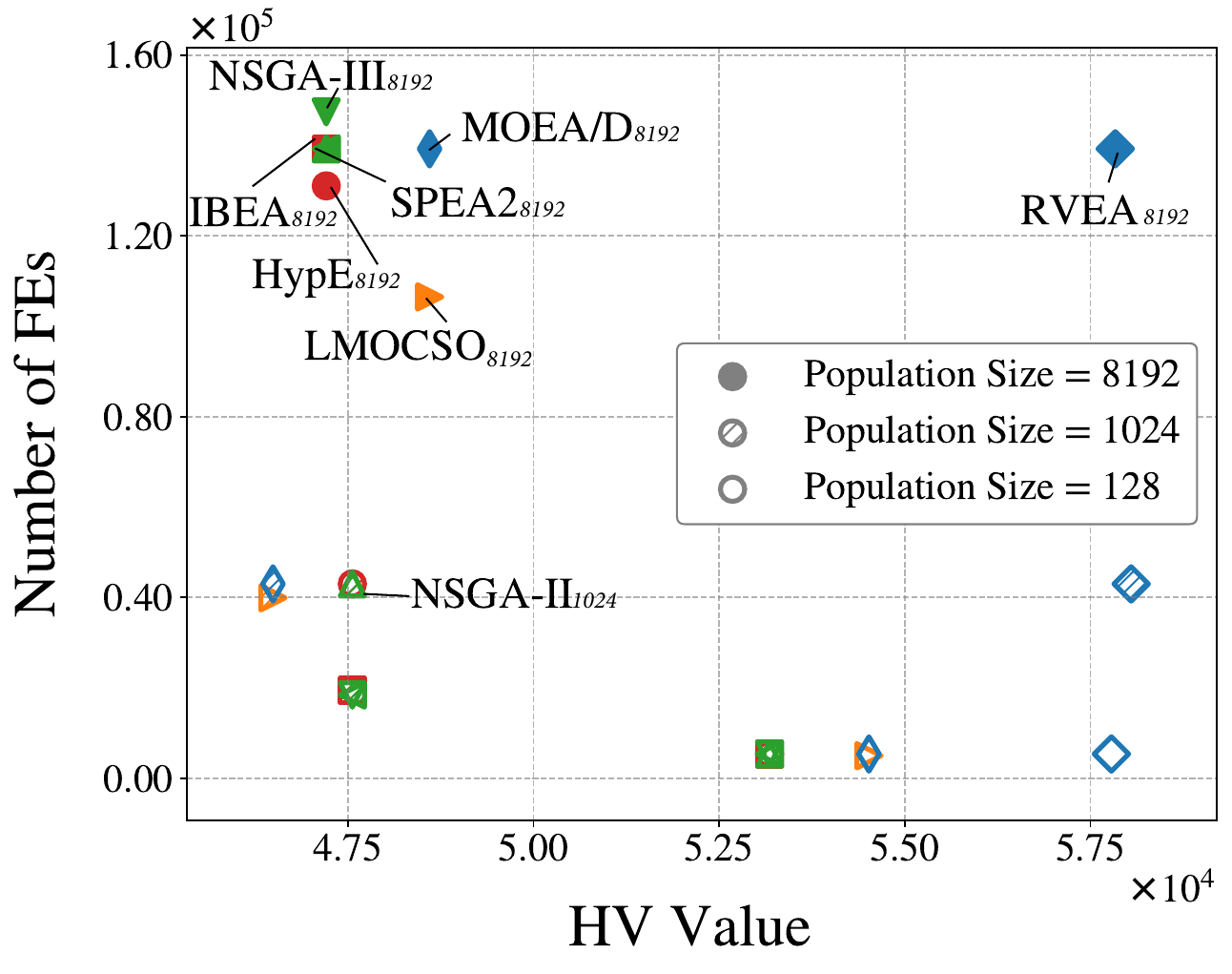}
        \centering
        \vspace{-0.5cm}
        \subcaption{MoSwimmer}
    \end{minipage}
    \hfill

    \vspace{0.2cm}

    \begin{minipage}[b]{0.235\textwidth}
        \centering
        \includegraphics[width=\textwidth]{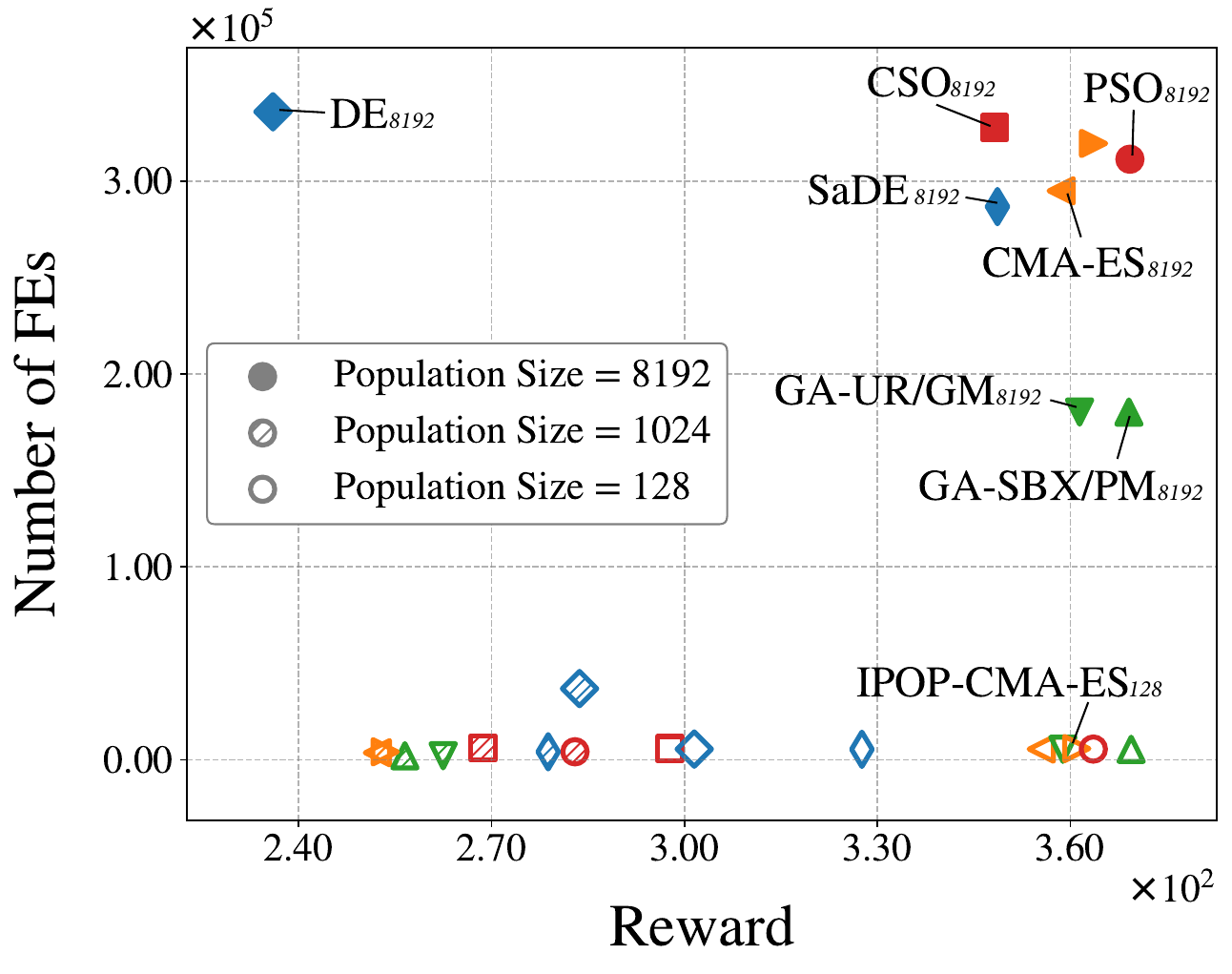}
        \centering
        \vspace{-0.5cm}
        \subcaption{Swimmer}
    \end{minipage}
    \hspace{2cm}
    \begin{minipage}[b]{0.235\textwidth}
        \centering
        \includegraphics[width=\textwidth]{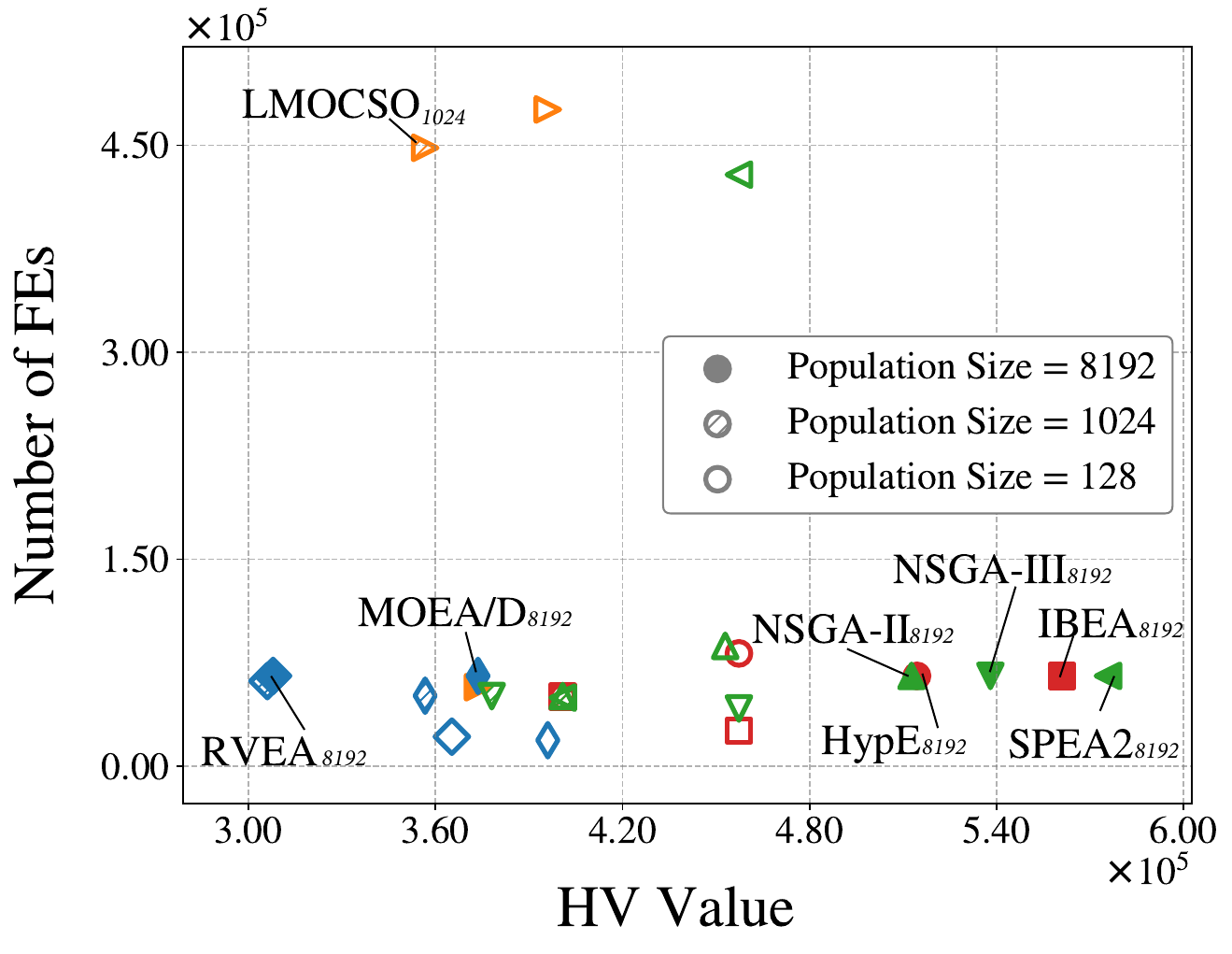}
        \centering
        \vspace{-0.5cm}
        \subcaption{MoWalker2d}
    \end{minipage}

    \caption{Performance comparison of EAs tested on neuroevolution tasks under varying population sizes with an NVIDIA RTX-2080-Ti GPU, evaluated in terms of solution quality and number of FEs completed within 600 seconds. Higher reward/HV values denote better performance. Results represent averaged performance values across 10 independent runs. Marker styles indicate population scales: hollow symbols for small populations (128), forward-slash-filled symbols for medium populations (1024), and solid symbols for large populations (8192). Different marker shapes distinguish between algorithms.}
    \label{fig:s-varying-popsize-neuro-mo}
\end{figure}

\end{document}